\documentclass{report}
\usepackage[final,nonatbib]{nips_2018}
\usepackage{natbib}
\usepackage[utf8]{inputenc}
\usepackage{paralist}
\usepackage[dvipsnames]{xcolor}
\definecolor{linkColor}{rgb}{0.18,0.39,0.62}
\usepackage[colorlinks=true,linkcolor=linkColor,citecolor=linkColor,filecolor=linkColor,urlcolor=linkColor]{hyperref}       

\usepackage{url}
\usepackage{graphicx}
\graphicspath{{fig/}}
\title{Vision-Language Pre-training: \\ Basics, Recent Advances, and Future Trends}

\author{
{\bf Zhe Gan, Linjie Li, Chunyuan Li, Lijuan Wang, Zicheng Liu, Jianfeng Gao}\\
Microsoft Corporation\\
{\tt\small \{zhgan,linjli,chunyl,lijuanw,zliu,jfgao\}@microsoft.com}}

\usepackage{mwe}
\usepackage{amsmath}
\usepackage{amssymb}
\usepackage{wasysym}
\usepackage{float}
\usepackage{bbm}
\usepackage{enumitem}
\usepackage{multirow}
\usepackage{arydshln}
\usepackage{amssymb}
\usepackage{pifont}
\usepackage{booktabs}
\usepackage{subcaption}
\usepackage[edges]{forest}
\newsavebox{\measurebox}
\definecolor{paired-light-blue}{RGB}{198, 219, 239}
\definecolor{paired-dark-blue}{RGB}{49, 130, 188}
\definecolor{paired-light-orange}{RGB}{251, 208, 162}
\definecolor{paired-dark-orange}{RGB}{230, 85, 12}
\definecolor{paired-light-green}{RGB}{199, 233, 193}
\definecolor{paired-dark-green}{RGB}{49, 163, 83}
\definecolor{paired-light-purple}{RGB}{218, 218, 235}
\definecolor{paired-dark-purple}{RGB}{117, 107, 176}
\definecolor{paired-light-gray}{RGB}{217, 217, 217}
\definecolor{paired-dark-gray}{RGB}{99, 99, 99}
\definecolor{paired-light-pink}{RGB}{222, 158, 214}
\definecolor{paired-dark-pink}{RGB}{123, 65, 115}
\definecolor{paired-light-red}{RGB}{231, 150, 156}
\definecolor{paired-dark-red}{RGB}{131, 60, 56}
\definecolor{paired-light-yellow}{RGB}{231, 204, 149}
\definecolor{paired-dark-yellow}{RGB}{141, 109, 49}
\tikzset{%
    parent/.style =          {align=center,text width=1cm,rounded corners=3pt, line width=0.3mm, fill=gray!10,draw=gray!80},
    child/.style =           {align=center,text width=2.3cm,rounded corners=3pt, fill=blue!10,draw=blue!80,line width=0.3mm},
    grandchild/.style =      {align=center,text width=2cm,rounded corners=3pt},
    greatgrandchild/.style = {align=center,text width=1.5cm,rounded corners=3pt},
    greatgrandchild2/.style = {align=center,text width=1.5cm,rounded corners=3pt},    
    referenceblock/.style =  {align=center,text width=1.5cm,rounded corners=2pt},
    data/.style =           {align=center,text width=2cm,rounded corners=3pt, fill=paired-light-blue!50,draw=paired-dark-blue!65,line width=0.3mm},
    data_wide/.style =           {align=center,text width=3cm,rounded corners=3pt, fill=paired-light-blue!50,draw=paired-dark-blue!65,line width=0.3mm},   
    data_work/.style =           {align=center, text width=4.5cm,rounded corners=3pt, fill=paired-light-blue!50,draw=blue!0,line width=0.3mm},  
    model/.style =           {align=center,text width=2cm,rounded corners=3pt, fill=paired-light-orange!50,draw=paired-dark-orange!65,line width=0.3mm},  
    model_more/.style =           {align=center,text width=4cm,rounded corners=3pt, fill=paired-light-orange!50,draw=paired-dark-orange!65,line width=0.3mm}, 
    model_work/.style =           {align=center,text width=4.5cm,rounded corners=3pt, fill=paired-light-orange!50,draw=red!0,line width=0.3mm},    
    pretraining/.style =           {align=center,text width=2cm,rounded corners=3pt, fill= paired-light-green!50,draw=paired-dark-green!75,line width=0.3mm}, 
    pretraining_wide/.style =           {align=center,text width=2.5cm,rounded corners=3pt, fill= paired-light-green!50,draw=paired-dark-green!75,line width=0.3mm}, 
    pretraining_more/.style =           {align=center,text width=4cm,rounded corners=3pt, fill= paired-light-green!50,draw=paired-dark-green!75,line width=0.3mm},   
    pretraining_work/.style =           {align=center,text width=4.5cm,rounded corners=3pt, fill= paired-light-green!50,draw= cyan!0,line width=0.3mm},      
    finetuning/.style =           {align=center,text width=2cm,rounded corners=3pt, fill= paired-light-purple!50,draw=paired-dark-purple!75,line width=0.3mm},   
    finetuning_work/.style =           {align=center,text width=4.5cm,rounded corners=3pt, fill= paired-light-purple!50,draw= orange!0,line width=0.3mm},        
    inference/.style =           {align=center,text width=2cm,rounded corners=3pt, fill= paired-light-red!35,draw=paired-light-red!90,line width=0.3mm},           
    inference_more/.style =           {align=center,text width=4cm,rounded corners=3pt, fill= paired-light-red!35,draw=paired-light-red!90,line width=0.3mm},
    inference_work/.style =           {align=center,text width=4.5cm,rounded corners=3pt, fill= paired-light-red!35,draw= magenta!0,line width=0.3mm},         
}

\newcommand\blfootnote[1]{%
  \begingroup
  \renewcommand\thefootnote{}\footnote{#1}%
  \addtocounter{footnote}{-1}%
  \endgroup
}

\definecolor{chp5orange}{RGB}{238, 135, 114}
\definecolor{chp5green}{RGB}{169, 209, 142}
\definecolor{chp5blue}{RGB}{115, 149, 211}

\begin{document}

\maketitle

\vspace{4cm}
\begin{abstract}
This paper surveys vision-language pre-training (VLP) methods for multimodal intelligence that have been developed in the last few years. We group these approaches into three categories: ($i$) VLP for image-text tasks, such as image captioning, image-text retrieval, visual question answering, and visual grounding; ($ii$) VLP for core computer vision tasks, such as (open-set) image classification, object detection, and segmentation; and ($iii$) VLP for video-text tasks, such as video captioning, video-text retrieval, and video question answering.  For each category, we present a comprehensive review of state-of-the-art methods, and discuss the progress that has been made and challenges still being faced, using specific systems and models as case studies. In addition, for each category, we discuss advanced topics being actively explored in the research community, such as big foundation models, unified modeling, in-context few-shot learning, knowledge, robustness, and computer vision in the wild, to name a few.
\end{abstract}

\blfootnote{$^\spadesuit$Zhe Gan and Jianfeng Gao initiated the project. Zhe Gan and Linjie Li took lead in the writing of Chapter~\ref{chp:intro}. Linjie Li and Jianfeng Gao took lead in the writing of Chapter~\ref{chp:basics}. Zhe Gan further took lead in the writing of Chapter~\ref{chp:vlp4imgtxt} and \ref{chp:conclusion}. Chunyuan Li took lead in the writing of Chapter~\ref{chp:vlp4vision}. Linjie Li further took lead in the writing of Chapter~\ref{chp:vlp4videotxt}. Lijuan Wang and Zicheng Liu took lead in the writing of Chapter~\ref{chp:industry}. All the authors provided project advice, and contributed to paper editing and proofreading.
}

\newcommand{\todo}[1]{{\color{red}{[{\bf TODO}: #1]}}}
\newcommand{\todoJG}[1]{{\color{blue}{[{\bf JG}: #1]}}}
\newcommand{\todoMG}[1]{{\color{purple}{[{\bf MG}: #1]}}}
\newcommand{\todoLL}[1]{{\color{cyan}{[{\bf LL}: #1]}}}

\newcommand{\figref}[1]{Fig.~\ref{#1}}
\newcommand{\eqnref}[1]{Eqn.~\ref{#1}}
\newcommand{\chref}[1]{Chapter~\ref{#1}}
\newcommand{\secref}[1]{Sec.~\ref{#1}}
\newcommand{\tabref}[1]{Table~\ref{#1}}
\newcommand{\ie}{{i.e.}}
\newcommand{\eg}{{e.g.}}
\newcommand{\etc}{{etc.}}
\newcommand{\cf}{{cf.}}

\newcommand{\dnfont}[1]{{\texttt{#1}}}  
\newcommand{\dafont}[1]{{\texttt{#1}}}  
\newcommand{\slotfont}[1]{\texttt{#1}}  
\newcommand{\valuefont}[1]{\textcolor{gray}{\texttt{#1}}}
\newcommand{\vecb}[1]{\mathbf{#1}}

\newcommand{\exbox}[1]{ 
{\begin{center}\fbox{%
    \begin{minipage}{.95\textwidth} 
      \centering #1
    \end{minipage}%
  }\end{center}} 
}

\newcommand{\defeq}{:=}
\newcommand{\E}{\mathbb{E}}
\newcommand{\Rset}{\mathbb{R}}
\newcommand{\mt}{{\operatorname{T}}}  			
\newcommand{\mi}{{-1}}  											
\newcommand{\argmin}{\operatorname{argmin}}
\newcommand{\argmax}{\operatorname{argmax}}

\newcommand{\Sset}{\mathcal{S}}
\newcommand{\Aset}{\mathcal{A}}
\newcommand{\Dset}{\mathcal{D}}

\newcommand{\mMSE}{\operatorname{MSE}}
\newcommand{\mFone}{\operatorname{F1}}
\newcommand{\mREC}{\operatorname{RECALL}}
\newcommand{\mACC}{\operatorname{ACCURACY}}
\newcommand{\mPRE}{\operatorname{PRECISION}}
\newcommand{\mAUC}{\operatorname{AUC}}
\newcommand{\1}{\mathbf{1}}

\newcommand{\av}{{\boldsymbol{a}}}
\newcommand{\bv}[0]{{\boldsymbol{b}}}
\newcommand{\cv}[0]{{\boldsymbol{c}}}
\newcommand{\dv}{\boldsymbol{d}}
\newcommand{\ev}[0]{{\boldsymbol{e}}\xspace}
\newcommand{\fv}[0]{{\boldsymbol{f}}}
\newcommand{\gv}[0]{{\boldsymbol{g}}}
\newcommand{\hv}[0]{{\boldsymbol{h}}}
\newcommand{\iv}[0]{{\boldsymbol{i}}\xspace}
\newcommand{\jv}[0]{{\boldsymbol{j}}\xspace}
\newcommand{\kv}[0]{{\boldsymbol{k}}\xspace}
\newcommand{\lv}[0]{{\boldsymbol{l}}}
\newcommand{\mv}[0]{{\boldsymbol{m}}}
\newcommand{\nv}[0]{{\boldsymbol{n}}\xspace}
\newcommand{\ov}[0]{{\boldsymbol{o}}\xspace}
\newcommand{\pv}[0]{{\boldsymbol{p}}}
\newcommand{\qv}[0]{{\boldsymbol{q}}}
\newcommand{\rv}{\boldsymbol{r}}
\newcommand{\sv}{{\boldsymbol{s}}}
\newcommand{\tv}[0]{{\boldsymbol{t}}}
\newcommand{\uv}{\boldsymbol{u}}
\newcommand{\vv}{\boldsymbol{v}}
\newcommand{\wv}{\boldsymbol{w}}
\newcommand{\xv}{\boldsymbol{x}}
\newcommand{\yv}{\boldsymbol{y}}
\newcommand{\zv}{\boldsymbol{z}}
\newcommand{\cdotv}{\boldsymbol{\cdot}}

\newcommand{\Amat}[0]{{{\bf A}}}
\newcommand{\Bmat}{{\bf B}}
\newcommand{\Cmat}{{\bf C}}
\newcommand{\Dmat}{{\bf D}}
\newcommand{\Emat}[0]{{{\bf E}}}
\newcommand{\Fmat}[0]{{{\bf F}}\xspace}
\newcommand{\Gmat}{{\bf G}}
\newcommand{\Hmat}{{\bf H}}
\newcommand{\Imat}{{\bf I}}
\newcommand{\Jmat}[0]{{{\bf J}}\xspace}
\newcommand{\Kmat}[0]{{{\bf K}}\xspace}
\newcommand{\Lmat}[0]{{{\bf L}}}
\newcommand{\Mmat}{{\bf M}}
\newcommand{\Nmat}[0]{{{\bf N}}\xspace}
\newcommand{\Omat}[0]{{{\bf O}}}
\newcommand{\Pmat}{{\bf P}}
\newcommand{\Qmat}[0]{{{\bf Q}}\xspace}
\newcommand{\Rmat}[0]{{{\bf R}}}
\newcommand{\Smat}[0]{{{\bf S}}}
\newcommand{\Tmat}[0]{{{\bf T}}}
\newcommand{\Umat}[0]{{{\bf U}}}
\newcommand{\Vmat}[0]{{{\bf V}}}
\newcommand{\Wmat}[0]{{{\bf W}}}
\newcommand{\Xmat}[0]{{{\bf X}}}
\newcommand{\Ymat}{{\bf Y}}
\newcommand{\Zmat}{{\bf Z}}

\newcommand{\Xcal}{\mathcal{X}}
\newcommand{\Ycal}{\mathcal{Y}}
\newcommand{\Ncal}{\mathcal{N}}
\newcommand{\Acal}{\mathcal{A}}
\newcommand{\Bcal}{\mathcal{B}}
\newcommand{\Dcal}{\mathcal{D}}
\newcommand{\Fcal}{\mathcal{F}}
\newcommand{\Tcal}{\mathcal{T}}
\newcommand{\Mcal}{\mathcal{M}}
\newcommand{\Lcal}{\mathcal{L}}
\newcommand{\Ocal}{\mathcal{O}}
\newcommand{\Pcal}{\mathcal{P}}
\newcommand{\Ical}{\mathcal{I}}
\newcommand{\Kcal}{\mathcal{K}}
\newcommand{\Gcal}{\mathcal{G}}
\newcommand{\Qcal}{\mathcal{Q}}
\newcommand{\Rcal}{\mathcal{R}}
\newcommand{\Scal}{\mathcal{S}}
\newcommand{\Hcal}{\mathcal{H}}
\newcommand{\Vcal}{\mathcal{V}}
\newcommand{\Zcal}{\mathcal{Z}}
\newcommand{\Ucal}{\mathcal{U}}
\newcommand{\Jcal}{\mathcal{J}}
\newcommand{\scal}{\mathcal{s}}

\newcommand{\alphav}{\boldsymbol{\alpha}}
\newcommand{\chiv}{\boldsymbol{\chi}}
\newcommand{\betav}[0]{{\boldsymbol{\beta}}}
\newcommand{\gammav}[0]{{\boldsymbol{\gamma}}\xspace}
\newcommand{\deltav}[0]{{\boldsymbol{\delta}}\xspace}
\newcommand{\epsilonv}{\boldsymbol{\epsilon}}
\newcommand{\zetav}{\boldsymbol{\zeta}}
\newcommand{\etav}{\boldsymbol{\eta}}
\newcommand{\ellv}[0]{{\boldsymbol{\ell}}}
\newcommand{\thetav}{\boldsymbol{\theta}}
\newcommand{\iotav}[0]{{\boldsymbol{\iota}}}
\newcommand{\kappav}[0]{{\boldsymbol{\kappa}}\xspace}
\newcommand{\lambdav}[0]{{\boldsymbol{\lambda}}}
\newcommand{\muv}[0]{{\boldsymbol{\mu}}}
\newcommand{\nuv}[0]{{\boldsymbol{\nu}}}
\newcommand{\xiv}[0]{{\boldsymbol{\xi}}}
\newcommand{\omicronv}[0]{{\boldsymbol{\omicron}}\xspace}
\newcommand{\piv}{\boldsymbol{\pi}}
\newcommand{\rhov}[0]{{\boldsymbol{\rho}}\xspace}
\newcommand{\sigmav}[0]{{\boldsymbol{\sigma}}}
\newcommand{\tauv}[0]{{\boldsymbol{\tau}}}
\newcommand{\upsilonv}[0]{{\boldsymbol{\upsilon}}\xspace}
\newcommand{\phiv}{\boldsymbol{\phi}}
\newcommand{\psiv}{\boldsymbol{\psi}}
\newcommand{\varthetav}{\boldsymbol{\vartheta}}
\newcommand{\omegav}[0]{{\boldsymbol{\omega}}}
\newcommand{\R}{\mathbb{R}}
\newcommand{\Z}{\mathbb{Z}}
\newcommand{\specialcell}[2][c]{%
  \begin{tabular}[#1]{@{}c@{}}#2\end{tabular}}
\newcommand{\specialcelll}[2][l]{%
  \begin{tabular}[#1]{@{}l@{}}#2\end{tabular}}

\tableofcontents
\newpage

\chapter{Introduction}
\label{chp:intro}

Humans perceive the world through many channels, such as images viewed by the eyes, or voices heard by the ears. Though any individual channel might be incomplete or noisy, humans can naturally align and fuse information collected from multiple channels in order to grasp the key concepts needed for a better understanding of the world.

One of the core aspirations in AI is to develop algorithms that endow computers with an ability to effectively learn from multimodal (or, multi-channel) data. This data is similar to sights and sounds attained from \emph{vision} and \emph{language} that help humans make sense of the world around us. For example, computers could mimic this ability by searching the most relevant images to a text query (or vice versa), and by describing the content of an image using natural language.

Vision-and-Language (VL), a popular research area that sits at the nexus of Computer Vision and Natural Language Processing (NLP), aims to achieve this goal. Inspired by the great success of language model pre-training in NLP (\emph{e.g.}, BERT~\citep{devlin2018bert}, RoBERTa~\citep{liu2019roberta}, T5~\citep{raffel2020exploring}, and GPT-3~\citep{brown2020language}), Vision-Language Pre-training (VLP) has recently attracted rapidly growing attention from both communities. With the promise to learn universal transferable visual and vision-language representations, VLP has become an increasingly central training paradigm for modern VL research.  

Recently, there are some related survey papers on VLP. \cite{zhang2020multimodal} focused on task-specific VL methods before the era of pre-training, and provided a concise discussion of VLP models. \cite{du2022survey,li2022vision} focused on VLP, but mainly on image-text tasks, without touch on video-text tasks. \cite{ruan2022survey} focused on VLP for video-text tasks. \cite{chen2022vlp} reviewed VLP methods for image-text and video-text tasks. However, the discussion is not in depth. The contributions of this survey paper are summarized as follows.
\begin{itemize}[leftmargin=*]
    \item We provide a comprehensive survey on modern VLP, not only covering its successful applications to traditional image-text and video-text tasks (\emph{e.g.}, image/video captioning, retrieval, and question answering), but also showing its great potential for core computer vision tasks (\emph{e.g.}, image classification, object detection and segmentation).
    \item We provide in-depth discussions on advanced topics at the frontier of VLP, ranging from big foundation models, unified modeling, in-context few-shot learning, knowledge-enhanced VLP, multilingual VLP, model robustness, model compression, to computer vision in the wild. 
    \item We picture the landscape of VL systems developed in research communities and released to public, demonstrating via case studies the progress we have made and the challenges we are facing.
\end{itemize}
 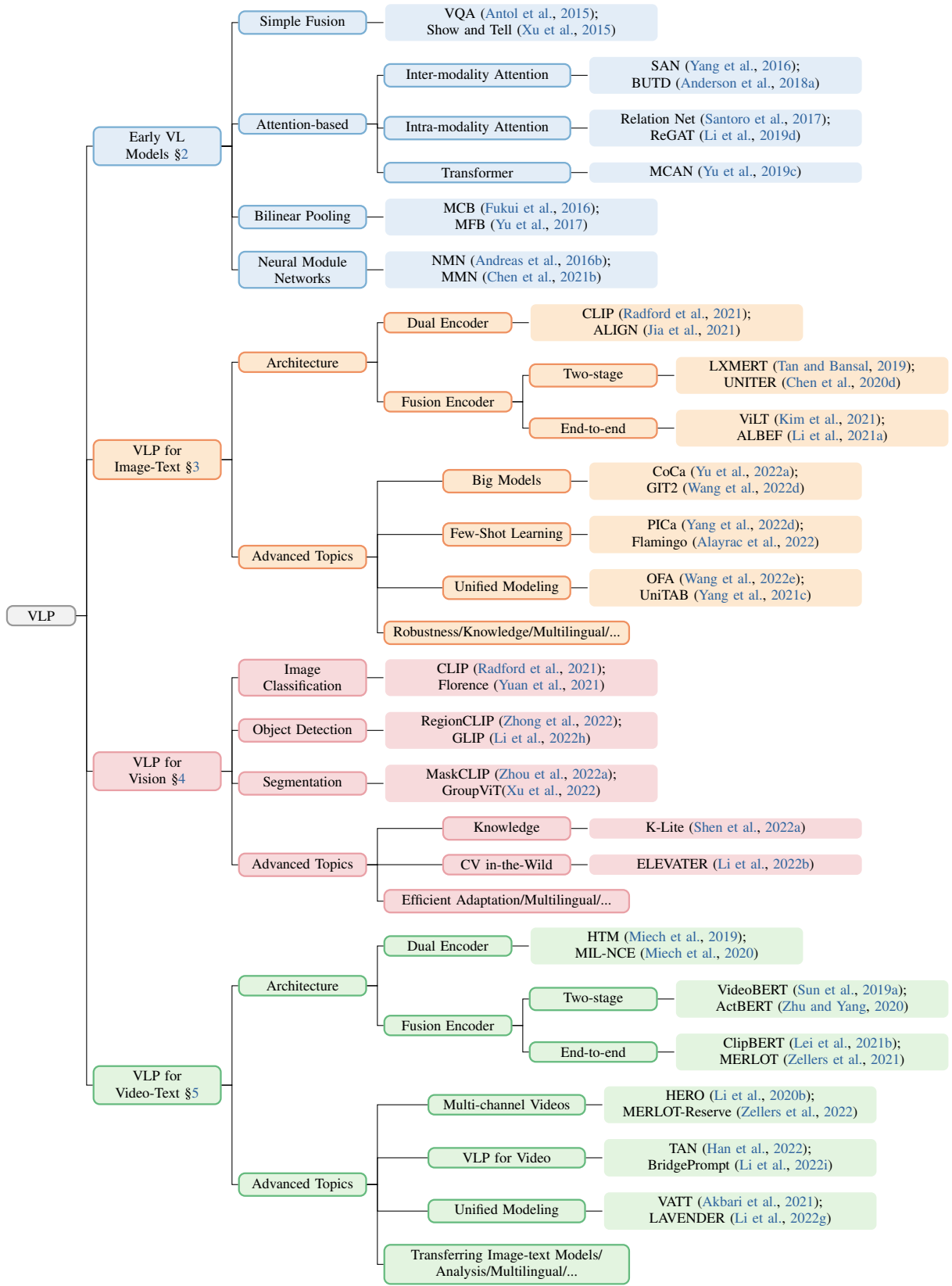
\begin{figure*}
\scriptsize
\hspace*{-30pt}
    \begin{forest}
        for tree={
            forked edges,
            grow'=0,
            draw,
            rounded corners,
            node options={align=center,},
            text width=2.7cm,
            s sep=6pt,
            calign=edge midpoint,
        },
        [VLP, fill=gray!45, parent
            [Early VL Models \S\ref{chp:basics}, for tree={ data}
                [Simple Fusion,  data
                    [VQA~\citep{antol2015vqa}; Show and Tell~\citep{xu2015show}, data_work]
                ]
                [Attention-based, data 
                    [Inter-modality Attention,  data_wide
                        [SAN~\citep{yang2016stacked}; BUTD~\citep{anderson2018bottom}, data_work]
                    ]
                    [Intra-modality Attention ,  data_wide 
                        [Relation Net~\citep{santoro2017simple};\\ ReGAT~\citep{li2019relation}, data_work]
                    ]
                    [Transformer,  data_wide
                        [MCAN~\citep{yu2019deep}, data_work]
                    ]
                ]
                [Bilinear Pooling,  data
                    [MCB~\citep{fukui2016multimodal}; MFB~\citep{yu2017multi}, data_work]
                ]
                [Neural Module\\Networks,  data
                    [NMN~\citep{andreas2016neural}; MMN~\citep{chen2021meta}, data_work]
                ]
            ]
            [VLP for Image-Text \S\ref{chp:vlp4imgtxt}, for tree={fill=red!45,model}
                [Architecture, model
                    [Dual Encoder, model
                        [CLIP~\citep{radford2021learning}; ALIGN~\citep{jia2021scaling}, model_work]
                    ]
                    [Fusion Encoder, model
                        [Two-stage, model
                            [LXMERT~\citep{tan-bansal-2019-lxmert}; UNITER~\citep{chen2020uniter}, model_work]
                        ]
                        [End-to-end, model
                            [ViLT~\citep{kim2021vilt}; ALBEF~\citep{li2021align}, model_work]
                        ]
                    ]
                ]
                [Advanced Topics, model
                    [Big Models, model
                        [CoCa~\citep{yu2022coca}; GIT2~\citep{wang2022git}, model_work]
                    ]
                    [Few-Shot Learning, model
                        [PICa~\citep{yang2021empirical}; Flamingo~\citep{alayrac2022flamingo}, model_work]
                    ]
                    [Unified Modeling, model
                        [OFA~\citep{wang2022ofa}; UniTAB~\citep{yang2021crossing}, model_work]
                    ]
                    [Robustness/Knowledge/Multilingual/..., model_more]
                ]
            ]
            [VLP for\\Vision \S\ref{chp:vlp4vision}, for tree={inference}
                [Image\\Classification, inference
                    [CLIP~\citep{radford2021learning}; Florence~\citep{yuan2021florence}, inference_work]
                ]
                [Object Detection, inference
                    [RegionCLIP~\citep{zhong2021regionclip}; \\ GLIP~\citep{li2021grounded}, inference_work]
                ]
                [Segmentation, inference
                    [MaskCLIP~\citep{zhou2021maskclip}; GroupViT\citep{xu2022groupvit}, inference_work]
                ]
                [Advanced Topics, inference
                    [Knowledge, inference
                        [K-Lite~\citep{shen2022k}, inference_work]
                    ]
                    [CV in-the-Wild, inference
                        [ELEVATER~\citep{li2022elevater}, inference_work]
                    ]
                    [Efficient Adaptation/Multilingual/..., inference_more]
                ]
            ]    
            [VLP for Video-Text \S\ref{chp:vlp4videotxt}, for tree={fill=blue!45, pretraining}
                [Architecture, pretraining
                    [Dual Encoder, pretraining
                        [HTM~\citep{miech2019howto100m}; MIL-NCE~\citep{miech19endtoend}, pretraining_work]
                    ]
                    [Fusion Encoder, pretraining
                        [Two-stage, pretraining
                            [VideoBERT~\citep{sun2019videobert}; ActBERT~\citep{zhu2020actbert}, pretraining_work]
                        ]
                        [End-to-end, pretraining
                            [ClipBERT~\citep{lei2021less}; MERLOT~\citep{zellers2021merlot}, pretraining_work]
                        ]
                    ]
                ]
                [Advanced Topics, pretraining
                    [Multi-channel Videos, pretraining_wide
                        [HERO~\citep{li2020hero}; \\ MERLOT-Reserve~\citep{zellers2022merlot}, pretraining_work]
                    ]
                    [VLP for Video, pretraining_wide
                        [TAN~\citep{han2022temporal}; \\ BridgePrompt~\citep{li2022bridge}, pretraining_work]
                    ]
                    [Unified Modeling, pretraining_wide
                        [VATT~\citep{akbari2021vatt}; LAVENDER~\citep{li2022lavender}, pretraining_work]
                    ]
                    [Transferring Image-text Models/\\Analysis/Multilingual/..., pretraining_more]
                ]
            ]  
        ]
    \end{forest}
    \caption{Overview of the paper structure, detailing Chapter \ref{chp:basics}-\ref{chp:vlp4videotxt}. 
    }
    \label{fig:paper_structure}
\end{figure*}

\section{Who Should Read this Paper?}
This paper is based on our CVPR 2022 tutorial\footnote{\url{https://vlp-tutorial.github.io/}}, with researchers in the computer vision and NLP communities as our primary target audience. It provides a detailed presentation of the important ideas and insights needed to understand modern VLP methods, and serves as a valuable resource for students, researchers, engineers, and practitioners that are interested in large-scale pre-training for VL representation learning and its applications in computer vision and multimodal tasks. The paper is structured as follows. 

\begin{itemize}[leftmargin=*]
    \item Chapter~\ref{chp:intro} introduces the landscape of VL research, and presents a historical view on the transition of VL research from task-specific methods to large-scale pre-training.
    \item Chapter~\ref{chp:basics} introduces early task-specific VL methods for visual question answering, image captioning, and image-text retrieval, which serve as the foundation to understand modern VLP methods. 
    \item Chapter~\ref{chp:vlp4imgtxt} describes VLP methods for image-text tasks, such as image captioning, image-text retrieval, visual question answering, and visual grounding. 
    \item Chapter~\ref{chp:vlp4vision} describes VLP methods for core computer vision tasks, including (open-vocabulary) image classification, object detection and segmentation.
    \item Chapter~\ref{chp:vlp4videotxt} describes VLP methods for video-text tasks, such as video captioning, video-text retrieval, and video question answering.
    \item Chapter~\ref{chp:industry} briefly reviews VL systems developed in industry and the challenges to deploy these VL systems in real-world settings.
    \item Chapter~\ref{chp:conclusion} concludes the paper and discusses research trends.
\end{itemize}

\paragraph{Relations between core chapters.} Chapter~\ref{chp:basics}-\ref{chp:vlp4videotxt} are the core chapters of this survey paper. An overview of the structure for these chapters are provided in Figure~\ref{fig:paper_structure}. As the wave of VLP starts with image-text tasks, we first provide a comprehensive review on the transition from early task-specific methods (Chapter~\ref{chp:basics}) to most recent VLP methods (Chapter~\ref{chp:vlp4imgtxt}) with image-text inputs. In Chapter~\ref{chp:vlp4vision}, we discuss how core computer vision tasks can be viewed as image-text tasks with open-vocabulary predictions, when powered by contrastively pre-trained image-text models (such as CLIP~\citep{radford2021learning}), and further enable computer vision in the wild~\citep{li2022elevater}. Extending image-text tasks to more modalities, we present how VLP methods can serve more applications with video-text inputs in Chapter~\ref{chp:vlp4videotxt}. 

\paragraph{How to read the paper.} Different readers have different backgrounds, and may have different purposes of reading this paper. Here, we provide a few guidance. 
\begin{itemize}[leftmargin=*]
    \item Each chapter is mostly self-contained. If you have a clear goal and a clear research direction that you want to focus on, then just jump to the corresponding chapter. For example, if you are interested in video-language pre-training, then you can directly jump to Chapter~\ref{chp:vlp4videotxt}.
    
    \item If you are a beginner in the VLP field, and are interested in getting a glimpse of the cutting-edge research of VLP, it is also highly suggested to read the whole paper chapter by chapter, as the paper provides a comprehensive literature review
    that helps you understand the VLP landscape.
    
    \item If you already have rich experience in VLP and are very familiar with the literature, feel free to jump to specific chapters you want to read. In particular, we include in each chapter a dedicated section to discuss advanced topics. For example, in Section~\ref{sec:chp3_adv_topics}, we have discussed big foundation models, unified image-text modeling, in-context few-shot learning, knowledge, robustness and probing analysis, \emph{etc}.
\end{itemize}


\section{Vision-and-Language: What Kinds of Problems?}

\begin{figure*}[t!]
  \centering
    \includegraphics[width=1.0\linewidth]{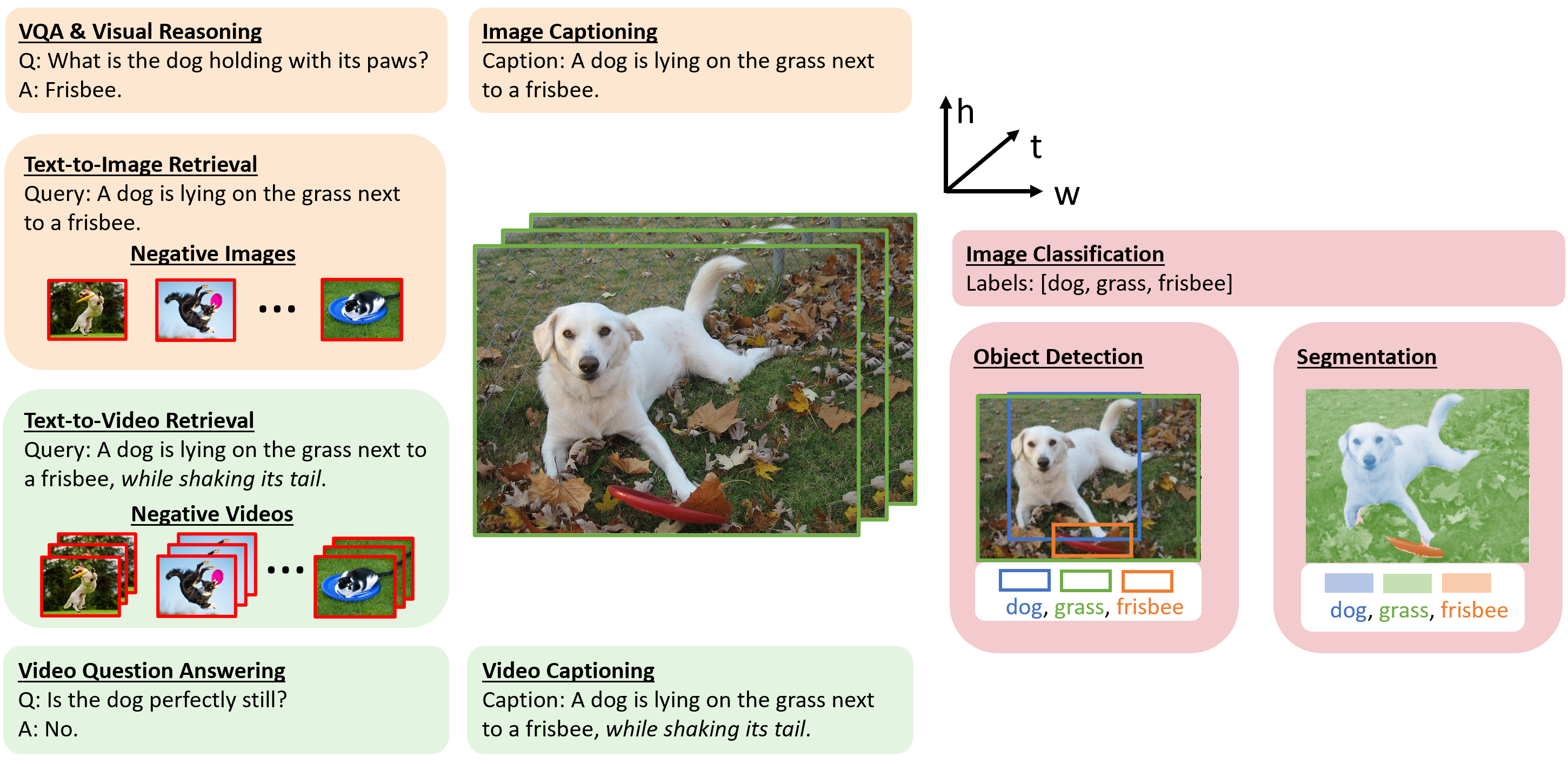}
  \caption{Illustration of representative tasks from three categories of VL problems covered in this paper:  \colorbox{paired-light-orange!50}{image-text tasks}, \colorbox{paired-light-red!50}{vision tasks as VL problems}, and \colorbox{paired-light-green!50}{video-text tasks}.}
  \label{fig:chp1_tasks}
\end{figure*}

We live in a multimodal world, and  our brains naturally learn to process multi-sense signals received from the environment to help us make sense of the world around us. More specifically, \emph{vision} is a large portion of how humans perceive, while \emph{language} is a large portion of how humans communicate. A multimodal AI system, by its definition, should have the ability to process such multimodal signals effectively and efficiently. Among the ever-growing literature on VL research, in this paper, we group VL problems into three categories, as detailed below.
\begin{itemize}[leftmargin=*]
    \item \textbf{Image-Text Tasks.} Arguably, the most important and well-studied tasks in VL research are image-text retrieval, image captioning~\citep{vinyals2015show}, and visual question answering (VQA)~\citep{antol2015vqa} (highlighted with orange in Figure~\ref{fig:chp1_tasks}). Centered around these tasks, many related tasks have been proposed and studied. 
    \begin{itemize}[leftmargin=*]
        \item \textbf{VQA and visual reasoning.} As extensions to visual question answering, researchers have developed datasets for visual reasoning~\citep{hudson2019gqa,suhr2018corpus}, visual commonsense reasoning~\citep{zellers2019recognition}, visual dialog~\citep{das2017visual},  knowledge-based VQA~\citep{marino2019ok}, scene-text-based VQA~\citep{singh2019towards}, \emph{etc}. The answers required in these these tasks can be open-ended free-form texts, or selected from multiple choices. 
        
        \item \textbf{Image captioning.} In addition to the setting where short single-sentence generation is required~\citep{lin2014microsoft}, researchers have also developed datasets for image paragraph captioning~\citep{krause2017hierarchical}, scene-text-based image captioning~\citep{sidorov2020textcaps}, visual storytelling~\citep{huang2016visual}, and so on. 
        
        \item \textbf{Image-text retrieval.} Popular image-text retrieval datasets are based on image captioning datasets~\citep{chen2015microsoftcoco, plummer2015flickr30k}. AI models are required to retrieve the most relevant text (or image) from a large corpus, given the image (or text) query. 
        
        \item \textbf{Visual grounding.} 
        Instead of text outputs, referring expression comprehension and phrase grounding~\citep{yu2016modeling,plummer2015flickr30k} requires bounding box outputs, where the model needs to predict the bounding box corresponding to the input text query. 
        
        \item \textbf{Text-to-image generation.} It can be considered as the dual task of image captioning, where the system is required to create a high-fidelity image based on the text input. A brief discussion on this task is provided in Section~\ref{sec:vlp4imggen}.
    \end{itemize}
    
    \item \textbf{Computer Vision Tasks as VL Problems.} Image classification, object detection, and segmentation  (highlighted with pink in Figure~\ref{fig:chp1_tasks}) are  core visual recognition tasks in computer vision. Traditionally, these tasks are considered as pure vision problems. As the advent of CLIP~\citep{radford2021learning} and ALIGN~\citep{jia2021scaling}, researchers have realized that language supervision can play an important role in computer vision tasks. First, the use of noisy image-text data crawled from web allows large-scale pre-training of vision encoders from scratch. Second, instead of treating the supervision signals (\emph{e.g.}, class labels) as one-hot vectors, we take the semantic meaning behind the labels into consideration and cast these computer vision tasks as VL problems. This  perspective generalizes the traditional close-set classification or detection models to recognizing unseen concepts in real-world applications, such as open-vocabulary object detection.
    
    \item \textbf{Video-Text Tasks.} Besides static images, videos are another important type of visual modality. Naturally, all aforementioned image-text tasks have their video-text counterparts, such as video captioning, retrieval, and question answering (highlighted with green in Figure~\ref{fig:chp1_tasks}). The uniqueness of video inputs, in comparison to images, requires an AI system to not only capture spatial information within a single video frame, but also capture the inherent temporal dependencies among video frames. 
\end{itemize}

While this paper provides a comprehensive survey of VLP, some of the important VL topics are not discussed.
For example, Vision-Language Navigation (VLN)~\citep{anderson2018vision}, another emerging topic at the intersection of VL research and embodied AI, is not covered in this paper.  


\begin{figure*}[t!]
  \centering
    \includegraphics[width=1.0\linewidth]{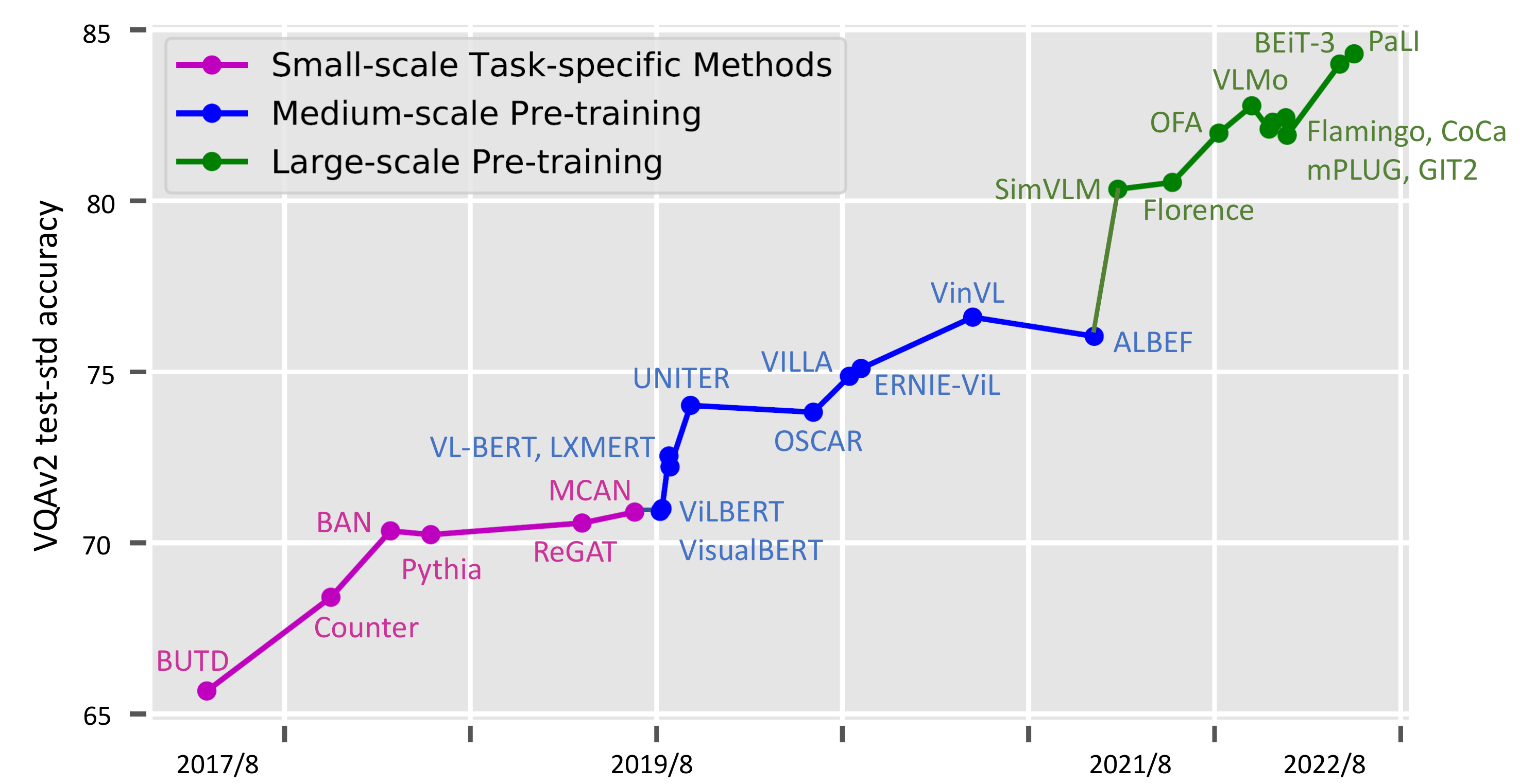}
  \caption{The transition from task-specific methods to large-scale pre-training, using the VQA task as a case study. Every time when there was a transition, we observe a big performance lift, \emph{e.g.}, from MCAN~\citep{yu2019deep} to UNITER~\citep{chen2020uniter}, and from ALBEF~\citep{li2021align} to SimVLM~\citep{wang2021simvlm}. Methods before August 2017 were not drawn; only some representative VLP works are shown to avoid the figure to be too crowded.}
  \label{fig:chp1_vlp_progress}
  \vspace{-2mm}
\end{figure*}

\section{The Transition From Task-Specific Methods to Large-Scale Pre-training}
From a historical perspective, the progress of VL research can be divided into three stages. In Figure~\ref{fig:chp1_vlp_progress}, we use the performance of the popular VQA task to illustrate the research transition from task-specific methods to medium-scale and large-scale pre-training.
\begin{itemize}[leftmargin=*]
    \item \textbf{Small-scale task-specific method design (2014/11-2019/8).} At this stage, many task-specific methods have been developed for image captioning and VQA. For example, an important line of work is to design various attention mechanisms based on pre-extracted visual features (\emph{e.g.}, ResNet~\citep{he2016deep}, Faster RCNN~\citep{ren2015faster}, C3D~\citep{tran2015learning}), pre-trained word embeddings (\emph{e.g.}, GLoVe~\citep{pennington2014glove}, word2vec~\citep{mikolov2013distributed}), and LSTM~\citep{hochreiter1997long}, as we will review in Chapter~\ref{chp:basics}. These attention method designs have been used to capture multimodal alignment, perform object relational reasoning, and model multi-step reasoning.
    
    \item \textbf{Medium-scale pre-training (2019/8-2021/8).} Inspired by the great success of BERT~\citep{devlin2018bert} in NLP, the VL field has gradually shifted to using Transformer-based multimodal fusion models that are pre-trained in medium-scale settings, \emph{e.g.}, using image-text datasets
    up to 4M images (roughly 10M image-text pairs in total), with model sizes ranging from 110M (BERT-base) to 340M (BERT-large). Typical examples of medium-scale VLP models include UNITER~\citep{chen2020uniter} and OSCAR~\citep{li2020oscar}, as will be described in Chapter~\ref{chp:vlp4imgtxt}.   
    
    \item \textbf{Large-scale pre-training (2021/8-now).} As the advent of CLIP~\citep{radford2021learning} and ALIGN~\citep{jia2021scaling} that aim to train  image-text dual encoders from noisy image-text pairs crawled from the web, large-scale VLP
    shows great promise and is becoming the foundation of 
    VL research. We have witnessed a boom of big multimodal foundation models, \emph{e.g.}, SimVLM~\citep{wang2021simvlm}, Florence~\citep{yuan2021florence}, Flamingo~\citep{alayrac2022flamingo}, CoCa~\citep{yu2022coca} and GIT~\citep{wang2022git}. 
    The high computational cost of VLP can be amortized via adapting the pre-trained models to a wide range of downstream tasks. The number of image-text pairs used for pre-training has increased to over 12B, with model sizes growing to 5B, as in GIT~\citep{wang2022git}. We provide some detailed discussion on big models in Section~\ref{sec:big_models}. 
\end{itemize}



\section{What is a Good VLP Model From an Overall Perspective?}
While VLP is an emerging field with many new exciting papers appearing, it remains less clear what is the north star we are pursuing as a community. We provide our perspective on the direction. We believe a good VLP model should: 

\begin{itemize}[leftmargin=*]
    \item \textbf{Achieve good performance on a wide range of downstream tasks.} The task coverage can be considered in a two-level granularity. First, the problem types are broad, for example, one model can perform on image-text tasks such as VQA, image captioning and text-to-image generation in Chapter~\ref{chp:vlp4imgtxt}, core computer vision tasks such as image classification, object detection and segmentation in Chapter~\ref{chp:vlp4vision}, video-text tasks such as video QA and captioning in Chapter~\ref{chp:vlp4videotxt}. Second, for each problem type, there is a broad coverage of datasets that represent different use scenarios. For example, \cite{li2022elevater} present 20 image classification datasets and 35 object detection datasets to illustrate various scenarios in the wild.
    
    \item \textbf{Adapt to new tasks with minimal cost.} The adaptation cost needs to be low when deploying a VLP model to a new task. Various efficiency metrics can be considered to measure the adaptation cost, including inference speed, GPU usage for further model weight update, the number of training samples, and the number of trainable parameters. This is an area not well defined yet, and there has been some early effort. For example, \cite{li2022elevater} provide a definition by decomposing the adaptation cost into sample-efficiency and parameter-efficiency.
\end{itemize}

To summarize, the north star of a good VLP model is a single unified model with fixed model weights (or, with inexpensive finetuning) that performs well on all the tasks above. This is an ambitious goal that the community is collectively working towards. Developing a central benchmark is itself an open research problem. We advocate for considering the following factors when benchmarking VLP models: the coverage of tasks, the performance on these tasks, and the cost of adaptation.

\section{Related Materials: Slide Decks and Pre-recorded Talks}

This survey paper extends what we present in CVPR tutorials by covering the most recent advances in the field. Below, we provide a list of slide decks and pre-recorded talks, that relate to the topics in each chapter, for references.

\begin{itemize}[leftmargin=*]
    \item \textbf{Chapter 2}:
    \begin{itemize}[leftmargin=*]
        \item \href{https://rohit497.github.io/Recent-Advances-in-Vision-and-Language-Research/slides/tutorial-part-2-vqa.pdf}{CVPR 2020 Tutorial: VQA and visual reasoning}~(\href{https://www.youtube.com/watch?v=n4mUriUrYR0}{Youtube}, \href{https://www.bilibili.com/video/BV1DV411r7B1/}{Bilibili})
        \item \href{https://rohit497.github.io/Recent-Advances-in-Vision-and-Language-Research/slides/tutorial-part-3-captioning.pdf}{CVPR 2020 Tutorial: Image captioning}~(\href{https://www.youtube.com/watch?v=Zn5uFGsq4j4}{Youtube},
    \href{https://www.bilibili.com/video/BV14V411k7Ea/}{Bilibili})
    \end{itemize}
        
    \item \textbf{Chapter 3}:
    \begin{itemize}[leftmargin=*]
        \item \href{https://datarelease.blob.core.windows.net/tutorial/VLP-Tutorial_2022/image_text_part1.pdf}{CVPR 2022 Tutorial: Overview of Image-Text Pre-training}~(\href{https://youtu.be/ce4lIytxfIo}{YouTube}, \href{https://www.bilibili.com/video/BV1d3411w7cZ/}{Bilibili})
        \item \href{https://datarelease.blob.core.windows.net/tutorial/VLP-Tutorial_2022/image_text_part2.pdf}{CVPR 2022 Tutorial: Unified Image-Text Modeling}~(\href{https://youtu.be/xVIGQP5t-Sk}{YouTube}, \href{https://www.bilibili.com/video/BV1uG411x7db/}{Bilibili})
        \item \href{https://datarelease.blob.core.windows.net/tutorial/VLP-Tutorial_2022/image_text_part3.pdf}{CVPR 2022 Tutorial: Advanced Topics in Image-Text Pre-training}~(\href{https://youtu.be/CqB6zLi3dFo}{YouTube}, \href{https://www.bilibili.com/video/BV1hW4y1z7T8/}{Bilibili})
        \item \href{https://datarelease.blob.core.windows.net/tutorial/VQA2VLN2021/VLP_part1.pdf}{CVPR 2021 Tutorial: Representations and Training Strategies for VLP}~(\href{https://youtu.be/ToP9jI0kBtw}{YouTube})
        \item \href{https://datarelease.blob.core.windows.net/tutorial/VQA2VLN2021/VLP_part2.pdf}{CVPR 2021 Tutorial: Robustness, Efficiency and Extensions for VLP}~(\href{https://youtu.be/5XRYFLPBA1U}{YouTube})
        \item \href{https://rohit497.github.io/Recent-Advances-in-Vision-and-Language-Research/slides/tutorial-part5-pretraining.pdf}{CVPR 2020 Tutorial: Self-supervised Image-Text Learning}~(\href{https://www.youtube.com/watch?v=C4UQWJcp7w4}{YouTube}, \href{https://www.bilibili.com/video/BV1oD4y1D7V2}{Bilibili})
    \end{itemize}
    
    \item \textbf{Chapter 4}:
    \begin{itemize}[leftmargin=*]
        \item \href{https://datarelease.blob.core.windows.net/tutorial/VLP-Tutorial_2022/vlp_for_v_part1.pdf}{CVPR 2022 Tutorial: VLP for Image Classification}~(\href{https://youtu.be/Tq7RWYWN2M0}{Youtube},
    \href{https://www.bilibili.com/video/BV1Ur4y1g76P/}{Bilibili})
    \item \href{https://datarelease.blob.core.windows.net/tutorial/VLP-Tutorial_2022/vlp_for_v_part2.pdf}{CVPR 2022 Tutorial: VLP for Object Detection}~(\href{https://youtu.be/XtZti41bMeY}{Youtube}, \href{https://www.bilibili.com/video/BV1ra411W7qV/}{Bilibili})
    \item \href{https://datarelease.blob.core.windows.net/tutorial/VLP-Tutorial_2022/vlp_for_v_part3.pdf}{CVPR 2022 Tutorial: Benchmarks for Computer Vision in the Wild} (\href{https://youtu.be/F519jcAppFA}{YouTube}, \href{https://www.bilibili.com/video/BV1FG411x7uB/}{Bilibili})
    \end{itemize}
    
    \item \textbf{Chapter 5}: 
    \begin{itemize}[leftmargin=*]
        \item \href{https://datarelease.blob.core.windows.net/tutorial/VLP-Tutorial_2022/video_text_part1.pdf}{CVPR 2022 Tutorial: Overview of Video-Text Pre-training}~(\href{https://youtu.be/LAeT1sBX6fc}{YouTube}, \href{https://www.bilibili.com/video/BV113411w7Tg/}{Bilibili})
        \item \href{https://datarelease.blob.core.windows.net/tutorial/VLP-Tutorial_2022/video_text_part2.pdf}{CVPR 2022 Tutorial: Learning from Multi-channel Videos: Methods and Benchmarks}~(\href{https://youtu.be/iTKvj1E8Re8}{YouTube}, \href{https://www.bilibili.com/video/BV1uU4y1X76o/}{Bilibili})
        \item \href{https://datarelease.blob.core.windows.net/tutorial/VLP-Tutorial_2022/video_text_part3.pdf}{CVPR 2022 Tutorial: Advanced Topics in Video-Text Pre-training}~(\href{https://youtu.be/WBK3PjOs1RA}{YouTube},  \href{https://www.bilibili.com/video/BV1gT41137cD/}{Bilibili})
        \item \href{https://datarelease.blob.core.windows.net/tutorial/VQA2VLN2021/VLP_part3.pdf}{CVPR 2021 Tutorial: Video-and-Language Pre-training}~(\href{https://youtu.be/19Z6cghHWMc}{Youtube})
    \end{itemize}
\end{itemize}

\chapter{Tasks, Benchmarks, and Early Models}
\label{chp:basics}


In Section~\ref{chp2-sec:task-datasets}, we first introduce major vision-language (VL) tasks and the benchmarks that are commonly used in the research community. We group these tasks into two categories. VL understanding tasks, such as image-text retrieval and visual question answering (VQA), require a VL model to \emph{select} the output from a given list of candidates. VL generation tasks, such as image captioning, require a VL model to \emph{generate} the output. 
In Section~\ref{chp2-sec:task-specific-models}, we take VQA as an example to present the VL models developed prior to the era of large-scale VLP.
Early VL models typically take a pipeline approach. First, 
the image features are extracted by a pre-trained visual encoder. The textual features are computed using a text encoder. 
Then, the cross-modal representations are obtained, by performing multimodal fusion on top of these features, for the final prediction. One of the major research focuses is on the \emph{attention} design for multimodal fusion, which we use to categorize these models and to reflect how task-specific models evolve over time. We show that early VL models eventually evolve into a Transformer-based architecture (\textit{e.g.}, MCAN~\citep{yu2019deep}), which is similar to some early VLP models (\textit{e.g.}, LXMERT~\citep{tan-bansal-2019-lxmert} and ViLBERT~\citep{lu2019vilbert}), as to be discussed in detail in Chapter~\ref{chp:vlp4imgtxt}.
In Section~\ref{chp2-sec:additional-topics}, we review  additional research topics for the development of early VL models, including bilinear pooling, compositional visual reasoning, and visual grounding. 

\section{Tasks and Benchmarks}
\label{chp2-sec:task-datasets}
\begin{figure*}[t!]
  \centering
    \includegraphics[width=1.0\linewidth]{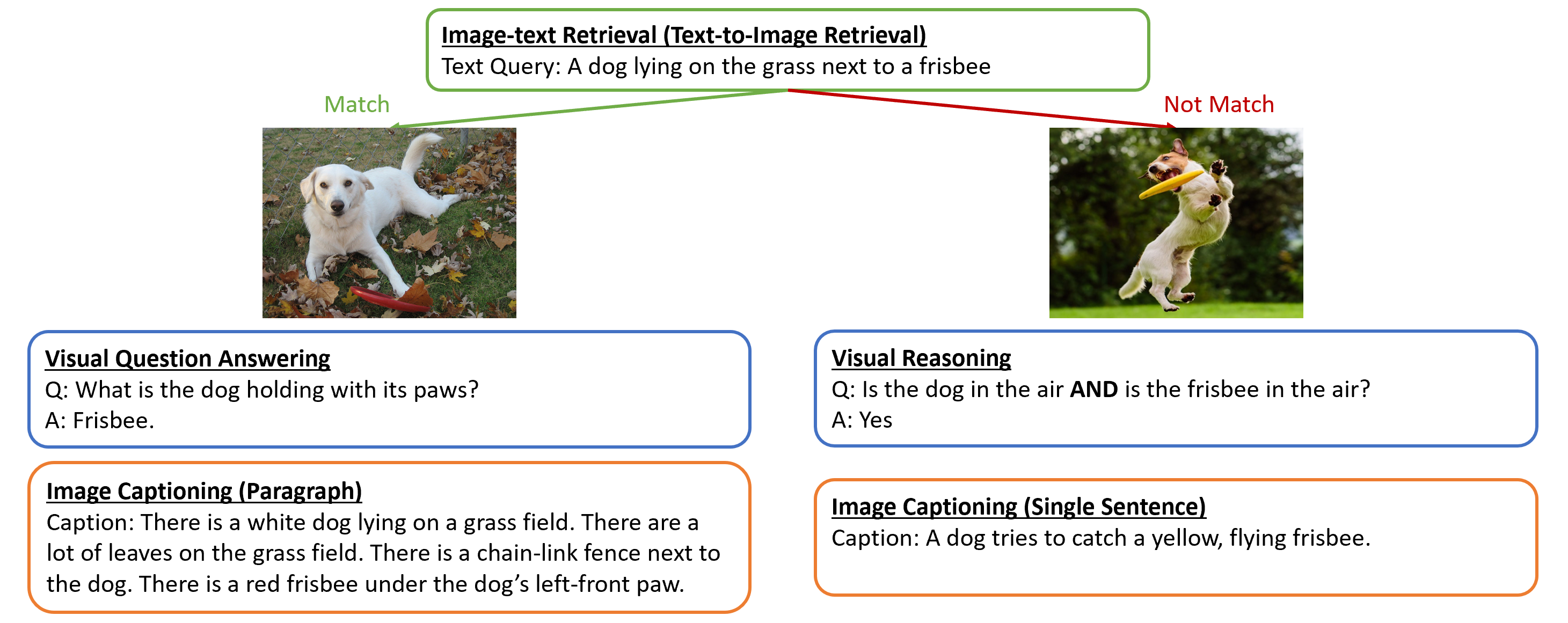}
  \caption{Illustration of representative vision-language tasks with image-text inputs: ($i$) image-text retrieval; ($ii$) visual question answering and visual reasoning; and ($iii$) image captioning with a single-sentence caption, or a more descriptive paragraph of captions.}
  \label{fig:chp2_tasks}
\end{figure*}

Casting as machine learning tasks, VL tasks can be formulated as $y = f(x;\theta)$, 
where we aim to learn a VL model $f$, parameterized by $\theta$, to generate output $y$ for input $x$. 
VL tasks can be categorized along two dimensions. 
\begin{itemize}[leftmargin=*]
    \item Depending on the modalities of $x$ and $y$, VL tasks can be grouped into image-text or video-text tasks. Here, we focus on popular image-text tasks and benchmarks, and defer the review of video-text tasks to Chapter~\ref{chp:vlp4videotxt}. 
    
    \item Depending on how $y$ is generated by $f$, VL tasks can be grouped into ($i$) \emph{understanding} tasks, such as image-text retrieval and visual question answering  (VQA), where $y$ is selected by $f$ from a given candidate list; and ($ii$) \emph{generation} tasks, such as image captioning and and text-to-image generation, where $y$ needs to be generated by $f$. In this section, we focus on three representative image-text tasks, including image-text retrieval, visual question answering (and its variant, visual reasoning), and image captioning. Examples of these tasks are illustrated in Figure~\ref{fig:chp2_tasks}. We introduce each task and representative benchmarks below.
\end{itemize}

\subsection{Image-text Retrieval}
Image-text retrieval can be categorized into two sub-tasks, including ($i$) text-to-image retrieval, which retrieves a relevant image given an input text query (illustrated in Figure~\ref{fig:chp2_tasks}), and ($ii$) image-to-text retrieval, which retrieves a textual description that can be grounded in the image query.  In both cases, the model needs to match the query to its relevant instances from a relatively large database (\textit{e.g.}, 1000-5000 images for a typical text-to-image retrieval task). Recall@K (K=1, 5, 10) is used as the evaluation metric. Popular datasets include COCO~\citep{chen2015microsoftcoco} and Flickr30K~\citep{plummer2015flickr30k}. 
\citet{sun2021lightningdot} propose to combine the training, validation and test sets of each dataset to form a larger candidate pool that can mimic a real-world text-to-image retrieval scenario which usually involves hundreds of thousands of images, and evaluate models in terms of both retrieval accuracy and inference speed.

\subsection{Visual Question Answering and Visual Reasoning}
\textbf{Visual Question Answering} (VQA)~\citep{antol2015vqa} is one of the most prominent VL tasks studied in the research community. Given an image-question pair, VQA requires the model to provide a correct answer to the question based on the image. There are two typically settings: ($i$) \textit{multiple-choice}, where a small set of answer choices (\textit{e.g.}, 4/5 answer choices) are provided, together with the image-question pair; and ($ii$) \textit{open-ended}, where the answer can be free-form that is not limited to any pre-defined answer candidates. 
However, to simplify the VQA tasks, most studies~\citep{antol2015vqa,anderson2018bottom,yu2019deep} treat both multiple-choice and open-ended VQA as classification problems. 
Specifically, the most frequent answers from the training set is selected to build an answer candidate set under open-ended setting. 
For example, the second version of VQA dataset, dubbed as VQAv2 \citep{goyal2017making}, contain approximately 3000 answers which can be used to form the list of candidates for all questions. 
As the VQA dataset contains 10 ground-truth answers per image-question pair, VQA score~\citep{antol2015vqa} is used to evaluate model performance. VQA score is defined as follows, considering the consensus among human annotators.
\begin{equation}
    \text{VQA score} = \min(\frac{\text{\# humans that provided that answer} }{3}, 1)\,.
\end{equation}
Recent studies have developed various VQA benchmarks. 
For example, Visual Dialog~\citep{das2017visual} extends single-round VQA to multi-round dialogue scenarios.
TextVQA~\citep{singh2019towards}, ST-VQA~\citep{biten2019scene} and OCR-VQA~\citep{mishraICDAR19} collect questions regarding scene texts in images. VizWiz-QA~\citep{gurari2018vizwiz} collects real-world VQA examples from visually-impaired people.
OK-VQA~\citep{marino2019ok} features questions based on both the image content and external knowledge. 
Another line of studies designs different diagnostic datasets based on the original VQA dataset~\citep{antol2015vqa,goyal2017making} to perform a stress test for VQA models. For instance, VQA-Rephrasing~\citep{shah2019cycle} exposes the brittleness of VQA models to linguistic variations in questions. VQA-CP~\citep{agrawal2018don} is designed to evaluate question-oriented language bias in VQA models.
\citet{agarwal2020towards} propose to study the robustness of VQA models against automated semantic image manipulations, and test the prediction consistency to questions on clean images and their corresponding manipulated images.

\textbf{Visual Reasoning} is a VL task, aiming to 
evaluate specific reasoning capabilities of a VL model. 
Most visual reasoning tasks are formulated as VQA. 
For example, GQA~\citep{hudson2019gqa} constructs large-scale rule-based questions that require multiple reasoning skills, spatial understanding and multi-step inference to produce answers.
VQA-LOL~\citep{gokhale2020vqa} generates questions via logical compositions and linguistic transformations over the VQAv2~\citep{goyal2017making} examples to examine model's ability of logical reasoning.
\citet{selvaraju2020squinting} develop a dataset containing perception-related sub-questions per question for a new reasoning split of the original VQA dataset~\citep{antol2015vqa,goyal2017making}. 
Visual Commonsense Reasoning (VCR)~\citep{zellers2019recognition} develop a multiple-choice question answering dataset that requires higher-order cognition and commonsense reasoning about the image content. 
Other visual reasoning datasets test a VL model's ability to match text and image content. For instance, NLVR$^2$~\citep{suhr2018corpus} requires the model to determine whether a natural language statement is true about a pair of input images. Visual Entailment~\citep{xie2019visual} asks the model to predict whether an image semantically entails its paired text. 

VQA score is used to evaluate models on all datasets derived from the VQA datasets~\citep{antol2015vqa,goyal2017making}. Accuracy is the default evaluation metric for all the other benchmarks.

\subsection{Image Captioning}
Image captioning is to generate a free-form textual caption for a given image. Captioning performance is usually evaluated on standard text generation metrics based on n-gram overlap, such as BLEU~\citep{papineni2002bleu}, METEOR~\citep{banerjee2005meteor}, ROUGE-L~\citep{lin2004rouge} and CIDEr~\citep{vedantam2015cider}. In addition, semantic content matching metrics, such as SPICE \citep{anderson2016spice}, are used to measure the similarity between model-generated text and references by extracting explicit semantic information units from text beyond n-grams. 

As shown in Figure~\ref{fig:chp2_tasks}, two kinds of captions are proposed for the image captioning task. Popular datasets, mostly designed with single-sentence captions, include COCO~\citep{chen2015microsoftcoco}, TextCaps~\citep{sidorov2020textcaps}, NoCaps~\citep{agrawal2019nocaps} and VizWiz-Captions~\citep{gurari2020captioning}. There have been less efforts~\citep{krause2017hierarchical} on building datasets with more descriptive, multi-sentence captions.  On the modeling side, most work~\citep{farhadi2010every,kulkarni2013babytalk,fang2015captions,anderson2018bottom} focus on the single-sentence captioning task.

\section{Task-specific VL Models}
\label{chp2-sec:task-specific-models}
Early VL models, which are developed before the era of large-scale VLP, usually tackle one specific VL task. In this section, we use VQA as the pivot task to review the architecture of these task-specific VL models.

\begin{figure*}[t!]
  \centering
    \includegraphics[width=1.0\linewidth]{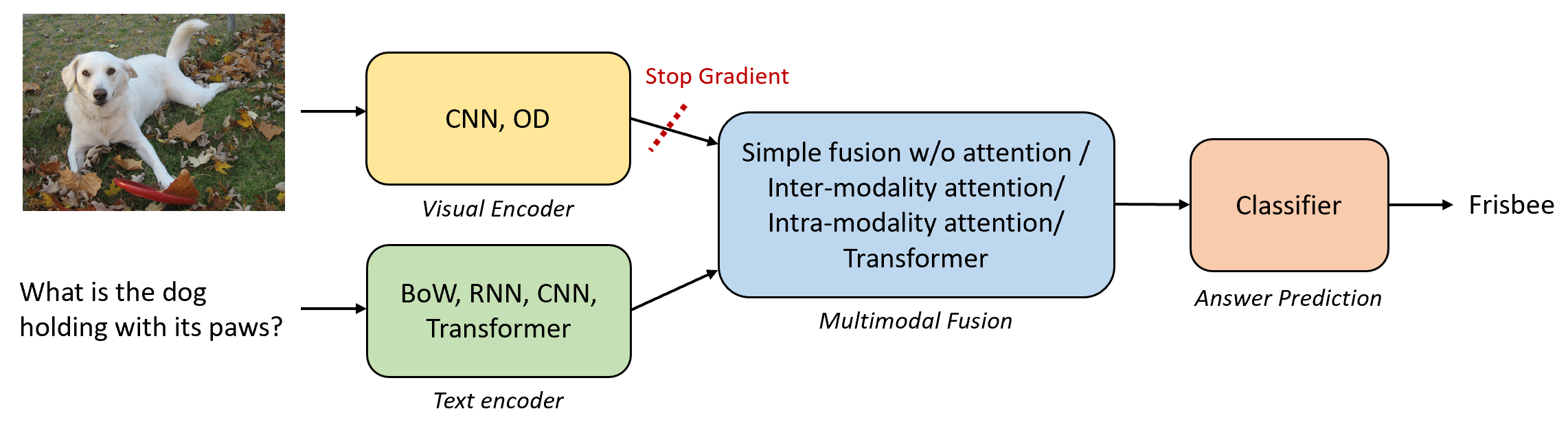}
  \caption{Illustration of a general framework for task-specific VQA models. In most cases, image features are extracted offline, with no gradient update to the visual encoder during model training.}
  \label{fig:chp2_vqa_framework}
\end{figure*}

\subsection{Model Architecture}
\paragraph{Overview.}
Given an image-question pair, a VQA model first 
extracts visual features $\vv=\{ \vv_1, \cdots, \vv_M\}$ via a \textit{visual encoder} and encodes the question input via a \textit{text encoder} into text features $\wv=\{ \wv_1, \cdots, \wv_N\}$. Here, $N$ can be the number of words in the question, or $N=1$ if a global textual representation is computed for the question. $M$ is the number of visual features for an image, which can be the number of image regions (\textit{e.g.}, $M \in [10, 100]$), or the number of grids (\textit{e.g.}, $M = 14\times14$), depending on the specific vision encoder being used. Likewise, $M=1$ when a global image representation is extracted. 
The text and visual features are then fed into a \textit{multimodal fusion module} to produce cross-modal representations, which are then fed into a task-specific output layer (\textit{e.g.,}, a classifier for the VQA task) to predict the answer. An illustration of this framework is shown in Figure~\ref{fig:chp2_vqa_framework}. 


\paragraph{Visual Encoder.} Most early VL methods~\citep{antol2015vqa,anderson2018bottom,yu2019deep} adopt a \textbf{two-stage training pipeline}, where visual features are first extracted from a pre-trained visual encoder. 
There are two types of visual encoders: ($i$) a plain convolutional neural network (CNN), and ($ii$) an object detector (OD). 
\begin{itemize}[leftmargin=3mm]
    \item
    \textbf{CNN.} Inspired by the success of CNN on image classification, early methods adopt CNN models (\emph{e.g.},  VGGNet~\citep{simonyan2014very}, AlexNet~\citep{krizhevsky2012imagenet}, GoogLeNet~\citep{szegedy2015going}, and ResNet~\citep{he2016deep}) pre-trained on ImageNet~\citep{deng2009imagenet} to extract visual features. 
    The very first VQA model \citep{antol2015vqa} experiments with \textbf{global visual features} from the last fully connected layer of VGGNet, which has been inherited by the immediate follow-up works~\citep{gao2015you,ren2015exploring,ma2016learning}. 
    To retain spatial information in the original images, researchers~\citep{yang2016stacked,Zhu_2016_CVPR,andreas2016neural,jabri2016revisiting} use \textbf{grid features} from earlier layers of pre-trained CNN models.
    Grid features represent the input image by a uniform grid of equally sized and shaped neural receptive fields, hence contain more local information than the holistic entire-image representation captured by the global visual features.
    \item \textbf{OD.} In contrast to the uniform grids, object detectors produce a set of salient image regions of varying size and aspect ratio. \textbf{Region features} are the pooled convolutional features extracted per region proposal. \cite{shih2016look} is the first work to exploit region features for VQA, where the regions are located using edges~\citep{zitnick2014edge}.  The most widely used OD model for VL research is a Faster R-CNN~\citep{ren2015faster} pre-trained on the Visual Genome (VG) dataset~\citep{krishna2016visual} from BUTD~\citep{anderson2018bottom}. 
\end{itemize}

\textbf{\emph{Discussion: from grids to regions, and back again.}}\, As discussed above, early explorations in VQA models~\citep{gao2015you,yang2016stacked,jabri2016revisiting} have witnessed the transition from holistic global visual features to grid features with a CNN visual encoder. Popularized by regional bottom-up features~\citep{anderson2018bottom}, OD models have soon dominated the design of visual encoder. Region features have become the de facto standard for VL tasks like VQA and image captioning in many follow-up works~\citep{Teney_2018_CVPR,gao2019dynamic,li2019relation, yu2019deep}.  However, 
\cite{jiang2020defense} argue that compared to the ``format'' of features (\textit{i.e.}, region vs. grids), the semantic content that visual features represent is more critical for their effectiveness. Grid features, extracted from the CNN backbone of an OD model trained on the same data as bottom-up features, can be equally performant, but with better efficiency, and can be more easily end-to-end finetuned than region features.


\paragraph{Text Encoder.} The input question is first tokenized into a sequence of words, and then encoded via a text encoder. 
Depending on how we view textual input, different neural models can be used for text encoding.
\begin{itemize}[leftmargin=*]
    \item \textbf{Bag-of-Words (BoW).} BoW-based methods~\citep[\emph{e.g.},][]{antol2015vqa,yu2015visual,jabri2016revisiting,shih2016look} independently encode each word in the input question, without considering dependencies between neighboring words. The sum or average of word embeddings (learned from scratch or extracted from the pre-trained word2vec~\citep{mikolov2013efficient}) are taken as the representation of the input question.
    \item \textbf{Recurrent Neural Networks (RNN).} RNN-based methods~\citep[\emph{e.g.},][]{ren2015exploring,malinowski2015ask,fukui2016multimodal,anderson2018bottom,Teney_2018_CVPR} intend to capture word dependencies and text structures.  
    The input words are one-hot encoded and passed through a word embedding layer (\textit{e.g.}, learned  from scratch or extracted from word2vec or initialized/concatenated with  GLoVe~\citep{pennington2014glove}). These word embeddings are further processed by an RNN-based text encoder (\textit{e.g.}, LSTM~\citep{hochreiter1997long} or GRU~\citep{cho2014learning}) to obtain the representation of the question.
    \item \textbf{Transformer.} Inspired by the success of Transformers~\citep{vaswani2017attention} (\emph{e.g.}, BERT~\citep{devlin2018bert}) with large-scale pre-training in NLP, researchers have used pre-trained BERT to extract question representations. This method has been integrated into several winning ensemble entries of the VQA Challenge~\citep{vqa2019winner,vqa2019runnerup}. 
\end{itemize}

In addition to what are discussed above, other text encoders, such as the CNN-based text encoder~\citep{ma2016learning} that is trained to recognize patterns in text (such as key phrases), have also been explored. A recent survey is \citet{minaee2021deep}.

\begin{figure*}[t!]
  \centering
    \includegraphics[width=1.0\linewidth]{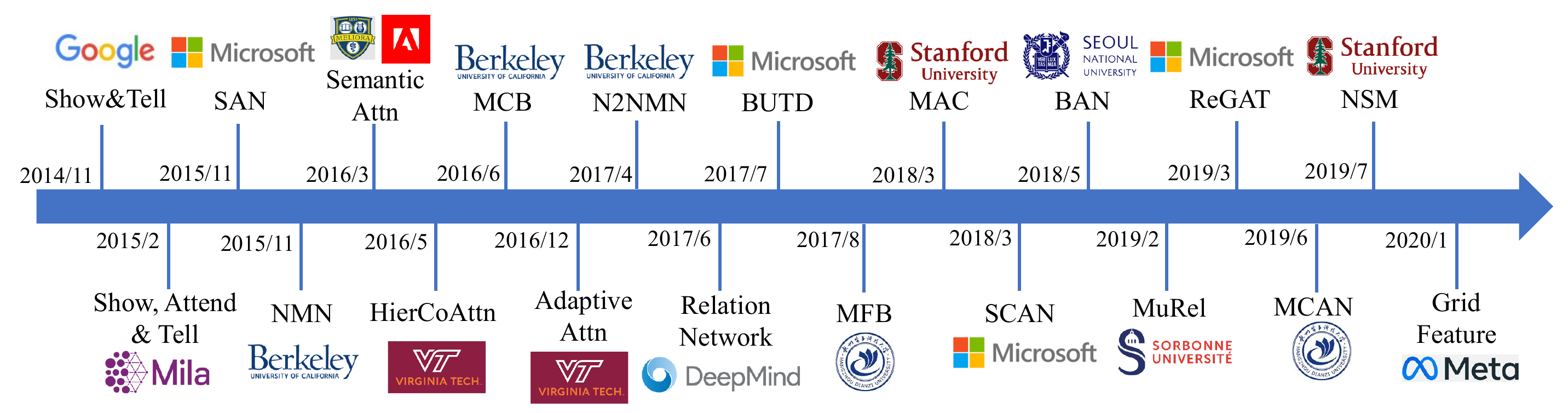}
  \caption{Early VL models developed along time. We mainly focus on the VQA task, and include methods for ($i$) inter-modality attention design for multimodal alignment (\emph{e.g.}, SAN~\citep{yang2016stacked} and BAN~\citep{kim2018bilinear}), ($ii$) intra-modality attention design for relational reasoning (\emph{e.g.}, Relation Network~\citep{santoro2017simple} and ReGAT~\citep{li2019relation}), ($iii$) bilinear pooling for better fusion (\emph{e.g.}, MCB~\citep{fukui2016multimodal} and MFB~\citep{yu2017multi}), ($iv$) the use of both inter- and intra-modality attention (\emph{e.g.}, MCAN~\citep{yu2018mattnet}), and ($v$) neural module network for compositional visual reasoning~\citep{andreas2016neural}. We also briefly include methods for image captioning and image-text retrieval. As there exist a vast number of literature on this topic, only some representative works are shown. }
  \label{fig:chp2_early_vl_models_along_time}
\end{figure*}

\paragraph{Multimodal Fusion Module.} Multimodal fusion aims at modeling interactions between visual features and text features. The design of multimodal fusion modules has always been the major topic in VL research, especially for task-specific VL models. We start the review with simple fusion methods (such as concatenation), followed by some of the most popular attention-based methods, which demonstrate how task-specific VL models evolve over time. For methods that are not based on attention, such as bilinear pooling~\citep{fukui2016multimodal}, we defer the discussion to Section~\ref{chp2-sec:additional-topics}.
\begin{itemize}[leftmargin=*]
    \item \textbf{Simple fusion without attention.} Image and text features are fused via element-wise product or sum, or concatenation~\citep{antol2015vqa,jabri2016revisiting}. More sophisticated designs refine the fused image-text features via LSTM~\citep{malinowski2015ask} or multimodal residual networks~\citep{kim2016multimodal}.
	\item \textbf{Inter-modality attention.} Inter-modality attention methods~\citep[\emph{e.g.},][]{yang2016stacked,lu2016hierarchical,nguyen2018improved} aim to capture \emph{multimodal alignment} between image and text inputs. Compared to simple fusion, attention models construct a more informative VL-joint representation since higher weights are put on the image regions that are more useful to solve the task. There are many works along this direction. We name a few below.
	Stacked Attention Network (SAN)~\citep{yang2016stacked} is the first that verifies the effectiveness of inter-modality attention in VQA, with question as query to attend image features. \citet{lu2016hierarchical} argue that attention on text is equally important as that on image, and develop a co-attention method to jointly perform question-guided image attention and image-guided text attention. BAN~\citep{kim2018bilinear} extends the idea of co-attention into bilinear attention, which considers every pair of question words and image regions. Stacking multiple inter-modality attention layers can also be viewed as a way to perform multi-step reasoning~\citep{yang2016stacked,gan2019multi}, where the attention distribution is refined layer by layer to focus on regions that are more relevant to the question. 
%
	\item \textbf{Intra-modality attention.} Intra-modality attention methods aim to perform \emph{relational reasoning} over image regions or question words. Considering the relations between object regions in image and dependencies between words in question, VQA performance can be improved by building graph structured representations~\citep{santoro2017simple, hu2019language}. For question, a graph built with words as nodes can be obtained through dependency parsing~\citep{teney2017graph}. For image, the graph with object regions as nodes can be built by leveraging external knowledge (\textit{e.g.}, scene graphs) and rule-based priors (\textit{e.g.}, estimating the relative positions of two objects with bounding box coordinates)~\citep{li2019relation}. Alternatively, one can also start with a fully-connected graph, and dynamically prune and refine the connections between nodes during model training~\citep{norcliffe2018learning,cadene2019murel}.
	\item \textbf{Transformer.}  Image (question) understanding can be achieved by not only attending to the other modality (through inter-modality attention), but also the related regions (other words) from the current modality (via intra-modality attention)~\citep{gao2019dynamic}. Based on the scaled dot-product attention in Transformer~\citep{vaswani2017attention}, MCAN~\citep{yu2019deep} uses the self-attention unit for intra-modal interactions (\textit{i.e.}, region-to-region or word-to-word) and the guided attention unit for dense inter-modal interactions (\textit{e.g.}, word-to-region). MCAN also adopts an encoder-decoder Transformer architecture, where the encoder with multiple layers of self-attention learns the self-attended question features, and the decoder uses the resulting question features to learn the attended image features with a stack of self-attention (on image features only) followed by guided-attention (with question feature as query to attend on image features). 
\end{itemize}

\paragraph{Task-specific Output Layer.} The cross-modal representations computed by the multimodal fusion module are fed to a task-specific output layer to generate model predictions. As VQA is usually modeled as a classification problem, the output layer is a classifier that consists of a fully-connected layer or a multi-layer perceptron followed by a softmax layer, to predict the answer.

\paragraph{Trends in VQA Models.} Now, we summarize the trends of model architecture designs in the VQA literature, detailed to each component. Figure~\ref{fig:chp2_early_vl_models_along_time} list some early VL models developed along time.
\begin{itemize}[leftmargin=*]
\item \textbf{Visual features} evolve in 4 stages: ($i$) \textit{global image features} with a holistic view of the entire image; ($ii$) \textit{grid features} that preserve local and spatial information with a uniform grid; ($iii$) \textit{region features} extracted from more salient object-centric image regions; and ($iv$) \textit{back to grid features} that can capture similar semantics when trained with object detection objective.
\item \textbf{Textual features} evolve in 3 stages: ($i$) \textit{bag-of-words} that encodes each word independently; ($ii$) \textit{RNNs} capturing word dependencies and text structures; and ($iii$) more powerful text representations with a \textit{pre-trained Transformer}.
\item \textbf{Multimodal fusion methods} evolve in 4 stages: ($i$) \textit{simple fusion without attention}; ($ii$) \textit{inter-modality attention} methods that model multimodal alignment between image and text inputs; ($iii$) \textit{intra-modality attention} methods that capture uni-modal relations; and ($iv$) \textit{Transformer-based models} that combine inter-modality and intra-modality attention.
\end{itemize}

\subsection{Case Study}
In this section, we present case studies of early VQA models.
\begin{figure*}[t!]
  \centering
    \includegraphics[width=1.0\linewidth]{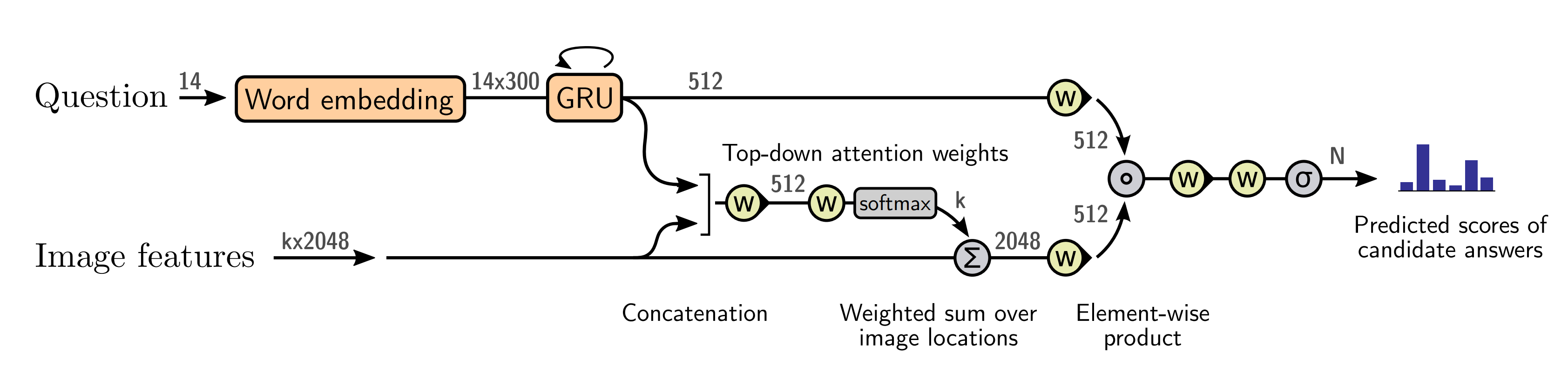}
  \caption{Overview of the BUTD model for VQA. Gray numbers indicate the dimensions of
the vector representations between layers. Yellow elements use learned parameters. Figure credit: \citet{anderson2018bottom}.}
  \label{fig:chp2_vqa_butd}
\end{figure*}

\paragraph{BUTD with Top-down Attention.} Given an image-question pair, regional bottom-up features $\vv=\{ \vv_1, \cdots, \vv_M\}$ ($M$ is the number of regions)\footnote{We use $M$ instead of $k$ from Figure~\ref{fig:chp2_vqa_butd} here to keep consistency throughout the paper.} are first extracted from an OD-based visual encoder, and the question feature $\wv$ is obtained with a word embedding layer followed by a GRU as the \textit{text encoder}.  Note that the question feature is a global textual representation with a single vector of dimension 512 as specified in Figure~\ref{fig:chp2_vqa_butd}. 

BUTD adopts inter-modality attention to attend the query question feature to each image region. Formally, the attention weight $a_i$ on each region $\vv_i$ is computed by an attention model $f_{\text{att}}$ and normalized with softmax operation:
\begin{align}
    e_i &= f_{\text{att}}(\vv_i, \wv) = \wv_a^T f_a([\vv_i, \wv])  \nonumber \\
    a_i &= \frac{\exp{(e_i)}}{\sum_{j=1}^M\exp(e_j)}\,, \label{eq:vqa_attn}
\end{align}
where $\wv_a$ is a learnable parameter vector, $f_a$ is a gated tanh layer. Once the attention weights are computed, the attended visual representation $\hat{\vv}$ is obtained via weighted sum over $\vv$.
\begin{align}
     \hat{\vv} = \sum_{i=1}^{M} a_i \vv_i\,.
\end{align}
Finally,  the cross-modal representation $\hv$ is obtained by
\begin{align}
    \hv = f_w(\wv) \circ f_v(\hat{\vv}) \,,
\end{align}
where $f_w$ and $f_v$ are gated tanh layers.
For answer prediction, a two-layer MLP is adopted as the classifier, with the cross-modal representation $\hv$ as the input. Binary cross-entropy is used as
the loss function to supervise the model training. 

\begin{figure*}[t!]
  \centering
    \includegraphics[width=1.0\linewidth]{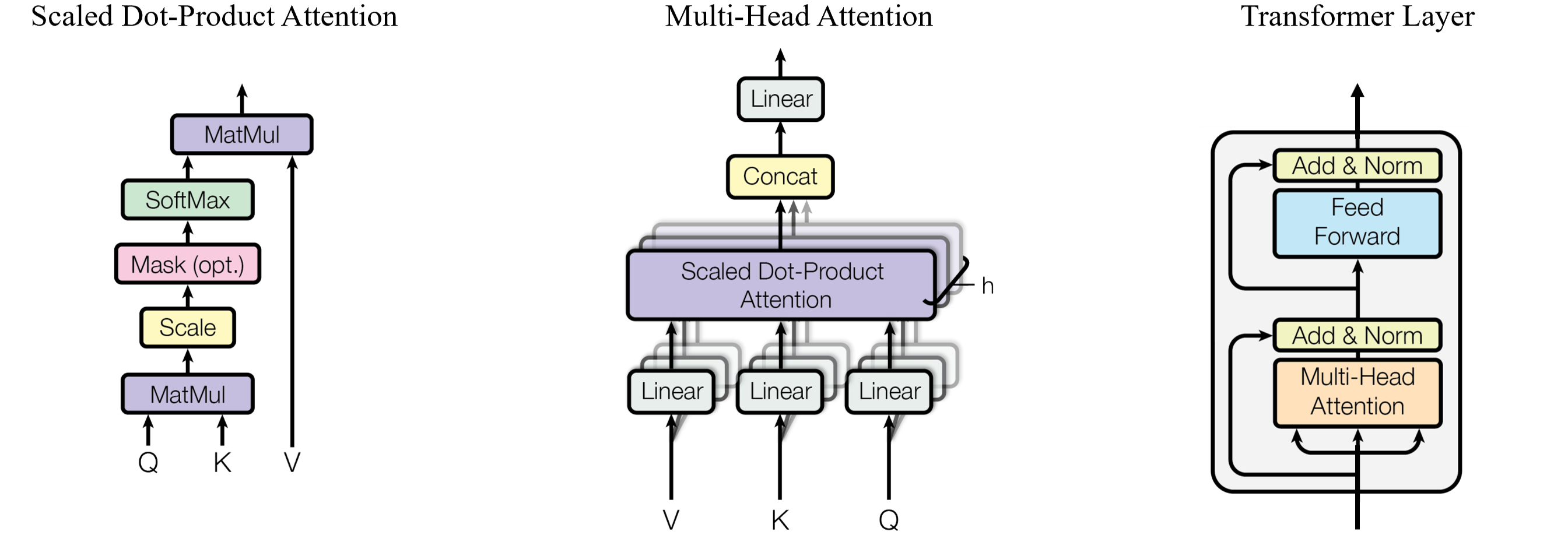}
  \caption{Overview of scaled dot-product attention (left), multi-head attention (middle) and Transformer layer (right). Figure credit: \citet{vaswani2017attention}.}
  \label{fig:chp2_transformer}
\end{figure*}

\paragraph{Transformer with Multi-head Scaled Dot-Product Attention.} The top-down attention introduced in BUTD is simple, in two aspects. On one hand, it is inter-modality attention only, while more advanced models~\citep{gao2019dynamic,yu2019deep} combine both inter-modality and intra-modality attention to learn better cross-modal representation. On the other hand, the attention mechanism is simple in that only question-to-region attention is used. Furthermore, the attention weights are learned with a single learnable parameter vector $\wv_a^T$ (which is usually referred as single-head attention in literature). 
Of late, the modern attention-based models~\citep{li2019relation,gao2019dynamic,yu2019deep} closely follow Transformer~\citep{vaswani2017attention} to adopt scaled dot-product attention, usually with multi-head. As Transformer architecture becomes the basis for VLP (and also the basic concept in the following chapters), we briefly review the multi-head scaled dot-product attention and the vanilla Transformer layer (shown in Figure~\ref{fig:chp2_transformer}).

\begin{itemize}[leftmargin=*]
    \item \textbf{Multi-head scaled dot-product attention.} With the inputs as three set of feature vectors, query $Q$, key $K$ and value $V$, scaled dot-product attention is defined as
\begin{align}
\text{Attention}(Q, K, V ) = \text{softmax}(\frac{QK^T}{\sqrt{d_k}})V\,,
\end{align}
where $d_k$ is the feature dimension of $Q$ and $K$. To extend it to multi-head attention (illustrated in the center of Figure~\ref{fig:chp2_transformer}), the queries, keys and values can be linearly projected $h$ times with different, learned linear projections to $d_k$, $d_k$ and $d_v$ dimensions, respectively. On each of these projected versions of queries, keys and values, the attention is performed in parallel, yielding $d_v$-dimensional output values. These are concatenated and once again projected to produce the final values. Compared to single-head attention, multi-head attention allows the model to jointly attend to information from different representation subspaces at different positions.

The scaled dot-product attention mechanism can be adopted for both inter-modality and intra-modality attention, depending on the inputs.  For example, word-to-region attention (\textit{inter-modality}) can be realized by using question features $\wv$ as query and visual features $\vv$ as key and value. When we set the query, key, value as the features from the same modality, it is considered as \textit{intra-modality} attention.
\item \textbf{Transformer layer.} As shown in the rightmost of Figure~\ref{fig:chp2_transformer}, a Transformer layer has two
sub-layers, ($i$) a multi-head attention layer, and ($ii$) a simple, position-wise fully connected feed-forward layer. A residual connection is added around each of the two sub-layers, followed by layer normalization. This Transformer layer is the building block of modern VLP models.
\end{itemize}


\subsection{Similar Trends in Captioning and Retrieval Models} 
In this subsection, we briefly review model architectures for image captioning and image-text retrieval, where similar trends to VQA models are observed.

\paragraph{Image Captioning.}
Early captioning models before deep learning use a modular architecture \citep[\emph{e.g.},][]{farhadi2010every,kulkarni2013babytalk,fang2015captions}, consisting of modules developed separately for detecting objects or concepts in images and generating captions using rules or machine learned models, respectively. Inspired by the Seq2Seq learning framework for machine translation \citep{sutskever2014sequence,bahdanau2014neural}, image captioning models nowadays adopt the encoder-decoder architecture. Specifically, a \textbf{visual encoder} is used to extract visual features and a \textbf{text decoder} generates a caption based on the visual features. To make the text decoder better exploit rich information in visual features, different \textbf{multimodal fusion} methods have been explored with or without attention.

\textbf{Case study:}\, We first use  the seminal ``Show, Attention and Tell'' model~\citep{xu2015show} as an example to review how a captioning model works. 
Grid features $\vv=\{ \vv_1, \cdots, \vv_M\}$ ($M=14\times14$ is the number of grids) are first extracted from a CNN-based visual encoder. A LSTM is used as the text decoder to produce a caption by generating one word at every time step conditioned on ($i$) the context vector $\zv_{t}$ at current time $t$, indicating relevant part of the image input; ($ii$) the current hidden state ($\hv_{t}$) of the LSTM; and ($iii$) previously generated words $\hat{y}_{1: t-1}$.  Here, we describe how the context vector is produced via attention. Similar to Equation~\ref{eq:vqa_attn}, an attention model $f_{\text{att}}$ followed by softmax normalization is adopted to compute the attention weight $a_{ti}$ for the $i$-th visual feature $\vv_i$ at time $t$. However, in the case of image captioning, instead of conditioning on the question feature $\wv$, the attention weights are conditioned on the previous hidden state $\hv_{t-1}$ of LSTM. Specifically,
\begin{align}
    e_{ti} &= f_{\text{att}}(\vv_i, \hv_{t-1}) \nonumber\\
    a_{ti} &= \frac{\exp{(e_{ti})}}{\sum_{j=1}^M\exp(e_{tj})}\,.
\end{align}
\citet{xu2015show} have explored two alternative mechanisms for $f_{\text{att}}$, which we refer the reader to the original paper for more details. 
After obtaining the attention weights, the context vector $\zv_{t}$ is computed via weighted sum of all visual features. That is, 
\begin{align}
    \zv_{t} = \sum_{i=1}^{M} a_{ti} \vv_i\,.
\end{align}
The output word probability at time $t$ can be calculated via an output layer $f_o$:
\begin{align}
   p(\hat{y}_t|\vv, \hat{y}_{1: t-1}) = f_o(\zv_{t}, \hv_{t}, \hat{y}_{t-1})\,.
\end{align}
During training, given the ground-truth caption sequence $y_{1: T}$, the following cross-entropy loss is minimized:
\begin{align}
   L_{XE}(\theta) = - \sum_{t=1}^{T}\log(p_{\theta}(y_t | \vv, y_{1: t-1}))\,,
\end{align}
where $\theta$ denotes all trainable parameters.

Next, we review how each component in task-specific captioning models evolves in recent years. 

\begin{itemize} [leftmargin=*]
    \item \textbf{Visual encoder.} Early studies~\citep{vinyals2015show,karpathy2015deep} adopt a CNN model as the image encoder to extract global visual features, and then quickly move to grid features~\citep{xu2015show,Yao_2017_ICCV}. Later, region features extracted from OD-based visual encoder become the default choice, since BUTD~\citep{anderson2018bottom} has shown bottom-up features much more effective for image captioning. And once again, \cite{jiang2020defense} also defend the use of grid features in terms of VQA and image captioning. More recently, fully Transformer-based captioning model~\citep{wang2022end,fang2022injecting} is built on top of grid features extracted from  Transformer-based visual encoder (\textit{e.g.}, Swin Transformer~\citep{liu2021swin}).
    
    \item \textbf{Text decoder.} RNN-based methods are widely adopted~\citep{mao2014deep,donahue2015long,pan2020x} before the emergence of Transformer.  CNN-based decoder has also been explored in \citet{Aneja_2018_CVPR}, showing on par performance but easier to train (\textit{e.g.}, better training efficiency, less likely to suffer from vanishing gradients), when compared with the prominent LSTM design. Of late, Transformer-based decoder~\citep{herdade2019image,li2019entangled,cornia2020meshed,luo2021dual} has become the most popular design choice.  
    
    \item \textbf{Multimodal fusion.} Early models without attention, directly input the global visual features to text decoder, either as the initial hidden state~\citep{xu2015show,vinyals2015show,karpathy2015deep} or to each step of the LSTM decoder~\citep{mao2014deep,donahue2015long}. Similar to the use of attention models in VQA, the encoder-decoder image captioning models are enhanced by incorporating inter-modality attention mechanism in the decoder~\citep[\emph{e.g.},][]{xu2015show,lu2017knowing,huang2019attention}, so that the caption can be generated based on the image regions/grids and concepts of interest. Intra-modality attention~\citep{you2016image,yao2019hierarchy,Yang_2019_CVPR} has also been explored for captioning, mostly focus on modeling object relational reasoning. For example, \citet{yao2018exploring} employ a graph convolutional network to integrate both semantic and spatial object relationships into visual encoder.  \citet{herdade2019image} build an object relation Transformer to explicitly incorporate information about the spatial relationship between input objects through geometric attention. 
\end{itemize}

 \begin{figure*}[t!]
  \centering
    \includegraphics[width=0.7\linewidth]{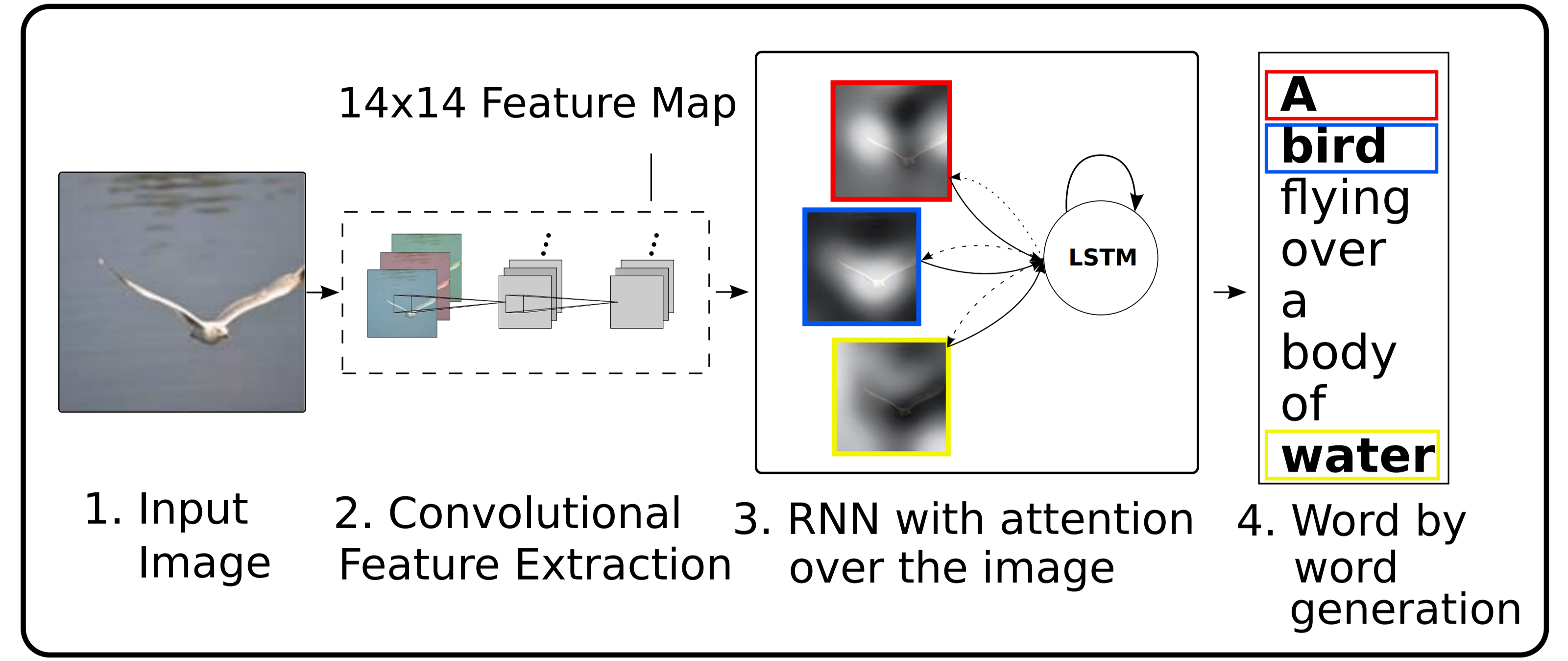}
  \caption{Overview of the seminal ``Show, Attend, and Tell'' model for image captioning. Figure credit: \citet{xu2015show}.}
  \label{fig:chp2_show_attend_tell}
\end{figure*}

\paragraph{Image-text Retrieval.}
 Image-text retrieval can be formulated as either a classification problem (\textit{i.e.}, to determine whether an image-text pair is matched or not), or a ranking problem (\textit{i.e.}, to rank all candidate instances based on their similarity to the query). The typical architecture of image-text retrieval models is similar to that of VQA models, which consists of a visual encoder, a text encoder, a multimodal fusion module and with or without a task-specific output layer on top of multimodal fusion module to project the cross-modal representations into similarity measure. Next, we discuss the evolution of the major components in details.

\begin{itemize} [leftmargin=*]

\item \textbf{Visual encoder.} We observe a similar transition with a plain CNN model, from global image features~\citep{kiros2014unifying,socher2014grounded,wang2016learning,klein2015associating} to grid features~\citep{huang2017instance,nam2017dual}.  Even before the first adoption of bottom-up features~\citep{anderson2018bottom} in \citet{Lee_2018_ECCV}, region features have been used to model finer-grained alignment between image and text. For example, \citet{karpathy2015deep} extract region features with R-CNN~\citep{girshick2014rich}; \citet{plummer2015flickr30k} leverage EdgeBox~\citep{zitnick2014edge} to generate region proposals; and \citet{niu2017hierarchical} further combine region features with global image features. 


\item \textbf{Text encoder.} Researchers have explored ($i$) BoW-based methods~\citep{klein2015associating,wang2016learning} by independently computing word embeddings; ($ii$) RNN-based architecture, such as LSTM~\citep{kiros2014unifying,socher2014grounded} and GRU~\citep{faghri2017vse++}; and ($iii$) CNN-based architecture~\citep{zheng2020dual}.

\item \textbf{Multimodal fusion.} There have been studies that focus on projecting global visual features and global text features into a common ``visual-semantic'' space~\citep{kiros2014unifying,socher2014grounded,wang2016learning}, where multimodal fusion is realized by simple dot product.  
Another paradigm of approaches examine more finer-grained alignment between regions in the image and words in the texts. The first attempt~\citep{karpathy2015deep} adopts inner product to fuse each word-region pair, and sum the similarity between aligned word and region pairs as the image-text similarity. The adoption of attention greatly enhances the performance of local-level matching methods. Lots of works~\citep{huang2017instance,nam2017dual,liu2019focus,zhang2020context} are devoted to designing better inter-modality attention. Perhaps the most prominent example is SCAN~\citep{Lee_2018_ECCV}, with cross-attention to not only use text as the query to attend to image regions, but also use the image query to attend to words.
Intra-modality attention mechanisms are also incorporated to enhance image/text representations. Image representations can be refined by position-focused attention module~\citep{wang2019position} or structured reasoning over object relationships with graph neural networks~\citep{Li_2019_ICCV}. Extending to text representations, \citet{chen2020expressing} design a word attention module and an object attention module to compute the self-attention weights of words and objects. \citet{liu2020graph} and \citet{diao2021similarity} apply graph neural network to both image and text inputs.
\end{itemize}

\section{Additional Topics}
\label{chp2-sec:additional-topics}
In this section, we review additional research topics for the development of early VL models, including bilinear pooling, compositional visual reasoning, and visual grounding.

\subsection{Bilinear Pooling}

Advanced attention design is a main theme for early VL research. Besides this, instead of simple concatenation and element-wise product for fusion, another line of work~\citep{fukui2016multimodal,kim2016hadamard,yu2017multi} aims to develop better methods for \emph{bilinear pooling}, \emph{i.e.}, how to fuse two vectors into a better representation.  

Specifically, \cite{fukui2016multimodal} proposed Multimodal Compact Bilinear (MCB) pooling, which is also the 2016 VQA challenge winner solution. However, the feature after Fourier transform is very high dimensional, which also makes MCB computation expensive. \cite{kim2016hadamard} proposed a simple Hadamard product for low-rank bilinear pooling, and \cite{yu2017multi} proposed Multimodal Factorized Bilinear (MFB) pooling. Other more advanced pooling methods include MUTAN~\citep{ben2017mutan} and BLOCK~\citep{ben2019block}, for example. In \cite{perez2018film}, the authors developed FiLM, a feature-wise linear modulation operator similar to conditional batch normalization, \emph{i.e.}, a general conditioning layer to inject language information (\emph{e.g.}, a question) into the image backbone (\emph{e.g.}, a convolutional neural network). 
 
This line of work is orthogonal to attention design, and typically they are used together to enhance each other. However, in the era of VLP, all these bilinear pooling and attention designs are largely replaced by, or converged to, the Transformer design. 

\subsection{Compositional Visual Reasoning}
Besides designing better attention methods to achieve stronger performance on standard VL tasks, such as VQA and image captioning, there are studies on \emph{compositional visual reasoning} that requires a model to learn a strong compositional generalization capability, \emph{i.e.}, understanding and answering compositional questions without seeing similar semantic compositions before. 
Below, we briefly review Neural Module Network (NMN)~\citep{andreas2016learning,andreas2016neural} that aims to perform such complex reasoning tasks. For evaluation, methods are typically tested on a diagnostic visual reasoning dataset called CLEVR~\citep{johnson2017clevr}, and a real-world visual reasoning dataset called GQA~\citep{hudson2019gqa}.

In order to answer a question about an image, NMN uses a set of pre-defined functions and explicitly encodes each function into a shallow neural network called a module. These modules are composed dynamically to build an instance-specific network for each input question. By first parsing the question into a program, and then executing the program via  dynamically composing an instance-specific network,
NMN excels in interpretability and compositionality by design, as each module is designed to accomplish a specific skill, and multiple
modules can be combined to perform a new task during inference.

Since NMN involves two steps, program synthesis and program execution, the original neural module network~\citep{andreas2016neural} cannot be trained end-to-end. IEP~\citep{hu2017learning} and N2NMN~\citep{johnson2017inferring} have successfully made NMN end-to-end trainable via reinforcement learning. Stack-NMN~\citep{hu2018explainable} makes a soft layout selection so that the whole model is fully differentiable.
Neural-Symbolic VQA~\citep{yi2018neural,mao2019neuro,vedantam2019probabilistic} performs symbolic reasoning by encoding images into scene graphs. \cite{chen2021meta} propose Meta Module Network, where only a general-purpose meta module is used for program execution recurrently. This meta module is able to take in function recipes and morph them into diverse instance modules dynamically. The instance modules are then woven into an execution graph for complex visual reasoning, inheriting the explainability and compositionality of NMN.

In addition to neural module networks, compositional attention networks~\citep{hudson2018compositional} and MuRel~\citep{cadene2019murel} have been proposed to realize multi-hop reasoning on complex questions. However, due to the pure attention design, these models are less interpretable. 
Also, Neural State Machine
(NSM)~\citep{hudson2019learning} is proposed. It first predicts a probabilistic scene graph, and then performs multi-hop reasoning over the graph for answer prediction, where the scene graph serves as a strong prior to the model.

In the recent VLP literature~\citep{tan-bansal-2019-lxmert,chen2020uniter,li2021align,dou2021empirical}, most methods use large-scale, Transformer-based monolithic networks. The research on compositional visual reasoning and neural module networks becomes less popular. But we believe that compositional generalization is an important topic worthy of further investigation even 
in the new era of large-scale pre-training. 

\subsection{Visual Grounding}
Now, we briefly discuss the visual grounding (VG) task. Different from the VL tasks introduced in Section~\ref{chp2-sec:task-datasets}, VG requires a model to ground a text query in the relevant object in the image, and predict bounding box coordinates. Likewise, we briefly review the popular benchmarks and representative task-specific models for VG. 

\begin{figure*}[t!]
    \centering
    \begin{subfigure}[t]{0.5\textwidth}
        \centering
        \includegraphics[height=3cm]{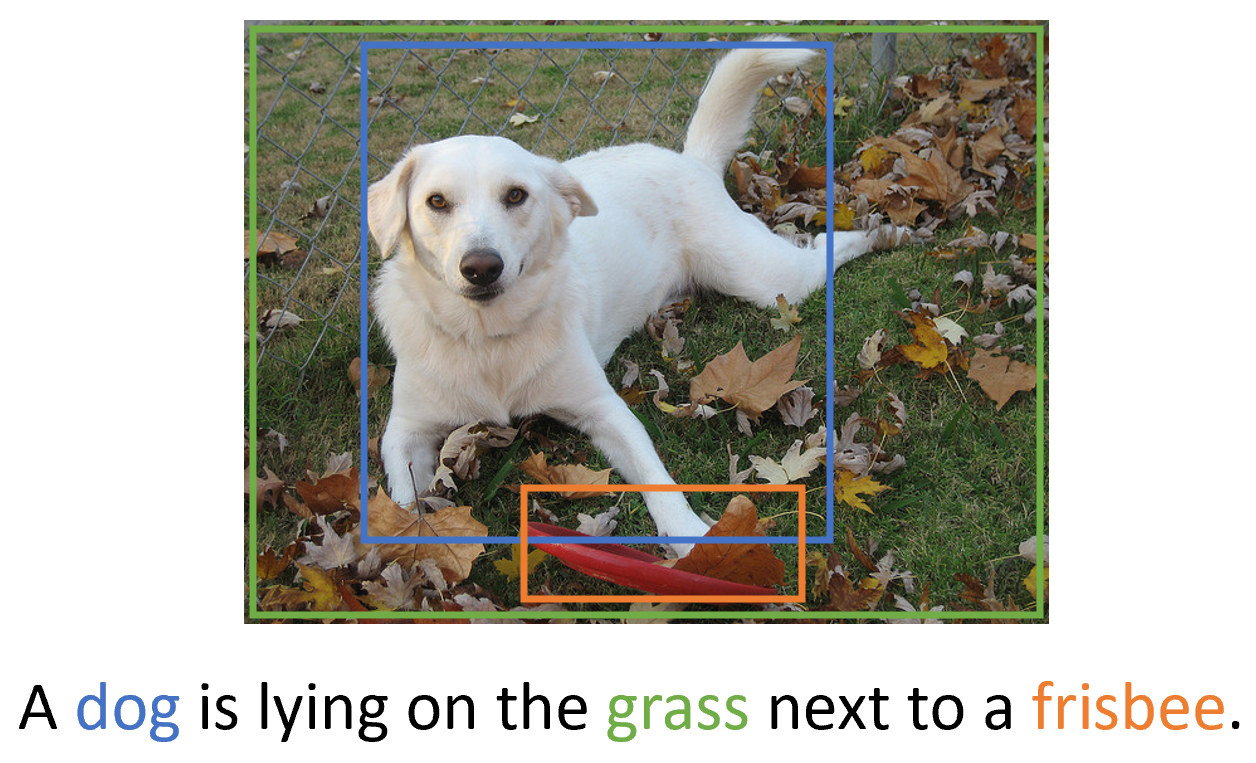}
        \caption{Phrase grounding.}
        \label{subfig:phrase_grounding}
    \end{subfigure}
    \begin{subfigure}[t]{0.4\textwidth}
        \centering
        \includegraphics[height=3cm]{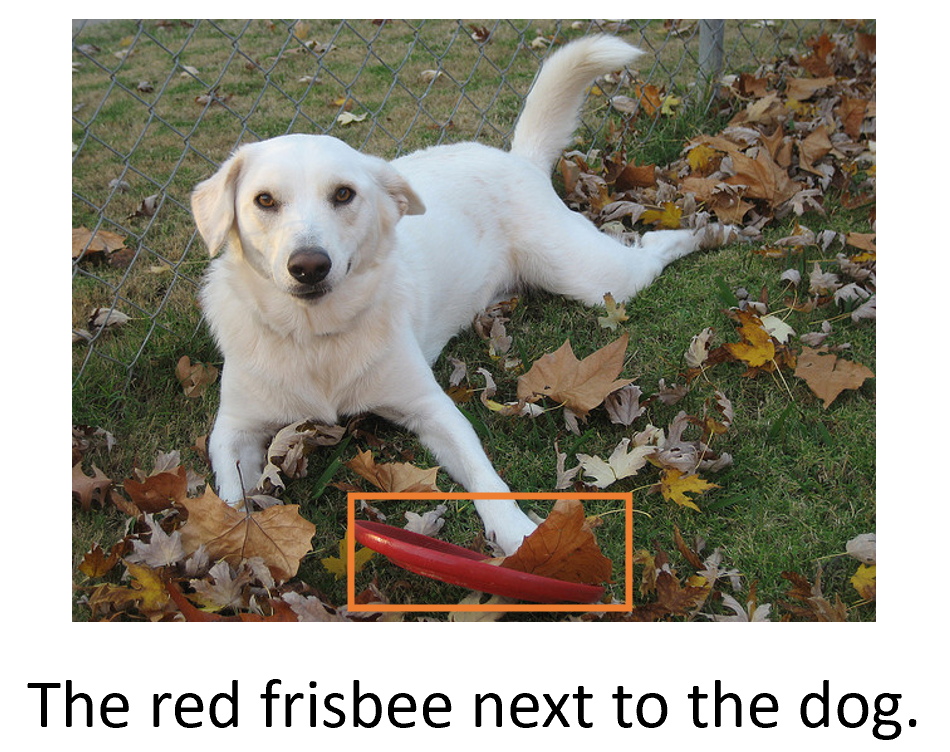}
        \caption{Referring expression comprehension.}
        \label{subfig:refexp}
    \end{subfigure}
    \caption{Visualization of two visual grounding tasks.}
    \label{fig:vg}
\end{figure*}

\paragraph{Task and Benchmark.} Two types of VG tasks are proposed in literature, phrase grounding and referring expression comprehension. 
\begin{itemize}[leftmargin=3mm]
\item \textbf{Phrase grounding} is introduced with the Flicker30K Entities dataset~\citep{plummer2015flickr30k}, in which multiple entities (phrases) in a sentence for an image are mapped to the boxes on the image to indicate the correspondences between them (Figure~\ref{subfig:phrase_grounding}). The task is to predict a bounding box for each entity. Recall@K is used to evaluate model performance and a predicted box for a given entity is considered correct if the intersection over union (IoU) between predicted and ground-truth bounding box is greater than or equal to 0.5. 
\item \textbf{Referring expression comprehension} is to localize the object in the input image that is referred to by an expression in text and return a bounding box around the object (Figure~\ref{subfig:refexp}). Three well-established datasets for this task are RefCOCO, RefCOCO+~\citep{yu2016modeling} and RefCOCOg~\citep{mao2016generation}. Similarly, a prediction is counted as a true positive, if the IOU is larger than or equal to 0.5. Accuracy is used as evaluation metric. 
\end{itemize}


\paragraph{Task-specific VG Models.} Early VL models for VG task can be generally grouped into two categories. One is two-stage methods~\citep{nagaraja2016modeling,kim2018bilinear}, which require to first generate object regions and then perform region-text matching via multimodal fusion to ground the query/referring expression. The region proposals are generated using either unsupervised methods~\citep{plummer2018conditional,wang2018learning} or a pre-trained object detector~\citep{yu2018mattnet,zhang2018grounding}. The other is one-stage models with end-to-end training~\citep{chen2018real,liao2020real}, where the bounding box proposal generation is guided by the text/phrase query. For example, \citet{yang2019fast} fuse a text query’s embedding (a single vector representation) into the YOLOv3 object detector~\citep{redmon2018yolov3}. The method is later improved  by using a recursive sub-query construction framework to reason between image and query for multiple rounds and reduces the referring ambiguity step by step~\citep{yang2020improving}. Lately, \citet{deng2021transvg} empirically show that complex fusion modules can be replaced by simple stack of Transformer encoder layers to achieve higher performance.

\chapter{VLP for Image-Text Tasks}
\label{chp:vlp4imgtxt}
Visual question answering (VQA)~\citep{antol2015vqa}, image captioning~\citep{vinyals2015show} and image-text retrieval~\citep{lin2014microsoft,plummer2015flickr30k} are arguably the three most widely studied image-text tasks in the literature. 
They require an AI system to comprehend both the input image and text contents. Inspired by the great success of language model pre-training~\citep{devlin2018bert,liu2019roberta,raffel2020exploring,brown2020language,he2020deberta}, coupled with the unification of architectures used in the NLP and computer vision communities~\citep{dosovitskiy2020image,carion2020end}, there has been a surging research interest in developing VLP methods for image-text tasks~\citep{tan-bansal-2019-lxmert,chen2020uniter,li2020oscar,zhang2021vinvl,kim2021vilt}. Specifically, large amounts of image-caption pairs are fed into a model that consumes both images and text to pre-train representations that encode rich multimodal knowledge and is helpful for downstream tasks. 
In this chapter, we present a systematic review of this new emerging training paradigm. Specifically, in Section~\ref{sec:chp3_glossary}, we provide an overview of representative VLP models, and divide them into several categories.
In Section~\ref{sec:chp3_model_architectures}, we describe the Transformer-based model architectures for VLP, and dissect the model designs along multiple dimensions including image encoder, text encoder, multimodal fusion, \emph{etc.}.
In Section~\ref{sec:chp3_pretrain_objectives} and \ref{sec:chp3_pretrain_data}, we introduce the commonly used pre-training objectives and pre-training datasets, respectively. 
In Section~\ref{sec:chp3_adv_topics}, we present a list of advanced research topics, including foundation models, multimodal few-shot learning, unified VL modeling, knowledge for VLP, robustness evaluation, model compression and so on. Lastly, in Section~\ref{sec:vlp4imggen}, we provide a brief discussion on text-to-image generation, another important image-text task that has received rapidly growing attention in the community. 

\section{Overview of VLP Models}
\label{sec:chp3_glossary}

Among the ever-growing literature, we broadly divide VLP methods into two categories: ($i$) dual encoder, and ($ii$) fusion encoder. Specifically, 
\begin{itemize}[leftmargin=*]
    \item For \textbf{dual encoder}, images and text are encoded separately, and modality interaction is only handled by a simple cosine similarity of the image and text feature vectors. This model architecture is effective for image retrieval tasks, and when scaled up, can be used to learn a strong image encoder from scratch via large-scale contrastive pre-training, as demonstrated by CLIP~\citep{radford2021learning} and ALIGN~\citep{jia2021scaling}. However, due to the lack of deep multimodal fusion, CLIP performs poorly on VQA and visual reasoning tasks. 
    \item For \textbf{fusion encoder}, besides the use of an image encoder and a text encoder, additional Transformer layers~\citep{vaswani2017attention} are typically employed to model the deep interaction between image and text representations. Prominent examples include UNITER~\citep{chen2020uniter}, VinVL~\citep{zhang2021vinvl}, SimVLM~\citep{wang2021simvlm}, and METER~\citep{dou2021empirical}. This fusion-encoder architecture achieves superior performance on the VQA and image captioning tasks, but can be very ineffective when applied to image retrieval, as it requires to encode all the possible image-text pairs (matched or not) to compute similarity scores for ranking. Recent work, such as ALBEF~\citep{li2021align}, UFO~\citep{wang2021ufo}, and VLMo~\citep{wang2021vlmo}, has also shown that it is possible to encapsulate both the \emph{dual encoder} and \emph{fusion encoder} design into one framework, so that the model is suitable for fast image retrieval, but at the same time can also be used for the VQA and image captioning tasks.
\end{itemize}

\begin{table*}[!t]
\resizebox{1.0\textwidth}{!}
{
  \begin{tabular}{l|cccc|c}
    \multirow{2}{*}{\bf Model} & \bf Vision & \bf Text  & \bf Multimodal & \multirow{2}{*}{\bf Decoder} & \bf Pre-training \\
    & \bf Encoder & \bf Encoder  & \bf Fusion &  & \bf Objectives \\
    \hline
    ViLBERT~\citep{lu2019vilbert} & 
\multirow{2}{*}{OD+Xformer}  & \multirow{2}{*}{Xformer}    & \multirow{2}{*}{Co-attn.} &  \multirow{10}{*}{\ding{55}$^\dagger$ } & MLM+ITM+MIM\\
LXMERT~\citep{tan-bansal-2019-lxmert} & &   & &   & MLM+ITM+MIM+VQA \\
    \cdashline{2-4}
    VisualBERT~\citep{li2019visualbert} & \multirow{8}{*}{OD}   & \multirow{8}{*}{Emb.}  & \multirow{8}{*}{Merged-attn.} &    & MLM+ITM\\
 VL-BERT~\citep{su2019vl} &  &  & &    & MLM+MIM\\
  UNITER~\citep{chen2020uniter} & &   & & & MLM+ITM+MIM+WRA \\
  OSCAR~\citep{li2020oscar} & &   & &  & MLM+ITM \\
  VILLA~\citep{gan2020large} & &   & & & MLM+ITM+MIM+WRA \\
  VinVL~\citep{zhang2021vinvl} & &  & &  & MLM+ITM\\
  UNIMO~\citep{li2020unimo} & &  & &  & MLM+ITM+MIM+ITC\\
  \cdashline{5-5}
   VL-T5~\citep{cho2021unifying}  &   &    & & \ding{51} & MLM+ITM+VQA+GC\\
  \hline
  PixelBERT~\citep{huang2020pixel} & \multirow{8}{*}{CNN} & \multirow{4}{*}{Emb.}  & \multirow{7}{*}{Merged-attn.} & \multirow{3}{*}{\ding{55}$^\dagger$ } & MLM+ITM\\
  SOHO~\citep{huang2021seeing} & &  & &  & MLM+ITM+MIM\\
  CLIP-ViL~\citep{shen2021much}  & & & &  & MLM+ITM+VQA\\
  \cdashline{5-5}
   SimVLM~\citep{wang2021simvlm} &   &   &    & \multirow{4}{*}{\ding{51}} & PrefixLM\\
     \cdashline{3-3}
   MDETR~\citep{kamath2021mdetr} &   & \multirow{2}{*}{Xformer} &  & & OD+TP+CA\\
   UniTAB~\citep{yang2021crossing} &  & &  & & Seq2Seq\\
   \cdashline{3-3}
   OFA~\citep{wang2022unifying} &  & \multirow{2}{*}{Emb.} &  & & Seq2Seq\\
   \cdashline{4-5}
   Flamingo~\citep{alayrac2022flamingo} &  &  & Cross-attn. & \ding{55}$^\dagger$  & LM \\
  \hline
  ViLT~\citep{kim2021vilt}& \multirow{1}{*}{Patch Emb.} & \multirow{3}{*}{Emb.}  & \multirow{5}{*}{Merged-attn.} & \multirow{10}{*}{\ding{55}$^\dagger$} & MLM+ITM\\
  \cdashline{2-2}
  Visual Parsing~\citep{xue2021probing} & \multirow{9}{*}{Xformer} &  & &   & MLM+ITM+MIM\\
  GIT~\citep{wang2022git} &  &   &  &  & LM\\
  \cdashline{3-3}
  VLMo~\citep{wang2021vlmo}& & \multirow{7}{*}{Xformer} & & & MLM+ITM+ITC\\
  BEiT-3~\citep{wang2022image}& & & & & MLM+MIM+MVLM\\
  \cdashline{4-4}
  ALBEF~\citep{li2021align} &   &  & \multirow{3}{*}{Cross-attn.} &   & MLM+ITM+ITC\\
  BLIP~\citep{li2022blip} &  &    &  &  & LM+ITM+ITC\\
  CoCa~\citep{yu2022coca} &  &    &  &  & LM+ITC\\
  \cdashline{4-4}
  METER~\citep{dou2021empirical} &  &    & \multirow{2}{*}{Co-attn.} &  & MLM+ITM\\
  FIBER~\citep{dou2022coarse} &  &    &  &  & LM+ITM+ITC\\
  \hline
  CLIP~\citep{radford2021learning}  & CNN/Xformer & \multirow{2}{*}{Xformer} & \multirow{2}{*}{None} &\multirow{2}{*}{\ding{55}}  & \multirow{2}{*}{ITC}\\
  ALIGN~\citep{jia2021scaling}  & CNN & &  &   & \\
  \end{tabular}
  }
  \caption{\textbf{Glossary of representative VLP models}. OD: object detector. Xformer: transformer. Emb.: embedding. MLM/MIM: masked language/image modeling. ITM: image-text matching. ITC: image-text contrastive learning. WRA: word-region alginment.  TP: token prediction. CA: contrastive alignment. GC: grounding+captioning. ($\dagger$) In many cases (\emph{e.g.}, Flamingo~\citep{alayrac2022flamingo}, CoCa~\citep{dou2021empirical}, and GIT~\citep{dou2021empirical}), the multimodal fusion module itself is also directly called (or serves as) the text decoder.}
  \label{tab:chp3_vlp_glossary}
  \vspace{-3mm}
\end{table*}

In this chapter, we mainly focus on the review of VLP methods based on the \emph{fusion-encoder} architecture, while postponing the detailed discussion of \emph{dual-encoder} models to Chapter~\ref{chp:vlp4vision}. Among fusion-encoder methods, we further divide them into two categories based on whether the model can be pre-trained end-to-end. This categorization also roughly reflects how the VLP methods evolve along time. Specifically, most early VLP methods~\citep{tan-bansal-2019-lxmert,su2019vl,chen2020uniter,li2020oscar,zhang2021vinvl} adopt a \textbf{two-stage pre-training pipeline}, where image region features are first extracted from a pre-trained object detector. More recently, \textbf{end-to-end pre-training} methods~\citep{huang2020pixel,kim2021vilt,li2021align} become popular, where image features are extracted from either convolutional neural networks (CNNs)~\citep{he2016deep}, vision Transformers (ViTs)~\citep{dosovitskiy2020image}, or only using image patch embeddings, and the model gradients can be back-propagated into the vision backbone for end-to-end training. End-to-end VLP methods have achieved new state of the art on all the major VL tasks.

\begin{itemize}[leftmargin=*]
    \item \textbf{OD-based VLP Models.} Early methods use pre-trained object detectors (ODs) to extract visual features. Among them, ViLBERT~\citep{lu2019vilbert} and LXMERT~\citep{tan-bansal-2019-lxmert} use co-attention for multimodal fusion, where two Transformers are applied respectively to region and text features, and another Transformer fuses the representations of the two modalities in a later stage. On the other hand, VisualBERT~\citep{li2019visualbert}, Unicoder-VL~\citep{li2020unicoder}, VL-BERT~\citep{su2019vl}, and UNITER~\citep{chen2020uniter} use a merged attention fusion module that feeds both region and text features into a single Transformer. The comparison between merged attention and co-attention is detailed in Section~\ref{sec:chp3_model_architectures}. OSCAR~\citep{li2020oscar} feeds additional image tags into the Transformer model, while VinVL~\citep{zhang2021vinvl} uses a stronger pre-trained OD for feature extraction, and demonstrates state-of-the-art performance across VL tasks. On the one hand, region features are object-level and semantic-rich; on the other hand, extracting region features can be time-consuming, and the pre-trained object detectors are usually frozen during pre-training, which may limit the capacity of VLP models.
    \item \textbf{End-to-End VLP Models.} Researchers have tried different ways to pre-train VL models in an end-to-end fashion. Specifically, we further divide them into two subcategories, based on how they encode images. 
        \begin{itemize}[leftmargin=*]
        \item \textbf{CNN-based Grid Features.} PixelBERT~\citep{huang2020pixel} and CLIP-ViL~\citep{shen2021much} feed grid features from CNNs and text directly into a Transformer. SOHO~\citep{huang2021seeing} first discretizes grid features using a learned vision dictionary, and then feeds the discretized features into their cross-modal module. While using grid features directly can be efficient, inconsistent optimizers are typically used for CNN and Transformer. For example, PixelBERT~\citep{huang2020pixel} and CLIP-ViL~\citep{shen2021much} use AdamW~\citep{loshchilov2018decoupled} for Transformer and SGD for CNN. 
        \item \textbf{ViT-based Patch Features.} Vision Transformers (ViTs) have been an increasingly active research topic in computer vision, motivating researchers to develop ViT-based VLP models. Among them, ViLT~\citep{kim2021vilt} directly feeds image patch features and text token embeddings into a pre-trained ViT model, and then pre-train the model on image-text datasets. ViTCAP~\citep{fang2022injecting} further extends ViLT for image captioning tasks. This has also led to follow-up works such as UFO~\citep{wang2021ufo} and VLMo~\citep{wang2021vlmo}, where UFO~\citep{wang2021ufo} uses the same Transformer to perform image/text encoding and multimodal fusion all together, while in VLMo~\citep{wang2021vlmo}, additional mixture-of-modality-experts layers are included. Besides this, visual parsing~\citep{xue2021probing}, ALBEF~\citep{li2021align}, METER~\citep{dou2021empirical}, BLIP~\citep{li2022blip}, X-VLM~\citep{zeng2021multi} and FIBER~\citep{dou2022coarse} all use ViT as their image encoder (\emph{e.g.}, plain ViT and Swin Transformer~\citep{liu2021swin}), and design different objectives for model pre-training.
        \end{itemize}
\end{itemize}

\begin{figure*}[t!]
  \centering
    \includegraphics[width=1.0\linewidth]{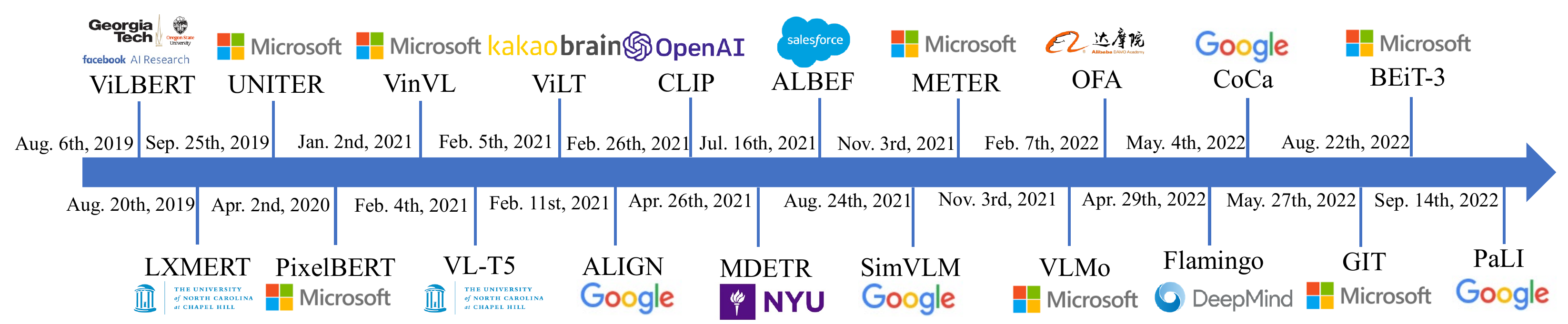}
  \caption{VLP models developed for image-text tasks along time. Due to space constraint, only some representative works are shown.}
  \label{fig:chp3_vlp_along_time}
\end{figure*}

We present a glossary of representative VLP models in Table~\ref{tab:chp3_vlp_glossary}, where models are dissected along multiple dimensions. In Figure~\ref{fig:chp3_vlp_along_time}, we show how these VLP models evolve along time. 

\begin{figure*}[t!]
  \centering
    \includegraphics[width=1.0\linewidth]{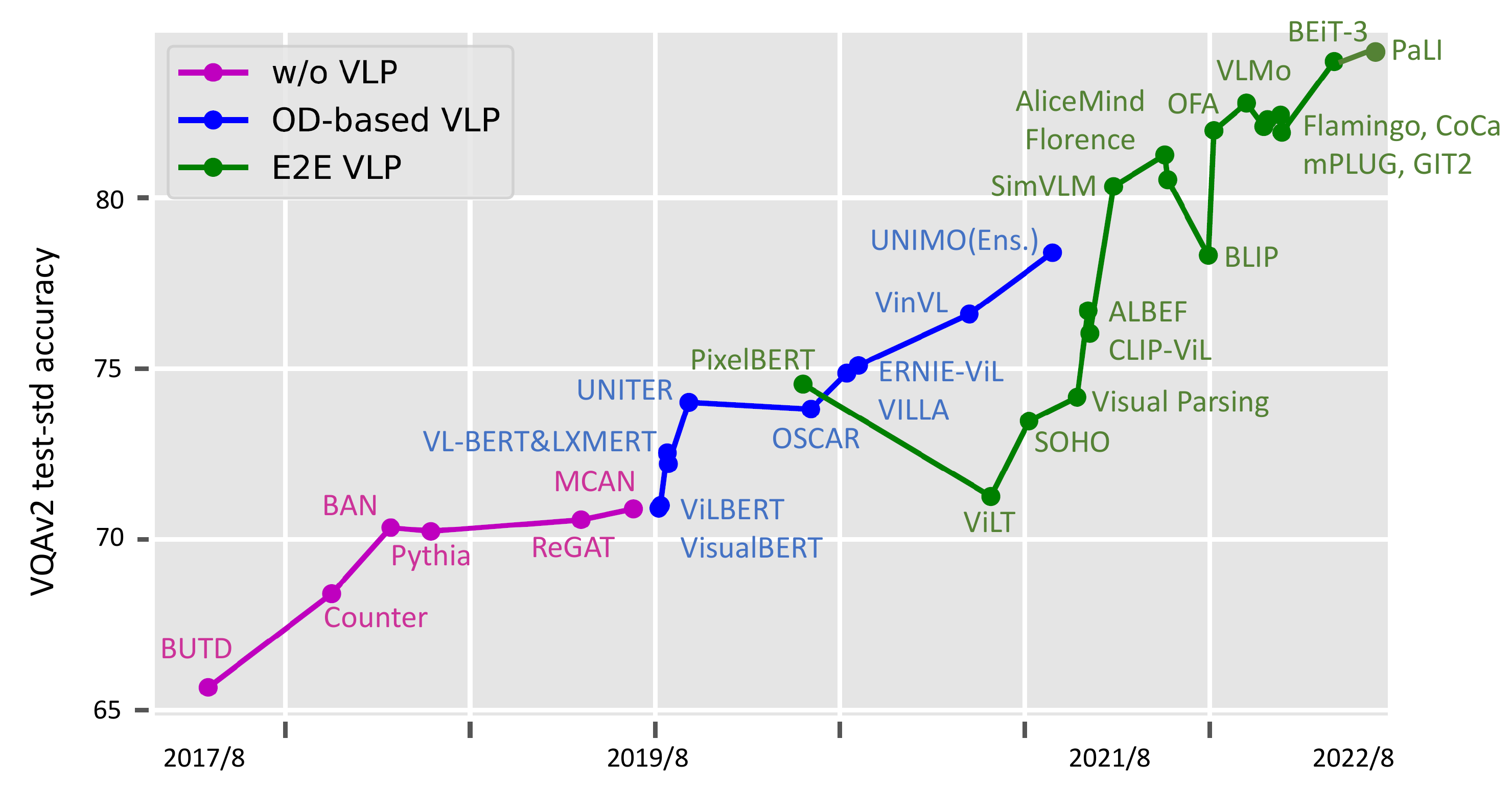}
  \caption{Research progress driven by large-scale VLP, using the VQA task as a case study. From August 2017 to August 2019, many task-specific methods have been developed. Since August 2019, OD-based VLP models have become popular. Later on, due to the emergence of vision Transformer~\citep{dosovitskiy2020image}, end-to-end VLP models have become the mainstream. During the last one year, we have witnessed a boom of big multimodal foundation models, \emph{e.g.}, SimVLM~\citep{wang2021simvlm}, Florence~\citep{yuan2021florence}, Flamingo~\citep{alayrac2022flamingo}, CoCa~\citep{yu2022coca}, GIT~\citep{wang2022git}, and BEiT-3~\citep{wang2022image}.  }
  \label{fig:chp3_vqa_summary}
\end{figure*}

\paragraph{Research Progress Driven by VLP.} Now, we use the VQA task as a case study to illustrate the research progress driven by large-scale VLP (see Figure~\ref{fig:chp3_vqa_summary}).

\begin{itemize}[leftmargin=*]
    \item \textbf{From August 2017 to August 2019}, many task-specific methods have been developed, ranging from the use of object-centric visual features, advanced attention mechanism design, object relational modeling, to the use of Transformer. The corresponding VQA accuracy has been boosted from $\approx$66\% to $\approx$71\%. 
    \item \textbf{From August 2019 to August 2021}, vision-language pre-training (VLP) has become the mainstream. It first started from OD-based VLP models, boosting the VQA accuracy from $\approx$71\% to $\approx$78\%; then end-to-end VLP methods based on convolutional networks and vision Transformer dominate the field.  
    \item \textbf{From August 2021 to August 2022}, we have witnessed a boom of big multimodal foundation models, \emph{e.g.}, SimVLM~\citep{wang2021simvlm}, Florence~\citep{yuan2021florence}, Flamingo~\citep{alayrac2022flamingo}, CoCa~\citep{yu2022coca}, GIT~\citep{wang2022git}, and BEiT-3~\citep{wang2022image}. When these models are scaled up in terms of both model size and pre-training dataset size, the VQA performance is further boosted from $\approx$80\% to $\approx$84\%.
\end{itemize}

\section{Model Architectures}~\label{sec:chp3_model_architectures}
\begin{figure*}[t!]
  \centering
    \includegraphics[width=1.0\linewidth]{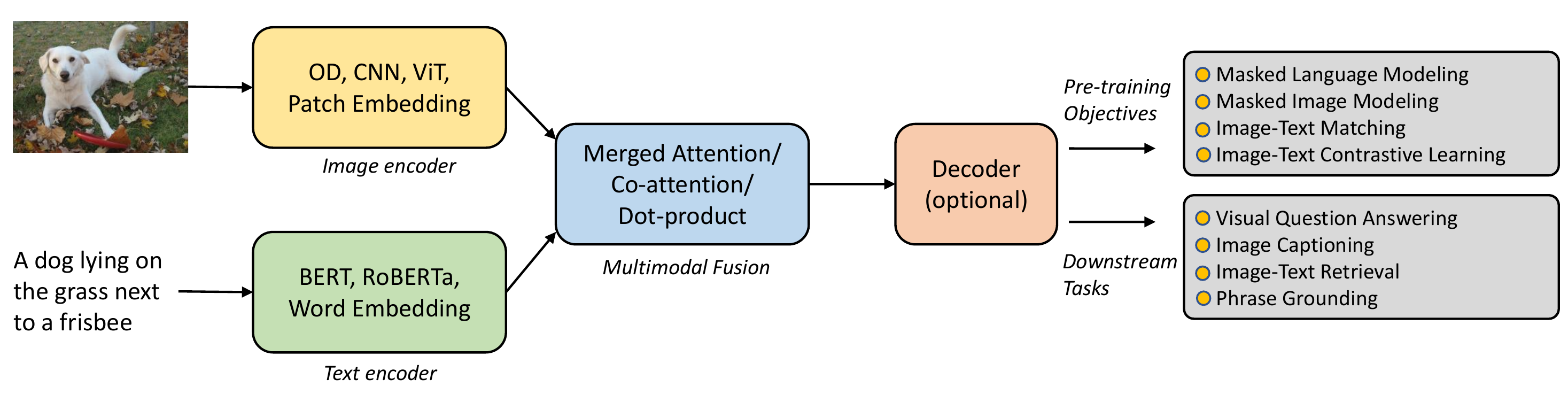}
  \caption{Illustration of a general framework for Transformer-based vision-language models.}
  \label{fig:chp3_framework}
\end{figure*}

\textbf{Overview.}
Given an image-text pair, a VL model first extracts text features $\wv=\{ \wv_1, \cdots, \wv_N\}$ and visual features $\vv=\{ \vv_1, \cdots, \vv_M\}$ via a \textit{text encoder} and a \textit{vision encoder}, respectively. Here, $N$ is the number of tokens in a sentence, and $M$ is the number of visual features for an image, which can be the number of image regions/grids/patches, depending on the specific vision encoder being used. 
The text and visual features are then fed into a \textit{multimodal fusion module} to produce cross-modal representations, which are then optionally fed into a \textit{decoder} before generating the final outputs. An illustration of this general framework is shown in Figure~\ref{fig:chp3_framework}.

In many cases, there are no clear boundaries among image/text backbones, multimodal fusion module, and the decoder. In this paper, we refer to the part of the model that only takes image/text features as input as the corresponding \emph{vision/text encoder}, and the part of the model that takes both image and text features as input as the \emph{multimodal fusion module}. Besides this, if there are additional modules that take the multimodal features as input to generate the output, we call it \emph{decoder}.

\paragraph{Vision Encoder.} As discussed in Section~\ref{sec:chp3_glossary}, there are three types of vision encoders: ($i$) an object detector (OD), ($ii$) a plain CNN, and ($iii$) a vision Transformer. 
\begin{itemize}[leftmargin=*]
    \item \textbf{OD.} The most widely used object detector for VL research is the Faster R-CNN~\citep{ren2015faster} pre-trained on the Visual Genome (VG) dataset~\citep{krishna2016visual} as in BUTD~\citep{anderson2018bottom}. 
    In VinVL~\citep{zhang2021vinvl}, a stronger OD model based on the ResNeXt-152 C4 architecture is pre-trained on multiple public OD datasets (including COCO~\citep{chen2015microsoftcoco}, OpenImages~\citep{kuznetsova2020open}, Objects365~\citep{shao2019objects365} and VG), and significant performance boost is observed across a wide range of VL tasks by using this stronger OD model. Additional care is taken to encode the location information of image regions, which is typically represented as a 7-dimensional vector.\footnote{[$x_1,y_1,x_2,y_2,w,h,w*h$] (normalized top/left/bottom/right coordinates, width, height, and area)} Both visual and location features are then fed through a fully-connected layer, to be projected into the same embedding space. The final embedding for each region is obtained by summing up the two FC outputs and then passing through a layer normalization layer.
    \item \textbf{CNN.} 
    In PixelBERT~\citep{huang2020pixel} and SOHO~\citep{huang2021seeing}, ResNet-50, ResNet-101 and ResNeXt-152 pre-trained from ImageNet classification are adopted. In CLIP-ViL~\citep{shen2021much}, ResNet-50, ResNet-101, and ResNet-50x4 pre-trained from CLIP~\citep{radford2021learning} are used. In SimVLM~\citep{wang2021simvlm}, they use the first three blocks (excluding the Conv stem) of ResNet-101 and ResNet-152 for their base and large models, respectively, and a larger variant of ResNet-152 with more channels for the huge model. Typically, it is observed that a stronger CNN backbone results in stronger downstream performance.
    \item \textbf{ViT.} Following~\cite{dosovitskiy2020image}, an image is first split into image patches, which are then flattened into vectors and linearly projected to obtain patch embeddings. A learnable special token \texttt{[CLS]} embedding is also prepended to the sequence. These patch embeddings, when summed up together with learnable 1D position embeddings and a potential image-type embedding, are sent into a multi-layer Transformer block to obtain the final output image features.  
    Different ViT variants have been studied for VLP, such as plain ViT~\citep{dosovitskiy2020image}, DeiT~\citep{touvron2020deit}, BEiT~\citep{bao2021beit}, Swin Transformer~\citep{liu2021swin}, and  CLIP-ViT~\citep{radford2021learning}, to name a few.
\end{itemize}

In a nutshell, no matter what vision encoder is used, the input image is represented as a set of feature vectors $\vv=\{ \vv_1, \cdots, \vv_M\}$.


\paragraph{Text Encoder.} Following BERT~\citep{devlin2018bert} and RoBERTa~\citep{liu2019roberta}, VLP models~\citep{tan-bansal-2019-lxmert,li2019visualbert,lu2019vilbert,su2019vl,chen2020uniter,li2020oscar} first segment the input sentence into a sequence of subwords~\citep{sennrich2016neural}, and then insert two special tokens at the beginning and the end of the sentence to generate the input text sequence. After we obtain the text embeddings, existing works either feed them directly to the multimodal fusion module~\citep{li2019visualbert,chen2020uniter}, or to several text-specific layers~\citep{tan-bansal-2019-lxmert,lu2019vilbert} before the fusion. For the former, the fusion module is typically initialized with BERT, and the role of text encoding and multimodal fusion is therefore entangled and absorbed in a single BERT model, and in this case, we consider text encoder as the word embedding layer. 

Language model (LM) pre-training has demonstrated impressive performance across tasks and different pre-trained LMs have been proposed. In METER~\citep{dou2021empirical}, the authors have  studied the use of BERT~\citep{devlin2018bert}, RoBERTa~\citep{liu2019roberta}, ELECTRA~\citep{clark2020electra}, ALBERT~\citep{lan2019albert}, and DeBERTa~\citep{he2020deberta} for text encoding. In Flamingo~\citep{alayrac2022flamingo}, a huge pre-trained LM with 70B parameters~\citep{hoffmann2022training} is used as the text encoder, and kept frozen during the VLP process for multimodal few-shot learning.
In a nutshell, no matter what text encoder is used, the input text is represented as a set of feature vectors $\wv=\{ \wv_1, \cdots, \wv_N\}$.

\begin{figure*}[t!]
  \centering
    \includegraphics[width=0.7\linewidth]{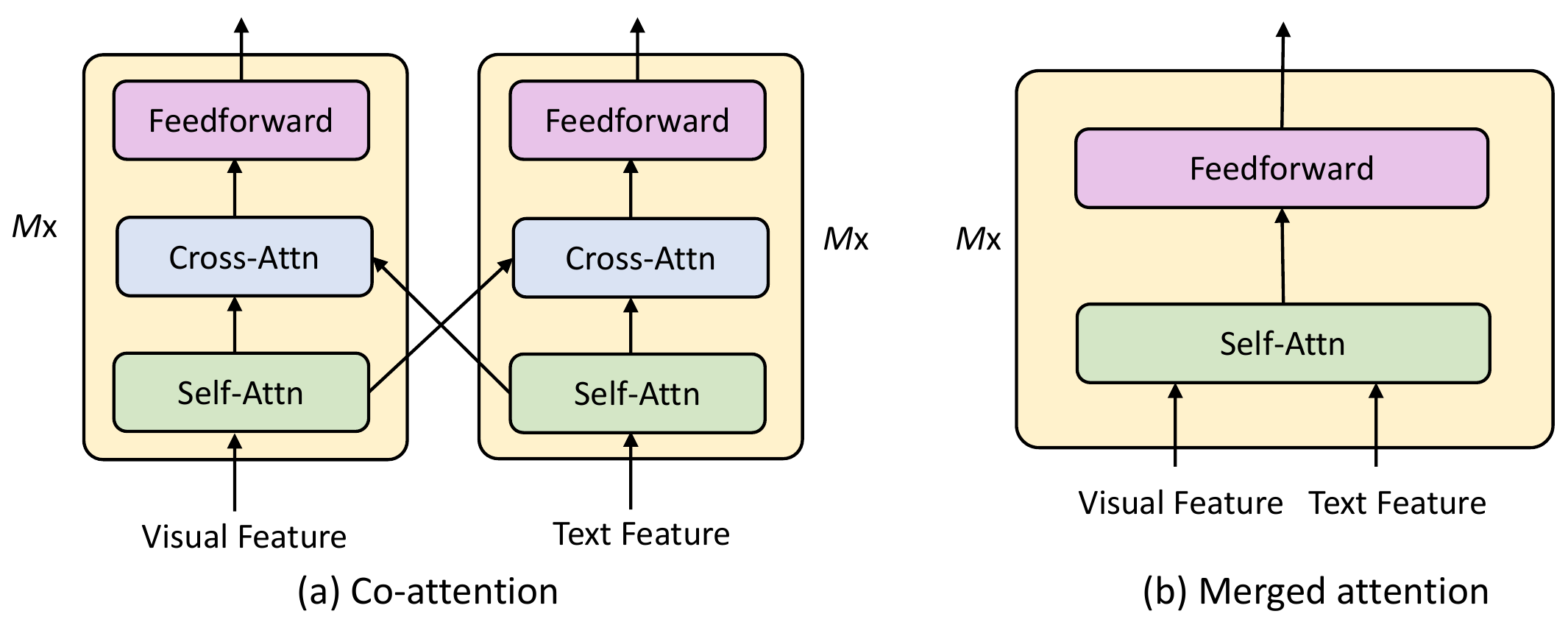}
  \caption{Co-attention and merged attention design for multimodal fusion.}
  \label{fig:chp3_fusion}
  \vspace{-3mm}
\end{figure*}

\paragraph{Multimodal Fusion.} 
For \emph{dual encoders} like CLIP~\citep{radford2021learning} and ALIGN~\citep{jia2021scaling}, fusion is performed via a dot-product between two global image and text feature vectors. 
For \emph{fusion encoder}, it takes both $\vv=\{ \vv_1, \cdots, \vv_M\}$ and $\wv=\{ \wv_1, \cdots, \wv_N\}$ as input, and learns contextualized multimodal representations denoted as $\tilde{\vv}=\{ \tilde{\vv}_1, \cdots, \tilde{\vv}_M\}$ and $\tilde{\wv}=\{ \tilde{\wv}_1, \cdots, \tilde{\wv}_N\}$. There are mainly two types of fusion modules, namely, \textit{merged attention} and \textit{co-attention}~\citep{hendricks2021decoupling}, shown in Figure~\ref{fig:chp3_fusion}. Specifically,
\begin{itemize}[leftmargin=*]
    \item In a \textbf{merged attention} module, the text and visual features are simply concatenated together, and then fed into a single Transformer block. This design has been used in many previous works, such as VisualBERT~\citep{li2019visualbert}, Unicoder-VL~\citep{li2020unicoder}, VLP~\citep{zhou2020unified}, VL-BERT~\citep{su2019vl}, UNITER~\citep{chen2020uniter}, OSCAR~\citep{li2020oscar}, VinVL~\citep{zhang2021vinvl}, ViLT~\citep{kim2021vilt}, GIT~\citep{wang2022git}, \emph{etc}.
    \item In a \textbf{co-attention} module, on the other hand, the text and visual features are fed into different Transformer blocks independently, and techniques such as cross-attention are used to enable cross-modal interaction. This design has been used in LXMERT~\citep{tan-bansal-2019-lxmert}, ViLBERT~\citep{lu2019vilbert}, ERNIE-ViL~\citep{yu2021ernie}, METER~\citep{dou2021empirical}, \emph{etc}. Also, in many models, only image-to-text cross-attention modules are used, such as ALBEF~\citep{li2021align}, BLIP~\citep{li2022blip}, CoCa~\cite{yu2022coca}, and  Flamingo~\citep{alayrac2022flamingo}.
\end{itemize}

For region-based VLP models, as shown in \cite{bugliarello-etal-2020-multimodal}, the \textit{merged attention} and \textit{co-attention} models can achieve comparable performance. Yet, the \textit{merged attention} module is more parameter-efficient, as the same set of parameters are used for both modalities. For end-to-end VLP models, as shown in METER~\citep{dou2021empirical}, co-attention performs better. However, there are no conclusive decision on which one is better, and it is largely an empirical choice for model design. 

In mPLUG~\citep{li2022mplug}, a combination of merged attention and co-attention is used for multimodal fusion; while in BLIP~\citep{li2022blip} and FIBER~\citep{dou2022coarse}, fusion is performed via simply inserting cross-attention modules inside the image and text backbones, which can be more lightweight and efficient. In MLP-ViL~\citep{nie2021mlp}, the authors study the use of MLP architectures for multimodal fusion.

\paragraph{\emph{Discussion: unified modeling with a shared backbone.}} Transformer has now become a universal computation engine~\citep{lu2021pretrained}. In UFO~\citep{wang2021ufo}, the authors have tried to use the same shared Transformer backbone for image/text encoding and multimodal fusion. In MS-CLIP~\citep{you2022learning} and VATT~\citep{akbari2021vatt}, the same shared backbone is used for contrastive pre-training across multiple modalities. In VLMo~\citep{wang2021vlmo}, additional mixture-of-modality-experts layers are further added, while the same self-attention layers are shared for image/text encoding and multimodal fusion. This mixture-of-expert design has achieved strong performance across multiple VL tasks.

\paragraph{Encoder-Only vs. Encoder-Decoder.} Most VLP models adopt the encoder-only architecture, where the cross-modal representations are directly fed into a MLP-based output layer to generate the final outputs. This encoder-only design naturally fits VL understanding tasks such as VQA and visual reasoning. When used for image captioning, the same encoder acts as a decoder to generate the output captions token by token by using a causal mask.

\begin{figure*}
  \centering
    \includegraphics[width=0.7\linewidth]{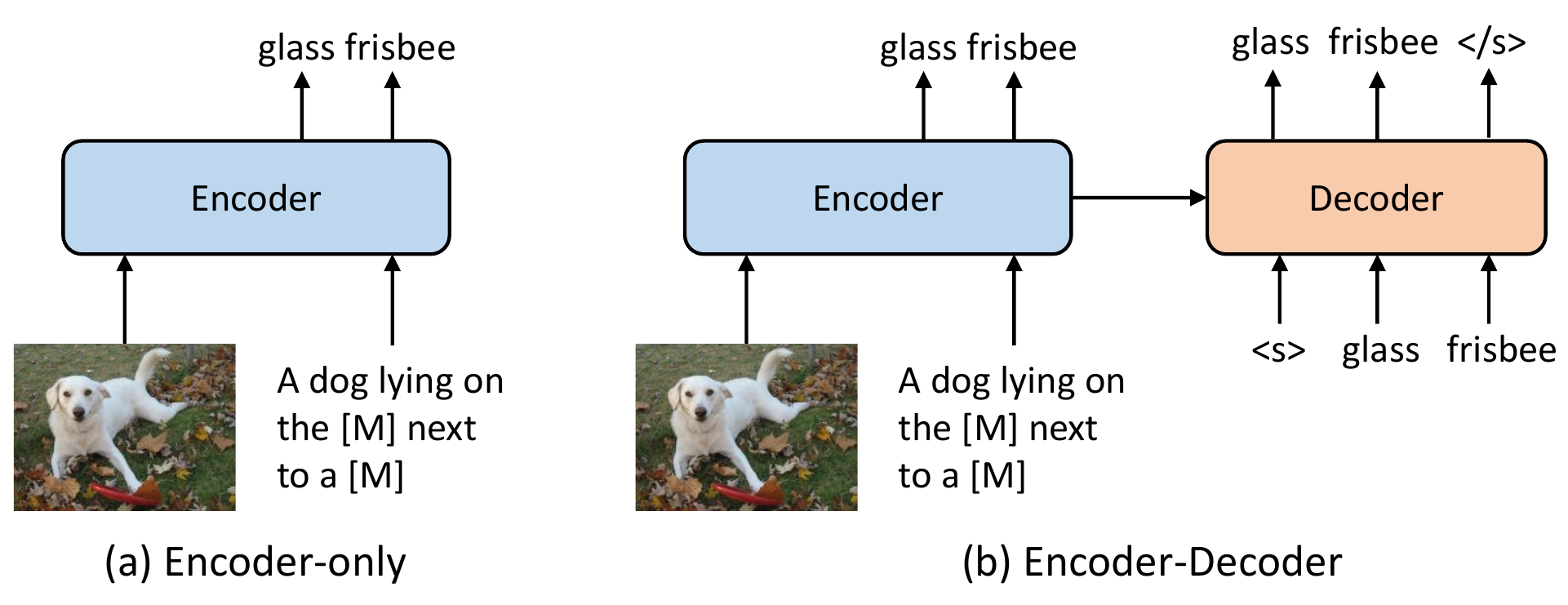}
  \caption{Comparison between Encoder-Only and Encoder-Decoder model architectures.}
  \label{fig:chp3_enc_vs_enc_dec}
\end{figure*}

Recently, inspired from T5~\citep{raffel2020exploring} and BART~\citep{lewis2019bart} in the NLP literature, VL-T5~\citep{cho2021unifying}, SimVLM~\citep{wang2021simvlm}, UniTAB~\citep{yang2021crossing}, OFA~\citep{wang2022unifying} and DaVinci~\citep{diao2022prefix}, on the other hand, advocate the use of a Transformer-based encoder-decoder architecture, where the cross-modal representations are first fed into a decoder and then to an output layer. In these models, the decoder attends to both the encoder representations and the previously generated tokens, producing the outputs autoregressively. The use of an encoder-decoder architecture can enable the unification of various image-text tasks and zero-shot/few-shot learning of VLP models (see Section~\ref{sec: unified_modeling} for more detailed discussions), and is also a natural fit for generation tasks. In MDETR~\citep{kamath2021mdetr}, the authors also adopt an encoder-decoder architecture, but the decoder is designed to generate bounding boxes in parallel, following the seminal work of DETR~\citep{carion2020end}. An illustrative comparison between encoder-only and encoder-decoder architectures is provided in Figure~\ref{fig:chp3_enc_vs_enc_dec}.

\section{Pre-training Objectives}~\label{sec:chp3_pretrain_objectives}
Now, we introduce how to design pre-training tasks. We will first review masked language modeling and image-text matching, which have been used extensively in almost every VLP model. Then, we will shift our focus to image-text contrastive learning and various masked image modeling tasks. 

\paragraph{Masked Language Modeling (MLM).} 
The MLM objective is first introduced in language model pre-training~\citep[\textit{e.g.,}][]{devlin2018bert,liu2019roberta}. In VLP, MLM with image-text pairs has also proven to be useful. Denote the mask indices as $\mathbf{m}\in \mathbb{N}^m$.\footnote{$\mathbb{N}$ is the natural numbers, $m$ is the number of masked tokens, and $\mathbf{m}$ is the set of masked indices.}
In MLM, given an image-text pair, we randomly mask out the input words with probability of 15\%, and replace the masked ones $\tilde{\mathbf{w}}_\mathbf{m}$ with special token \texttt{[MASK]}.\footnote{Following BERT~\citep{devlin2018bert}, this 15\% is typically decomposed into 10\% random words, 10\% unchanged, and 80\% \texttt{[MASK]}.}
The goal is to predict these masked tokens based on their surrounding words $\tilde{\mathbf{w}}_{\setminus \mathbf{m}}$ and the paired image $\tilde{\mathbf{v}}$, by minimizing the negative log-likelihood:
\begin{equation}
    \mathcal{L}_{\text{MLM}}(\theta) = -\mathbb{E}_{(\tilde{\mathbf{w}}, \tilde{\mathbf{v}})\sim D} \log P_{\theta}(\tilde{\mathbf{w}}_\mathbf{m} | \tilde{\mathbf{w}}_{\setminus \mathbf{m}}, \tilde{\mathbf{v}})\,,
\end{equation}
where $\theta$ denotes the trainable parameters. 
Each pair $(\tilde{\mathbf{w}}, \tilde{\mathbf{v}})$ is sampled from the whole training set $D$. There are several MLM variants used in VLP. Specifically,
\begin{itemize}[leftmargin=*]
    \item \textbf{Seq-MLM}: In order to adapt the pre-trained model for image captioning, it is observed~\citep{zhou2020unified,wang2021ufo} that adding a seq2seq \emph{causal mask} during pre-training is beneficial. That is, in Seq-MLM, the model can only use its preceding context to predict the masked token, which is consistent to the way the model performs image captioning during inference.  
    \item \textbf{LM}: Direct language modeling is used in BLIP~\citep{li2022blip} and CoCa~\citep{yu2022coca} for VLP. The model predicts the caption given an image token-by-token autoregressively. 
    \item \textbf{Prefix-LM}: Using the encoder-decoder framework as in SimVLM~\citep{wang2021simvlm}, a PrefixLM pre-training objective is proposed, where a sentence is first split into two parts, and the bi-directional attention is enabled on the prefix sequence and the input image, while a causal attention mask is adopted on the remaining tokens. 
\end{itemize}

\paragraph{Image-Text Matching (ITM).} In ITM, given a batch of matched or mismatched image-caption pairs, the model needs to identify which images and captions correspond to each other. Most VLP models treat image-text matching as a binary classification problem. Specifically, a special token (\emph{i.e.}, $\texttt{[CLS]}$) is
appended at the beginning of the input sentence to learn a global cross-modal representation. We then feed the model with either a matched or mismatched image-caption pair $\langle \tilde{\mathbf{v}}, \tilde{\mathbf{w}} \rangle$ with equal probability, and a classifier is added on top of the $\texttt{[CLS]}$ token to predict
a binary label $y$, indicating whether the sampled image-caption pair is matched. Specifically, denote the output score as $s_{\theta}(\tilde{\mathbf{w}}, \tilde{\mathbf{v}})$, We apply the binary cross-entropy loss for optimization:
\begin{equation}
    \mathcal{L}_{\text{ITM}}(\theta) = - \mathbb{E}_{(\tilde{\mathbf{w}}, \tilde{\mathbf{v}})\sim D} [y \log s_{\theta}(\tilde{\mathbf{w}}, \tilde{\mathbf{v}}) + (1-y) \log (1-s_{\theta}(\tilde{\mathbf{w}}, \tilde{\mathbf{v}}))] )\,.
\end{equation}
Besides randomly sampling a negative image-text pair, harder negative pairs can also be mined from an image-text contrastive loss introduced below, which has been shown to be effective in improving the downstream performance, as reported in ALBEF~\citep{li2021align}, VLMo~\citep{wang2021vlmo}, and FIBER~\citep{dou2022coarse}. 

\paragraph{Image-Text Contrastive Learning (ITC).} Early VLP models, such as UNITER~\citep{chen2020uniter} and VinVL~\citep{zhang2021vinvl}, do not use ITC for pre-training (one exception is LightningDOT~\citep{sun2021lightningdot}). Though the ITC loss is widely studied before VLP~\citep{frome2013devise}, in the context of end-to-end VLP, it is mostly popularized by CLIP~\citep{radford2021learning} and ALIGN~\citep{jia2021scaling} to pre-train a dual encoder. Later on, it is also used to pre-train a fusion encoder as in ALBEF~\citep{li2021align}. Note  that this ITC loss is used on top of the outputs of image and text encoders, before multimodal fusion (\emph{i.e.}, the use of $\wv$ and $\vv$, instead of $\tilde{\wv}$ and $\tilde{\vv}$). Specifically, given a batch of $N$ image-text pairs, ITC aims to predict the $N$ matched pairs from all the $N^2$ possible image-text pairs. With a little bit abuse of notation, let $\{\vv_i\}_{i=1}^N$ and $\{\wv_i\}_{i=1}^N$ denote respectively the normalized image vectors and text vectors in a training batch. To compute image-to-text and text-to-image similarities, we have: 
\begin{align}
    s_{i,j}^{i2t} = \vv_i^\top \wv_j,&\,\, s_{i,j}^{t2i} = \wv_i^\top \vv_j \,,\\
    \mathcal{L}_{\text{ITC}}^{i2t}(\theta) = -\frac{1}{N} \sum_{i=1}^N\log\frac{\exp(s_{i,i}^{i2t}/\sigma)}{\sum_{j=1}^N \exp (s_{i,j}^{i2t}/\sigma)},&\,\,\,\, \mathcal{L}_{\text{ITC}}^{t2i}(\theta) = -\frac{1}{N} \sum_{i=1}^N\log\frac{\exp(s_{i,i}^{t2i}/\sigma)}{\sum_{j=1}^N \exp (s_{i,j}^{t2i}/\sigma)}\,,
\end{align}
where $\sigma$ is a learned temperature hyper-parameter,  $\mathcal{L}_{\text{ITC}}^{i2t}$ and $\mathcal{L}_{\text{ITC}}^{t2i}$ are image-to-text and text-to-image contrastive loss, respectively. The ITC loss can be further enhanced via triple contrastive learning (\emph{i.e.}, TCL~\citep{yang2022vision}), a multimodal learnable codebook (\emph{i.e.}, CODIS~\citep{duan2022multi}), or a loop interaction between ITC and ITM (\emph{i.e.}, LoopITR~\citep{lei2022loopitr}).

\paragraph{Masked Image Modeling (MIM).} Similar to the MLM objective, researchers have studied various masked image modeling (MIM) tasks for pre-training. Specifically, the model is trained to reconstruct the masked patches or regions $\tilde{\mathbf{v}}_{\mathbf{m}}$ given the remaining visible patches or regions $\tilde{\mathbf{v}}_{\setminus \mathbf{m}}$ and all the words $\tilde{\mathbf{w}}$ as 
\begin{equation}
    \mathcal{L}_{\text{MIM}}(\theta) = \mathbb{E}_{(\tilde{\mathbf{w}}, \tilde{\mathbf{v}})\sim D} P_{\theta}(\tilde{\mathbf{v}}_\mathbf{m} | \tilde{\mathbf{v}}_{\setminus \mathbf{m}}, \tilde{\mathbf{w}})\,.
\end{equation}
The designs of MIM can be divided into two categories.
\begin{itemize}[leftmargin=*]
    \item \textbf{For OD-based VLP models}, \emph{e.g.}, LXMERT~\citep{tan-bansal-2019-lxmert} and UNITER~\citep{chen2020uniter}, some of the input regions are randomly masked (\emph{i.e.}, the visual features of the masked regions are replaced by zeros), and the model is trained to regress the original region features via minimizing the mean squared error loss. Researchers~\citep{tan-bansal-2019-lxmert,lu2019vilbert,chen2020uniter} have also tried to first generate object labels for each region using a pre-trained object detector, which can contain high-level semantic information, and the model is trained to predict the object labels for the masked regions instead of the original region features. 
    \item \textbf{For end-to-end VLP models}, \emph{e.g.}, ViLT~\citep{kim2021vilt} and METER~\citep{dou2021empirical}, researchers have investigated the use of masked patch regression/classification for masked image modeling. Specifically, 
        \begin{itemize}[leftmargin=*]
            \item For \textbf{MIM with discrete VQ tokens}, inspired by BEiT~\citep{bao2021beit}, discrete VQ tokens are first extracted for the input patches, and the model is then trained to reconstruct the discrete tokens. Specifically, the VQ-VAE~\citep{van2017neural} model in DALL-E~\citep{ramesh2021dalle} is first used to tokenize each image into a sequence of discrete tokens. Each image is resized so that the number of patches is equal to the number of tokens, and thus each patch corresponds to a discrete token. Then, we randomly mask 15\% of the patches and feed the masked image patches to the model as before, but now the model is trained to predict the discrete tokens instead of the masked patches.
            \item For \textbf{MIM with in-batch negatives}, by imitating MLM which uses a text vocabulary, the model is trained to reconstruct input patches by using a dynamical vocabulary constructed with in-batch negatives. Concretely, at each training step, we sample a batch of image-caption pairs $\{ \langle \vv^k, \wv^k \rangle \}_{k=1}^B$, where $B$ is the batch size. We treat all the patches in $\{ \mathbf{v}^k \}_{k=1}^B$ as candidate patches. For each masked patch, we mask 15\% of the input patches. The model needs to select the original patch within this candidate set. The model is trained to maximize its probability similar to noise contrastive estimation~\citep{nce1}.
        \end{itemize}
\end{itemize}

Notably, recent state-of-the-art VLP models (\emph{e.g.}, VinVL~\citep{zhang2021vinvl}, ALBEF~\citep{li2021align}, VLMo~\citep{wang2021vlmo}) do not apply MIM during pre-training, and in ViLT~\citep{kim2021vilt} and METER~\citep{dou2021empirical}, the authors also demonstrate that MIM is not helpful for downstream performance. However, there are also recent works that adopt masked vision-language modeling (as in MaskVLM~\citep{kwon2022masked} and VL-BEiT~\citep{bao2022vl}), which try to randomly mask patches/tokens while keeping the other modality intact. 

\paragraph{Other Pre-training Tasks.}
Besides these typical pre-training tasks introduced above, researchers have also investigated other possibilities. For example, 
\begin{itemize}[leftmargin=*]
    \item UNITER~\citep{chen2020uniter} proposes a word-region alignment objective that tries to align the image and text features using optimal transport~\citep{xie2020fast,chen2019improving,chen2020graph}.
    \item In E2E-VLP~\citep{xu2021e2e}, MDETR~\citep{kamath2021mdetr}, GLIP~\citep{li2021grounded}, and X-VLM~\citep{zeng2021multi}, bounding box prediction from object detection and phrase grounding is directly used as a fine-grained pre-training task.
\end{itemize}

\begin{figure*}
  \centering
\begin{subfigure}{0.48\linewidth}
    \includegraphics[width=1.0\linewidth]{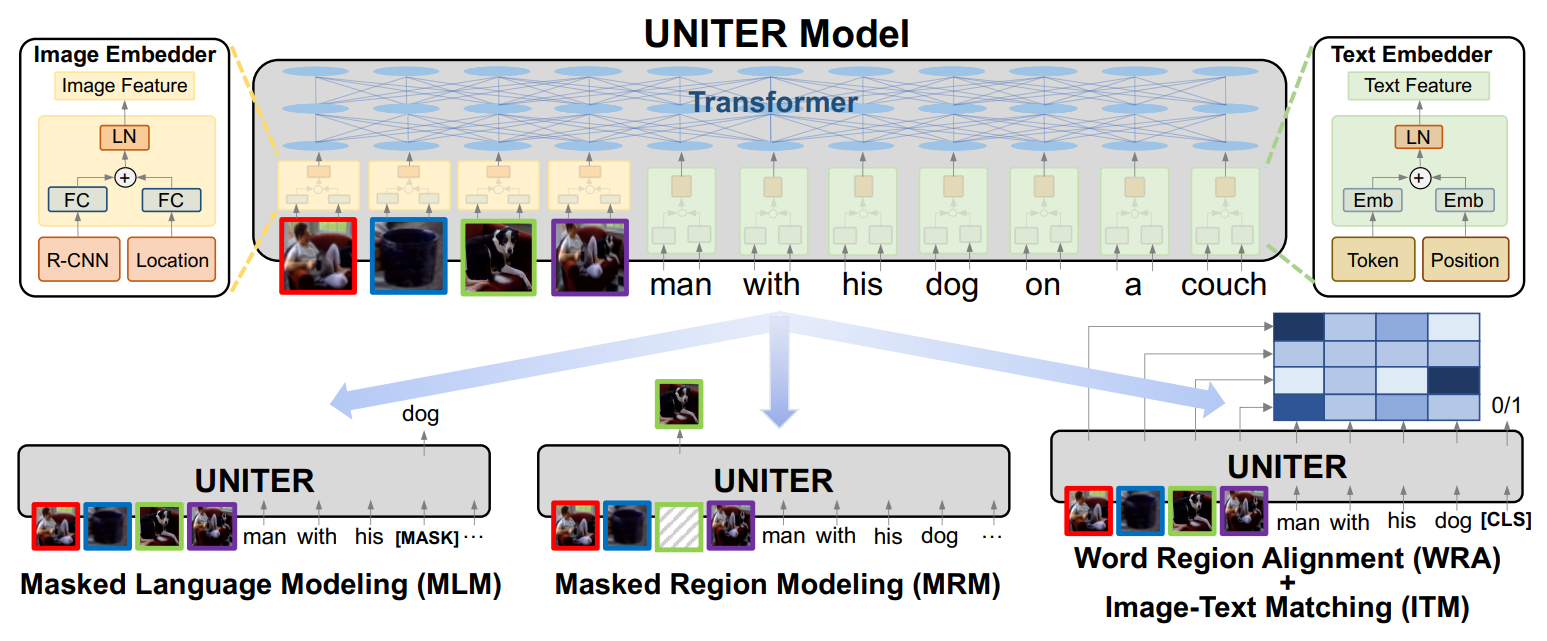}
    \caption{UNITER}
    \label{fig:chp3_uniter}
  \end{subfigure}
   \begin{subfigure}{0.48\linewidth}
    \includegraphics[width=1.0\linewidth]{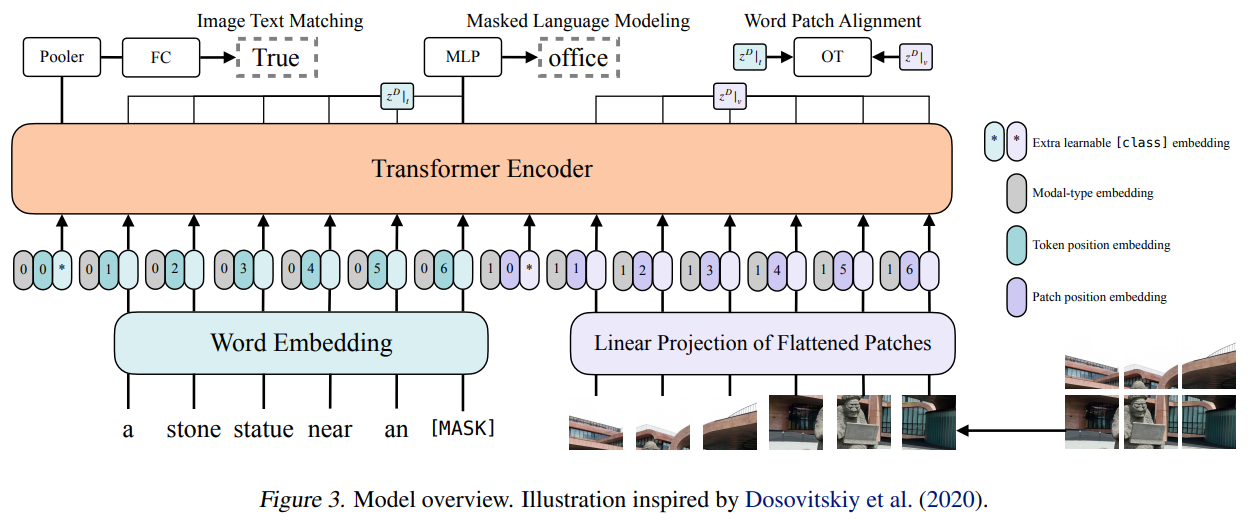}
    \caption{ViLT}
    \label{fig:chp3_vilt}
  \end{subfigure}\\
  \begin{subfigure}{0.48\linewidth}
    \includegraphics[width=1.0\linewidth]{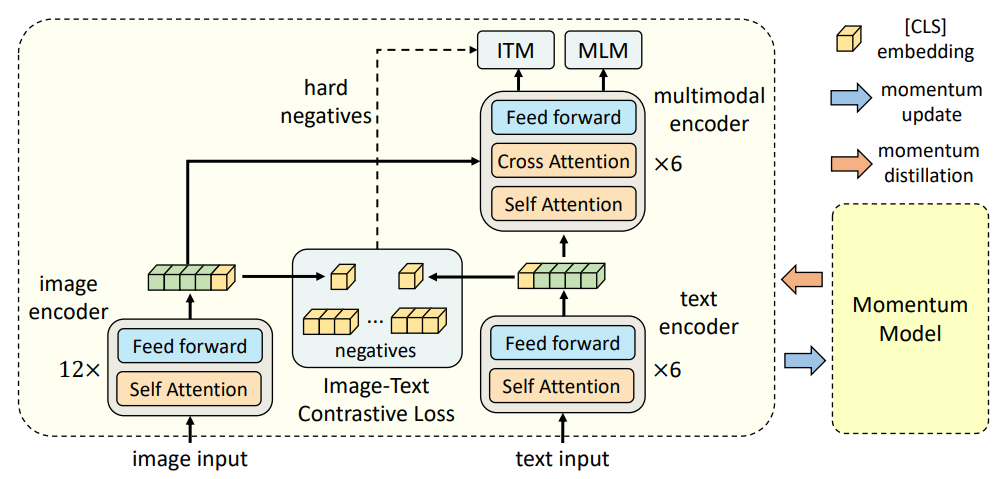}
    \caption{ALBEF}
    \label{fig:chp3_albef}
  \end{subfigure}
  \begin{subfigure}{0.48\linewidth}
    \includegraphics[width=1.0\linewidth]{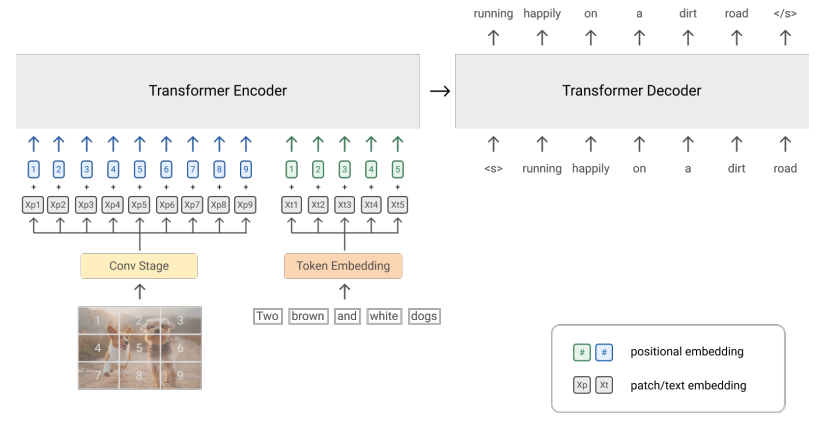}
    \caption{SimVLM}
    \label{fig:chp3_simvlm}
  \end{subfigure}
  \caption{Overview of four representative VLP models for image-text tasks: (a) UNITER~\citep{chen2020uniter}, (b) ViLT~\citep{kim2021vilt}, (c) ALBEF~\citep{li2021align},and (d) SimVLM~\citep{wang2021simvlm}. Figures are from the corresponding papers.
  \label{fig:chp3_case_study}}
  \vspace{-2mm}
\end{figure*}

\paragraph{Case Study.} Until now, we have introduced the general model architecture and popular pre-training tasks in the image-text literature. To provide the readers with more concrete examples, we select four representative models as case studies, including ($i$) UNITER~\citep{chen2020uniter}, an OD-based 
image-text model; 
($ii$) ViLT~\citep{kim2021vilt}, a minimal end-to-end image-text model that builds upon vision Transformer; 
($iii$) ALBEF~\citep{li2021align}, an end-to-end image-text model that uses both contrastive and generative objectives for pre-training, and ($iv$) SimVLM~\citep{wang2021simvlm}, the first large-scale pre-trained encoder-decoder image-text model with simple PrefixLM as the pre-training objective. Below, we briefly review their architectures and pre-training tasks. 
\begin{itemize}[leftmargin=*]
    \item \textbf{UNITER.} The architecture of UNITER is shown in Figure~\ref{fig:chp3_uniter}. The image is encoded by an offline pre-trained OD model to extract regional features. Together with positional embeddings, these image features are then concatenated with word embeddings from the input text, followed by several Transformer layers for multimodal fusion. The model is pre-trained via commonly used tasks including masked language modeling, image-text matching, and masked region modeling. The authors also provide a word-region alignment loss via the use of optimal transport. The multimodal Transformer is initialized via the pre-trained BERT-base or BERT-large model. 
    \item \textbf{ViLT.} Figure~\ref{fig:chp3_vilt} illustrates the model architecture of ViLT, which is the simplest image-text model one can imagine. The image is divided into patches, and encoded via patch embeddings, and the text is encoded via word token embeddings. These features are concatenated and sent to a Transformer, which is initialized via supervised pre-trained plain vision Transformer on ImageNet22k. Pre-training was performed via masked language modeling, image-text matching, matched patch modeling, and word-patch alignment.
    \item \textbf{ALBEF.} As shown in Figure~\ref{fig:chp3_albef}, ALBEF adopts a general VLP architecture which has also been extensively studied in METER~\citep{dou2021empirical}. Specifically, a vision Transformer is used to encode the image, the first 6 layers of a BERT model is used to encode the text, followed by multimodal fusion via the last 6 layers of the BERT model. The key innovation lies in the use of contrastive objectives during pre-training, which is introduced in CLIP, but has not been used for fusion-encoder-based image-text models. By incorporating the contrastive loss into pre-training, fast image-text retrieval via simple dot-product of two feature vectors can be achieved, while VQA and image captioning tasks that require deep multimodal fusion can also be tackled via the top fusion layers. 
    \item \textbf{SimVLM.} Lastly, we briefly discuss SimVLM, as shown in Figure~\ref{fig:chp3_simvlm}. CLIP and ALIGN are the first two large-scale pre-trained dual encoders, which can be only applied to (zero-shot) image classification and retrieval, while SimVLM is the first large-scale pre-trained encoder-decoder model that can be used for tasks that require deep multimodal fusion. Furthermore, the pre-training objective has been simplified as a single PrefixLM loss. The model shows great promise for training big image-text models, and a further detailed discussion is delayed to Section~\ref{sec:big_models}. 
\end{itemize}

\begin{table}[!t]
\centering
  \begin{tabular}{c|cccc|c}
    & \bf COCO & \bf VG & \bf CC3M & \bf SBU & \bf Total \\
    \hline
\#Images & 113K & 108K & 3.1M & 875K & 4.2M\\
\#Captions & 567K & 5.4M & 3.1M & 875K & 10M \\
  \end{tabular}
  \caption{Statistics of the pre-training datasets used in a typical academic setting.}
  \label{tab:chp3_vlp_data_stats}
  \vspace{-3mm}
\end{table}

\section{Pre-training Datasets}
\label{sec:chp3_pretrain_data}

In the era of large-scale multimodal pre-training, \emph{data is oxygen.} Besides model architecture and pre-training task innovations, another important factor to influence downstream performance is the \emph{pre-training datasets}. Below, we briefly review the pre-training datasets used in two settings: ($i$) the academic setting, which is more affordable for universities with limited computation resources, and ($ii$) the industrial setting, where large-scale web-crawled datasets are used for pre-training. 

\noindent \textbf{Academic Setting.} In a typical academic setting, VLP models are pre-trained on a collection of four commonly used image-caption datasets: COCO~\citep{chen2015microsoftcoco}, Visual Genome (VG)~\citep{krishna2016visual}, Conceptual Captions (CC3M)~\citep{sharma2018conceptual}, and SBU Captions~\citep{ordonez2011im2text}. The statistics of these pre-training datasets is shown in Table~\ref{tab:chp3_vlp_data_stats}. The first two are considered as ``in-domain'' datasets, as most VL downstream tasks are built on top of them; while the latter two are treated as ``out-of-domain'' datasets. Recently, CC12M~\citep{changpinyo2021conceptual} has also been frequently used for pre-training, which is a larger version of the original CC3M dataset by relaxing the data collection pipeline used in CC3M.

Localized Narratives~\citep{pont2020connecting} is another new image-text dataset, where the annotators are asked to describe an image with their voice while simultaneously hovering their mouse over the region they are describing; as a result, each image corresponds to a long paragraph. 
This dataset has recently been used for image-text pre-training in FLAVA~\citep{singh2022flava}. Besides image-caption datasets, existing works, such as UNIMO~\citep{li2020unimo}, UNIMO-2, and VL-BEiT~\citep{bao2022vl}, also propose to use image-only and text-only datasets for multimodal pre-training. 

\begin{figure*}[t!]
  \centering
    \includegraphics[width=1.0\linewidth]{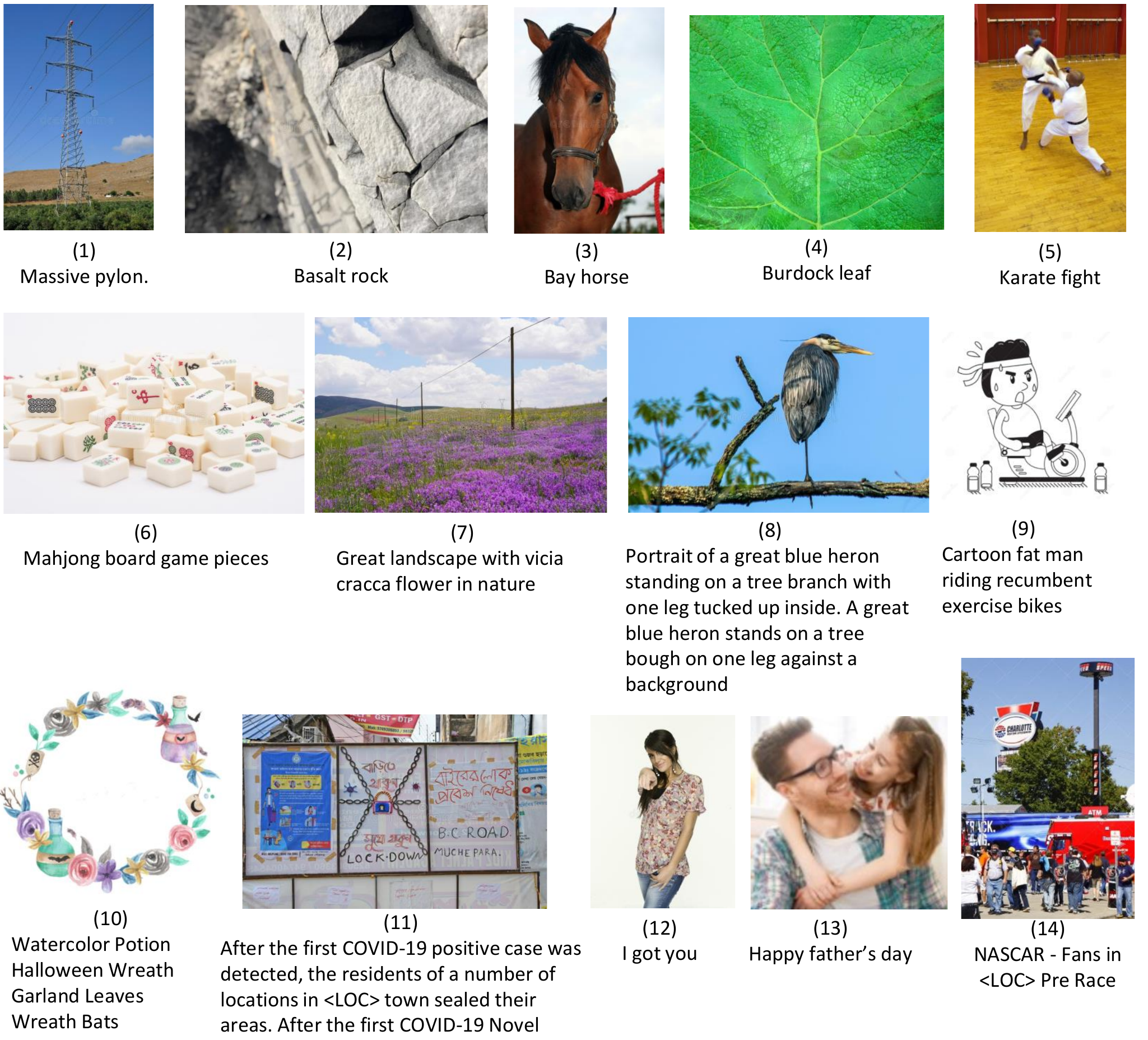}
  \caption{Examples of how the web-crawled image-text datasets look like. Figure credit: LEMON~\citep{hu2021scaling}.}
  \label{fig:chp3_visual_examples}
\end{figure*}

\paragraph{Industrial Setting.} 
In what follows, we brief  some of the web-crawled image-text datasets used in the industrial setting.
\begin{itemize}[leftmargin=*]
\item The dataset used in \textbf{CLIP}~\citep{radford2021learning} consists of $400$ million image-text pairs, which is built upon a set of $500,000$ queries. The queries include all the words occurring at least $100$ times in
the English version of Wikipedia and are augmented with bi-grams. The image-text pairs are searched such that the text includes one of the queries. The final results are also balanced to include up to $20,000$ image-text pairs per query.
\item The dataset used in \textbf{ALIGN}~\citep{jia2021scaling} has $1.8$ billion image-text pairs. Later works such as  SimVLM~\citep{wang2021simvlm} and CoCa~\citep{yu2022coca} also uses this dataset. The data collection pipeline is similar to that used in Conceptual Captions~\citep{sharma2018conceptual,changpinyo2021conceptual}, but most cleaning steps are relaxed. Only some rule-based filters are applied, such as image size, alt-text frequencies, and rare words.
\item The Wikipedia-based Image-Text Dataset (\textbf{WIT})~\citep{srinivasan2021wit} is composed of $11.5$ million unique images and $37.6$ million texts. Different from the aforementioned datasets, it features multilingual texts across $108$ languages. The images and texts are collected from the Wikipedia content pages. It provides texts from multiple sources, such as reference, attribution and
alt-texts, and texts in different languages for the same image.
\item \textbf{WenLan}~\citep{huo2021wenlan} consists of $30$ million image-text pairs. The web-collected pairs have gone through an elaborate cleaning process. For each data source, topic
models are used to extract topic words, and the topic distribution is analyzed to select desired contents.
\item \textbf{LAION-400M/5B}~\citep{schuhmann2021laion} has $400$ million or $5$ billion image-text pairs, and are recently released to public. Instead of applying human designed heuristics in data cleaning, this dataset relies on the CLIP~\citep{radford2021learning} model to filter image-text pairs, where the cosine similarity scores between image and text embeddings are calculated and filtered by threshold $0.3$. 
\item \textbf{RedCaps}~\citep{desai2021redcaps} comprises 12 million image-text pairs from 350 subreddits. It contains everyday things that users like to share on social
media, \emph{e.g.}, hobbies and pets. Captions often contain specific and fine-grained descriptions.
\item The dataset used in \textbf{Florence}~\citep{yuan2021florence} and \textbf{GIT}~\citep{wang2022git} contains 800 million image-text pairs, which include ALT200M introduced in LEMON~\citep{hu2021scaling}. This dataset has been scaled up to include 12 billion web-crawled image-text pairs.
\end{itemize}

The datasets such as CC3M~\citep{sharma2018conceptual}, CC12M~\citep{changpinyo2021conceptual}, WIT~\citep{srinivasan2021wit}, RedCaps~\citep{desai2021redcaps} and LAION-400M/5B~\citep{schuhmann2021laion} are released to public with the image URL and associated meta files. Other datasets are proprietary. 

Now, we provide some visual examples in Figure~\ref{fig:chp3_visual_examples} to show how the web-crawled datasets look like (examples are from ALT200M~\citep{hu2021scaling}). While some of the alt attributes are descriptive sentences that can serve as good training targets, \emph{e.g.}, Figure~\ref{fig:chp3_visual_examples} (7), (8), (9), it is noted that some texts are not semantically well-formed, \emph{e.g.}, Figure~\ref{fig:chp3_visual_examples} (10). Some texts are very short phrases containing only 2 to 4 words, \emph{e.g.}, Figure~\ref{fig:chp3_visual_examples} (1) - (6). Some texts do not precisely describe the image content, but refer to external knowledge or information. For example, Figure~\ref{fig:chp3_visual_examples} (12) shows a woman pointing at the camera, but the text is ``I got you''. 
The text steam in  Figure~\ref{fig:chp3_visual_examples} (11) is likely to be extracted from news. 
The quality of the textual data does present some challenges for the model to learn from noisy supervision. However, there are indeed a large variety of (fine-grained) visual objects present in the images and texts, such as burdock leaf, karate, mahjong, and great blue heron. Compared to human-annotated datasets, these web-collected data provide much richer training resources, especially for long-tailed concepts. 

\paragraph{Data Guidance.} Collecting massive image-text pairs at larger-scale has driven the development of foundation models in VLP (see Section~\ref{sec:big_models} for discussions about big models). However, 
these large, mostly uncurated, web-scraped datasets are usually collected with little oversight.
A recent data audit~\citep{birhane2021multimodal} to large-scale datasets (\emph{e.g.}, LAION-400M) uncovered a wide range of inappropriate content including pornographic imagery, racist slurs, and harmful social stereotypes (\textit{e.g.}, stereotypical
representations of people described as lawyers, flight attendants, homemakers, \emph{etc}.). One should always keep in mind during model development that training models on this data risks reflecting or even scaling up the underlying problems. In addition, it is crucial to follow responsible AI practices~\citep{mitchell2019model,gebru2021datasheets,pushkarna2022data} to transparently document and share information about datasets and models.

\section{Advanced Topics}
\label{sec:chp3_adv_topics}

\begin{figure*}
  \centering
    \includegraphics[width=1.0\linewidth]{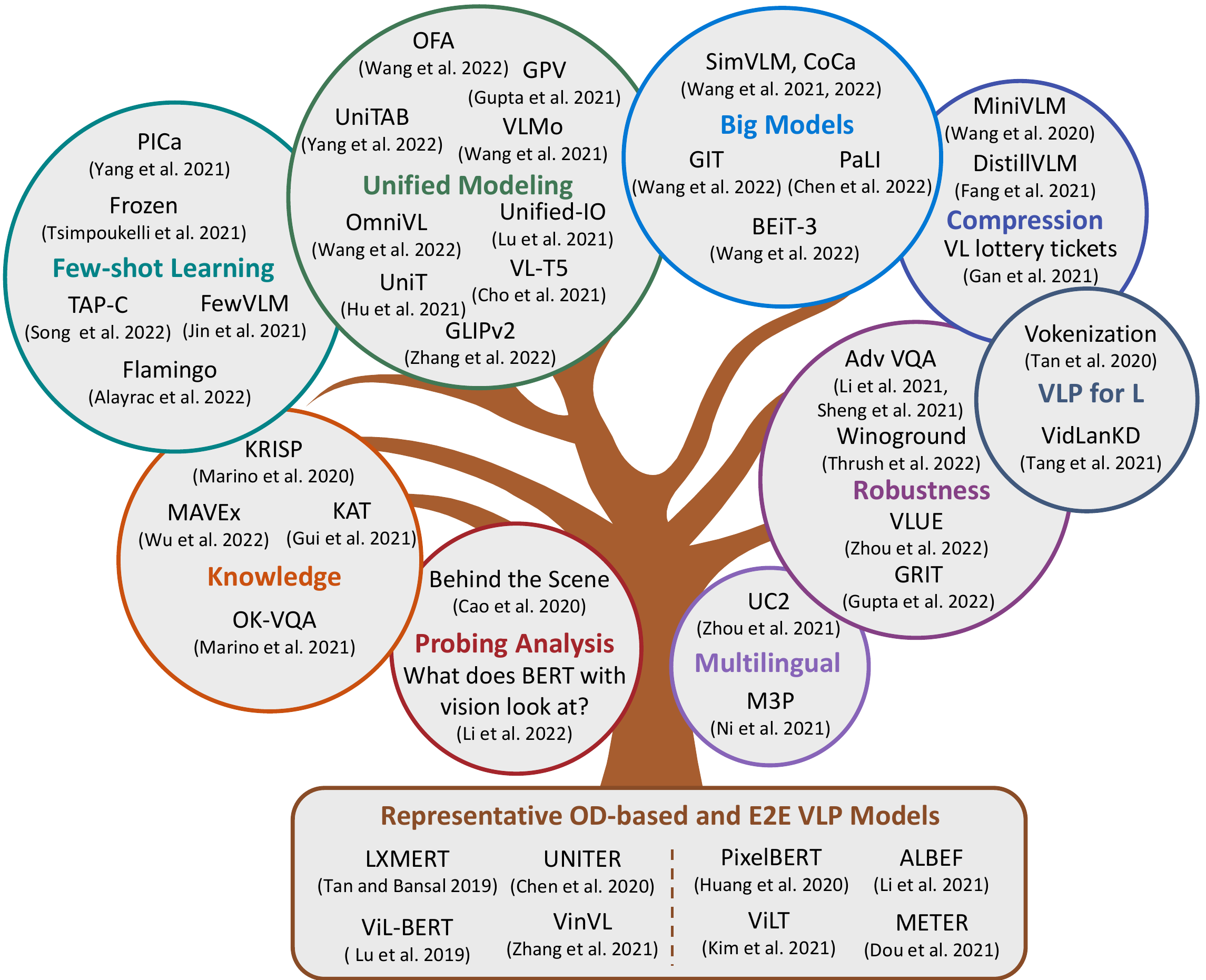}
  \caption{Advanced topics in VLP for image-text tasks. In each topic, we only list a few representative works due to limited space. Also, if some models are shown in one topic, it will not be shown in another topic to avoid repetition. For example, Flamingo~\citep{alayrac2022flamingo} is shown in the topic of Few-Shot Learning, therefore, not shown in the topic of Big Models. }
  \label{fig:chp3_advanced_topics}
\end{figure*}

As the literature on image-text-based VLP is growing rapidly, many other interesting research topics have emerged, as summarized in Figure~\ref{fig:chp3_advanced_topics}. Below, we provide a brief discussion on each individual topic, \emph{e.g.}, big models, few-shot learning, unified modeling, robustness evaluation, \emph{etc}. 

\subsection{Big Models} \label{sec:big_models}
\begin{figure*}[t!]
  \centering
    \includegraphics[width=0.85\linewidth]{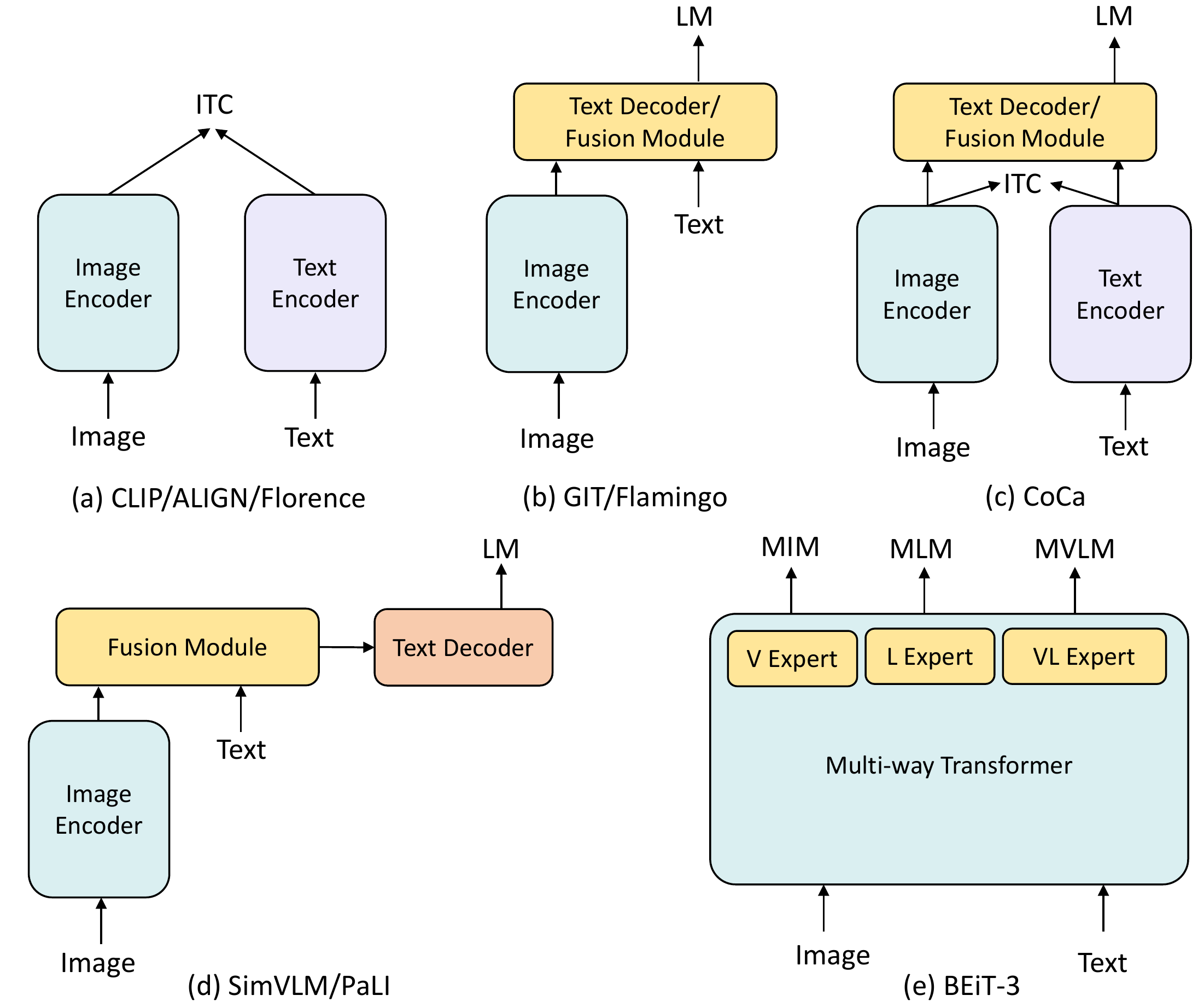}
  \caption{\textbf{Illustration of how the recent big VLP models look like.} (a) \emph{contrastive} pre-training, including models like CLIP~\citep{radford2021learning}, ALIGN~\citep{jia2021scaling}, Florence~\citep{yuan2021florence}, BASIC~\citep{pham2021combined}, \emph{etc}. (b) \emph{generative} pre-training, including models like GIT~\citep{wang2022git} and Flamingo~\citep{alayrac2022flamingo}. LEMON~\citep{hu2021scaling} and most previous OD-based VLP models also adopt this model architecture, but uses additional pre-training losses such as MLM and ITM. (c) Joint \emph{contrastive} and \emph{generative} pre-training, such as CoCa~\citep{yu2022coca}. METER~\citep{dou2021empirical} also uses this model architecture, but is pre-trained with MLM and ITM instead.
  Base-sized models such as ALBEF~\citep{li2021align} and FIBER~\citep{dou2022coarse} also adopt both ITC and MLM losses. (d) \emph{generative} pre-training with an encoder-decoder architecture, including models like SimVLM~\citep{wang2021simvlm} and PaLI~\citep{chen2022pali}. (e) VL-BEiT~\citep{bao2022vl} and BEiT-3~\citep{wang2022image} performs unified masked data modeling with a multi-way Transformer design. }
  \label{fig:chp3_big_models_architecture}
\end{figure*}

\emph{Scale} is believed to be important to achieve state-of-the-art performance and build general-purpose foundation models. As observed in the NLP field, larger language models are being pre-trained, from 340M-sized BERT-large model~\citep{devlin2018bert}, to GPT-3~\citep{brown2020language} with 175B parameters, and the more recent PaLM~\citep{chowdhery2022palm} with 540B parameters. A similar trend is observed in the VLP field. In Table~\ref{tab:chp3_big_models}, we summarize some of the recent big VLP models in terms of model size, pre-training dataset size, and pre-training tasks. Some observations are summarized below. 
\begin{itemize}[leftmargin=*]
\item Most big VLP models are obtained via either \emph{contrastive} pre-training or \emph{generative} pre-training, or a combination of both. An illustration of how these big models look like is provided in Figure~\ref{fig:chp3_big_models_architecture}. The use of ITC enables \emph{fast} image-text retrieval and open-set image classification, while the use of MLM or LM after the fusion module powers multimodal understanding tasks such as image captioning and VQA.
\item The current big VLP models typically contain roughly 1B parameters, pre-trained over roughly 1B-10B image-text pairs. 
\item Flamingo~\citep{alayrac2022flamingo} adopts a large \emph{frozen} language model (70B in size) to keep the in-context few-shot learning capability inherited from the pre-trained language model, while GIT~\citep{wang2022git} adopts a large contrastively pre-trained image encoder instead, with a relatively small text decoder.
\item Both Flamingo~\citep{alayrac2022flamingo} and GIT~\citep{wang2022git} first pre-train an image encoder via contrastive learning, and then perform generative pre-training. However, the image encoder and text decoder are both kept frozen in Flamingo~\citep{alayrac2022flamingo}; while the text decoder is randomly initialized in GIT~\citep{wang2022git}, and the image encoder is not kept frozen during the generative pre-training phase. 
\item Instead of performing contrastive and generative pre-training separately, CoCa~\citep{yu2022coca} performs a joint contrastive and generative pre-training in one stage.
\item By using only masked data modeling and a multi-way Transformer design, BEiT-3~\citep{wang2022image} achieves state-of-the-art performance on VQA and other VL tasks.
\end{itemize}

\begin{table*}[!t]
\resizebox{1.0\textwidth}{!}
{
  \begin{tabular}{l|cccc|c|c}
    \multirow{2}{*}{\bf Model} & \multicolumn{4}{c|}{\bf Model Size}  & \multirow{2}{*}{\bf PT dataset size}  & \multirow{2}{*}{\bf PT Tasks} \\
    \cmidrule(lr){2-5}
     & Image Enc. & Text Enc.$^\dagger$ & Fusion$^\dagger$ & Total & & \\
    \midrule
  CLIP ViT-L/14~\citep{radford2021learning}  & 302M & 123M & 0 & 425M  & 400M & ITC\\
  ALIGN~\citep{jia2021scaling}  & 480M & 340M & 0  & 820M  & 1.8B & ITC\\
  Florence~\citep{yuan2021florence}  & 637M & 256M & 0  & 893M  & 900M & ITC\\
  \midrule
  SimVLM-huge~\citep{wang2021simvlm}  & 300M & 39M  & 600M  & 939M & 1.8B & PrefixLM\\
  METER-huge~\citep{dou2021empirical}  & 637M & 125M & 220M  & 982M & 900M+20M$^1$ & MLM+ITM\\
  LEMON~\citep{hu2021scaling}  & 147M$^2$ & 39M & 636M  & 822M & 200M & MLM\\
  Flamingo~\citep{alayrac2022flamingo}  & 200M & 70B & 10B  & 80.2B & 2.1B+27M$^3$ & LM\\
  GIT~\citep{wang2022git}  & 637M & 40M & 70M  & 747M & 800M & LM\\
  GIT2~\citep{wang2022git}  & 4.8B & 40M & 260M  & 5.1B & 12.9B & LM\\
  CoCa~\citep{yu2022coca}  & 1B & 477M & 623M  & 2.1B & 1.8B+3B$^4$ & ITC+LM\\
  \multirow{2}{*}{BEiT-3~\citep{wang2022image}}  & \multirow{2}{*}{692M$^5$} & \multirow{2}{*}{692M$^5$} & \multirow{2}{*}{52M$^5$}  & \multirow{2}{*}{1.9B} & \multirow{2}{*}{21M+14M$^6$} & MIM+MLM\\
  & & & & & & +MVLM \\
  \multirow{2}{*}{PaLI~\citep{chen2022pali}}  & \multirow{2}{*}{3.9B} & \multirow{2}{*}{40M} & \multirow{2}{*}{13B}  & \multirow{2}{*}{16.9B} & \multirow{2}{*}{1.6B} & LM+VQA$^7$\\
  & & & & & & +OCR+OD \\
  \end{tabular}
  }
  \caption{\textbf{A summary of recent big VLP models} in terms of model size, pre-training dataset size, and pre-training tasks. Note, that some of the numbers shown in this table are based on our best estimate. $^1$: 20M image-text pairs are used for VLP, while 900M data is used to pre-train the Florence image encoder. $^2$: This is the model size of an object detector as used in VinVL~\citep{zhang2021vinvl}. $^3$: 2.1B image-text data plus 27M video-text data. $^4$: 1.8B image-text data plus 3B image-tag data before filtering. $^5$: shared attention blocks contain another 317M parameters. $^6$: 21M image-text pairs plus 14M images from ImageNet-21K (additional 160GB documents are omitted here). $^7$: A complete set of pre-training tasks for PaLI~\citep{chen2022pali} include LM, PrefixLM, VQA, VQG, OCR, and OD. $^\dagger$: In our context, a module that takes both image and text features as input is considered as the fusion module, and a module that only takes text as input is considered as text encoder. Sometimes, the fusion module is called a text decoder in the literature, such as in SimVLM~\citep{wang2021simvlm}, Flamingo~\citep{alayrac2022flamingo}, and GIT~\citep{wang2022git}. ITC: image-text contrastive loss. ITM: image-text matching. MLM/LM: (masked) language modeling. MIM: masked image modeling. MVLM: masked vision-language modeling. }
  \label{tab:chp3_big_models}
  \vspace{-3mm}
\end{table*}

\subsection{In-Context Few-Shot Learning} \label{sec:few_shot}
Achieving state-of-the-art performance via full model finetuning is good. It is more desirable to train a model that can quickly adapt to different downstream tasks via only providing a few in-context examples. In the context of language model pre-training, such capability has been demonstrated in GPT-3~\citep{brown2020language} via large-scale pre-training on massive text corpora. Inspired by this, researchers have also started to investigate multimodal in-context few-shot learning. Below, we mainly discuss three pieces of work: Frozen~\citep{tsimpoukelli2021multimodal}, PICa~\citep{yang2021empirical}, and Flamingo~\citep{alayrac2022flamingo}..  
\begin{itemize}[leftmargin=*]
    \item Frozen~\citep{tsimpoukelli2021multimodal} is the pioneering work on this topic. It shows that by using a large \emph{frozen} language model and learning an image encoder to align the embedding space of images and text via a simple image captioning task, strong in-context few-shot learning performance can be obtained. However, an image is encoded using only two global vectors, which are not sufficient to capture all visual information of the image. Further, the frozen language model is only 7B in model size, which may not be large enough. 
    
    \item In order to retain the strong in-context few-shot learning capability of the 175B-sized GPT-3~\citep{brown2020language}, PICa~\citep{yang2021empirical} proposes to prompt GPT-3 via the use of image captions for multimodal few-shot learning, since GPT-3 can only read text but not images. With such a simple approach, 4-shot prompting can already outperform supervised SoTA on the challenging OK-VQA benchmark that requires external knowledge to answer a question about an input image correctly. However, its performance improvement on the VQAv2 dataset is limited, since captions cannot capture every detail of an image, and fine-grained visual information can be lost. Recently, in a similar spirit, VidIL~\citep{wang2022language} is proposed to perform few-shot video-language learning via inheriting the in-context learning capability from GPT-3 as well.
    
    \item To address the above challenges, Flamingo~\citep{alayrac2022flamingo} proposes to use both a contrastively pre-trained \emph{frozen} image encoder and a large \emph{frozen} language model, and insert gated cross-attention modules to bridge these two frozen models. By large-scale pre-training and using a 70B-sized frozen language model, SoTA in-context few-shot learning results are reported.
\end{itemize}
      
Besides relying on large language models, researchers have also explored other approaches for few-shot learning. In FewVLM~\citep{jin2021good}, the authors propose to train a VL-T5-like model~\citep{cho2021unifying} with PrefixLM and MLM, and found that PrefixLM is helpful for zero/few-shot image captioning, while MLM is good for zero/few-shot VQA. In TAP-C~\citep{song2022clip}, the authors show that CLIP~\citep{radford2021learning} can be a few-shot learner for VQA and visual entailment tasks. For VQA, the authors propose to reformulate it as an image-text retrieval task; while for visual entailment, captions and hypothesis (text-text pairs) are used in training, while image and hypothesis (image-text pairs) are used at inference.

\paragraph{Zero-shot Image Captioning.} A crucial benefit of training big VLP models is the potential of achieving zero-shot generalization. In image-text tasks, while zero-shot retrieval can be readily powered by the use of contrastive loss during pre-training, \emph{zero-shot} image captioning has been rarely evaluated, largely due to that the zero-shot performance is poor as the model is pre-trained on web-scale noisy image-text pairs. Quantitative evaluation of zero-shot captioning is provided in SimVLM~\citep{wang2021simvlm} and FewVLM~\citep{jin2021good}, and qualitative visual examples are provided in LEMON~\citep{hu2021scaling} and CM3~\citep{aghajanyan2022cm3}. Zero-shot image captioning can also be achieved via the use of CLIP and GPT-2 together, as discussed in MAGIC~\citep{su2022language} and ZeroCap~\citep{tewel2022zerocap}.

\begin{figure*}[t!]
  \centering
    \includegraphics[width=1.0\linewidth]{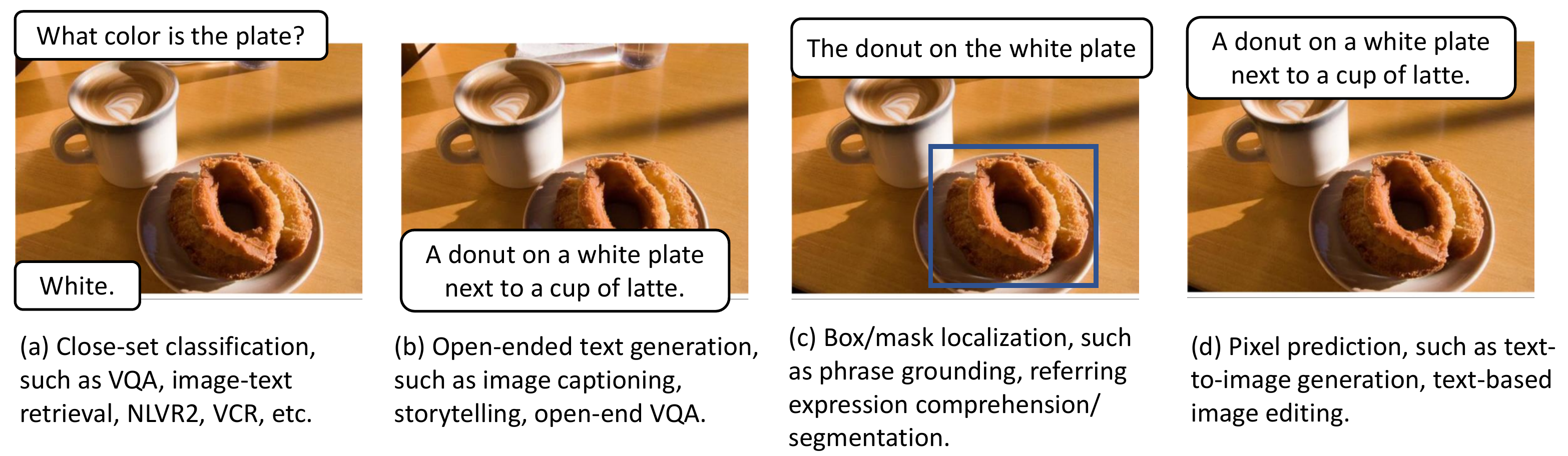}
  \caption{\textbf{The spectrum of image-text tasks that researchers have tried to unify.} (a) Close-set classification, such as VQA, image-text retrieval, NLVR2, \emph{etc}. (b) Open-ended text generation, such as image captioning, visual storytelling, and open-ended VQA. (c) Box/mask localization, such as phrase grounding, referring expression comprehension/segmentation, and grounded captioning. (d) Pixel prediction, such as text-to-image generation, text-based image editing, \emph{etc}. Figure credit: from Zhengyuan Yang's CVPR 2022 tutorial slides on unified image-text modeling.}
  \label{fig:chp3_unified_vl_modeling}
\end{figure*}

\subsection{Unified Image-Text Modeling} \label{sec: unified_modeling}
As shown in Figure~\ref{fig:chp3_unified_vl_modeling},
image-text tasks can be roughly divided into four categories: ($i$) Close-set classification, such as VQA, image-text retrieval, and visual reasoning; ($ii$) Open-ended text generation, such as image captioning, visual storytelling, and free-form open-ended VQA; ($iii$) Box/mask localization, such as phrase grounding, referring expression comprehension/segmentation, and grounded captioning; and ($iv$) Pixel prediction, such as text-to-image generation and text-based image editing. How to design a unified image-text model that can support all these downstream tasks becomes an increasingly important topic. We provide a brief summary of current attempts towards this goal below.
\begin{itemize}[leftmargin=*]
    \item \textbf{Unifying image-text tasks as text generation.} Borrowing ideas from T5~\citep{raffel2020exploring} and BART~\citep{lewis2019bart}, VL-T5~\citep{cho2021unifying} proposes to use a sequence-to-sequence (seq2seq) encoder-decoder framework to unify different VL tasks as text generation, so that different tasks can be directly supported without introducing task-specific heads. Since pre-trained object detectors are used to (pre-)extract bounding boxes and the corresponding regional features, the box prediction task in phrase grounding and referring expression comprehension becomes a region index classification problem. However, the fact that the model cannot be end-to-end pre-trained results in sub-optimal downstream performance.
    SimVLM~\citep{wang2021simvlm} proposes a simple end-to-end seq2seq learning framework, and considers VQA as a text generation task as in VL-T5, and performs large-scale pre-training. 
    
    \item \textbf{Unifying text generation and box prediction as language modeling.} The approaches above have unified certain image-text tasks (\emph{e.g.}, VQA, visual reasoning and image captioning) as text generation. However, bounding box coordinates cannot be directly predicted. 
    By quantizing bounding box coordinates as discrete tokens, Pix2Seq~\citep{chen2021pix2seq} and Pix2SeqV2~\citep{chen2022unified} propose to treat object detection (OD) as a language modeling task using a seq2seq framework. Inspired by this, in UniTAB~\citep{yang2021crossing}, the authors have tried to unify text generation and bounding box prediction into a single Transformer encoder-decoder architecture via representing each bounding box using a set of discrete tokens, which enables UniTAB to approach different VL tasks with a single set of parameters, generate desired text and box outputs together, and meanwhile detect the alignments between words and boxes.
    
    \item \textbf{Unifying text generation and image generation as language modeling.} Through the use of VQ-VAE~\citep{van2017neural,razavi2019generating}, images can also be represented as a sequence of discrete image tokens. Therefore, image generation can be naturally regarded as a language modeling task.  Recent works, such as Taming Transformer~\citep{esser2021taming}, DALL-E~\citep{ramesh2021dalle}, and Parti~\citep{yu2022scaling}, have shown that this approach can generate high-quality realistic images. Inspired by this, recent work shows that image generation and text generation (\emph{e.g.}, image captioning) can be unified, such as ERINE-ViLG~\cite{zhang2021ernie}, L-Verse~\citep{kim2021verse}, and DU-VLG~\citep{huang2022vlg}. Furthermore, DaVinci~\citep{diao2022prefix} combines a prefix image modeling task and a prefix language modeling (as used in SimVLM~\citep{wang2021simvlm}) for pre-training. \cite{aghajanyan2022cm3} introduce CM3, a causally masked generative model pre-trained over a large corpus of structured multi-modal documents that can contain both text and image tokens (from a pre-trained VQVAE-GAN). After pre-training, the authors show that the model can generate images unconditionally, conditioned on text, and learn to perform image captioning in a zero-shot setting.
    
    \item \textbf{Unifying text generation, box prediction and image generation all together.} In OFA~\citep{wang2022ofa}, the authors propose to unify text generation, box prediction, and image generation all together, by combining the ideas of Pix2Seq~\citep{chen2021pix2seq} and VQ-VAE~\citep{van2017neural}. Using the same idea, Unified-IO~\citep{lu2022unified} further supports modalities as diverse as images, masks, key points, boxes, and text, and tasks as varied as depth estimation, inpainting, semantic segmentation, captioning, and reading comprehension. However, the performance of Unified-IO on downstream tasks is not satisfactory at its current stage.
    
    \item \textbf{Unifying localization and VL understanding.} Serializing bounding boxes as token sequences allows the design of a unified model to tackle all tasks without introducing task-specific heads. This is appealing. However, the downstream object detection (OD) performance is either not evaluated, or still lagging behind the state of the art by a large margin. There is another line of work that tries to unify localization and VL understanding but still uses additional OD heads to output bounding boxes. Prominent examples include GPV-1~\citep{gupta2022towards}, MDETR~\citep{kamath2021mdetr}, UniT~\citep{hu2021unit}, GLIPv2~\citep{zhang2022glipv2}, and FIBER~\citep{dou2022coarse}. Specifically, GPV-1~\citep{gupta2022towards} and GPV-2~\citep{kamath2022webly} advocate the concept of \emph{general-purpose} vision systems. MDETR~\citep{kamath2021mdetr} and GLIP~\citep{li2021grounded} propose to unify object detection and phrase grounding for grounded pre-training, which further inspires GLIPv2~\citep{zhang2022glipv2} to unify localization and VL understanding.
    FIBER~\citep{dou2022coarse} provides another solution to tackle both localization and VL understanding tasks, by designing a new fusion-in-the-backbone architecture, and a new pre-training strategy, \emph{i.e.}, first performing coarse-grained pre-training on image-text data, followed by fine-grained pre-training on image-text-box data. 
\end{itemize}

Besides unifying different tasks within one framework, there are also works on designing a unified Transformer. For example, UFO~\citep{wang2021ufo} develops a unified Transformer that can be flexibly used as dual encoder and fusion encoder. VLMo~\citep{wang2021vlmo} proposes to further introduce additional modality-specific experts, and its scaled-up version BEiT-3~\citep{wang2022image} has recently achieved state-of-the-art results on VQA and other VL tasks.

\subsection{Knowledge} \label{sec:knowledge}

We mainly focus on knowledge-requiring VQA tasks that require external knowledge in addition to the image content to answer a question correctly. Below, we divide the discussion into three parts.
\begin{itemize}[leftmargin=*]
    \item \textbf{Datasets.}
    The earliest explicit knowledge-based VQA datasets are KB-VQA~\citep{wang2015explicit} and FVQA~\citep{wang2017fvqa}. However, the knowledge required in these datasets is retained in the same knowledge graphs that are used to generate the dataset. KVQA~\citep{shah2019kvqa} is based on images in Wikipedia articles. OK-VQA~\citep{marino2019ok} is a recent popular VQA dataset that requires external, open-domain knowledge to answer a question given an input image. More recently, WebQA~\citep{chang2022webqa} is collected using web queries, and A-OKVQA~\citep{schwenk2022okvqa} is a crowdsourced dataset composed of a diverse set of questions requiring a broader base of commonsense and world knowledge to answer. 
    
    \item \textbf{Knowledge sources.} There are two categories of knowledge sources: ($i$) \emph{explicit} structured symbolic knowledge bases such as Wikipedia, ConceptNet, WordNet, and Google images; and ($ii$) \emph{implicit} unstructured knowledge bases, \emph{i.e.}, large-scale pre-trained language models such as GPT-3~\citep{brown2020language}, where rich encyclopedia and commonsense knowledge has been encoded.
    
    \item \textbf{Methods.} Most studies followed a two-step approach to tackle the knowledge-based VQA tasks, \emph{i.e.}, first retrieve knowledge from external resources, and then reason over the selected knowledge, the input image, and question for answer prediction. Below, we mainly discuss methods designed for OK-VQA. Specifically, \cite{shevchenko2021reasoning} propose to build a knowledge base with knowledge embeddings, and then inject these knowledge embeddings into VLP models. KRISP~\citep{marino2021krisp} propose to retrieve the implicit knowledge stored in pre-trained language models as a supplementary knowledge resource to the structured knowledge base. MAVEx~\citep{wu2022multi} presents an answer validation approach to make better use of the noisy retrieved knowledge. More recently, PICa~\citep{yang2021empirical} shows that by prompting GPT-3 via the use of image captioning and in-context few-shot learning, state-of-the-art results can be obtained. This approach has been further enhanced in KAT~\citep{gui2021kat} by additionally retrieving  knowledge from explicit knowledge bases. 
\end{itemize}

Besides knowledge-based VQA that explicitly requires external knowledge to solve the tasks, there also exist models such as ERINE-ViL~\citep{yu2021ernie} and ROSITA~\citep{cui2021rosita} that use knowledge encoded inside the scene graphs to improve performance on standard VL tasks (\emph{e.g.}, VQAv2 and image-text retrieval). By pre-training on large-scale image-text data, the recent GIT work~\citep{wang2022git} shows that rich multimodal knowledge about the visual world has been encoded in the model weights, and the pre-trained model can readily recognize
scene text, tables/charts, food, logos, landmarks, characters, products, \emph{etc.}, and output these knowledge in natural language format when finetuned on the TextCaps dataset~\citep{sidorov2020textcaps}. A related survey on knowledge-intensive NLP tasks is \citet{yin2022survey}.

\subsection{Robustness and Probing Analysis} \label{sec:robustness}
In the majority of the VLP literature, models are evaluated on standard benchmarks such as VQAv2~\citep{goyal2017making}, image captioning, NLVR2~\citep{suhr2018corpus}, visual entailment~\citep{xie2019visual}, image-text retrieval, referring expression comprehension~\citep{yu2018mattnet}, \emph{etc}. These benchmarks have driven tremendous progress in the field (\emph{e.g.}, see Figure~\ref{fig:chp3_vqa_summary}), and some big VLP models have even surpassed human performance on some of these tasks. Although this progress is meaningful and exciting, we should not focus solely on topping the leaderboard, and should avoid both \emph{over-claiming} and \emph{under-claiming} the capabilities learned by the models (will be detailed in the discussion below). To date, it remains unclear how robust these pre-trained models are. In what follows, we review popular approaches to robustness analysis along multiple dimensions: ($i$) diagnostic tests; ($ii$) challenging sets that test out-of-distribution (OOD) generalization; ($iii$) human-adversarial attacks; and ($iv$) probing analysis.

\paragraph{Diagnostic Tests.} Diagnostic tests aim to verify one specific capability or one specific type of robustness of VLP models. For example,  \cite{li2020closer} has conducted a host of thorough evaluations of OD-based VLP models, including ($i$) robustness against \emph{linguistic variation} via VQA-Rephrasings~\citep{shah2019cycle}; ($ii$) robustness against \emph{logical reasoning} via VQA-LOL~\citep{gokhale2020vqa}; and ($iii$) robustness against \emph{visual content manipulation} via IV-VQA and CV-VQA~\citep{agarwal2020towards}.  CLEVR~\citep{johnson2017clevr} is a diagnostic dataset for testing compositional visual reasoning. GQA~\citep{hudson2019gqa} provides large-scale rule-based questions from ground-truth scene graphs of real-world images to test VQA model’s ability on positional reasoning and relational reasoning.
Winoground~\citep{thrush2022winoground} is a carefully curated dateset to probe VLP models' visio-linguistic compositionality on an image-text matching task. Furthermore, 
\cite{parcalabescu2020seeing} propose to test VL models on counting tasks.  The Visual Commonsense Tests (ViComTe) dataset~\citep{zhang2022visual} is created to test to what degree unimodal (language-only) and multimodal (image and language) models capture a broad range of visually salient attributes. VALSE~\citep{parcalabescu2021valse} is proposed to test VLP models centered on linguistic phenomena. CARET~\citep{jimenez2022carets} is proposed to systematically measure consistency and robustness of modern VQA models through  six fine-grained capability tests. 

\paragraph{Out-of-distribution Generalization.} VL models are typically assessed by measuring their performance on unseen data that comes from the same distribution as the training data. However, this assumption does not hold when deploying VL systems in practice. One of the most popular VQA datasets that is designed to test the out-of-distribution (OOD) generalization of VL models is VQA-CP~\citep{agrawal2018don}. It is constructed via reshuffling examples in VQAv2. GQA-OOD~\citep{kervadec2021roses} improves from VQA-CP and is based on the GQA dataset, and proposes to evaluate the performance differences between in-distribution and out-of-distribution split. Besides VQA, VLUE~\citep{zhou2022vlue} also creates OOD test sets for other VL tasks, including image-text retrieval, image captioning, and visual grounding. \cite{gupta2022grit} introduce the GRIT benchmark, which aims to test the performance, robustness, and calibration of a vision system across 7 vision and VL tasks, multiple data sources, and diverse concepts. Recently, \cite{agrawal2022rethinking} perform a comprehensive study on the OOD generalization capability of modern VLP models by conducting cross-dataset evaluations. 

\paragraph{Human-Adversarial Attacks.} To build a benchmark that can organically evolve over time, \cite{li2021adversarial,sheng2021human} introduce Adversarial VQA datasets that are collected iteratively via an adversarial human-and-model-in-the-loop procedure~\citep{nie2019adversarial}. Interestingly, they find that during dataset collection, non-expert annotators can easily attack modern VLP models successfully. These VLP models also achieve far worse performance on the new benchmark than on standard VQAv2 dataset. More recently, \cite{bitton2022winogavil} introduce WinoGAViL, which is an online game to collect VL associations, used as a dynamic benchmark to evaluate state-of-the-art VLP models. On one hand, these benchmarks are valuable as they successfully demonstrate the weaknesses of the SoTA VLP models, and shed new light on robustness studies in the community. On the other hand, we also need to be careful not to \emph{under-claim} the capabilities learned by the models, as these datasets are specially collected to fool these models.

\paragraph{Probing Analysis.} Besides testing VLP models on various benchmarks for robustness analysis, there also exists a line of work that aims to probe and understand what has been learned in the VLP models~\citep{cao2020behind,li2020does,salin2022vision}, such as cross-modal input ablation test~\citep{frank2021vision}, verb understanding~\citep{hendricks2021probing},  bias analysis~\citep{srinivasan2021worst}, the decoupling of the role of data, attention, and losses in VLP models~\citep{hendricks2021decoupling}, to name a few.

\subsection{VL for Language, Model Compression, Multilingual VLP, and Beyond}

\paragraph{VL for Language.} With the advent of VLP models like CLIP~\citep{radford2021learning} and ALIGN~\citep{jia2021scaling}, it has now been widely accepted that image-text data can be used to learn strong image encoders from scratch, and enable zero-shot image classification capabilities. On the other hand, human language is grounded in visual knowledge like colors, sizes, and shapes. A natural question to ask is whether image-text data can also help learn better language representations. Vokenization~\citep{tan2020vokenization} and its follow-up work iACE~\citep{lu2022imagination} propose to concatenate tokens and token-related images as vokens to enrich learned language representations. In VidLanKD~\citep{tang2021vidlankd}, the authors show that it is beneficial to use video-distilled knowledge transfer to improve language understanding tasks that involve world knowledge, physical reasoning, and temporal reasoning. Similarly, VaLM~\citep{wang2022visually} proposes to visually-augment text tokens with retrieved relevant images from CLIP~\citep{radford2021learning}, and use a visual knowledge fusion layer to enable multimodal grounded language modeling. VaLM shows substantial gains on object color and size reasoning, when compared with a text-only baseline.   

\paragraph{Model Compression.} Model compression, especially for Transformer-based models, is an important research topic in NLP. Methods such as knowledge distillation~\citep{sun2019patient,jiao2019tinybert,sun2020contrastive} and pruning~\citep{chen2020lottery} have been widely studied. In the context of VLP, MiniVLM~\citep{wang2020minivlm} investigates how to design a compact vision-language model, and proposes to pre-train an efficient low-cost object detector offline to replace the commonly used object detector as in BUTD~\citep{anderson2018bottom}, saving computation costs without decreasing performance much. DistilVLM~\citep{fang2021compressing} proposes to perform knowledge distillation for VLP. In the VL lottery ticket paper~\citep{gan2021playing}, the authors study the over-parameterization of VLP models via the lens of lottery ticket hypothesis. 

\paragraph{Efficient Adaptation.} In the NLP literature~\citep{liu2021pre}, techniques such as adapter~\citep{houlsby2019parameter}, prompt/prefix tuning~\citep{li2021prefix,lester2021power}, multi-task learning~\citep{liu2019multi}, and LoRA~\citep{hu2021lora} have been proposed for parameter-efficient adaptation of large language models for downstream tasks. By adding a few adapter layers into the Transformer backbone, or adding some learnable continuous prompt vectors, while freezing the pre-trained language model backbone, comparable or even better performance can be obtained in downstream tasks. The same idea has also been investigated in VL-Adapter~\citep{sung2022vl}, where a few adapter and prompt-tuning variants have been carefully tested. More recently, \cite{sung2022lst} propose ladder side-tuning for parameter and memory efficient transfer learning, which has also been applied to image-text tasks. 

\paragraph{Multilingual VLP.} Most VLP studies focus on English-only VL benchmarks, which leaves multilingual VLP a relatively less explored territory. To enable VLP models to support multiple languages, UC2~\citep{zhou2021uc2} and M3P~\citep{ni2021m3p} propose to add multilingual text encoders, and show how to use both English-only and multilingual data for joint pre-training. In MURAL~\citep{jain2021mural}, the authors pre-train a dual encoder that solves both image-text matching and translation pair matching tasks. By incorporating billions of translation pairs, MURAL~\citep{jain2021mural} extends ALIGN~\citep{jia2021scaling} to multilingual scenarios. More recently, in CCLM~\citep{zeng2022cross}, the authors introduce cross-view language modeling that unifies cross-lingual cross-modal pre-training using the ALBEF~\citep{li2021align} model architecture, and claim that CCLM is the first
multi-lingual multi-modal model that surpasses the translate-test performance of representative English VL models by zero-shot cross-lingual transfer. The most recent  model is PaLI~\citep{chen2022pali}, which can be considered a multilingual version of SimVLM~\citep{wang2021simvlm}. 

\paragraph{Unsupervised VLP.} Inspired by unsupervised machine translation, in~\cite{li2020unsupervised}, the authors investigate whether a strong VLP model can be learned without parallel image-text data. They propose to conduct masked-modeling-based pre-training on text-only and image-only data, and introduce object tags detected by an OD model to serve as anchor points to bridge the two modalities. \cite{zhou2022unsupervised} suggest that using tags alone is not sufficient, and propose to first construct a weakly
aligned image-text corpus via a retrieval-based approach,
then apply a set of multi-granular alignment pre-training
tasks to bridge the gap between the two
modalities. On one hand, exploring VLP under an unsupervised setting looks appealing and is a valid research problem; on the other hand, paired image-text data is actually not very difficult to collect and scale up, and there already exist many such large-scale image-text datasets as discussed in Section~\ref{sec:chp3_pretrain_data}, suggesting that unsupervised VLP may not be an urgent problem in practice. It would be interesting to investigate how image-only and text-only data can help improve the downstream performance of a model pre-trained on image-text data.  

\paragraph{Socratic Models.} Large foundation models are ubiquitous nowadays. Different models store different forms of knowledge across different domains. What if these large foundation models have a way to communicate and cooperate with each other? Will composing different foundation models in a zero-shot or few-shot manner enable new capabilities? To answer this, \cite{zeng2022socratic} propose the concept of Socratic Models, which use language as the representation by which inter-domain foundation models can jointly be used for inference. Several models belong to this category. For example, PICa~\citep{yang2021empirical} uses the cooperation of  VinVL (a SoTA image captioning model)~\citep{zhang2021vinvl} and GPT-3 for few-shot knowledge-based VQA. MAGIC~\citep{su2022language} uses a CLIP-induced score to regularize the language generation of GPT-2 so that the zero-shot generated caption is semantically related to the given image. BEST~\citep{xie2022visual} uses the cooperation of Florence~\citep{yuan2021florence} and GPT-3 for  visual storytelling and image paragraph captioning. \cite{wang2022language} propose the cooperation of CLIP, BLIP~\citep{li2022blip}, and GPT-3 for few-shot video-language learning. Flamingo~\citep{alayrac2022flamingo} uses a \emph{frozen} image encoder and a big \emph{frozen} language decoder, builds the connection between them via inserting cross-attention blocks, and performs large-scale pre-training. 

\paragraph{More Applications.} Besides VLP for standard VL tasks, VLP has also been applied to tackle ($i$) TextVQA~\citep{singh2019towards} and TextCaps~\citep{sidorov2020textcaps} tasks that require an AI system to comprehend scene text in order to perform VQA and captioning, such as TAP~\citep{yang2021tap} and LaTr~\citep{biten2022latr}; ($ii$) visual dialog~\citep{das2017visual} that requires an AI system to chat about an input image, such as VisDial-BERT~\citep{murahari2020large} and  VD-BERT~\citep{wang2020vd}; ($iii$) fashion-domain tasks, such as Kaleido-BERT~\citep{zhuge2021kaleido} and FashionVLP~\citep{goenka2022fashionvlp}; and ($iv$) vision-language navigation (VLN), such as PREVALENT~\citep{hao2020towards} and VLN-BERT~\citep{hong2021vln}, to name a few. A detailed literature review on VLN can be found in \cite{gu2022vision}.

\section{Text-to-Image Generation}
\label{sec:vlp4imggen}

Another important image-text task that is not covered in this chapter yet is Text-to-Image (T2I) generation, which aims to produce an image that correctly reflects the meaning of a textual description, and can be viewed as the inverse of image captioning \citep{chen2015microsoftcoco}. 

\begin{figure*}[t!]
  \centering
    \includegraphics[width=1.0\linewidth]{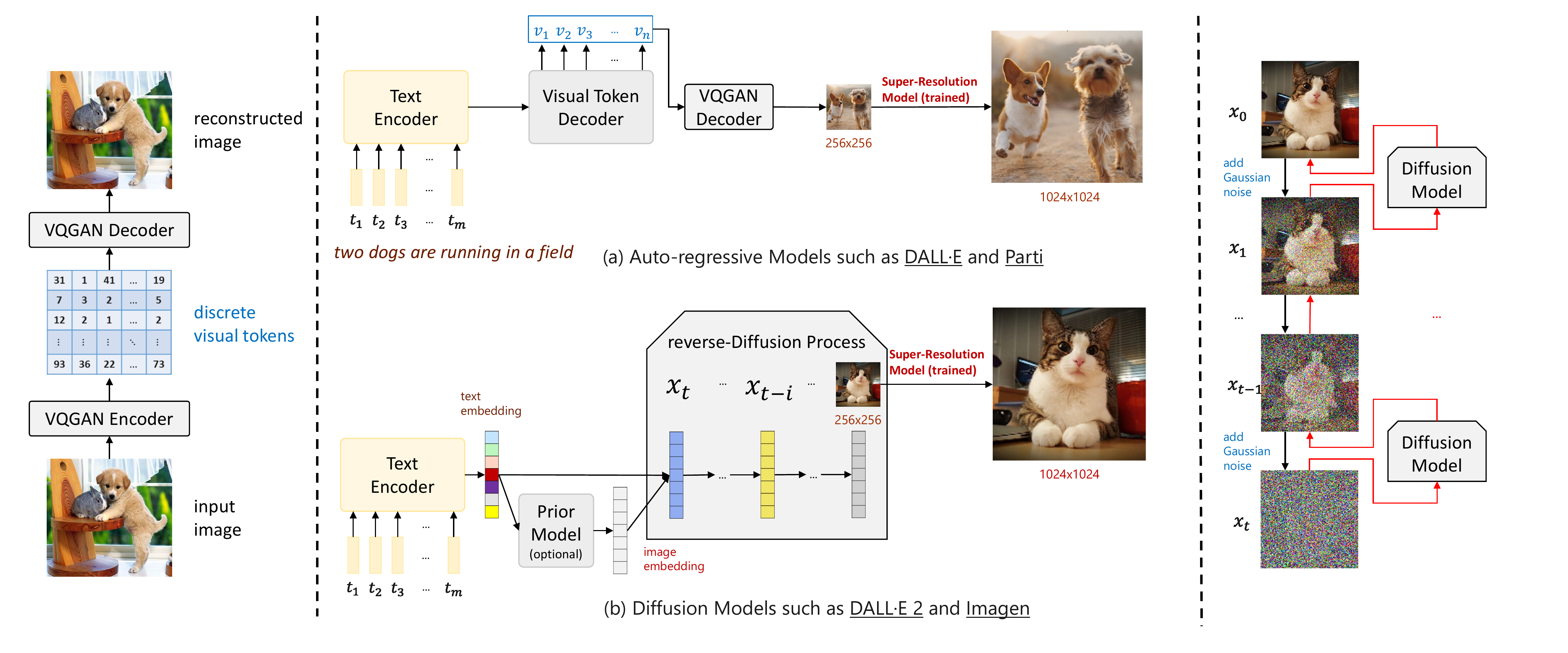}
  \caption{An illustration of discrete-token-based auto-regressive models (\emph{e.g.}, DALL-E~\citep{ramesh2021dalle} and Parti~\citep{yu2022scaling}) and diffusion-based models (\emph{e.g.}, DALL-E 2~\citep{ramesh2022hierarchical} and Imagen~\citep{saharia2022photorealistic}) for high-quality text-to-image generation with large-scale pre-training. Figure credit: from Chenfei Wu.}
  \label{fig:chp3_t2i_models}
\end{figure*}

\paragraph{Before VLP.}
As the pioneering work in T2I generation, \cite{mansimov2016GeneratingICLRl} shows that recurrent variational auto-encoder could generate novel visual scenes conditioned on image captions; however, the generated image quality is not satisfactory. Research on T2I generation was then greatly advanced with the prosperity of generative adversarial networks (GANs). \cite{reed2016Generative} extended conditional GANs to T2I generation, and has been shown to work on restricted datasets (\emph{e.g.}, Oxford-102 Flowers and CUB-200 Birds) with relatively small image resolutions (64x64). In recent years, this field has made remarkable progress thanks to the improved multimodal encoding (\emph{e.g.}, StackGAN~\citep{zhang2017stackgan}, StackGAN++~\citep{zhang2018stackgan++}), novel attention mechanisms (\emph{e.g.}, AttnGAN~\citep{xu2018attngan}, SEGAN~\citep{tan2019semantics}, ControlGAN~\citep{li2019controllable}), the use of cycle structure (\emph{e.g.}, MirrorGAN~\citep{qiao2019mirrorgan}), \emph{etc}. 

To extend the success of GANs to limited-data regime, it is common to use pre-training, \emph{i.e.}, initializing the optimization process by pre-trained GAN models on some large datasets~\citep{grigoryev2022and}. However, most GAN-based pre-training is conducted on image datasets only, which did not leverage image-text pairs used for vision-language pre-training (VLP), except for recent work using the CLIP model in GAN-based methods such as LAFITE \citep{zhou2022LAFITE}, which demonstrates the first work on training T2I generation models without using text data explicitly. 

\paragraph{In the Context of VLP.}
While GAN-based methods are still popular for image synthesis, there is a new paradigm shift for T2I generation. In the context of VLP, we classify these methods into two categories: ($i$) VQ-token-based auto-regressive methods (\emph{e.g.}, DALL-E~\citep{ramesh2021dalle} and Parti~\citep{yu2022scaling}), and ($ii$) diffusion-based methods (\emph{e.g.}, DALL-E 2~\citep{ramesh2022hierarchical} and Imagen~\citep{saharia2022photorealistic}). An illustration of these methods is provided in Figure~\ref{fig:chp3_t2i_models}. Below, we provide a brief review on these recent works.

\subsection{VQ-token-based Auto-regressive Methods}
\paragraph{Discrete Token Representation.} 
In 2017, VQ-VAE~\citep{van2017neural} was proposed, which provides a simple yet powerful generative model that learns discrete representations for high-quality image reconstruction. Later on, in VQ-VAE-2~\citep{razavi2019generating}, researchers show that high-fidelty and high-resolution images can be generated. 
With the prevalence of Transformer model \citep{vaswani2017attention} which have achieved impressive improvements in domains such as language models \citep{devlin2018bert} and image generative pre-training~\citep{chen2020generative}, the modeling of VQ token sequence is also naturally handled by Transformer~\citep{esser2021taming}.

\paragraph{Auto-regressive Modeling.}
These recent advances fueled by Transformer suggest a possible route for T2I generation and the potential to benefit from large-scale VLP. Specifically, DALL-E~\citep{ramesh2021dalle} demonstrates that training a large-scale auto-regressive Transformer on numerous image-text pairs can result in a high-fidelity generative model with controllable synthesis results through text prompts. NUWA \citep{wu2021n} presents a unified multimodal pre-trained model that allows to generate or manipulate visual data (\emph{i.e.}, images and videos) with a 3D transformer encoder-decoder framework and a 3D Nearby Attention (3DNA) mechanism. In NUWA-Inifinity~\citep{wu2022nuwa}, the authors further propose an autoregressive over autoregressive generation method for high-resolution infinite visual synthesis, which is capable of generating images of arbitrary aspect ratio. Parti~\citep{yu2022scaling} adopts a similar Transformer-based encoder-decoder architecture, and trains the model in scale, and demonstrates impressive image generation results. Make-A-Scene~\citep{gafni2022make} proposes to use an additional segmentation map (can be generated or not) as additional input to further aid the image generation process.
Other examples include CogView~\citep{ding2021cogview} and CogView2~\citep{ding2022cogview2}, which is also similar to DALL-E~\citep{ramesh2021dalle}.

\begin{figure*}[t!]
  \centering
    \includegraphics[width=1.0\linewidth]{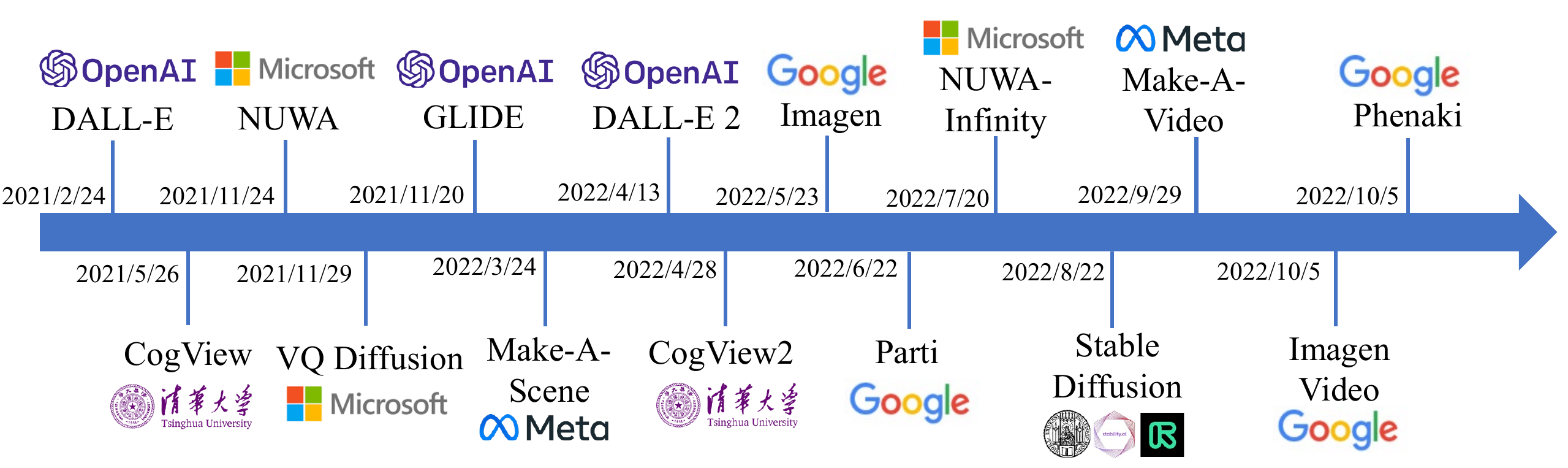}
  \caption{Auto-regressive and diffusion-based text-to-image/video models developed over time. Only some representative works are shown.}
  \label{fig:chp3_t2i_along_time}
\end{figure*}

\paragraph{Bi-Directional Image-Text Generation.}
ERINE-ViLG~\citep{zhang2021ernie}, L-Verse~\citep{kim2021verse} and OFA~\citep{wang2022unifying} demonstrate that large-scale generative joint pre-training for both text and image tokens (from VQ-VAE) can be finetuned on diverse downstream tasks, such as style learning (domain-specific text-to-image), super-resolution (image-to-image), image captioning (image-to-text), and even text-image retrieval, \emph{etc}.

\subsection{Diffusion-based Methods} 
\paragraph{Continuous Diffusion.}
Recently, diffusion models such as denoising diffusion probabilistic models (DDPM)~\citep{ho2020denoising} have achieved great successes in image generation tasks. Recent works~\citep{dhariwal2021diffusion} have demonstrated even higher quality image synthesis compared to VQ-token-based models and GANs. Furthermore, a recent denoising diffusion implicit model (DDIM)~\citep{song2020denoising} further accelerates the sampling procedure and enables nearly perfect inversion. 
We refer the readers to \citet{yang2022diffusion} for a comprehensive survey of diffusion models.

To extend diffusion-based methods for T2I generation, GLIDE \citep{nichol2021glide} adopts \emph{continuous} diffusion, and compares CLIP guidance and classifier-free guidance in diffusion models, and concludes that a diffusion model of 3.5 billion parameters with classifier-free guidance outperforms DALL-E in terms of human evaluation. More recently, DALL-E 2~\citep{ramesh2022hierarchical}, Imagen~\citep{saharia2022photorealistic} and Stable Diffusion (a scaled-up version of Latent Diffusion~\citep{rombach2022high}) have pushed this line of work to a new level, especially due to the open-source efforts of Stable Diffusion. Instead of performing diffusion in the pixel space as in DALL-E 2~\citep{ramesh2022hierarchical} and Imagen~\citep{saharia2022photorealistic}, the Latent Diffusion model~\citep{rombach2022high} proposes to perform diffusion in the continuous latent space instead.

\paragraph{Discrete Diffusion.}
By combining VQ-token-based and diffusion-based methods, recent works such as ImageBART~\citep{esser2021imagebart} and VQ-Diffusion~\citep{gu2021vector} propose to model the latent \emph{discrete} code space of a VQ-VAE \citep{razavi2019generating} by learning a parametric model using a conditional variant of DDPM, for the task of T2I generation.

\paragraph{Text-to-Video Generation.}
The field is progressing at a rapid speed. Not just satisfied at text-to-image generation, recent works, such as Make-A-Video~\citep{singer2022make}, Imagen Video~\citep{ho2022imagen}, and Phenaki~\citep{villegas2022phenaki}, have significantly lifted the quality of text-to-video generation to a new level. 


\chapter{VLP for Core Vision Tasks}
\label{chp:vlp4vision}

Computer vision has become ubiquitous in our society, with applications in visual search, image understanding, mapping, medicine, and self-driving cars. Core to many of these applications are visual recognition tasks such as image classification and object detection. The primary goal of these tasks is to assign a semantically meaningful concept to the visual instance such as images or regions. Traditional computer vision systems are trained to predict a fixed set of predetermined concepts, such as the image class labels on ImageNet~\citep{deng2009imagenet}/JFT300M, the object categories on COCO~\citep{lin2014microsoft}, and so on. Although close-to-human performance has been reported on these tasks, the restricted form of a close-set of concepts limits models' generality and usability, since additional labeled data is needed to specify semantic concepts that are unseen in training data. In this chapter, we describe how recent advances in VLP tackle the core visual recognition problems. Section~\ref{sec:vlp4vision_overview} provides the overview rational on the paradigm shift. This is exemplified by the three vision problems, including image classification  in Section~\ref{sec:vlp_vision_image_classification}, object detection in Section~\ref{sec:vlp_vision_od} and image segmentation in Section~\ref{sec:vlp_vision_segmentation}. Section~\ref{sec:vlp_vision_trend} outlines the trend of computer vision in the wild, and Section~\ref{sec:vlp_vision_advanced_topics} summarizes the chapter with a discussion on advanced topics.

\section{Overview}
\label{sec:vlp4vision_overview}

Recent state-of-the-art computer vision systems are trained from free-form natural language supervision, ranging from simple object category names to descriptive captions. These language-augmented visual models have shown strong transfer ability. We believe that two following factors contribute to the paradigm shift.

\begin{minipage}{1.0\textwidth}
\centering

\begin{enumerate}[label=(\arabic*),leftmargin=5.5mm]
\item  {\it Open-set recognition is enabled due to the problem reformulation from classification to retrieval.}  Traditional classification formulation defines and learns a fixed set of embedding vectors, each of which represents an object category. It is infeasible for the models to predict and transfer beyond this close-set of concepts. The alternative is to cast image classification as an image-to-text retrieval task, where one searches for an image (or regions in the image) the matched concepts. Parametric models such as neural nets are employed to encode both images and language (concepts) and perform dense retrieval to retrieve an image from its relevant concepts.

\item {\it Model generality and usability is improved since the form of language supervision allows a wide range of visual concepts to be represented.} 
The fixed set of visual concepts is an over-simplified representation of visual concepts, due to the compactness requirement in a classification head. In contrast, the newly introduced text encoder in the retrieval formulation is capable of dealing with a much larger concept pool. Natural language is semantically richer than any set of concept labels (\emph{e.g.,} object categories).
The text sequence form of language also allows to represent external knowledge (\emph{e.g.}, from WordNet and Wikipedia) in the same format as image captions and concept labels, further boosting the concept coverage. 

\end{enumerate}

\end{minipage}

In this chapter, we illustrate the paradigm shift by presenting case studies on three prominent computer vision tasks, image classification (IC), object detection (OD), and segmentation. We review UniCL~\citep{yang2022unicl}, CLIP~\citep{radford2021learning}, ALIGN~\citep{jia2021scaling} for IC,  ViLD~\citep{gu2021zero}, RegionCLIP~\citep{zhong2021regionclip}, GLIP~\citep{li2021grounded} for OD, and LSeg~\citep{li2022language}, OpenSeg~\citep{ghiasi2021open}, DenseCLIP~\citep{rao2021denseclip} for image segmentation.

\begin{figure*}[t!]
  \centering
    \includegraphics[width=1.0\linewidth]{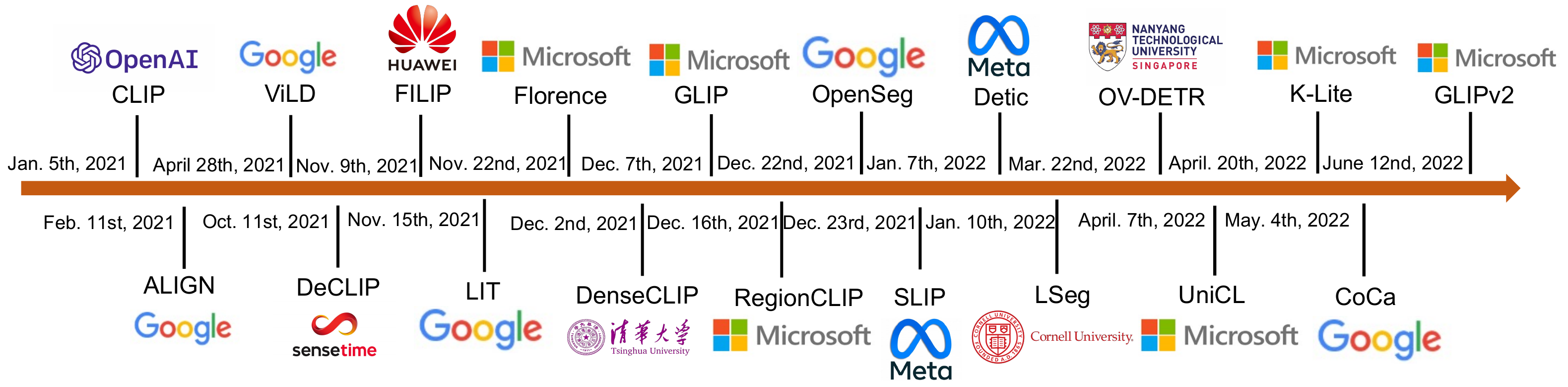}
  \caption{VLP models developed for core computer vision problems along time. Due to space constraint, only some representative works are shown. This figure was created by Jianwei Yang in CVPR 2022 Tutorial.}
  \label{fig:chp4_timeline}
\end{figure*}

We present a glossary of representative VLP models in Table~\ref{tab:chp4_core_vision_glossary}, where models are described along multiple dimensions. 
In Figure~\ref{fig:chp4_timeline}, we show how these VLP models evolve along time. %
This line of research equips computer vision models with the capability of open-set visual recognition, opening the possibilities of building generalizable computer vision systems with a strong task-level transfer ability, and thus paving the way towards {\it Computer Vision in the Wild (CVinW)}\footnote{\url{https://computer-vision-in-the-wild.github.io}}~\citep{li2022elevater}.


\begin{table*}[!t]
\resizebox{1.0\textwidth}{!}
{
  \begin{tabular}{l|c|ccc|c}
    {\bf Model} & \bf  {\bf Problems}  & \multicolumn{3}{c|}{\bf Pre-training Data}    &    \bf Pre-training Objectives \\ \cline{3-6} 
    \hline
        &    & \bf Image-Text &  \bf Image-label  & \bf Knowledge     & \\        
  CLIP~\citep{radford2021learning}  &\multirow{8}{*}{IC} & WebIT (400M) &  &   & \multirow{2}{*}{ITC}\\
  ALIGN~\citep{jia2021scaling}  &  & WebIT (1.8B) &  &      &\\ \cdashline{6-6}
  FILIP~\citep{yao2021filip}  &  &  YFCC15M, WebIT (300M) &  &     & ITC+WRA\\
  DeCLIP~\citep{li2021supervision}  &  & YFCC15M, WebIT (88M)  &  &      & ITC+SSL+TP\\  
  SLIP~\citep{mu2021slip}  &  & YFCC15M &  &   &    ITC+SSL \\ \cdashline{4-4}\cdashline{6-6}
  Florence~\citep{yuan2021florence}  & &  WebIT 500M & Image-tag 300M &      & \multirow{3}{*}{UniCL} \\   
  UniCL~\citep{yang2022unicl}  &  & CC15M, YFCC15M  &  ImageNet-21K &      &\\   \cdashline{5-5}
  K-Lite~\citep{shen2022k}  &  & CC15M, YFCC15M & ImageNet-21K & Wiktionary, WordNet &    \\  
 \hline
         &    & \bf Image-Text &  \bf OD Annotation & \bf Phrase grounding &   \\   
  ViLD~\citep{gu2021zero}  &\multirow{8}{*}{OD} &  &  LVIS (120K) &   &   \multirow{2}{*}{ITC+ distillation} \\
  RegionCLIP~\citep{zhong2021regionclip}  &  & CC3M &  &   &    \\  \cdashline{6-6}
  GLIP~\citep{li2021grounded}   &  & CC3M+12M,SBU & Combined OD &  Gold-G &     Object-Phrase Alignment\\  
Detic~\citep{zhou2022detecting}  &  & ImageNet & LVIS &   &   Non-prediction-based Loss   \\  
PromptDet~\citep{feng2022promptdet} & & LAION-400M & LVIS &   &   Regional Prompt Learning  \\  
OWL-ViT~\citep{minderer2022simple}   &  & WebIT (3.6B)  & Combined OD (2M) &   &   Bipartite Matching Loss   \\  
OV-DETR~\citep{zang2022open} &  & & COCO / LVIS &   &   Binary Matching Loss \\  
X-DETR~\citep{cai2022x} & & CC,SBU, LocNar & COCO & Gold-G  &  Object-Language Alignment   \\  
   \hline
           &    & \bf Image-Text &  \bf Seg Annotation  & \bf Phrase grounding  & \\   
  LSeg~\citep{li2022language}  &\multirow{6}{*}{Seg} &   & COCO/VOC &\multirow{3}{*}{}  & Mask Prediction + Grounding  \\
  OpenSeg~\citep{ghiasi2021open}  &  & LocNar & COCO  &     & Word-Pixel Matching \\  
  CLIPSeg~\citep{luddecke2022image}  &  & &   &  PhraseCut  & Visual Prompt Engineering\\ 
  MaskCLIP~\citep{zhou2021maskclip}   &  & &  COCO/VOC &    &  Pseudo Label Distillation\\  
  DenseCLIP~\citep{rao2021denseclip}   &  & & ADE20K/COCO &     & Pixel-Text Matching\\  
  GroupViT~\citep{xu2022groupvit}   &  & CC12M/YFCC15M &  &     & Multi Label Contrastive\\    
  \end{tabular}
  }
  \caption{\textbf{Glossary of representative VLP models for core vision tasks}. For data scale, we report \# image-text pairs, including both image-label and image-caption. IC: image classification. OD: object detection.
  LocNar: Localized Narratives.
  Golden-G is the mixed golden ground-truth grounding data processed in MDETR~\citep{kamath2021mdetr}.
  ITC: image-text contrastive learning. WRA: word-region alignment. TP: Token Prediction. SSL: Self-supervised learning.}
  \label{tab:chp4_core_vision_glossary}
\end{table*}

\section{VLP for Image Classification}
\label{sec:vlp_vision_image_classification}

We first define a triplet-wise data format $\Scal  = \{ (\xv_n, \tv_n, y_n) \}_{n=1}^N$, where $\xv \in \Xcal$  is the image, and $\tv \in \Tcal$ is its corresponding language description (ranging from simple tokens such as category names to free-form text sequences such as captions), and $y \in \Ycal$ is a label indicating the index of the grouped or unique language description in the dataset. Given these triplet data instances, our goal is to learn generic and rich visual-semantic representations, so that an image $\xv$ is predicted to correctly align with its language description $\tv$, \emph{i.e.}, image classification.  

\begin{figure*}[t!]
  \centering
    \includegraphics[width=0.9\linewidth]{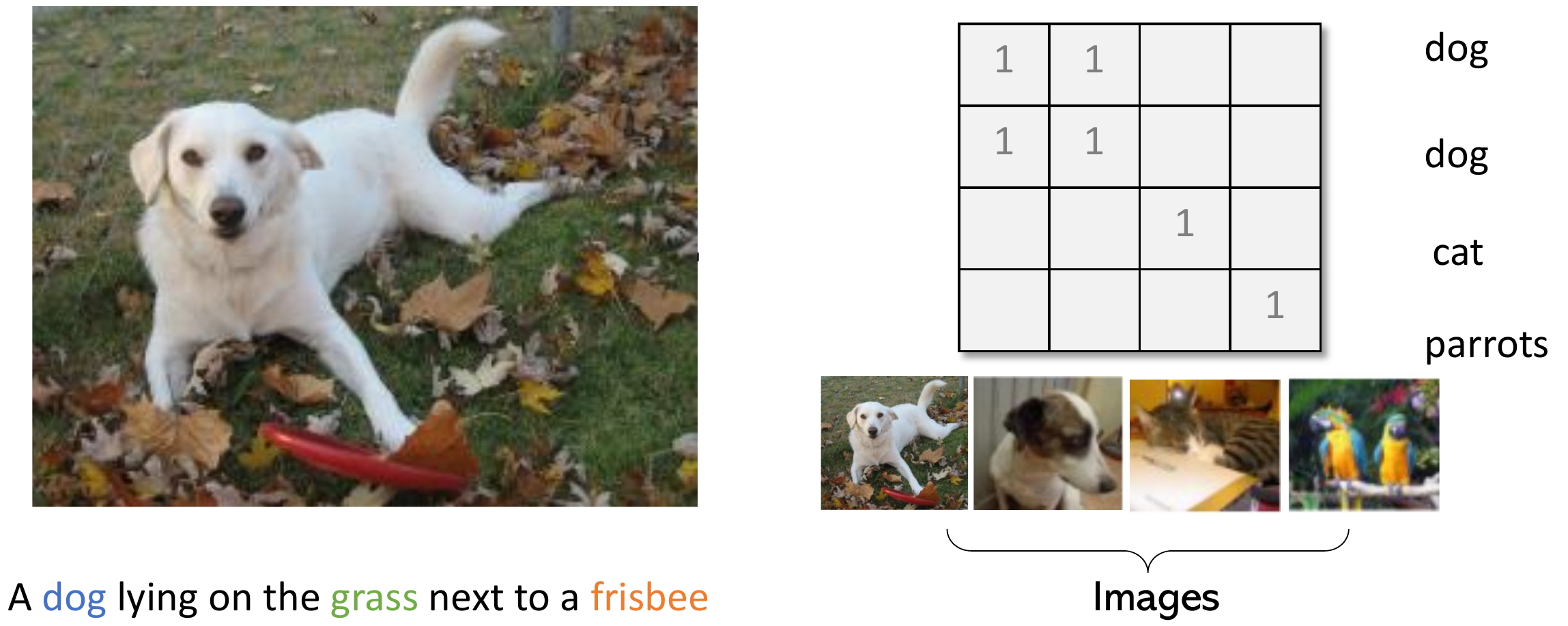}
  \caption{Image-caption matching for image classification.}
  \label{fig:chp4_ic}
\end{figure*}

For each image $\xv$, an image encoder model $f_{\thetav}$ parameterized by $\thetav$ first represents $\xv$ as a visual feature vector $ \Tilde{\vv}  \in \R^{P\times 1}$: $ \Tilde{\vv} = f_{\thetav}(\xv)$. For each language description $\tv \in \Tcal$, we encode it with a text encoder $f_{\phiv}(\tv)$ parameterized by $\phiv$ to get its feature vector $ \Tilde{\uv}  \in \R^{P \times 1}: \Tilde{\uv}  = f_{\phiv}(\tv)$. Note that $ \Tilde{\vv}$ and $ \Tilde{\uv} $ is the vector representation of the entire image and sentence, respectively.
For $i$-th image $\xv_i$ and $j$-th language description $\tv_j$ in a batch $\Bcal$, we normalize their feature vectors in a hyper-sphere using $ \uv_i = \frac{    f_{\thetav}(\xv_i)  }{   \| f_{\thetav}(\xv_i)  \|} $ and $ \vv_j = \frac{  f_{\phiv}(\tv_j)   }{ \| f_{\phiv}(\tv_j)\| }  $, and their similarity is calculated as $s_{ij}  = \uv_i^{T} \vv_j  $. Figure~\ref{fig:chp4_ic} shows an example image-text pair and a batch of four image-text pairs.

\paragraph{UniCL.} A bidirectional superivsed contrastive learning objective is defined based on the matching between images and language descriptions~\citep{yang2022unicl}:
%
\begin{align}\label{eq:obj_unicl}
\vspace{2mm}
	\min_{ \{ \thetav, \phiv \} } ~~ \Lcal_{\text{UniCL}} 	= & \Lcal_{i2t} + \Lcal_{t2i}, \text{with} 
	\\
	\label{eq:obj_i2t_label_unicl}
	\Lcal_{i2t}	= & - \sum_{ i \in \Bcal } \frac{1}{ |\Pcal(i)|  }  \sum_{ k \in \Pcal(i) }
\log \frac{ \exp(\tau \uv_{i}^T \vv_k)  }{\sum_{ j \in \Bcal}  \exp(\tau \uv_{i}^T \vv_{j})  } \,,
    \\
    \label{eq:obj_t2i_label_unicl}
	\Lcal_{t2i}	= & - \sum_{ j \in \Bcal } \frac{1}{ |\Qcal(j)|  }  \sum_{ k \in \Qcal(j) }
\log \frac{ \exp(\tau \uv_{k}^T \vv_j )  }{\sum_{ i \in \Bcal}  \exp(\tau \uv_{i}^T \vv_{j} )  }\,,
\end{align}
where $\Pcal(i) = \{ k | k \in \Bcal, y_k = y_i\}$ and $\Qcal(j) = \{ k | k \in \Bcal, y_k = y_j\} $, and $\tau$ is a temperature hyper-parameter, controlling the strength of penalties on hard negative samples. In Figure~\ref{fig:chp4_ic}, there are two images that share the same language/concept ``dog'', the corresponding elements in the target matrix for contrastive learning is labelled as positive, based on UniCL formulation. By scaling up UniCL to 800M training samples, it leads to Microsoft Florence model~\citep{yuan2021florence}, which leads to the SoTA performance on many tasks by then. 

\paragraph{CLIP/ALIGN.} 
CLIP~\citep{radford2021learning} and ALIGN~\citep{jia2021scaling} assume that there are only one-to-one mappings between an image and its paired caption in a batch, \emph{i.e.,} $ \Pcal(i) = \{i\}$ and $ \Qcal(j) = \{ j \} $. The CLIP training objective is
\begin{align}\label{eq:obj_clip}
\vspace{2mm}
	\min_{ \{ \thetav, \phiv \} } ~~ \Lcal_{\text{CLIP}} 	= & \Lcal_{i2t} + \Lcal_{t2i}, \text{with} 
	\\
	\label{eq:obj_i2t_label_clip}
    \Lcal_{i2t}	= & - \sum_{ i \in \Bcal } 
\log \frac{ \exp(\tau \uv_{i} \vv_i )  }{\sum_{ j \in \Bcal}  \exp(\tau \uv_{i} \vv_{j} )  }\,,
    \\
    \label{eq:obj_t2i_label_clip}
	\Lcal_{t2i}	= & - \sum_{ j \in \Bcal } 
\log \frac{ \exp(\tau \uv_{j} \vv_j )  }{\sum_{ i \in \Bcal}  \exp(\tau \uv_{i} \vv_{j} )  }\,.
\end{align}
For the example in Figure~\ref{fig:chp4_ic}, CLIP or ALIGN only considers the on-diagonal elements as positive, and all off-diagonal elements as negative. Ideally,  CLIP or ALIGN should be applied to image-text pairs without duplication in either modality.

\paragraph{Connections to Traditional Classification Formulation.} 
Note that $\Lcal_{\text{UniCL}}$  in  \eqref{eq:obj_unicl} is closely related to the standard cross-entropy loss used in supervised image classification problems. Specifically, the image-to-language contrastive term in~\eqref{eq:obj_i2t_label_unicl} recovers cross-entropy as a special case, when the following three conditions are satisfied.  
$(i)$ The text encoder $f_{\phiv}$ is represented as a simple linear embedding layer $\Wmat$. 
$(ii)$ The batch size $|\Bcal|$ is sufficiently larger than the number of classes $K$, so that all the class embedding vectors are used in contrastive learning, when stochastic sampling is used for training. $(iii)$ $\tau=1$, and $\ell_2$ normalization is excluded, so that  $\Tilde{\uv} = \uv$ and   $\Tilde{\vv}=\vv$.  In practice, all of these conditions can be easily satisfied, and~\eqref{eq:obj_i2t_label_unicl} becomes 
\begin{align}\label{eq:objective_ce}
\min_{ \{ \thetav, \Wmat \} } ~~ \Lcal_{\text{IC}} =   \sum_{i \in \Bcal}
\log \frac{ \exp(\wv_{\hat{y}}  \Tilde{\vv}_i )  }{\sum_{k=1}^K \exp(\wv_{k} \Tilde{\vv}_i )  }\,,
\end{align}
where $\hat{y}$ is the ground-truth label for the $i$-th image in the batch.

\paragraph{Other Language-Image Pre-training Methods for IC.} Learning a vision backbone from web-scale image-text pairs is an emerging research topic. There are an increasing number of papers recently, aiming to improve zero-shot/few-shot performance of IC in the wild.

\begin{itemize}[leftmargin=3.0mm]

\item {\bf Improved Contrastive Pre-training Objectives.} FILIP~\citep{yao2021filip} bootstraps the fine-grained region-word correspondences. PyramidCLIP~\citep{gao2022pyramidclip} constructs an input pyramid with different semantic levels, and aligns two modalities in the form of hierarchy via both intra and cross-level alignment. Prefix conditioning~\citep{saito2022prefix} introduces the use of  prefixed prompt to combine image-caption and image-label data, based on the data type. CyCLIP~\citep{goel2022cyclip} shows that consistent representations can be learned by explicitly symmetrizing the similarity between the two mismatched image-text pairs (cross-modal consistency), and the similarity between the image-image pair and the text-text pair (in-modal consistency). 

\item {\bf Self-supervised + Contrastive Objectives.} DeCLIP~\citep{li2021supervision} comprehensively investigates multiple single-modality self-supervision signals in image-text pairs.  SLIP~\citep{mu2021slip} studies the integration of image-to-image self-supervised learning and image-to-text contrastive learning. Masked image/language modeling is also combined with image-to-text contrastive learning, such as  MultiMAE~\citep{bachmann2022multimae} and M3AE~\citep{geng2022multimodal}.

\item {\bf Frozen Models.} LiT~\citep{zhai2021lit} introduces the ``contrastive-tuning'' method, showing that locking the pre-trained image encoder and tuning text encoder works
best for zero-shot transfer. Flamingo~\citep{alayrac2022flamingo} leverages pre-trained models from each single modality, and continues to pre-train the cross-modal module to achieve impressive image classification performance using in-context learning.

\item {\bf Scaling.} Due to the promising results of web-scale pre-training for computer vision tasks, there is a trend to explore the scaling success of VLP models.  BASIC~\citep{pham2021combined} is proposed to scale up the contrastive learning framework of CLIP and ALIGN in three dimensions: data size, model size, and batch size, achieving 85.7\% zero-shot accuracy on ImageNet. LIMoE~\citep{mustafa2022multimodal} is a sparse mixture of experts model capable of language-image multimodal learning.
Pathways Language and Image model (PaLI)~\citep{chen2022pali} finds that joint scaling of the vision and language components is important. Since existing Transformer language models are much larger than their vision counterparts, PaLI trains the largest ViT to date to quantify the
benefits from even larger-capacity vision models, based on large multilingual mix of pre-training task and a new image-text training set containing 10B images and texts in over 100 languages.

\end{itemize}

In the literature, there are two different experiment settings to evaluate the open-set image classification ability of pre-trained models.

\begin{itemize}[leftmargin=3.0mm]

\item {\bf Class-level Transfer in a Single Domain.} The traditional zero-shot transfer evaluation for image classification has been studied for decades, where a manual split is pre-defined in a given visual domain, ensuring that evaluation concepts are not observed in training. Examples include Animal with Attributes (AwA)~\citep{lampert2013attribute}, Caltech-UCSD Birds-200 (CUB)~\citep{wah2011caltech}, SUN~\citep{patterson2012sun}, aPY~\citep{farhadi2009describing}, and ZS-ImageNet~\citep{rohrbach2011evaluating,fu2016semi}. 
 
\item {\bf Task-level Transfer.} To demonstrate the strong usability and generality of CLIP, \cite{radford2021learning} directly apply the pre-trained checkpoint to recognize any concepts in around 30 public image classification datasets in the community. Impressive results are reported, though the model has ``never observed'' the images from these downstream datasets. It quickly popularizes the zero-shot task transfer evaluation for computer vision foundation models. 
Many CLIP variants~\citep{li2021supervision,gao2022pyramidclip,yang2022unicl} are proposed. But these works perform evaluation using different 
downstream datasets, making their results not comparable.
The recent {\it Image Classification in the Wild (ICinW)} benchmark is an attempt to formalizing the task-level evaluation with 20 public datasets~\citep{li2022elevater}.

\end{itemize}

\begin{figure*}[t!]
  \centering
    \includegraphics[width=0.99\linewidth]{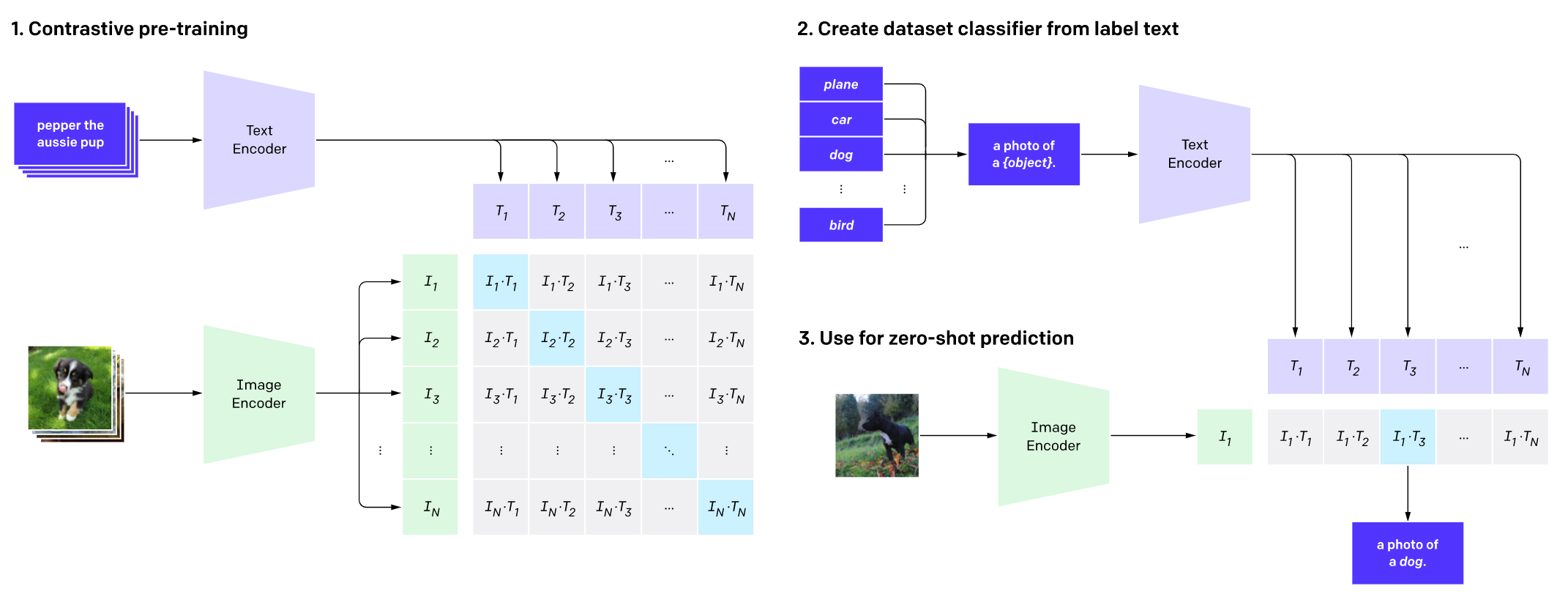}
    \includegraphics[width=0.99\linewidth]{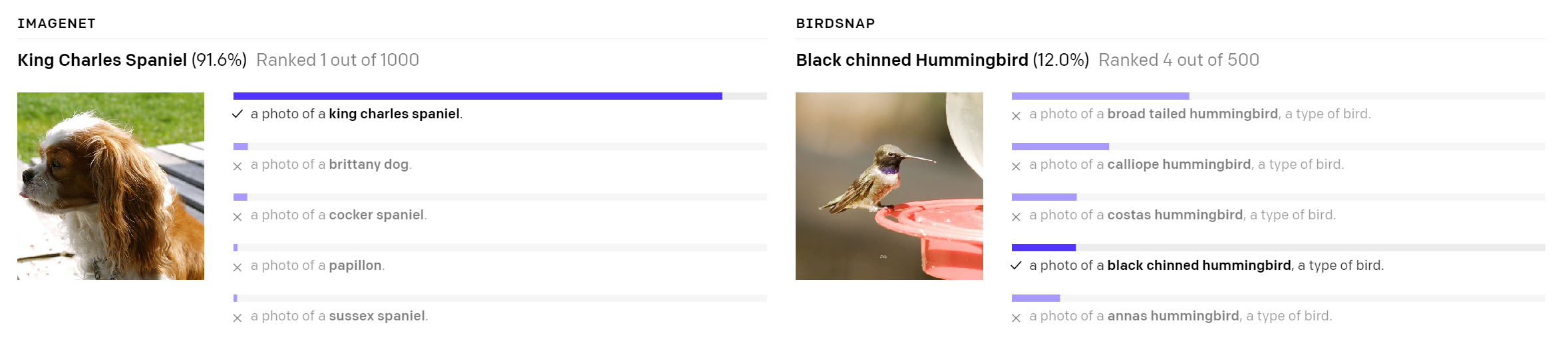}
    \includegraphics[width=0.99\linewidth]{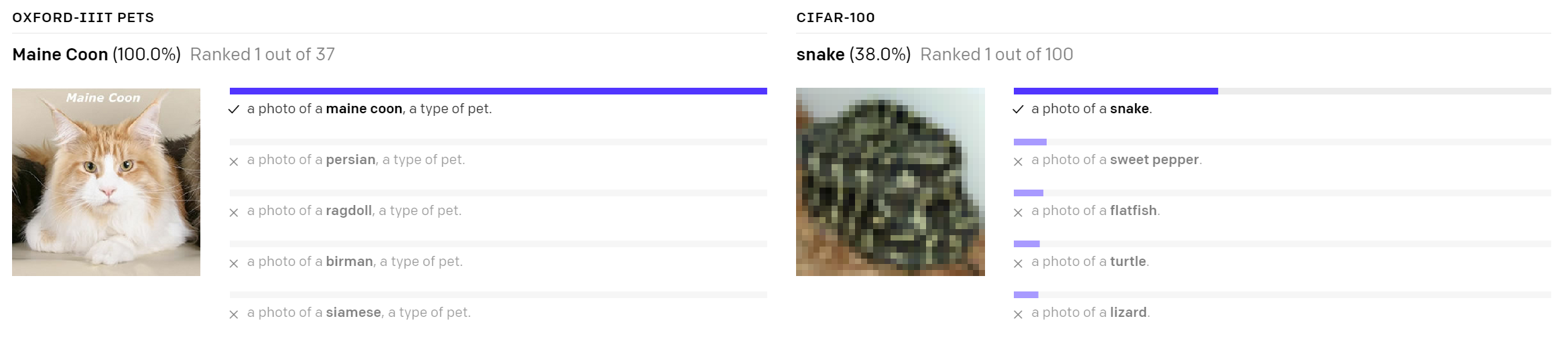}
  \caption{{\bf Top}: CLIP pre-trains an image encoder and a text encoder to predict which images are paired with which texts in a dataset/batch. This behavior allows us to turn CLIP into a zero-shot classifier. We convert all the classes into captions such as ``a photo of a dog'' and predict the class of the caption CLIP estimates best pairs with a given image. {\bf Bottom}: predictions of zero-shot CLIP classifiers on examples from four datasets. This figure was created in~\cite{radford2021learning}.}
  \label{fig:chp4_clip}
\end{figure*}

\paragraph{Use Cases of Language-Image Models for IC.} In Figure~\ref{fig:chp4_clip}, we illustrate how the image-text contrastive-trained model like CLIP can be used for zero-shot image classification. Given a new IC dataset/task with a set of concept/category names, each concept is converted into caption by augmenting it with various text templates. The caption is used as prompt for the text encoder to extract concept representations. The query image is fed into the image encoder to extract the visual representation, which is used to compute the similarity with respective to all concepts. The one with highest similarity yields the predicted concept. In the bottom of Figure~\ref{fig:chp4_clip}, four cases are illustrated, one is from ImageNet and others are from ICinW that represent real-world IC scenarios. 


\section{VLP for Object Detection}
\label{sec:vlp_vision_od}
A typical object detection task contains two sub-tasks. 
$(i)$ {\it Localization} aims to locate the presence of objects in an image and indicate the position with a bounding box.
$(ii)$ {\it Recognition} determines what object categories are present in the region of interest (or bounding box). 
The recognition task is similar to the image classification task (Section~\ref{sec:vlp_vision_image_classification}), except that classification is performed on the entire image in IC but on individual regions/boxes in OD. Therefore, by following the reformulation that converts classification to retrieval as described in Section~\ref{sec:vlp_vision_image_classification}, one may improve OD models' transfer ability for open-set recognition. 
Specifically, each region/box feature is fed into two prediction heads, \emph{i.e.}, a box classifier and a box regressor, which are trained with the classification loss $\Lcal_{\text{cls}}$ and the localization loss $\Lcal_{\text{loc}}$, respectively:
\begin{equation}\label{eqn:loss}
    \Lcal_{\text{OD}} = \Lcal_{\text{cls}} + \Lcal_{\text{loc}}\,.
\end{equation}

\begin{figure*}[t!]
  \centering
    \includegraphics[width=0.9\linewidth]{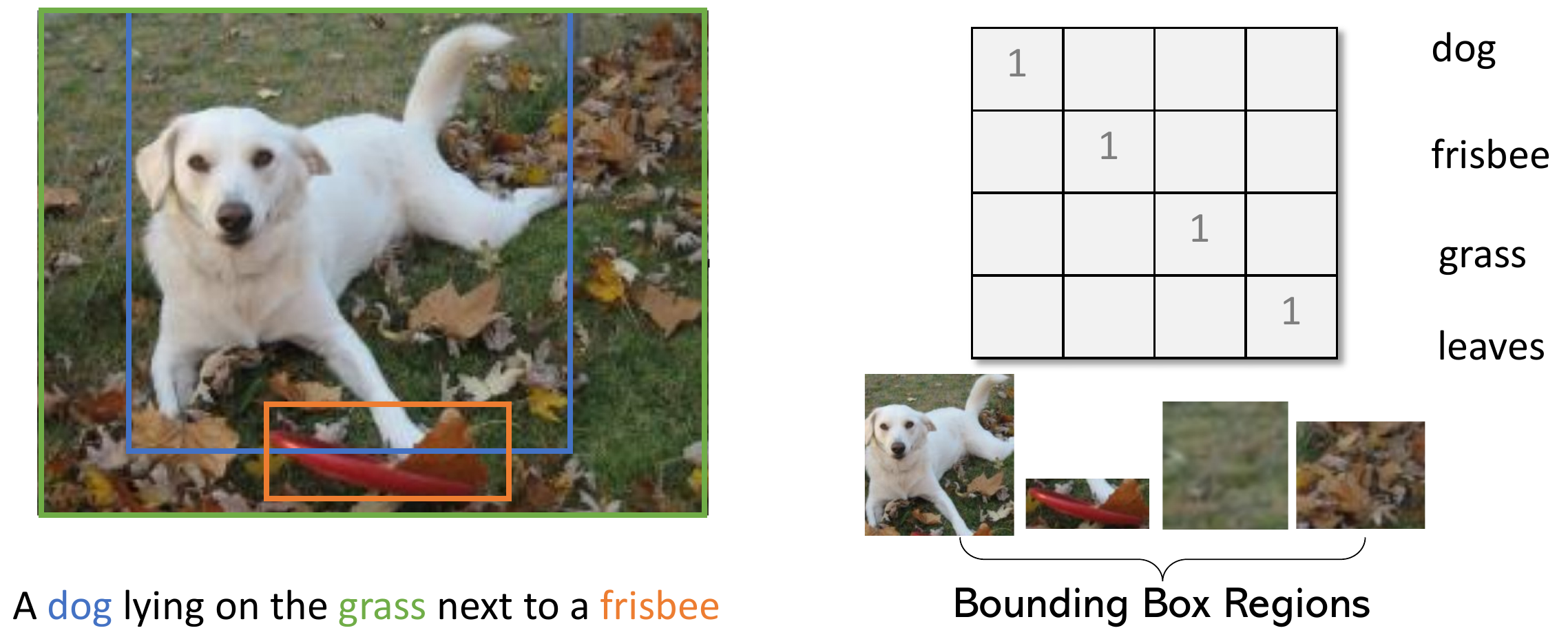}
  \caption{Region-phrase matching for object detection.}
  \label{fig:chp4_od}
\end{figure*}

\subsection{One-stage Models}

In the traditional OD formulation, the box classifier is implemented using a simple linear layer, and the classification loss $\Lcal_{\text{cls}}$ can be written as:
\begin{equation}\label{eqn:cls_logits}
    \Omat \!=\! f_{\thetav}(\xv), \,\,   \Smat_{\text{cls}} \!=\! \Omat \Wmat^{\top}, \,\,  \Lcal_{\text{cls}} \!=\! \Mcal(\Smat_{\text{cls}}; \Tmat)\,.
\end{equation}
Here,\footnote{$M$ is the number of region/box features, $d$ is the visual feature hidden dimension, $K$ is the number of object classes, and we ignore the bias in the box classifier for simplicity.} $\Omat \in \R^{M\times d}$ are the object/region/box features of the input image, $\Wmat \in \R^{K\times d}$ is the weight matrix of the box classifier, $\Smat_{\text{cls}} \in \R^{M\times K}$ are the output classification logits, $\Tmat \in \{0,1\}^{M\times K}$ is the target, and $\Mcal(\Smat; \Tmat)$ is the loss measure, \emph{e.g.}, focal loss in one-stage OD models.


Instead of classifying each region/box into $K$ classes, GLIP~\citep{li2021grounded} reformulates OD as a phrase grounding task, by grounding/aligning each region in $K$ phrases in a text prompt $\tv$. The alignment scores $\Smat_{\text{ground}}$ are computed between regions in image $\xv$ and words in the prompt $\tv$:
\begin{equation}\label{eqn:ground_logits}
    \Omat \!=\! f_{\thetav}(\xv), \,\, \Pmat \!=\! f_{\phiv}(\tv), \,\, \Smat_{\text{ground}} \!=\! \Omat \Pmat^{\top}\,,
\end{equation}
where $\Pmat \in \R^{L \times d}$ are the contextualized word/token features from the language encoder, and $L$ is the length of language prompt $\tv$. $\Pmat$ plays a similar role to the weight matrix $\Wmat$ in \eqref{eqn:cls_logits}. 
The grounding model, consisting of both the image encoder $f_{\thetav}$ and the language encoder $ f_{\phiv}$, is trained end-to-end by minimizing the loss defined in \eqref{eqn:loss} \& \eqref{eqn:cls_logits}, with a simple replacement of the classification logits $\Smat_{\text{cls}}$ in \eqref{eqn:cls_logits} with the region-word alignment scores $\Smat_{\text{ground}}$ in \eqref{eqn:ground_logits}. In Figure~\ref{fig:chp4_od}, we show an example of  $\Smat_{\text{ground}}$  computed for 4 region-word pairs. Note that all the bounding box proposals used to compute $\Smat_{\text{ground}}$ are extracted from one image. The matched pairs are assigned higher scores than the mismatched pairs. 

\subsection{Two-stage Models}
By distilling the knowledge from the CLIP/ALIGN model into a two-stage detector, ViLD~\citep{gu2021zero} and RegionCLIP~\citep{zhong2021regionclip} are proposed for zero-shot object detection.

In a two-stage detector, a separate region proposal network (RPN) with loss $\Lcal_{\text{rpn}}$ is used to distinguish foreground from background. Since $\Lcal_{\text{rpn}}$ does not use semantic information of object classes, it can be merged into the localization loss $\Lcal_{\text{loc}}$ in \eqref{eqn:loss}. In RegionCLIP, RPN is used to propose image regions for all images in a batch to produce $N$ image regions in total. The set of image regions is denoted by $\{\rv_i\}_{i=1}^N$. Given the proposed regions, the visual representation $\vv_i$ of region $\rv_i$ is produced by the visual encoder with a feature pooling method, such as RoIAlign.

RegionCLIP also builds a large pool of candidate concepts for image regions, which are often different from the concepts for full images. These concepts are in the form of natural language, and encoded into semantic representations $\{\uv_k\}_{k=1,...,K}$ by a pre-trained text encoder $\mathcal{L}$, where $K$ denotes the size of the concept pool. 

By leveraging the pre-trained CLIP, the object concept $\uv$ with the highest matching score is selected as the pseudo label for each region $\rv$, and thus constructing the positive pairs of $\{\uv,\vv\}$. A similar contrastive learning framework with an additional distillation loss is used to train the OD models.

\begin{figure*}[t!]
  \centering
    \includegraphics[width=0.99\linewidth]{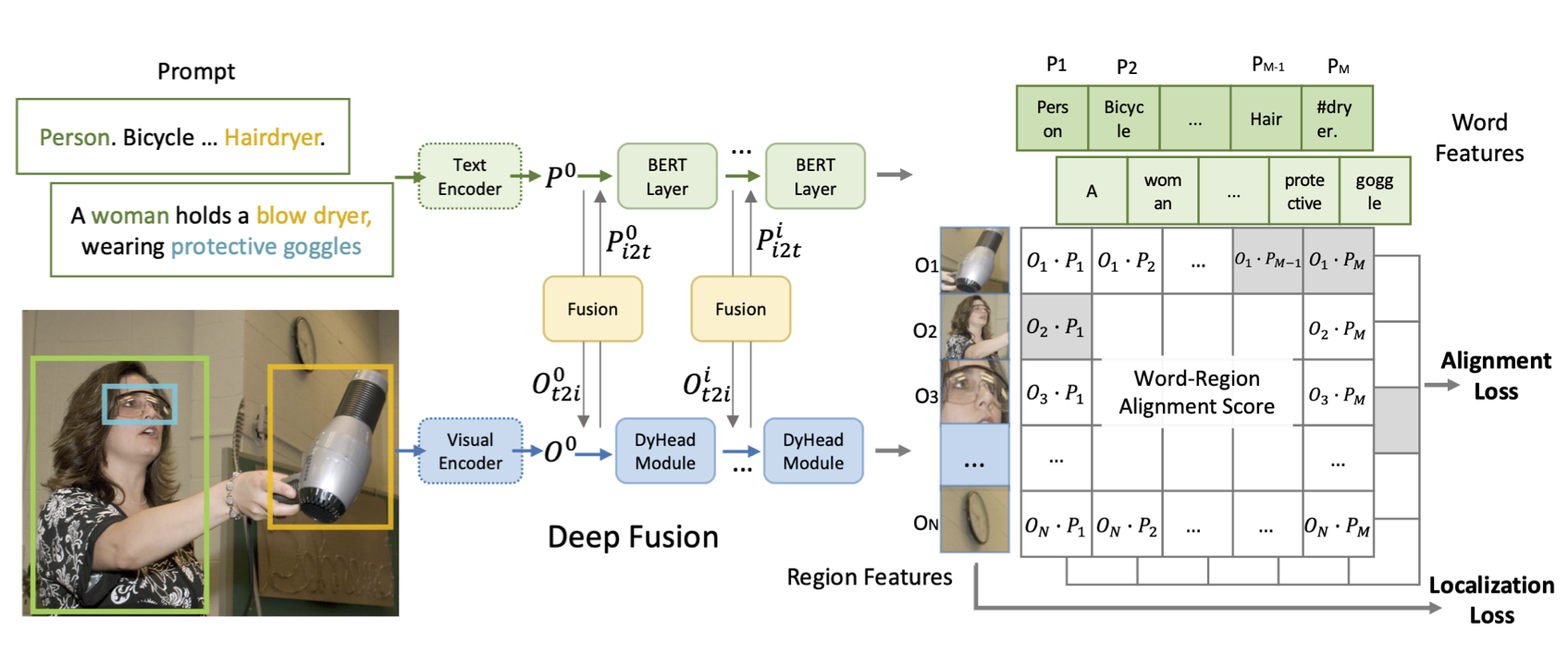}
    \includegraphics[width=0.8\linewidth]{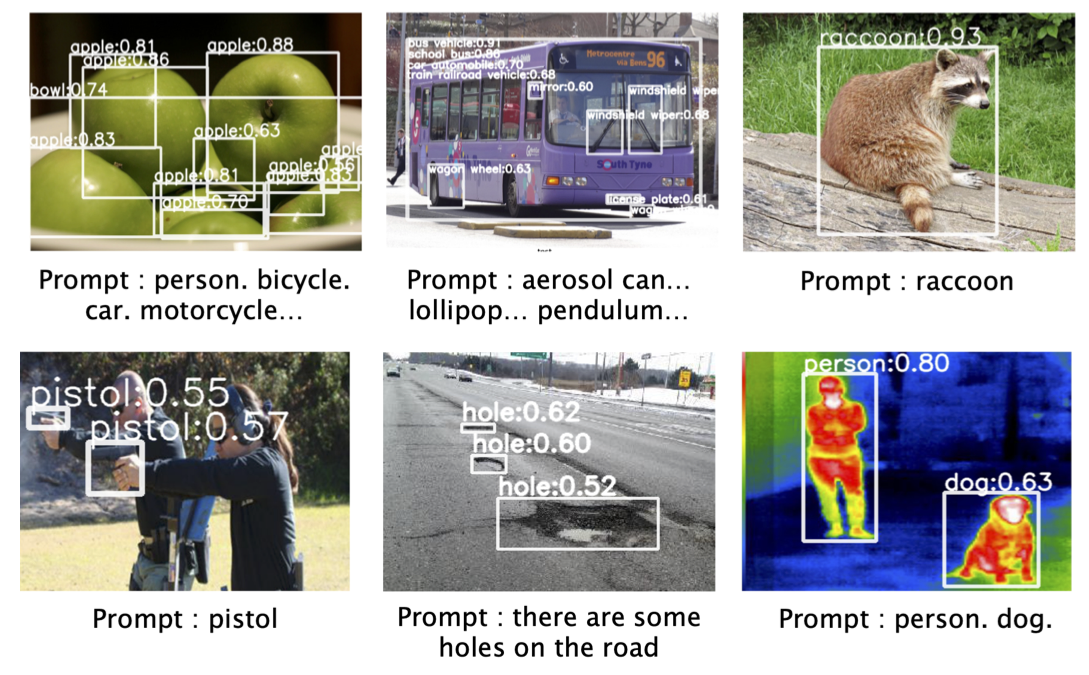}
  \caption{{\bf Top}: GLIP pre-trains an image encoder, a text encoder and a fusion module to predict which image box regions are paired with which words/phrase of the text prompt. This behavior allows us to turn GLIP into a zero-shot OD detector. We convert all of a dataset’s classes into captions by concatenation and predict the words/phrases of the caption that GLIP estimates best pairs with a given box. {\bf Bottom}: predictions of zero-shot GLIP object detector on examples from six datasets in ODinW~\citep{li2022elevater}. This figure was created in~\cite{li2021grounded}.}
  \label{fig:chp4_glip}
\end{figure*}

\paragraph{Other Language-Image Pre-training Methods for OD.} Learning generic open-set object detectors from image-text pairs is an increasingly popular topic.
Similar to GLIP, MDETR~\citep{kamath2021mdetr} reformulates detection as
a phrase grounding problem, and uses a single text query for the whole image.
FIBER~\citep{dou2022coarse} improves GLIP by ($i$) using a coarse-to-fine pre-training pipeline, and ($ii$) performing fusion in the backbone rather than in the OD head as in GLIP. 
OVR-CNN~\citep{zareian2021open} fine-tunes an image-text model to detection on a limited vocabulary and relies on image-text pre-training for generalization to an open vocabulary setting.
Detic~\citep{zhou2022detecting} improves long-tail
detection performance with weak supervision by training only the classification head on the examples where only image-level annotations are available.
Other con-current works include OV-DETR~\citep{zang2022open}, 
X-DETR~\citep{cai2022x},
FindIT~\citep{kuo2022findit}, 
PromptDet~\citep{feng2022promptdet},
and 
OWL-ViT~\citep{minderer2022simple}.

In the literature, there are two different experiment settings to evaluate the open-set object detection ability of pre-trained OD models.

\begin{minipage}{1.0\textwidth}
\centering
\hspace{0mm}
\begin{itemize}[leftmargin=3.0mm]

\item {\bf Class-level Transfer in a Single Domain}. One common zero-shot transfer evaluation for object detection follows the setting in~\cite{zareian2021open}, where a manual split is pre-defined for a given visual domain, ensuring no concept overlap between training and evaluation. For example, on LVIS~\citep{gupta2019lvis}, 866 frequent and common categories are treated as the base categories for training, and 337 rare categories are held out as the novel categories for evaluation. On COCO, there is a split with 48 base categories and 17 novel categories, removing 15 categories without a synset in the WordNet hierarchy.
 
\item {\bf Task-level Transfer}. This is an increasingly popular setting, where the pre-trained OD model is evaluated on multiple datasets in a zero-shot setting. For example, inspired by CLIP, the LVIS-trained model is evaluated on 3 datasets, including PASCAL VOC, COCO and Objects365 in ViLD~\citep{gu2021zero}. 
The recent {\it Object Detection in the Wild (ODinW)} benchmark generalizes the task-level evaluation to a more comprehensive regime, 13 datasets initiated in~\cite{li2021grounded} and 35 datasets formalized in~\cite{li2022elevater}, respectively.

\end{itemize}
\end{minipage}

\paragraph{Use Cases of Language-Image Models for OD.} In Figure~\ref{fig:chp4_glip}, we illustrate how the region-phrase matching models like GLIP can be used for zero-shot object detection. Given a new OD dataset/task with a set of concept/category names, all concepts are converted into caption by concatenation, add some simple user customized text prompt. The caption is used as prompt for the text encoder to extract concept representations. The query image is fed into the image encoder to extract the dense visual representation, which is used to compute the similarity with respect to all concepts via a deep fusion module. The similarity above the a given threshold yields the predicted results: the box of interest and the matched concept.
In the bottom of Figure~\ref{fig:chp4_glip}, six use cases are illustrated, all of them are from ODinW benchmark that represent the real-world OD scenarios. 

\begin{figure*}[t!]
  \centering
    \includegraphics[width=0.9\linewidth]{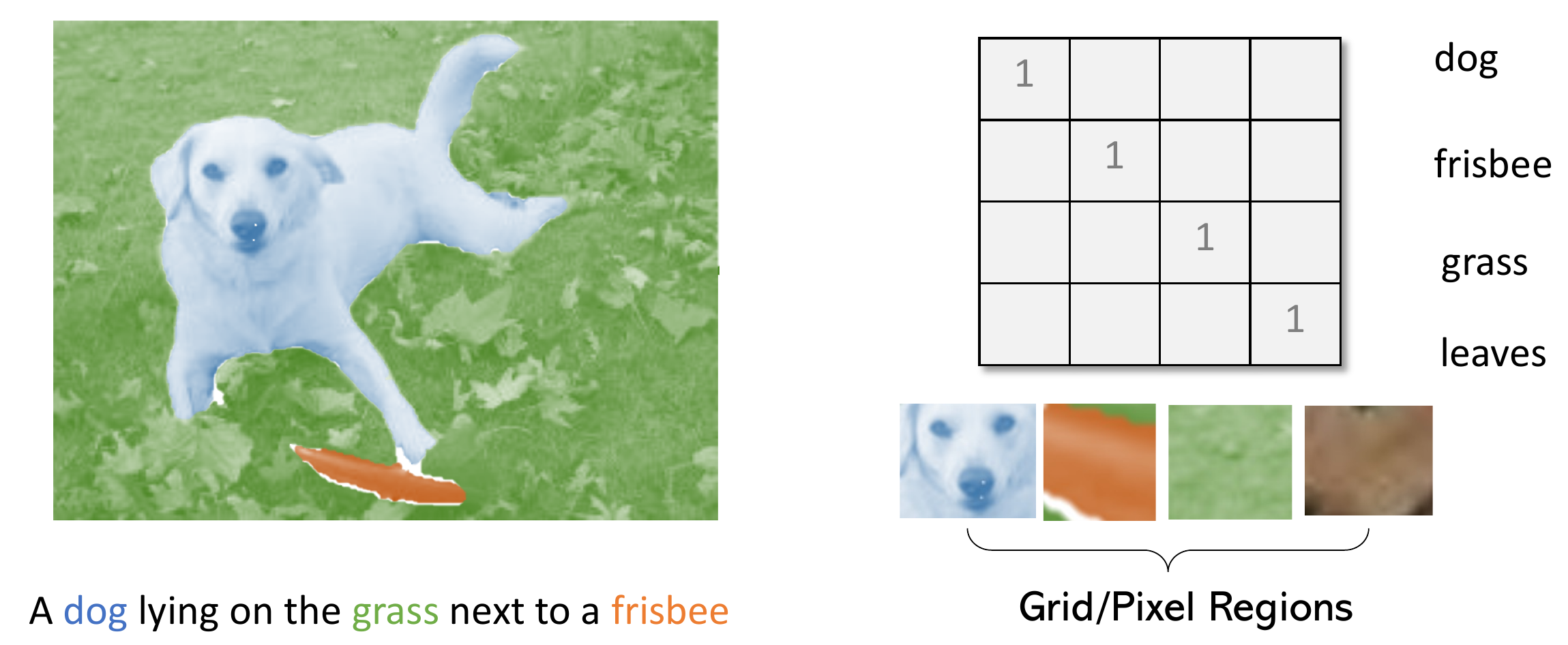}
  \caption{Pixel-phrase matching for segmentation.}
  \label{fig:chp4_seg}
\end{figure*}

\section{VLP for Segmentation}
\label{sec:vlp_vision_segmentation}
Image segmentation involves grouping image pixels and assigning a class label to each pixel of an image.
We use Language driven Semantic segmentation (LSeg)~\citep{li2022language} as an example to illustrate the image segmentation process, where textual categories and image pixels are embedded into a common space, and each pixel is assigned to a semantic category.

For any semantic segmentation task with a set of $K$ class labels, the text encoder embeds them into a continuous vector space $\R^d$, producing an embedding matrix for all classes $\Pmat = [ \pv_1, \cdots, \pv_K] \in \R^{K\times d}$ as outputs. For an image $\xv$, an image encoder encodes it into a dense grid representation $\Omat \in \R^{H \times W \times d}$, where $H$ and $W$ specify the spatial size of the feature map. The word-grid similarity tensor is computed as dot product $\Smat_{\text{seg}} = \Omat \Pmat^{\top} \in \R^{(H\times W) \times K}$. 
In Figure~\ref{fig:chp4_seg}, we show a simplified example of  $\Smat_{\text{seg}}$, which is computed on 4 word-grid pairs. 
Note that all the grid features to compute $\Smat_{\text{seg}}$ are extracted from one image. The matched pairs are assigned higher scores than the mismatched pairs.

For a given position pair, we minimize a per-grid Softmax with cross-entropy loss (with temperature
scaling) as is standard in semantic segmentation. In LSeg, a dense prediction Transformer~\citep{ranftl2021vision} is used to decode the features, and a final spatial regularization block spatially regularizes and cleans up the predictions.

Due to rich semantics in image-text pair data, there are many other works that use language-image models for segmentation, as detailed below.

\begin{minipage}{1.0\textwidth}
\centering
\hspace{0mm}
\begin{itemize}[leftmargin=3.0mm]

\item {\bf CLIP-based Segmentation.} 
Many segmentation models directly adapt the pre-trained CLIP to pixel-level visual recognition tasks, including PhraseCut~\citep{wu2020phrasecut}, OpenSeg~\citep{ghiasi2021open}, CLIPSeg~\citep{luddecke2022image}, ZS-Seg~\citep{xu2021simple}, MaskCLIP~\citep{zhou2021maskclip}, DenseCLIP~\citep{rao2021denseclip} and MaskCLIP~\citep{ding2022open}. 
OpenSeg~\citep{ghiasi2021open} also performs model learning with class agnostic mask annotations for generating mask proposals. 

\item {\bf Training from scratch.} GroupViT~\citep{xu2022groupvit} is a new hierarchical grouping Transformer architecture that exploits the global self-attention mechanism of Transformers to partition input images into progressively larger arbitrary-shaped groups. It is pre-trained with a multi-label image-text contrastive loss on around 12M image-text pairs. Since GroupViT automatically groups images into semantically-similar segments, its output can be easily transferred to semantic segmentation without fine-tuning.

\end{itemize}
\end{minipage}


\section{Trends: From Close-set to Open-set, to in-the-Wild}
\label{sec:vlp_vision_trend}
In the above three subsections, we have described how we might extend a close-set recognition model to perform three open-set recognition tasks: image classification, object detection and segmentation. 
The solution is to utilize a parametric function such as neural language models to represent categories, instead of traditional non-parametric representations such as one-hot vector embedding. 
Though it endows the functionality of open-set recognition, the model is still lacking the power to perform well on a large range of downstream tasks in the wild, where both the visual appearance of input images and the semantics of output categories often vary significantly from one application to another.

\begin{figure}[t!] 
  \begin{center}
  \vspace{0.5mm}
    \includegraphics[width=0.92\textwidth]{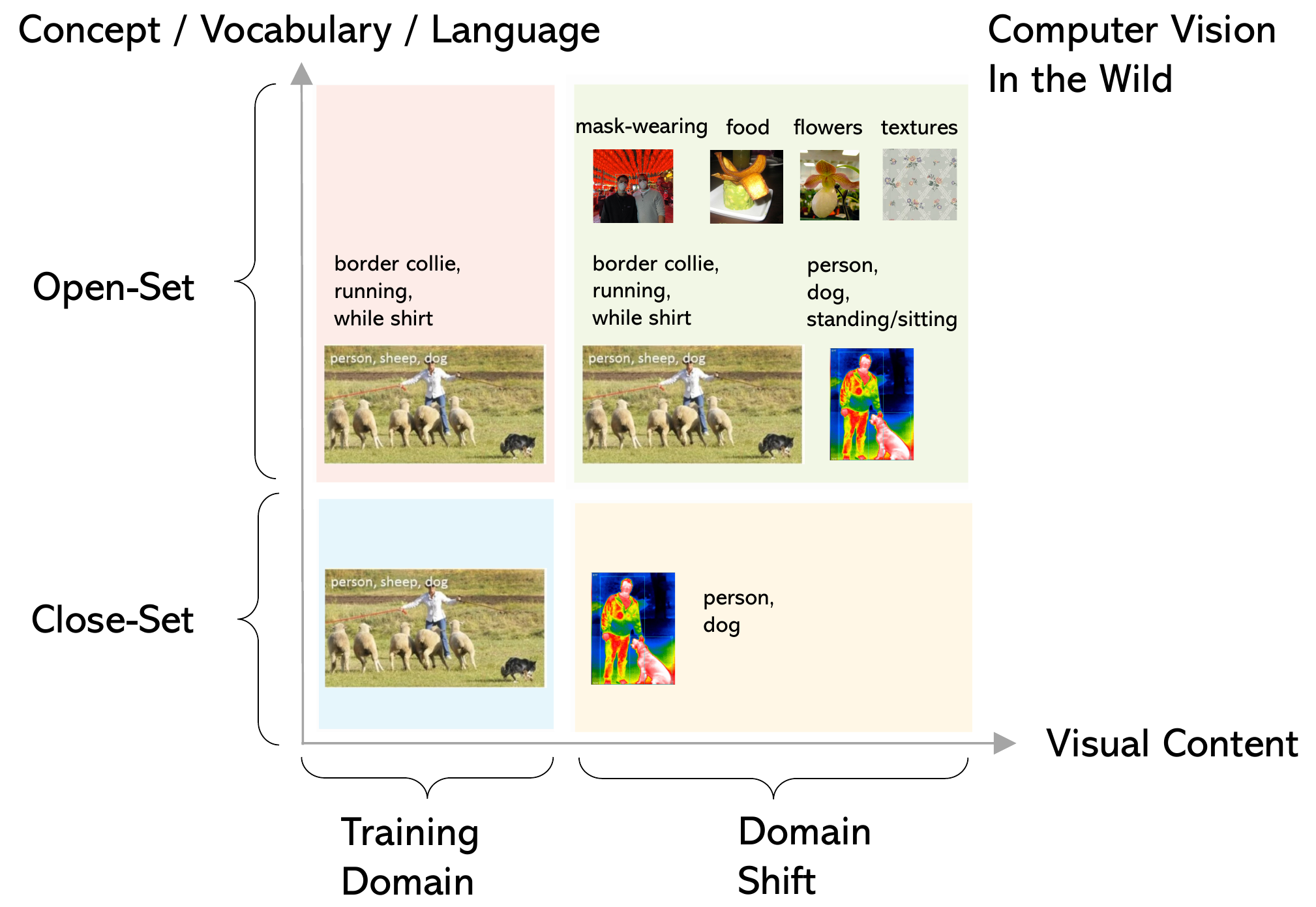}
  \end{center}
    \vspace{-2mm}
    \caption{Illustration on the setting of ``Computer Vision in the Wild (CVinW)'', in comparison with other settings. The 2D space is constructed with two dimensions: input image and output concept. The 2D chart is divided into four quadrants, based on the requirements between the model development stage and  the model evaluation stage. For the example provided in the standard setting, the natural image with concept ``person, sheep, dog'' is presented. Figure from~\cite{li2022elevater}.}
    \vspace{-2mm}
    \label{fig:chp4_setting_cvinw}
\end{figure}

\begin{figure*}[t!]
  \centering
    \includegraphics[width=0.9\linewidth]{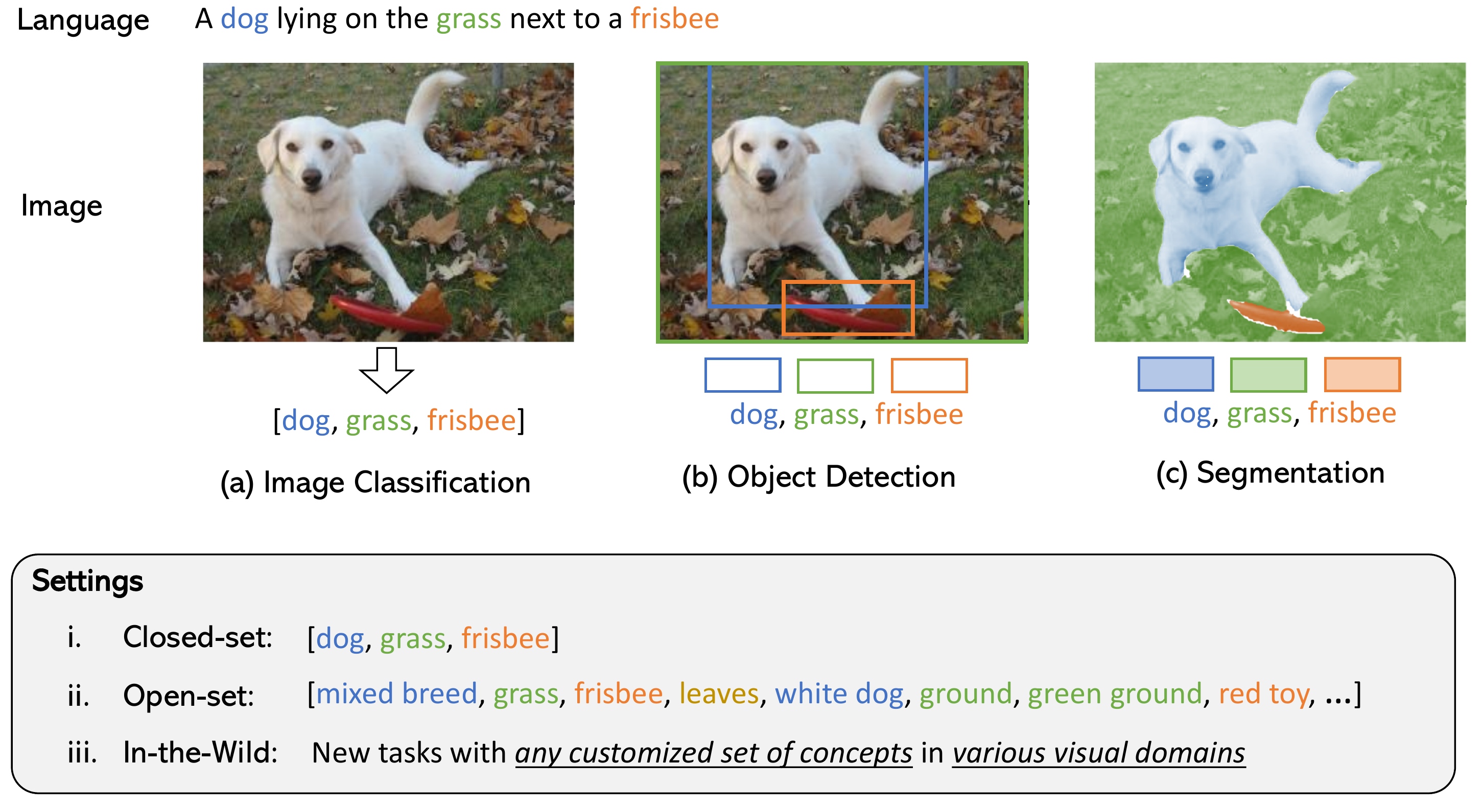}
  \caption{Illustration of different vision recognition settings.}
  \label{fig:chp4_settings}
\end{figure*}

In Figure~\ref{fig:chp4_setting_cvinw}, we use the definitions in~\cite{li2022elevater} to compare four settings studied in the computer vision community: traditional close-set recognition setting (bottom-left quadrant), open-set recognition setting (top-left quadrant), domain adaptation or out-of-distribution setting (bottom-right quadrant), and the CVinW setting (top-right quadrant). It is clear that CVinW considers variations in both visual domains and concept domains.  In fact, any visual recognition task can be naturally defined using a customized set of concepts and a given visual domain. From this perspective, CVinW considers {\bf task-level transfer}, which is beyond the {\bf concept/class-level transfer} that often appears in the traditional open-set recognition setting. In Figure~\ref{fig:chp4_settings}, we use the same image above to illustrate the difference among these settings.

The goal of developing foundation models for computer vision in the wild is two-fold:

\begin{itemize}[leftmargin=3.0mm]

\item {\bf The ability to transfer to a wide range of new downstream tasks.} It means the application scenarios of the foundation models are broad. The well-established datasets such as ImageNet and COCO are two representative close-set tasks for image classification and object detection, respectively. In real-world settings, both the visual domain and the concept sets can vary significantly, beyond ImageNet and COCO. The effectiveness of a foundation model is better measured by its applicability than by its performance on any specific tasks.

\item {\bf The adaptation cost of task-transfer is low.}
One major advantage of pre-trained foundation models is the promise that they can transfer to downstream tasks {\it effortlessly} (or in an inexpensive manner). It means that model adaptation efficiency is an important factor to measure the usability of a foundation model.
Good foundation models should be deployed with minimum adaptation effort. To measure the cost of adaptation, \cite{li2022elevater} define the adaptation cost in two orthogonal dimensions: sample-efficiency (measured by the number of training examples), and parameter-efficiency (measured by the number of trainable parameters). The  established datasets such as ImageNet and COCO do not provide the best evaluation setting for foundation models. To achieve SoTA performance on these datasets, it often requires full-model fine-tuning on the full-shots, a setting that comes with high adaptation cost.
As a north star, one foundation model with fixed weights should zero-shot transfer well to many downstream tasks.

\end{itemize}

The above goal of developing foundation models can be achieved on a range of computer vision tasks individually or jointly. When achieved individually, the setup is to build one separate foundation model for each problem. Most VLP models described in this chapter fall into this category. When achieved jointly, the setup is to build one unified foundation model across all tasks. Computer vision tasks require image processing at different levels of granularity (image, region, pixels), rendering the cross-task unification challenging. It remains an appealing open research topic to build one AI system that can leverage vision-language data at different levels of granularity, seeking the best trade-off between data scale and semantics-richness.


\section{Advanced Topics}
\label{sec:vlp_vision_advanced_topics}
As the literature on VLP for core computer vision problems is growing rapidly, an increasingly large number of papers and interesting research topics have emerged, as summarized in Figure~\ref{fig:chp4_tree}. Below, we provide a brief discussion on a few important topics, \emph{e.g.}, knowledge-augmented visual models, multilingual language-image models, efficient and robust model adaptation, benchmark, \emph{etc}. 

\begin{itemize}[leftmargin=3.0mm]

\item {\bf Knowledge-Augmented Visual Models}. The text encoder is arguably the most unique component in the recent-developed language-augmented computer vision systems. It is thereby important to improve text encoding for core visual recognition tasks. K-LITE~\citep{shen2022k} enriches entities in natural language with the WordNet/Wikipedia knowledge base, presenting a scalable way to transfer to a large range of new tasks in a zero-shot and few-shot manner for image classification and object detection. Compared with CLIP/UniCL/GLIP, K-LITE is much more sample-efficient in pre-training. \cite{tian2021vl} explore to leverage external knowledge to improve long-tailed visual recognition within individual domains, which falls into the category of class-level transfer.

\item {\bf Multilingual Language-Image Contrast}. The success of image-text contrastive learning using English captions has inspired the use of other language sources.
MURAL~\citep{jain2021mural} is pre-trained on multilingual image-text pairs from scratch, with an image-to-text contrastive loss and a text-to-text contrastive loss among different languages.
By distilling from the original English CLIP, \cite{carlsson_multilingualclip} train a language-specific encoder while keeping its image encoder fixed.
Other multilingual/bilingual/monolingual variants of contrastive language-image models include
Korean~\citep{ko2022large},
Italian~\citep{bianchi2021contrastive}, Russian~\citep{shonenkov2022ruclip} and Chinese~\citep{gu2022wukong}.

\item {\bf Efficient Adaptation Methods.} With the model size growing, it raises the problem of how to efficiently adapt a pre-trained model to a large range of downstream tasks. There are studies on sample-efficiency (\emph{e.g.}, zero-shot and few-shot) and parameter-efficiency (\emph{e.g.}, prompt tuning, linear probing and full model fine-tuning). 
For VLP models, it provides a unique opportunity to leverage the text encoders for model adaptation, including
Conditional Prompt Learning~\citep{zhou2022conditional},
Color Prompt Tuning (CPT)~\citep{yao2021cpt},
VL-Adapter~\citep{sung2022vl}, and
CLIP Adapter~\citep{gao2021clip}.
A comprehensive study on parameter efficiency can be found in ~\citet{he2022parameter}.

\begin{figure*}
  \centering
    \includegraphics[width=1.0\linewidth]{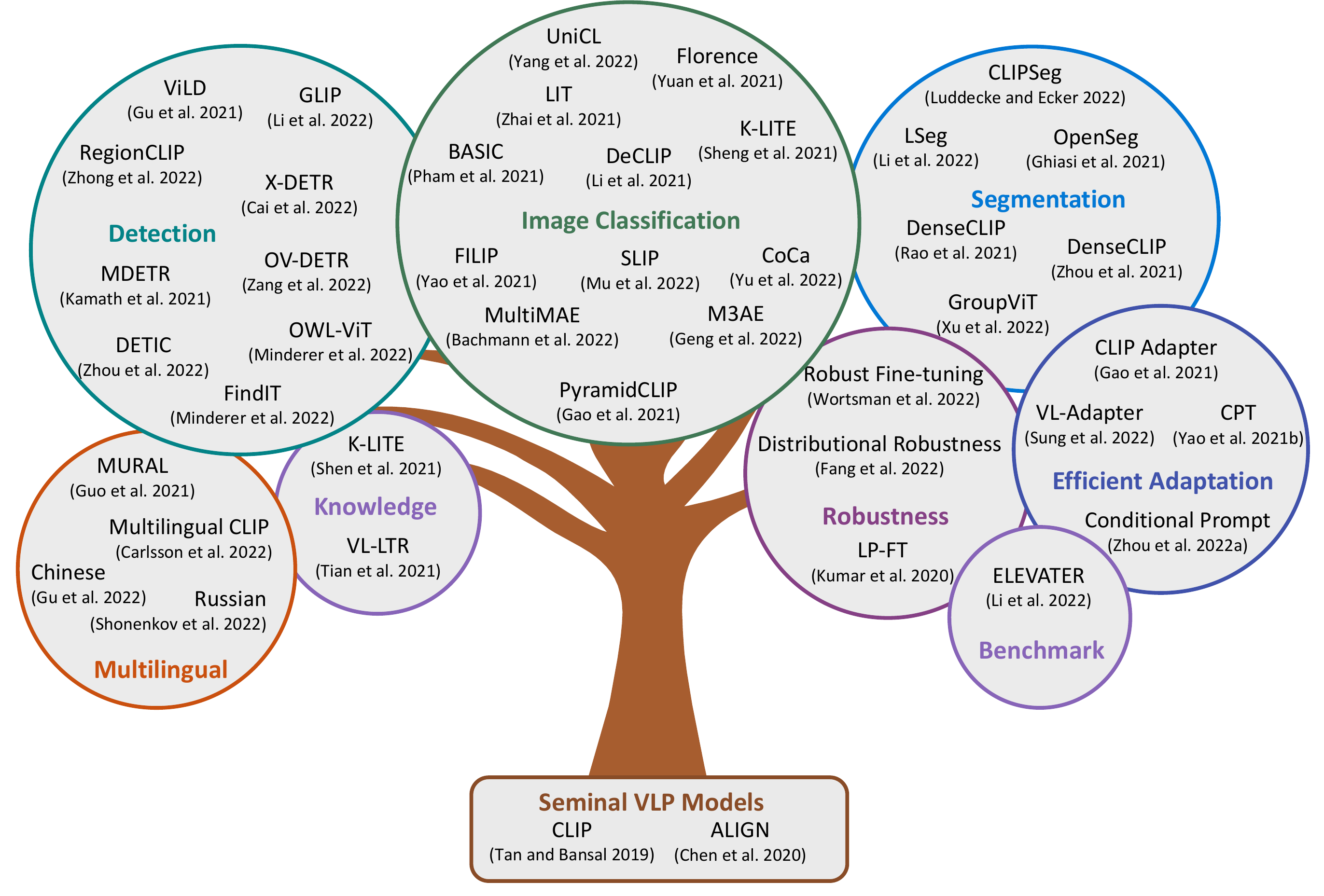}
  \caption{Research topics and papers in VLP for core computer vision tasks.}
  \label{fig:chp4_tree}
\end{figure*}
\item {\bf Robustness.}
\cite{wortsman2022robust} study robust fine-tuning of zero-shot models. \cite{fang2022data} report that
data determines distributional robustness in CLIP.
Fine-tuning CLIP can distort pre-trained features and underperform in out-of-distribution settings~\citep{kumar2022fine}.
The original CLIP paper reports that few-shot is worse than zero-shot CLIP when the number of shots is small. 
In contrast, \cite{li2022elevater} show that few-shot CLIP is always better than zero-shot CLIP when the pre-trained text encoder is properly used in model adaptation.
  
\item {\bf Benchmark.}  
It remains challenging to efficiently transfer and fairly evaluate pre-trained language-augmented visual models to downstream datasets and tasks. ELEVATER~\citep{li2022elevater} provides an evaluation platform for language-augmented visual models. ELEVATER contains a suite of datasets and an easy-to-use toolkit to evaluate task-level transfer ability of pre-trained visual models, which differs from the traditional benchmarks for evaluating class-level zero-shot transfer. 
It is used for the ICinW and ODinW challenges described above, serving as a common playground for computer vision in the wild.

\item {\bf Open-set Visual Relationship Recognition.}  The idea of open-set recognition has been extended to more visual recognition tasks, such as relation detection. 
Relational Language-Image Pre-training (RLIP)~\citep{yuan2022rlip}  improves zero-shot, few-shot and fine-tuning Human-Object-Interaction (HOI) detection performance, and the robustness to learning from noisy annotations.

\item {\bf Open-set Video Classification.}  Multi-modal Open-Vocabulary video classification (MOV)~\citep{qian2022multimodal} is proposed to use the vision encoder from pre-trained text-image models with minimal modifications to encode video, optical flow and audio spectrogram, and design a cross-modal fusion mechanism to aggregate complimentary multi-modal information.  X-CLIP~\citep{ni2022expanding} adapts the pre-trained text-image models to video recognition. It uses a cross-frame attention mechanism that explicitly exchanges information across frames, and a video-specific prompting scheme to leverage video content information for generating discriminative textual prompts.

\end{itemize}

We refer the readers who are interested in the literature on Computer Vision in the Wild (\emph{i.e.}, VLP for core vision tasks) to the up-to-date CVinW reading list at \url{https://github.com/Computer-Vision-in-the-Wild/CVinW_Readings}.


\chapter{VLP for Video-Text Tasks}
\label{chp:vlp4videotxt}
 
Videos contain multiple modalities in nature, and have been used as an epitome to test how AI systems perceive the world. 
In this chapter, we provide a systematic review on vision-language pre-training (VLP) for video-text tasks. We start with an introduction to popular video-text tasks in Section~\ref{sec:vid-txt-task}. In Section~\ref{sec:vid-txt-model}, we review the architecture of a typical video-text model, which consists of a video encoder, a text encoder and a multimodal fusion module.  
We divide the representative video-language models into two categories: ($i$) dual encoder, where video and text are encoded separately and a light multimodal fusion layer or operation (\emph{e.g.}, dot product) is used to fuse the video and text features; and ($ii$) fusion encoder, where on top of the video encoder and text encoder, multiple additional Transformer layers are usually adopted to capture deep interactions between the video and text features.
Section~\ref{sec:vid-txt-pretrain-tasks} and Section~\ref{sec:vid-txt-pretrain-data} present, respectively, the popular pre-training tasks adopted in literature and the datasets available for large-scale video-text pre-training. 
In Section~\ref{sec:vid-txt-advanced}, we discuss advanced topics and research trends in video-text pre-training, such as comprehensive video-text benchmarks and learning from multi-channel videos.

\section{Video-Text Tasks}\label{sec:vid-txt-task}
We introduce three popular video-text tasks: text-to-video retrieval, video question answering, and video captioning. Examples of these tasks are illustrated in Figure~\ref{fig:chp5_tasks}.

\subsection{Text-to-Video Retrieval} 
\begin{figure*}[t!]
  \centering
    \includegraphics[width=1.0\linewidth]{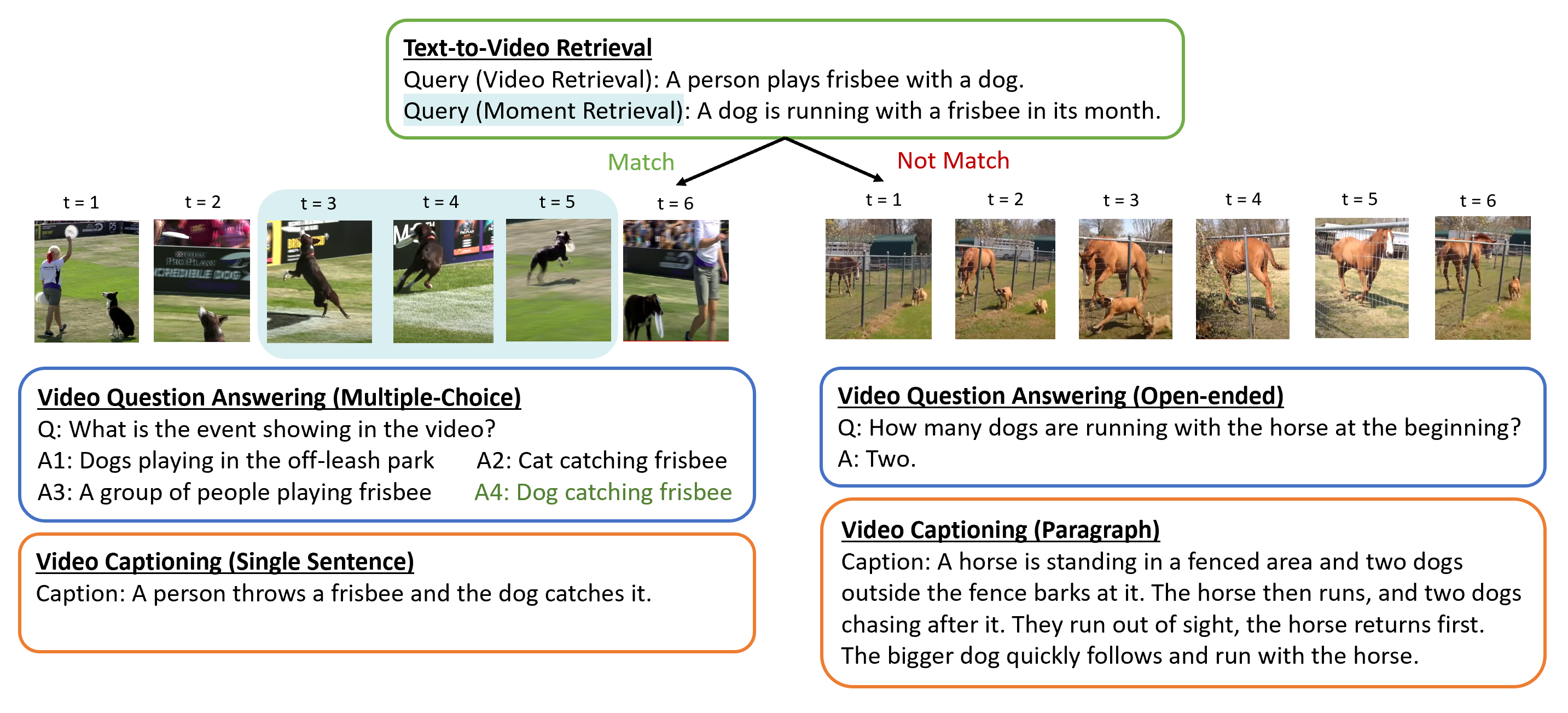}
  \caption{Illustration of representative video-text tasks: ($i$) text-to-video retrieval, including video retrieval and moment retrieval; ($ii$) video question answering in both multiple-choice and open-ended settings; and ($iii$) video captioning with a single-sentence caption or a paragraph of captions.}
  \label{fig:chp5_tasks}
\end{figure*}

The text-to-video retrieval task is to retrieve a relevant video or video segment given a natural language query, from a large video corpus. The task can be further categorized into three types, depending on its settings.
\begin{itemize}[leftmargin=3.0mm]
    \item \textbf{Video retrieval (VR)}~\citep{chen-dolan-2011-collecting,xu2016msr} aims to retrieve a video from a large video corpus. In this setting, the text queries are supposed to give an overview description of a video. Take the example shown in Figure~\ref{fig:chp5_tasks}, ``a person plays frisbee with a dog'' summarizes the event happening in the first video. This is analogous to text-to-image retrieval, and Recall@K (K=1, 5, 10, 100) is used as the evaluation metric.
    \item \textbf{Single Video Moment retrieval (SVMR)}~\citep{regneri-etal-2013-grounding,krishna2017dense} is to ground the text query in a specific time interval of a given video. The text query is only relevant to 
    a specific segment of the whole video. For example, in Figure~\ref{fig:chp5_tasks}, ``a dog is running with a frisbee in its month'' can only be grounded in the visual content at $t=3,4,5$ in the first video. Again, Recall@K (K=1, 5, 10, 100) is used as the evaluation metric, with a constraint on temporal intersection over union (tIoU) between the ground truth and the predicted proposals (\textit{e.g.}, tIoU$\geq$0.5/0.7). 
    \item \textbf{Video Corpus Moment Retrieval (VCMR)}~\citep{lei2020tvr,li2020hero} further extends the pool of relevant video segments from a single video to a large video corpus. It can be viewed as the combination of VR and SVMR. An AI model is required to not only retrieve the relevant video from the video corpus, but also localize the video segment in the retrieved video so that it can be described by the text query. For example, given the query ``a dog is running with a frisbee in its month'', the model needs to correctly match it to the first video and then ground the text query in the video segment from $t=3$ to $t=5$. Similarly, VCMR is evaluated using Recall@K (K=1, 5, 10, 100) with tIoU$\geq$0.5/0.7.
\end{itemize}

Most VLP models~\citep{miech2019howto100m,bain2021frozen} are evaluated on VR. Popular VR datasets include ($i$) MSVD~\citep{chen-dolan-2011-collecting}, MSRVTT~\citep{xu2016msr},  LSMDC~\citep{rohrbach2015lsmdc}, YouCook2~\citep{zhou2018youcook2} and VATEX~\citep{wang2019vatex} for single-sentence-to-video retrieval; and ($ii$) DiDeMo~\citep{hendricks2017didemo} and ActivityNet Captions~\citep{krishna2017dense-caption} for paragraph-to-video retrieval. 
The paragraph-to-video retrieval datasets are transformed from the datasets collected for the more challenging SVMR or VCMR tasks. 
In DiDeMo and ActivityNet Captions, each sentence of a paragraph is annotated with relevant time intervals. 
More recently, TVR~\citep{lei2020tvr} and How2R~\citep{li2020hero} are proposed to incorporate additional dialogue/subtitle information to perform VCMR with multi-channel video inputs. 

\subsection{Video Question Answering} 
Given a video-question pair, video question answering (QA) requires an AI model to answer the question based on the video content. There are two settings,  both are evaluated in accuracy.

\begin{itemize}[leftmargin=3.0mm]
\item \textbf{Multiple-Choice} Video QA~\citep{jang2017tgif-qa}: A model needs to identify the correct answer from a list of fixed, small number of answer candidates (\textit{e.g.}, 4-5 answer candidates). As the answer is constrained to a finite set, the task is often formulated as classification. Popular datasets include TGIF-Action, TGIF-Transition~\citep{jang2017tgif-qa}, TVQA~\citep{lei2018tvqa}, TVQA+~\citep{lei2019tvqa}, How2QA~\citep{li2020hero} and Drama-QA~\citep{choi2021dramaqa}. 
In the literature, video-to-text retrieval tasks with a small number of text candidates are often regarded as a multiple-choice QA task, such as LSMDC-MC~\citep{torabi2016lsmdc-fib} and MSRVTT-MC~\citep{yu2018js-fusion}. 
More recently, different video reasoning datasets have been proposed, mostly in the format of multiple-choice QA. Examples include VIOLIN~\citep{liu2020violin} for video-and-language inference, VLEP~\citep{lei2020vlep} for future event prediction in videos, NExT-QA~\cite{xiao2021next} to test on causal action reasoning, and STAR~\citep{wu2021star_situated_reasoning} to test on 4 types of situated reasoning: interaction, sequence, prediction and feasibility. 

\item \textbf{Open-ended} Video QA~\citep{xu2017msrvtt-qa}: the correct answer can be free-form, constructed by words from the whole word vocabulary. The common practice is to fist form a finite set of answer vocabulary by selecting the most frequent answers from the training split, and formulate it as a classification task. Popular datasets in this setting include LSMDC-FiB~\citep{torabi2016lsmdc-fib}, TGIF-Frame~\citep{jang2017tgif-qa}, MSRVTT-QA, MSVD-QA~\citep{xu2017msrvtt-qa},  ActivityNetQA~\citep{yu2019activitynet} and iVQA~\citep{yang2021just}.
\end{itemize}

\subsection{Video Captioning}
The task is to generate a natural language description for a given video, which is the only generation task among the three. The caption is expected to comprehensively describe the content of the video, including the events or objects of interest, the evolution of the events or object behaviors along time, and the relations among them. Most popular benchmarks \citep{chen-dolan-2011-collecting,xu2016msr,wang2019vatex} require generating a \textbf{single-sentence caption} to describe the overall video content. 
Although a single sentence may be enough to summarize the event happening in short videos, 
descriptions of longer videos are often \textbf{multiple-sentence paragraphs}, as in the dense captioning benchmark~\citep{krishna2017dense-caption}. 
Recently, multi-modal video captioning datasets (\textit{e.g.,} TVC~\citep{lei2020tvr}) are proposed with captions describing both visual scenes and dialogues/subtitles in videos. 
Captioning performance is evaluated using standard text generation metrics, such as  BLEU~\citep{papineni2002bleu}, METEOR~\citep{banerjee2005meteor}, ROUGE-L~\citep{lin2004rouge} and CIDEr~\citep{vedantam2015cider}.

\begin{figure*}[t!]
  \centering
    \includegraphics[width=1.0\linewidth]{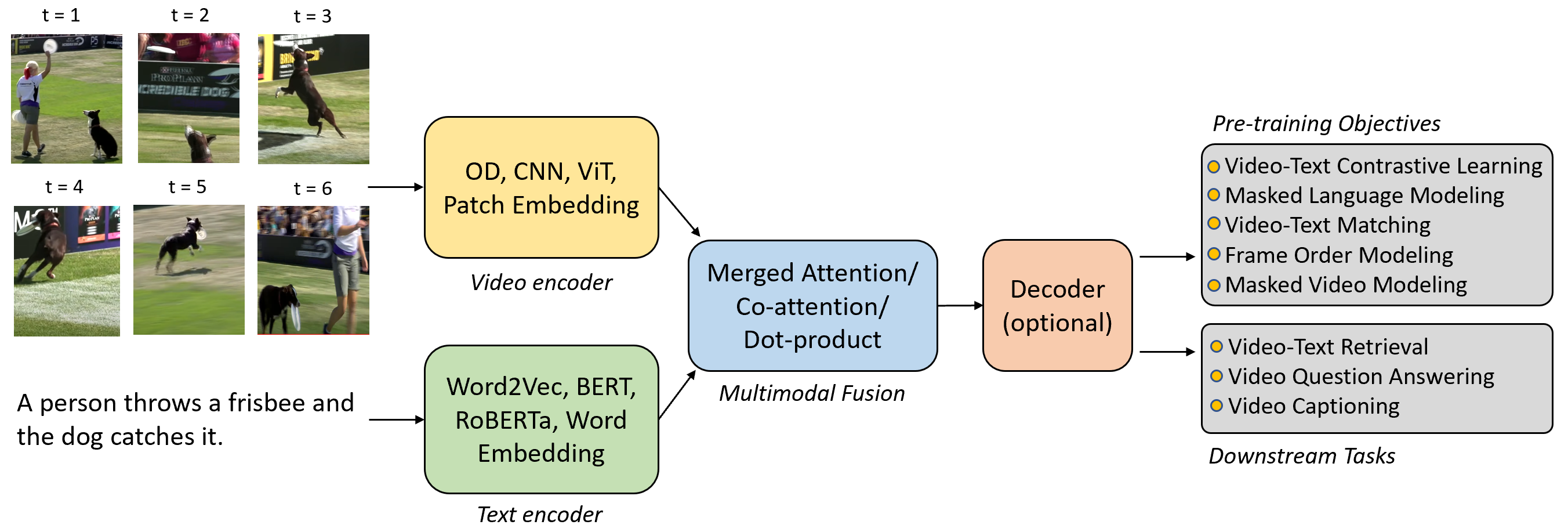}
  \caption{Illustration of a general framework for Transformer-based video-language models.}
  \label{fig:chp5_framework}
\end{figure*}

\section{Model Architectures}~\label{sec:vid-txt-model}
\noindent\textbf{Overview.} 
Given a pair of text sentence $\wv$ and video $\vv$, a typical video-text model first extracts a sequence of text features $\wv=\{ \wv_1, \cdots, \wv_N\}$ and visual features $\vv = \{\vv_1, \cdots, \vv_M\}$ via a \textit{text encoder} and a \textit{video encoder}, respectively. Here, $N$ is the number of tokens in a sentence, and $M$ is the number of visual features for a video, which can be the number of frames/regions/patches, depending on the specific vision encoder being used.  
A \textit{multimodal fusion module} 
projects these features into a shared embedding space to produce cross-modal representations. We broadly divide video-text models into two categories, based on the design of the multimodal fusion module:
\begin{itemize}[leftmargin=3.0mm]
    \item \textbf{Dual Encoder}, where video and text are encoded separately, and the interaction between video and text features is modeled using a light-weight operation (\emph{e.g.}, dot product or cosine similarity). This design is favorable in text-to-video retrieval for fast search~\citep{bain2021frozen}, and also widely adopted for promoting better video representations via contrastive video-text pre-training~\citep{miech2019howto100m}.
    However, such shallow cross-modal interactions are not effective enough for video QA and captioning tasks, as shown in Support-Set~\citep{patrick2020support}. Therefore, an additional text decoder is needed for caption generation. 
    \item \textbf{Fusion Encoder}, where on top of the video encoder and text encoder, additional Transformer layers~\citep{vaswani2017attention} are adopted to capture fine-grained interactions between video and text features. Preeminent works with deep fusion encoder, such as VideoBERT~\citep{sun2019videobert}, UniVL~\citep{luo2020univl}, ClipBERT~\citep{lei2021less}, and MERLOT~\citep{zellers2021merlot}, show strong performance on the video QA and captioning tasks. 
    While still achieving competitive performance on text-to-video retrieval tasks, fusion encoders are computationally more expensive than dual encoders~\citep{bain2021frozen}. 
\end{itemize}

\begin{figure*}[t!]
  \centering
    \includegraphics[width=1.0\linewidth]{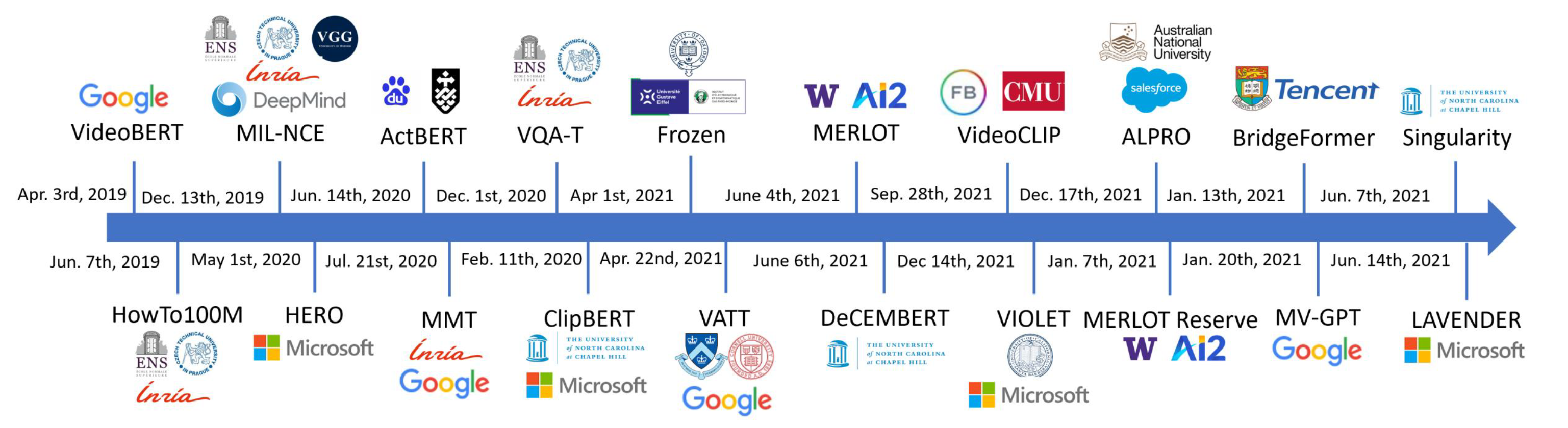}
  \caption{VLP models developed for video-text tasks along time. Due to space constraint, only some representative works are shown.}
  \label{fig:chp5_vlp_along_time}
\end{figure*}

The final outputs of a video-text model are either generated directly via an output layer that operates on the cross-modal representations produced by the multimodal fusion module (for \textit{encoder-only} models), or a decoder that is added in between the multimodal fusion module and the output layer (for \textit{encoder-decoder} models). An illustration of this framework is shown in Figure~\ref{fig:chp5_framework}. Table~\ref{tab:vlp_glossary_video_text} summarizes the representative VLP models for video-text tasks, including fusion encoder models (the upper block) and dual encoder models (the lower block). 
In Figure~\ref{fig:chp5_vlp_along_time}, we further show how these VLP models evolve along time.  
Next, we review each component in detail.

 \begin{table*}[!t]
\resizebox{1.0\textwidth}{!}
{
  \begin{tabular}{l|lccccc}
     \bf \multirow{2}{*}{Model}  & \bf Multimodal & \bf Vision & \bf Text  & \bf \multirow{2}{*}{Decoder} & \bf \multirow{2}{*}{E2E} & \bf Pre-training \\
     & \bf Fusion & \bf Encoder & \bf Encoder & & & \bf Objectives\\
    \hline
VideoBERT~\citep{sun2019videobert} &  & 3D CNN  & Emb.    &  \ding{55} & \ding{55} & MLM+VTM+MVM\\
ActBERT~\citep{zhu2020actbert} &  & OD  & Emb. &  \ding{55}  & \ding{55} & MLM+VTM+MVM\\
HERO~\citep{li2020hero} &  & 2D+3D CNN & Emb.     &  \ding{55} & \ding{55} & \specialcell{MLM+VTM+FOM\\+MFM}\\
UniVL~\citep{luo2020univl} &  & \specialcell{2D+3D CNN+Xformer} & Xformer  &  \ding{51} & \ding{55} & \specialcell{VTC+MLM+VTM\\+MFM+LM}\\
VQA-T~\citep{yang2021just}  & & 3D CNN & Xformer  &  \ding{55} & \ding{55} & MLM+VTC\\
TACo~\citep{yang2021taco}  & \multirow{1}{*}{Xformer}  & \specialcell{2D+3D CNN} & Xformer     &  \ding{55} & \ding{55} & VTC\\
\cmidrule(lr){3-7}
ClipBERT~\citep{lei2021less} &  & 2D CNN
  & Emb.  & \ding{55} & \ding{51} & MLM+VTM\\

MERLOT~\citep{zellers2021merlot}  &  & \specialcell{2D CNN+Xformer}
& Xformer     &  \ding{55} & \ding{51} & MLM+VTC+FOM\\
MV-GPT~\citep{seo2022end}  &  & Xformer & Emb.     &  \ding{51} & \ding{51} & MLM+LM\\
LAVENDER~\citep{li2022lavender} & & Xformer & Xformer & \ding{55} & \ding{51} & MLM+VTM as MLM\\
Singularity~\citep{lei2022revealing} & & Xformer & Xformer & \ding{55} & \ding{51} & MLM+VTM+VTC\\
\midrule

HTM~\citep{miech2019howto100m} &  & 3D CNN  & Word2Vec &  \ding{55} & \ding{51} & VTC\\
MIL-NCE~\citep{miech19endtoend} &  & 3D CNN  & Word2Vec &  \ding{55} & \ding{51} & VTC\\
Support Set~\citep{patrick2020support} & \multirow{1}{*}{Dot Product} & \specialcell{2D+3D CNN+Xformer} & Xformer    &  \ding{51} & \ding{55} & VTC+LM\\
 VideoCLIP~\citep{xu2021videoclip} &  & \specialcell{3D CNN+Xformer} & Xformer   &  \ding{55} & \ding{55} & VTC\\
  Frozen~\citep{bain2021frozen} & & Xformer & Xformer & \ding{55} & \ding{51} & VTC\\
    
  \end{tabular}
  }
  \caption{\textbf{Glossary of representative VLP models for video-text tasks}. E2E: end-to-end. CNN: convolutional neural netowrks. OD: object detector. Xformer: transformer. Emb.: embedding. MLM/MFM/MVM: masked language/frame/video modeling. VTM: video-text matching. VTC: video-text contrastive learning. FOM: frame order modeling. LM: language modeling.
  }
  \label{tab:vlp_glossary_video_text}
\end{table*}

\paragraph{Video Encoder.} Unlike static images, a video clip consists of a sequence of frames/images that evolve over time. Hence, the video encoder needs to capture not only spatial information from each frame, but also temporal dynamics across frames. Over time, the video encoder evolves from multiple offline feature extractors, to a single video encoder learned in an end-to-end manner. The change in video encoder also reflects the general trend in VLP for video-text tasks, \emph{i.e.,} from two-stage pre-training to end-to-end pre-training, similar to image-text models in Chapter~\ref{chp:vlp4imgtxt}. 
 
 \begin{itemize}[leftmargin=3.0mm]
 \item \textbf{Multiple offline feature extractors.} Early methods~\citep{sun2019videobert,zhu2020actbert,li2020hero} use a combination of fixed video feature extractors, such as 2D CNNs pre-trained for image classification (\emph{e.g.}, ResNet~\citep{he2016deep}), 3D CNNs pre-trained for action recognition (\emph{e.g.}, I3D~\citep{carreira2017quo}),
 and object detection models (\emph{e.g.},  Faster RCNN~\citep{girshick2015fast}). These video features are further processed to have a similar format to text inputs or projected into the same high-dimensional space as text representations. For example, VideoBERT~\citep{sun2019videobert} generates a sequence of ``visual tokens'' (in analogy to textual tokens) by applying hierarchical vector quantization to the pre-extracted video features from S3D~\citep{zhang2018s3d} pre-trained on Kinetics~\citep{kay2017kinetics}. 
ActBERT~\citep{zhu2020actbert} represents a video by combining a sequence of action features from a 3D CNN and a sequence of regional object features from Faster R-CNN. The learnable embedding of a special token (\texttt{[ACT]} for action and \texttt{[REGION]} for object) is then added to the features before being fed to the multimodal fusion module. 
HERO~\citep{li2020hero} concatenates 3D Slowfast~\citep{feichtenhofer2019slowfast} features and 2D ResNet-101 features extracted at the same frame rate as the video representation. The concatenated video features are projected into a hidden space via a fully-connected layer, and then a positional embedding, which encodes the temporal order of input frame features, is added.
 
\item \textbf{Single video encoder learned in an end-to-end manner.} Although models based on pre-extracted video features achieve strong performance, these fixed features are somewhat disconnected with the target video-text tasks/domains. Offline feature extractors are often trained on pure vision tasks in different domains. To address this issue, researchers try to refine the video encoder during video-text pre-training~\citep{miech19endtoend,lei2021less,zellers2021merlot} in an end-to-end (E2E) manner. Instead of using multiple video encoders, which renders excessive computational demands, a single video encoder is used. For example, HTM~\citep{miech19endtoend} learns video representations from scratch with a randomly initialized I3D~\citep{carreira2017quo}. In ClipBERT~\citep{lei2021less}, ResNet-50 pre-trained for object detection~\citep{jiang2020defense} along with a temporal mean pooling is used to generate video representations.  With advances in vision Transformers (ViTs), recent E2E models adopt a fully Transformer-based architecture. Frozen~\citep{bain2021frozen} inserts several space-time self-attention blocks into a pre-trained ViT~\citep{dosovitskiy2020image} to learn a global video representation via contrastive video-text pre-training. MV-GPT~\citep{seo2022end} and LAVENDER~\citep{li2022lavender} directly encode video inputs via a video vision Transformer (\emph{e.g.,} ViViT~\citep{arnab2021vivit} and a video Swin Transformer~\citep{liu2021video-swin}). 
\end{itemize}

\paragraph{Text Encoder.} Text inputs are first tokenized into a sequence of tokens to obtain the token embeddings. Before the wide adoption of BERT-like models for video-text pre-training, early dual encoder models~\citep{miech2019howto100m,miech19endtoend} utilize the pre-trained word2vec embeddings~\citep{mikolov2013efficient}, followed by a max-pooling operation 
to obtain the overall sentence representation. Most recent works follow the standard text pre-processing steps of BERT to tokenize a text into a sequence of WordPieces~\citep{wu2016google}, with two special tokens (\texttt{[CLS]} and \texttt{[SEP]}) inserted at the beginning and the end of the sequence, respectively. 
A word embedding layer, consisting of token embedding,  position embedding and layer normalization layers, is used to embed these tokens to vectors in a high-dimensional continuous space. For dual encoder models, the learned  embeddings are the feature vectors produced by a deep Transformer network~\citep{patrick2020support,bain2021frozen,xu2021videoclip}. For fusion encoder models, they are either directly fed into a multimodal fusion module~\citep{tang-etal-2021-decembert,xu2021vlm}, where the word embedding layer is the only text-specific model component or processed by several Transformer layers for text encoding before multimodal fusion~\citep{yang2021just,yang2021taco,seo2022end}.

\paragraph{Multimodal Fusion.} For \textit{dual encoder} models like HTM~\citep{miech2019howto100m} and MIL-NCE~\citep{miech19endtoend}, the global video/text representations extracted from video/text encoders are aligned in a common semantic space via a lightweight inner product. 
For \textit{fusion encoder} models, the most popular design is \textit{merged attention} (illustrated in Figure~\ref{fig:chp3_fusion} (b) of Chapter~\ref{sec:chp3_model_architectures}), where the text and video features are simply concatenated and then fed into a single Transformer block. In a recent study~\citep{lei2022revealing}, cross attention modules are inserted to the top few Transformer layers between self-attention and feed-forward layers, to enable text features to attend to a variable-length visual feature sequence. This is similar to \textit{co-attention}  (illustrated in Figure~\ref{fig:chp3_fusion} (a) of Chapter~\ref{sec:chp3_model_architectures}). But it is asymmetric in that only video-to-text cross-attention modules are used.

 
\paragraph{Encoder-Only vs. Encoder-Decoder.} 
Similar to image-text models, most existing video-text models adopt an encoder-only architecture, which directly generate the final outputs from the cross-modal representations via an output layer. UniVL~\citep{luo2020univl},  MV-GPT~\citep{seo2022end}, Support Set~\citep{patrick2020support} are exemplary works with an encoder-decoder architecture, where a decoder is added between the encoder and the output layer. In these works, the decoder is pre-trained and used for generating video captions autoregressively on downstream tasks. A general comparison of encoder-only and encoder-decoder architectures for image-text inputs are shown in Figure~\ref{fig:chp3_enc_vs_enc_dec}. This illustrative comparison can be directly applied to video-text inputs, simply by replacing the input image with a sequence of input video frames.

\section{Pre-training Tasks}\label{sec:vid-txt-pretrain-tasks}

In this section, we review pre-training tasks adopted in video-text pre-training. We first introduce the popular pre-training tasks. For example, dual encoder models are typically optimized via \textit{Video-Text Contrastive Learning}. For fusion encoder models, two popular pre-training tasks are \textit{Masked Language Modeling} and \textit{Video-Text Matching}. Then, we move to pre-training tasks that are designed to model the unique characteristics of video inputs, such as \textit{Frame Order Modeling} and different variants of \textit{Masked Video Modeling}.

\paragraph{Video-Text Contrastive Learning (VTC).} 
In VTC, the model aims to learn the correspondence between video and text. VTC is widely adopted to train the dual-encoder models~\citep{miech2019howto100m,bain2021frozen}, where video and text inputs are fused via a lightweight inner product. This simple dot product is also used to compute the video-to-text and text-to-video similarities in VTC for dual-encoder models. Specifically, given a batch of $N$ video-text pairs, VTC aims to predict the $N$ matched pairs from all the $N^2$ possible video-text pairs.
\begin{align}
    s_{i,j}^{v2t} = \vv_i^\top \wv_j,&\,\, s_{i,j}^{t2v} = \wv_i^\top \vv_j \,,\\
    \mathcal{L}_{\text{VTC}}^{v2t}(\theta) = -\frac{1}{N} \sum_{i=1}^N\log\frac{\exp(s_{i,i}^{i2t}/\sigma)}{\sum_{j=1}^N \exp (s_{i,j}^{v2t}/\sigma)},&\,\,\,\, \mathcal{L}_{\text{VTC}}^{t2v}(\theta) = -\frac{1}{N} \sum_{i=1}^N\log\frac{\exp(s_{i,i}^{t2v}/\sigma)}{\sum_{j=1}^N \exp (s_{i,j}^{t2v}/\sigma)}\,, \nonumber
\end{align}
where $\{\vv_i\}_{i=1}^N$ and $\{\wv_i\}_{i=1}^N$ are the normalized video vectors and text vectors in a training batch,  $\sigma$ is a learned temperature hyper-parameter,  $\mathcal{L}_{\text{VTC}}^{v2t}$ and $\mathcal{L}_{\text{VTC}}^{t2v}$ are video-to-text and text-to-video contrastive loss, respectively. 

The naive formulation of VTC assumes there exist correct alignments between video and text in pre-training data, which is not always the case.  A great challenge in large-scale contrastive pre-training on existing video-text data is the inherent misalignment between visual frames and speech-transcribed subtitles. 
To address the visually misaligned narrations,  MIL-NCE~\citep{miech19endtoend} is proposed to combine multiple instance learning with contrastive learning to use the weak and noisy training signals in narrated videos. 
VideoCLIP~\citep{xu2021videoclip} constructs temporally overlapped pairs of video and text clips of varying length, in contrast to fixed length in~\citet{miech2019howto100m,miech19endtoend}, to increase the quality and quantity of pre-training corpus. In addition, they contrast not only different clips from the same video, but also harder negatives that are similar to the in-batch clips, retrieved from other videos.

Moreover, the conventional contrastive learning computes the loss after aggregating all the words in the text and frames in the video. In TACo~\citep{yang2021taco}, the authors propose to make it token-aware, where the lose is computed using only a subset of words (\textit{e.g.}, nouns and verbs), to improve the grounding of individual words in the video. TACo combines token-aware VTC with the naive VTC, applied to the dual-encoder architecture, and further adds a third VTC loss enhanced by deep multimodal fusion. Specifically, the similarity between video and text input is the multimodal fusion output for the \texttt{[CLS]} token, computed by the Transformer block operating on top of the dual encoders, which is exactly the fusion-encoder architecture. To reduce the complexity in computing the fusion-encoder VTC loss, they adopt a cascade sampling strategy to only sample a small subset of hard negatives based on the token-aware VTC and the naive VTC loss. 

\paragraph{Masked Language Modeling (MLM).} MLM is a direct adoption of the one used for language model pre-training, except that the inputs are video-text pairs. Formally, the inputs for MLM include: ($i$) sub-word tokens from an input sentence $\wv$; ($ii$) the visual inputs (\textit{e.g.}, frame patches/features) $\vv$ aligned with $\wv$; and ($iii$) mask indices $\mathbf{m} \in \mathbb{N}^M$. 
$\mathbb{N}$ is a natural number, $M$ is the number of masked tokens, and $\mathbf{m}$ is the set of masked indices. 
In practice, we randomly mask out input words with a probability of 15\%, and replace the masked tokens $\mathbf{w}^{\mathbf{m}}$ with the special token \texttt{[MASK]}. Following BERT, the 15\% randomly masked-out words are further decomposed into 10\% random words, 10\% unchanged, and 80\% \texttt{[MASK]}.
The goal is to predict these masked words based on the observation of their surrounding words $\mathbf{w}^{\setminus \mathbf{m}}$ and the visual inputs $\mathbf{v}$ aligned to the sentence, by minimizing the negative log-likelihood:
\begin{equation}
    \mathcal{L}_{\text{MLM}}(\theta) = -\mathbb{E}_{D} \log P_{\theta}(\mathbf{w}^{\mathbf{m}} | \mathbf{w}^{\setminus \mathbf{m}}, \mathbf{v})\,,
\end{equation}
where $\theta$ denotes trainable parameters. 
Each pair $(\mathbf{w}, \mathbf{v})$ is sampled from the training set $D$.

Similar to  image-text pre-training, video-text pre-training methods also use a variant of MLM, language modeling, 
where captions are generated token-by-token autoregressively, as in UniVL~\citep{luo2020univl} and Support-Set~\citep{patrick2020support}.  
Specific to video-text pre-training, speech-transcribed texts are usually less formal with utterances or repetitively mentioning the key objects. 
To avoid masking on un-grounded words, MERLOT~\citep{zellers2021merlot} provides a simple heuristic solution to mask words based on the learned attention weights and empirically verifies its advantages over random masking. 

\paragraph{Video-Text Matching (VTM).} 
In VTM, the model is given a batch of positive video-text pairs and negative video-text pairs, which are constructed by replacing the video/text inputs in positive video-text pairs.
The goal of VTM is to identify positive pairs of videos and texts. 
VTM is often formulated as a binary classification task.  Specifically, a special token (\emph{i.e.}, $\texttt{[CLS]}$) is inserted at the beginning of the input sentence, whose learned vector representation is used as the cross-modal representation of the input video-text pair. 
We then feed the model with either a matched or mismatched video-text pair $\langle \mathbf{v}, \mathbf{w} \rangle$ with equal probability, and learn a classifier to 
predict binary label $y$, indicating whether the sampled video-text pair is positive or negative. 
Specifically, denoting the output score by $s_{\theta}(\mathbf{w}, \mathbf{v})$, we apply the binary cross-entropy loss for optimization:
\begin{equation}
    \mathcal{L}_{\text{VTM}}(\theta) = - \mathbb{E}_{(\mathbf{w}, \mathbf{v})\sim D} [y \log s_{\theta}(\mathbf{w}, \mathbf{v}) + (1-y) \log (1-s_{\theta}(\mathbf{w}, \mathbf{v}))] )\,.
\end{equation}

Different variations of VTM have been proposed to
capture alignments along the temporal dimension of different levels of granularity.
For example, HERO \citep{li2020hero} considers both the global alignment (predicting whether a text matches the input video) and local temporal alignment (retrieving the moment where the text should be localized in the video clip), which is proven effective for downstream video corpus moment retrieval.

\paragraph{Other Pre-training Tasks.} Besides the pre-training tasks discussed above, some attempts have been made to leverage the unique characteristics of video inputs for self-supervised pre-training.

\begin{itemize}[leftmargin=*]
\item \textbf{Frame Order Modeling (FOM).}\quad FOM is proposed to model the chronological order of events or actions happening in video. During training, we scramble a certain percentage of input frames (or frame features) randomly chosen, and the model is trained to explicitly recover the correct temporal order. Two variants are explored, including reconstructing absolute temporal order of these shuffled frames as in HERO~\citep{li2020hero}, and predicting the relative order between each pair of frames as in MERLOT~\citep{zellers2021merlot}. In both works, FOM is applied to the videos paired with temporally grounded texts, such as subtitles or ASR outputs. 
\begin{itemize}
    \item \textbf{FOM with absolute temporal order.}\quad At time $t$, let's denote the video frame inputs as $\vv^t$ and the temporally grounded sentence as $\wv^t$. The inputs to FOM are $(i)$ all subtitle sentences $\{\wv^t\}$; $(ii)$ visual frames $\{\vv^t\}$; and $(iii)$ the reorder indices $\mathbf{r}=\{r_i\}_{i=1}^R\in \mathbb{N}^R$, where $R$ is the number of reordered frames, and $\mathbf{r}$ is the set of reorder indices. During training, 15\%  of the frames are randomly selected to be shuffled, and the goal is to reconstruct their original order along the temporal dimension, denoted as $\mathbf{t} = \{t_i\}_{i = 1}^{R}$, where $t_i \in \{1, ..., N_v\}$. FOM is formulated as a classification problem, where $\mathbf{t}$ is the ground-truth labels of the reordered frames. The final objective is to minimize the negative log-likelihood: 
\begin{equation}
    \mathcal{L}_{\text{FOM}}(\theta) = -\mathbb{E}_{D} \textstyle{\sum_{i=1}^R} \log P_{\theta}([r_i, t_i])\,.
\end{equation}
\item \textbf{FOM with relative temporal order.}\quad During training, 40\% of the time, an integer $n$, indicating the number of frames to be randomly shuffled, is first randomly picked  from $[2, T]$, given $T$ input frames. Then, $n$ frames are chosen at random to be randomly scrambled. After shuffling, the frames together with the text inputs are fed into the model to learn the joint video-language representations. For a pair of frames at timestep $t_i$ and $t_j$ (after the shuffling), we concatenate their hidden states and pass the result through a two-layer MLP, predicting if $t_i < t_j$ or $t_i > t_j$. Similarly, FOM with relative temporal order can be optimized using a cross-entropy loss.
\end{itemize}

\item  \textbf{Masked Video Modeling (MVM).}\quad As the consecutive frames may contain similar spatial information, MVM is introduced to reconstruct high-level semantics or low-level details for a certain percentage of ``masked'' visual inputs (\textit{i.e.}, features or patches), given intact video tokens/features from neighboring frames and the paired textual description. Specifically, the model is trained to reconstruct the masked patches or features $\mathbf{v}_{\mathbf{m}}$ given the remaining visible patches or features $\mathbf{v}_{\setminus \mathbf{m}}$ and the paired text $\mathbf{w}$. That is, 
\begin{equation}
    \mathcal{L}_{\text{MVM}}(\theta) = \mathbb{E}_{(\mathbf{w}, \mathbf{v})\sim D} P_{\theta}(\mathbf{v}_\mathbf{m} | \mathbf{v}_{\setminus \mathbf{m}}, \mathbf{w})\,.
\end{equation}
Similar objectives have been proposed for image-text pre-training~\citep{chen2020uniter,kim2021vilt}, known as masked image modeling (MIM) as described in Chapter~\ref{sec:chp3_pretrain_objectives}.

\begin{itemize}
\item \textbf{MVM with in-batch negatives} is explored in HERO~\citep{li2020hero}, leveraging Noise Constrative Estimation loss~\citep{jozefowicz2016exploring} to supervise the model to identify the correct frame feature corresponding to the masked frames, compared to all negative distractors in the same batch. 

\item \textbf{MVM with discrete visual tokens}, is first introduced in VideoBERT~\citep{sun2019videobert}. VideoBERT tokenizes continuous S3D~\citep{zhang2018s3d} features extracted from input video frames into discrete ``visual tokens'' using hierarchical k-means. These visual tokens are then used as both the video inputs to the model and the prediction targets for MVM. MVM is formulated as a classification task that is performed in the same manner as MLM. Similarly, 15\% of the input visual tokens are randomly masked and the model is trained to recover these masked visual tokens. More recently, VIOLET~\citep{fu2021violet} draws inspirations from self-supervised learning methods on vision Transformers~\citep{bao2021beit,vimpac} to take advantages of pre-trained DALL-E~\citep{ramesh2021dalle} to extract discrete visual tokens as the MVM targets. VIOLET randomly mask out the raw input video frame patches, and train the model to predict the corresponding visual tokens for these masked patches in an end-to-end manner. 

\item \textbf{MVM with other visual targets}. In addition to discrete visual tokens, \citet{fu2022empirical} empirically examines 7 other reconstructive targets of MVM, from low-level pixel values and oriented gradients to high-level depth maps, optical flow predictions, various latent visual features from deep neural networks. Likewise, the raw input video frame patches are randomly masked, and the model training is supervised with $l_1$ loss between the MVM prediction and these continuous visual targets of the masked patches.
\end{itemize}
\end{itemize}

\begin{figure*}[t!]
\centering
\sbox{\measurebox}{%
  \begin{minipage}[b]{.25\textwidth}
  \vspace{18pt}
  \subfloat
    [MIL-NCE]
    {\label{fig:chp5_milnce}\includegraphics[width=\textwidth]{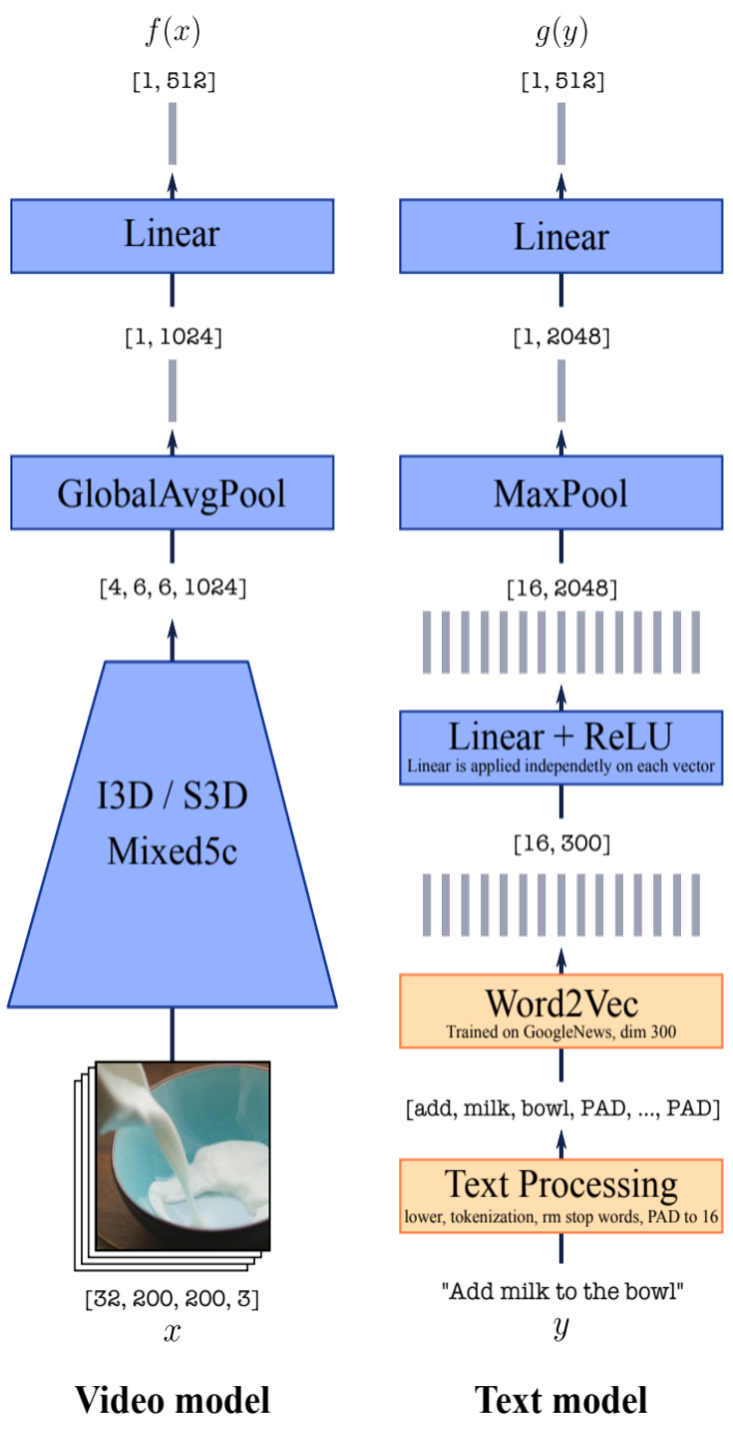}}
  \end{minipage}}
 \hfill
\usebox{\measurebox}\qquad
\begin{minipage}[b][\ht\measurebox][s]{.65\textwidth}
\centering
\subfloat
  [UniVL]
  {\label{fig:chp5_univl}\includegraphics[width=\textwidth]{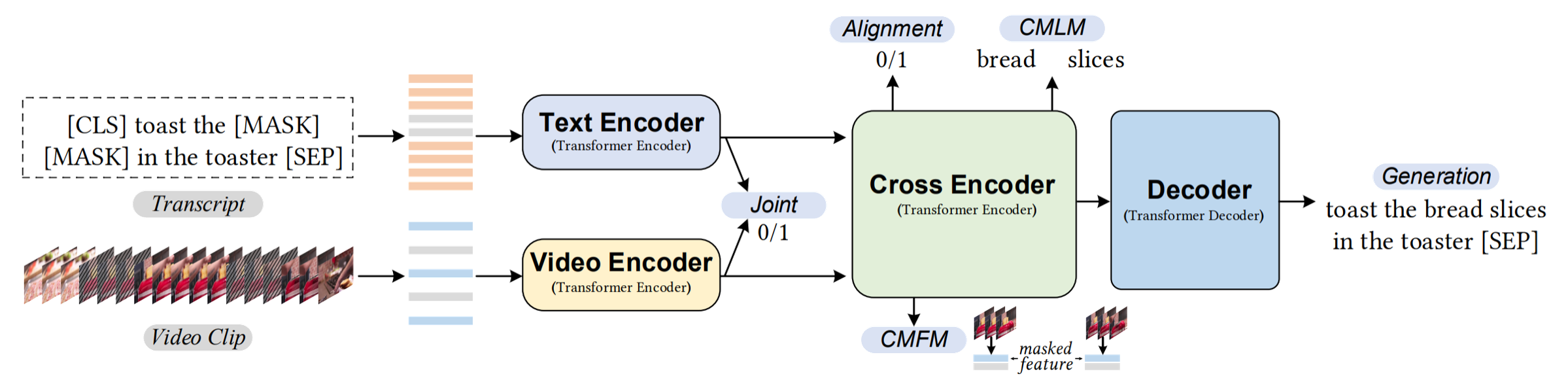}}


\subfloat
  [ClipBERT]
  {\label{fig:chp5_clipbert}\includegraphics[width=0.5\textwidth]{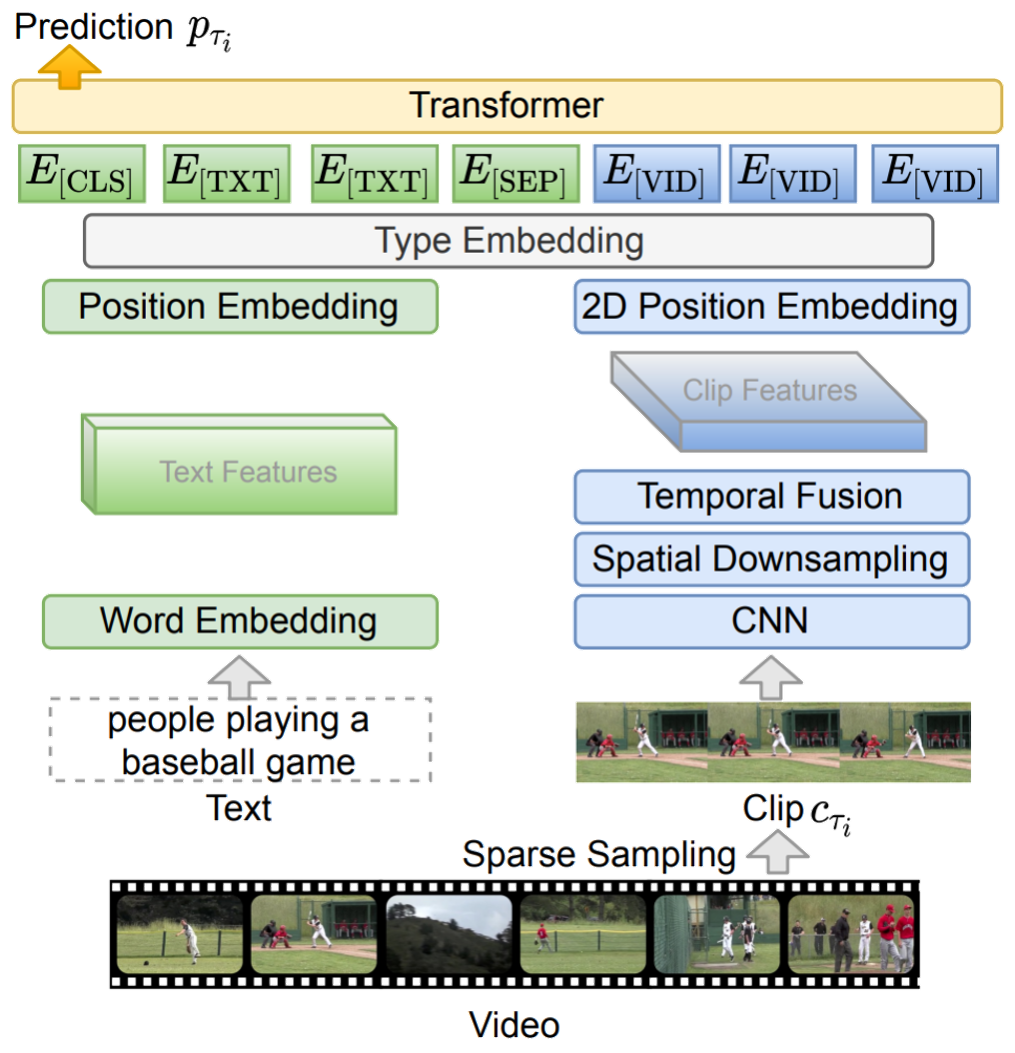}}
\end{minipage}
\caption{Overview of three representative VLP models for video-text tasks: (a) MIL-NCE~\citep{miech19endtoend}, (b) UniVL~\citep{luo2020univl} and (c) ClipBERT~\citep{lei2021less}. Figures are from the corresponding papers.}
\label{fig:chp5_case_study}
\end{figure*}

\paragraph{Case Study.} Until now, we have introduced the general model architecture and popular pre-training tasks in video-text  literature. To provide the readers with more concrete examples, we select three representative models as case studies, including ($i$) MIL-NCE~\citep{miech19endtoend}, a 
dual-encoder model; 
($ii$) UniVL~\citep{lu2022unified}, a fusion encoder model that offline extracts video features; 
and ($iii$) ClipBERT~\citep{lei2021less}, an end-to-end fusion encoder model that directly learns from raw video pixels. We briefly review their architectures and pre-training tasks. 
\begin{itemize}[leftmargin=*]
    \item \textbf{MIL-NCE.} The architecture of MIL-NCE is shown in Figure~\ref{fig:chp5_milnce}. Video is encoded by a 3D CNN backbone (\textit{e.g.}, I3D\citep{carreira2017quo} or S3D~\citep{zhang2018s3d}) to extract 3D grid features, which are then globally mean pooled to obtain the global video embedding. Text sentence is encoded with a pre-trained word2vec embedding, followed by a max-pooling operation to obtain the global text embedding. MIL-NCE is pre-trained with VTC, where the similarity between video-text pairs is measured by the dot product between the two global embeddings. 
    \item \textbf{UniVL.} Figure~\ref{fig:chp5_univl} illustrates the model architecture of UniVL, which contains two single Transformer encoders to embed video and text respectively, a cross-modal Transformer to model the interactions between text and video embeddings, and a Transformer decoder. UniVL follows a two-stage pipeline. First, a off-the-shelf feature extractor (\textit{e.g.}, S3D or ResNet-152\citep{he2016deep}) is used to extract video features from densely sampled frames. Then, these video features along with the accompanying text sentences are fed into UniVL to learn multimodal representations. UniVL is pre-trained with 5 tasks, VTC, VTM, MLM, MVM with in-batch negatives, and a language modeling task that pre-trains the decoder to generate token-by-token autoregressively. 
    \item \textbf{ClipBERT.} As shown in Figure~\ref{fig:chp5_clipbert}, ClipBERT adopts a fusion encoder architecture. A 2D CNN followed by a temporal mean pooling layer is used to encode sparsely sampled frames from each video clip. Text inputs are first encoded with a word embedding layer, and then sent to a multi-layer Transformer for multimodal fusion, along with video features. In comparison to previous two-stage pipelines~\citep{li2020hero,luo2020univl} that offline extract video features with 3D CNNs from densely sampled frames, the video encoding of ClipBERT is less computationally heavy, which makes end-to-end optimization feasible during pre-training and finetuning. ClipBERT is pre-trained with image-text matching and MLM on image-text datasets, COCO~\citep{chen2015microsoftcoco} and VG~\citep{krishna2017visual}. More discussions about leveraging image-text pairs to pre-train video-text models can be found in Section~\ref{sec:chp5_it2vt}.
\end{itemize}

\section{Pre-training Datasets}
\label{sec:vid-txt-pretrain-data}

\begin{figure*}[t!]
    \centering
    \begin{subfigure}[t]{\textwidth}
        \centering
        \includegraphics[width=\linewidth]{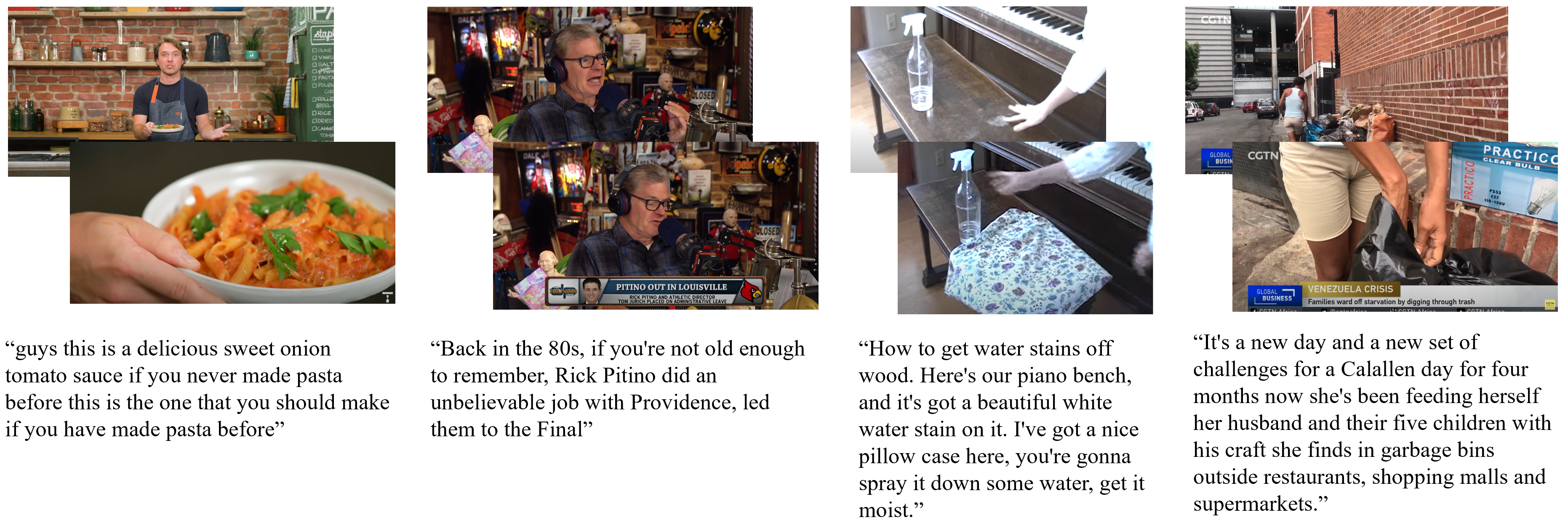}
        \caption{Examples of YouTube video clips, paired with ASR transcripts, sampled from YTTemporal-180M~\citep{zellers2021merlot}.}
        \label{subfig:merlot}
    \end{subfigure}
    \begin{subfigure}[t]{\textwidth}
        \centering
        \includegraphics[width=\linewidth]{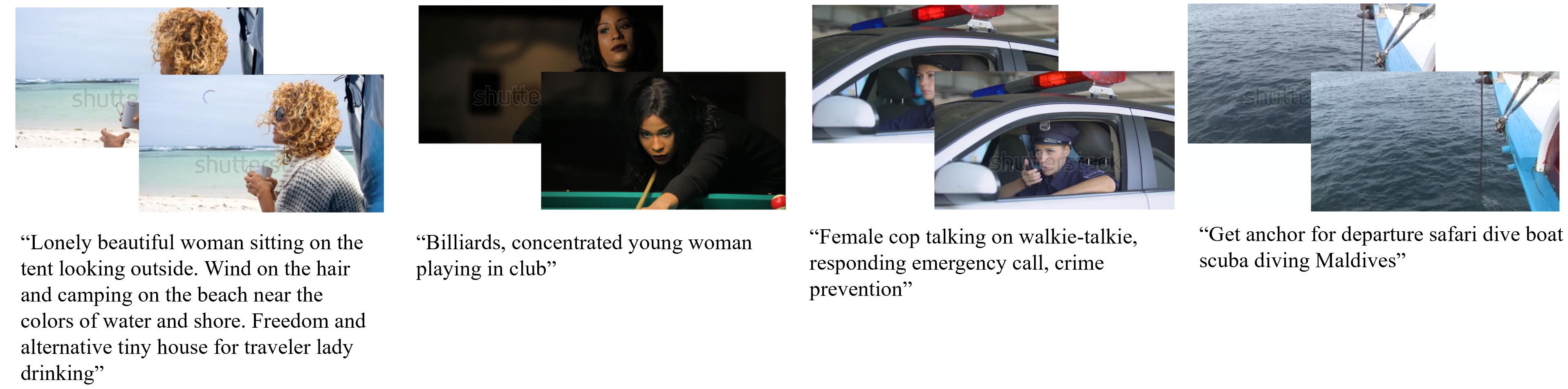}
        \caption{Examples of short videos, paired with alt-text descriptions. Figure credit: Frozen~\citep{bain2021frozen}.}
        \label{subfig:webvid}
    \end{subfigure}
    \caption{Visualization of exemplary video-text pre-training data.}
    \label{fig:chp5_pretrain_data_examples}
\end{figure*}

In contrast to the rapid progress on developing large-scale image-text pre-training datasets, video-text pre-training datasets are harder to collect and often noisier. Most of the video datasets~\citep{miech2019howto100m,zellers2021merlot,zellers2022merlot} stem from YouTube (Figure~\ref{subfig:merlot}). YouTube videos are usually long, with a duration of 6 minutes on average. 
The accompanying texts are generated by Automatic Speech Recognition (ASR), which are often inaccurate.
Alternatively, researchers~\citep{bain2021frozen,pan2020auto} have also tried to scrape video-text pairs from the web (Figure~\ref{subfig:webvid}).
Videos in such datasets are usually shorter (less than 1 minute), and the paired alt-text can describe the global video semantics.  
Below, we introduce these large-scale video-text datasets. 

\begin{itemize}[leftmargin=*]
\item The dataset used in \textbf{VideoBERT}~\citep{sun2019videobert} contains a set of 312K videos with a total duration of roughly 966 days. The videos are obtained by extracting publicly available videos from YouTube with topics related to ``cooking'' and ``recipe'', and then videos longer than 15 minutes are removed. YouTube's ASR toolkit\footnote{\url{https://developers.google.com/youtube/v3/docs/captions}} is utilized to get timestamped speech-transcribed texts from these videos. In the end, 120K videos with texts in English are kept for video-text pre-training. The remaining videos, although without paired texts in English, can be used for video-only pre-training tasks.

\item \textbf{HowTo100M}~\citep{miech2019howto100m} consists of 1.22M instructional videos from YouTube, covering human activities such as cooking, hand crafting, personal care, gardening, \emph{etc}. These videos are collected by searching YouTube videos with ``how to'' text queries (\textit{e.g.,} how to paint furniture). The text queries cover a set of refined ``visual tasks'' from WikiHow\footnote{\url{https://www.wikihow.com/}}, which involve human interactions with the physical world. The text accompanied with each video is also collected from YouTube, either written manually by the content creators or auto-generated by an ASR system. The original long videos are further cut into short clips with an average duration of 4 seconds, which produces 136M clip-text pairs in total. 

\item \textbf{HD-VILA-100M}~\citep{xue2022advancing} features high resolution (720p) videos from YouTube and consists of
100M video clip and sentence pairs from 3.3 million videos with 371.5K hours in total.  Before the emergence of HD-VILA-100M, previous video-text datasets, including both large-scale pre-training datasets~\citep{miech2019howto100m} and downstream benchmarks~\citep{xu2016msr,hendricks2017didemo}, are mostly 240p or 360p. Videos in HD-VILA-100M are collected from 15 popular YouTube categories (\textit{e.g.}, sports, music, autos). During collection, the authors ensure a balanced video clip
number in each category to ease the under-fitting problem. Using the off-the-shelf tool\footnote{\url{https://github.com/ottokart/punctuator2}}, the auto-generated subtitles in YouTube videos are split into complete sentences and aligned to their corresponding clips via Dynamic Time Warping using the timestamp of the original subtitles. 
After processing, each pair in HD-VILA100M consists of a video clip about 13.4 seconds on average and a sentence with 32.5 words on average.

\item \textbf{YTTemporal-180M}~\citep{zellers2021merlot} is derived from 6M YouTube videos, which cover diverse domains and topics. The authors first collect a large amount ($\sim$ 27M) of video candidate ids from YouTube, including instructional videos, lifestyle vlogs of everyday events and some auto-suggested videos by YouTube with topics like ``science'' or ``home improvement''. Then, these candidate videos are filtered using YouTube API and some pre-trained computer vision models. Specifically, they exclude those videos that do not have English ASR tracks, or are over 20 minutes long, or are not visually grounded, or whose thumbnails do not have objects (based on the predictions of a pre-trained image classification model). Like other YouTube-based pre-training datasets, the accompanying texts for these videos are produced using ASR tools, and later processed to add punctuation. The total 6M videos are cut into 180M short clips based on the predicted punctuation added to the ASR texts, which may suggest a sentence ending. This dataset is further augmented with the audio modality and scaled up to 1B (in \# frame-text-audio triplets), namely \textbf{YTTemporal-1B} in \citet{zellers2022merlot}.

\item \textbf{WebVid2.5M}~\citep{bain2021frozen} is inspired by the web-crawled image-text dataset Conceptual Captions (CC3M)~\citep{sharma2018conceptual}. Following a similar collection pipeline, a total of 2.5M text-video pairs were scraped from the same source as CC3M. Although more than 20x smaller than YouTube-based pre-training datasets, WebVid2.5M is of higher quality and widely adopted, in which the texts are manually generated captions, mostly well formed sentences, and can more precisely describe the visual scenes. This dataset has been recently enlarged to \textbf{WebVid10M} with 10M video-text pairs.

\item Auto-captions on GIF (\textbf{AutoGIF})~\citep{pan2020auto} crawls over 160M GIF videos from commercial GIF websites with text queries constructed by extracting objects, actions and subject-verb-object triplets from existing image/video benchmarks. The GIF videos can be viewed as video without audio channel, and it is usually as short as 3 seconds.

\item \textbf{TV Dataset} is first introduced in~\citet{lei2018tvqa}, in which video clips from 6 popular TV series across 3 genres (medical dramas, sitcoms and crime shows) are used to collect a downstream video-language question answering dataset. It consists of 22K video clips from 925 episodes. Each video clip is 60-90 seconds long, covering long-range scenes with complex character interactions and social/professional activities. The accompanying texts are  human-written subtitles, transcribed from the dialogue/conversation happening in the video.
\end{itemize}

\begin{figure*}[t!]
  \centering
    \includegraphics[width=1.0\linewidth]{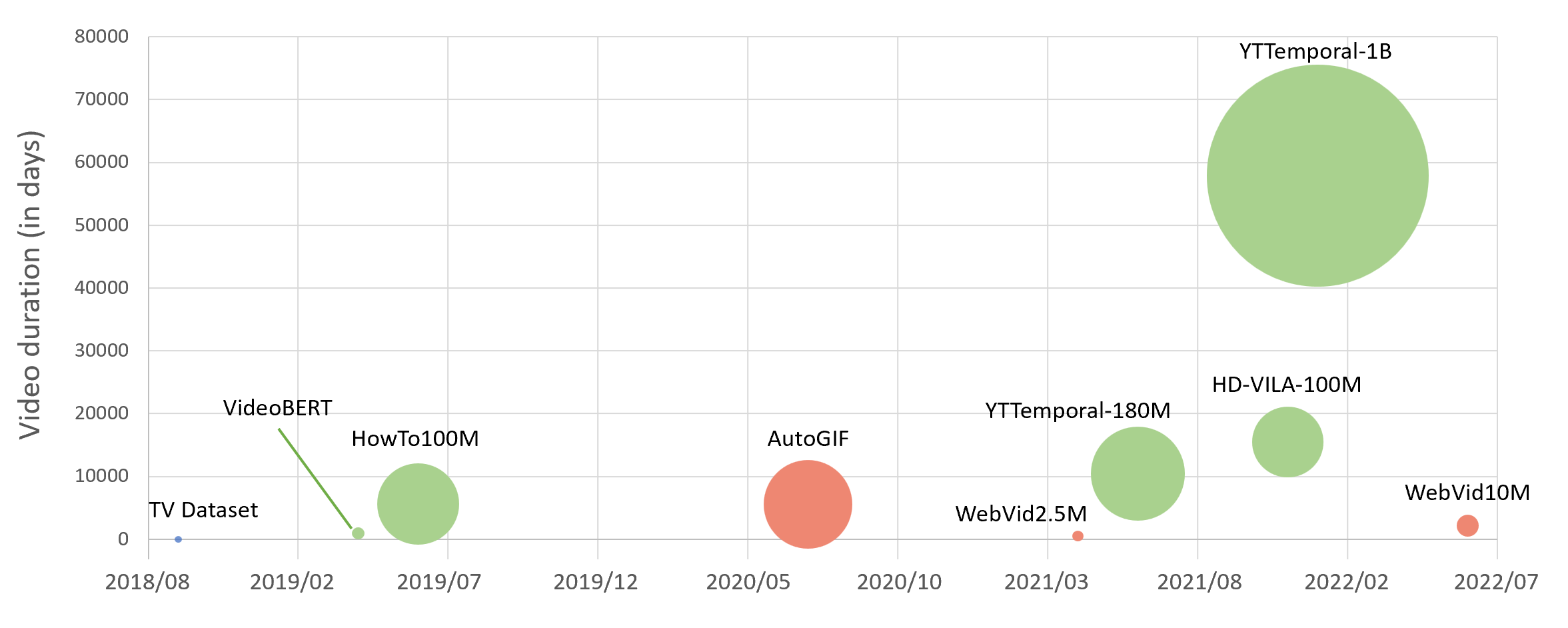}
  \caption{The evolution of video-text pre-training data along time. The x-axis indicates the year and the month that each dataset is released. The y-axis is the total video duration in number of days. The size of the circle indicates the number of video-text pairs in the dataset. We group popular datasets into ($i$) \colorbox{chp5green}{Youtube-based datasets}; ($ii$) \colorbox{chp5orange}{Datasets with short videos and alt-texts}, and ($iii$) \colorbox{chp5blue}{TV-show based dataset}~\citep{lei2018tvqa}.  Youtube-based datasets include the one used in VideoBERT~\citep{sun2019videobert}, HowTo100M~\citep{miech2019howto100m}, HD-VILA-100M~\citep{xue2022advancing}, YTTemporal-180M~\citep{zellers2021merlot} and YTTemporal-1B~\citep{zellers2022merlot}. Datasets with short videos and alt-texts cover AutoGIF~\citep{pan2020auto}, WebVid-2.5M and WebVid-10M~\citep{bain2021frozen}. }
  \label{fig:chp5_pretrain_data_scale}
\end{figure*}

We summarize the characteristics of existing video-text datasets from three perspectives:
\begin{itemize}[leftmargin=*]
\item \textbf{Video Source}: The TV Dataset~\citep{lei2018tvqa} is sourced from popular TV shows while all other datasets are crawled from Internet. It is worth noting that the large-scale datasets (\emph{e.g.}, HowTo100M~\citep{miech2019howto100m}, HD-VILA-100M~\citep{xue2022advancing}, YTTemporal~\citep{zellers2021merlot,zellers2022merlot}) are mostly based on YouTube videos.
\item \textbf{Accessibility}: HowTo100M and Webvid~\citep{bain2021frozen} are released to public with raw videos. Frames extracted at 3 fps are released in the TV dataset, due to copyright concerns. Media URLs are released in the datasets of YTTemporal, HD-VILA-100M and AutoGIF~\citep{pan2020auto}. 
\item \textbf{Scale}: YouTube-based datasets like the HowTo100M, HD-VILA-100M and YTTemporal datasets are  large scale, from 100M to 1B video clips. While other datasets, especially WebVid($<$10M videos) and the TV dataset ($<$22K videos), are smaller. The evolution in scale of popular video-text pre-training datasets along time are depicted in Figure~\ref{fig:chp5_pretrain_data_scale}.
\end{itemize}

As images can be considered as a special case of videos, with temporal size 1,  researchers~\citep{lei2021less,bain2021frozen} have explored to leverage image-text data for video-text pre-training. Popular image-text datasets, introduced in Chapter~\ref{sec:chp3_pretrain_data}, have been recently added to the video-text pre-training corpora, including COCO~\citep{chen2015microsoftcoco}, Visual Genome (VG)~\citep{krishna2016visual}, SBU Captions~\citep{ordonez2011im2text}, Conceptual Captions (CC3M)~\citep{sharma2018conceptual}, and CC12M~\citep{changpinyo2021conceptual}.

\section{Advanced Topics}\label{sec:vid-txt-advanced}
\begin{figure}
    \centering
    \includegraphics[width=0.95\textwidth]{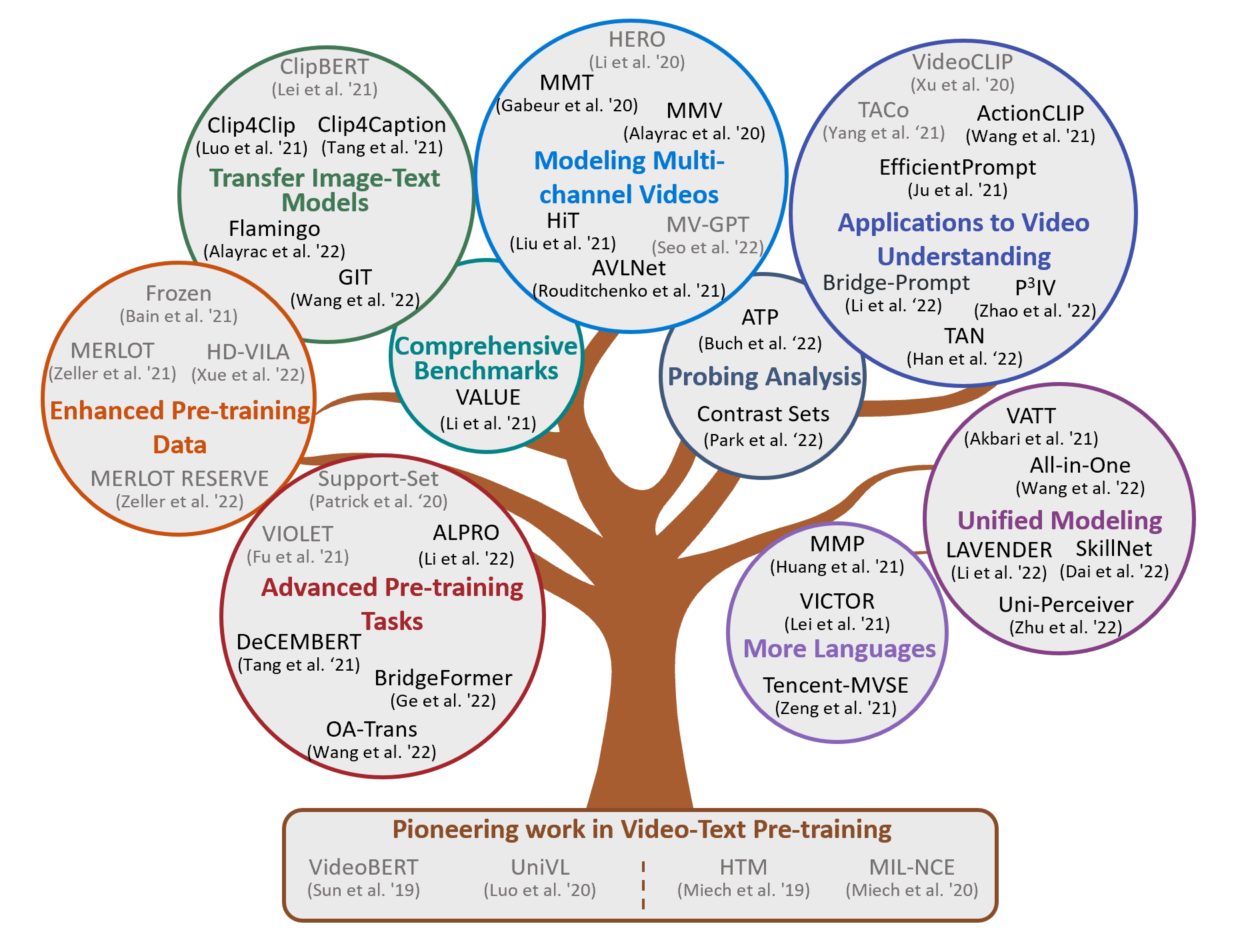}
    \caption{Advanced topics in VLP for video-text tasks. We gray out works that have been covered in previous sections.}
    \label{fig:chp5_advanced_topics}
\end{figure}

In this section, we discuss advanced research topics being explored and future directions, as summarized in Figure~\ref{fig:chp5_advanced_topics}.

\subsection{Advanced Pre-training Tasks}
Apart from the four common pre-training tasks discussed in Section~\ref{sec:vid-txt-pretrain-tasks}, other pre-training objectives have been explored to improve the model training on noisy video-text pairs. 
\citet{tang-etal-2021-decembert} add automatically-extracted dense region captions from the video frames as auxiliary text input, to provide informative visual cues to improve the learning of video and language associations. To alleviate the temporal misalignment issue, it incorporates an entropy minimization-based constrained attention loss to encourage the model to automatically focus on the correct captions from a pool of candidate ASR captions.  
\citet{li2022align} propose a new visually-grounded pre-training task, prompting entity modeling (PEM), to learn fine-grained region-entity alignment. The prediction targets for the PEM task are generated by an entity prompter module, trained with contrastive learning to produce the similarity between a video crop and text prompts instantiated with entity names. During training, the PEM task asks the model to predict the entity pseudo-labels (\textit{i.e.}, normalized similarity scores) for randomly-selected video crops. 
In BridgeFormer~\citep{ge2022bridging}, the authors exploit the rich semantics of text (\textit{i.e.}, nouns and verbs) to build question-answer pairs to form a question answering task as a pretext task, with which the model can be trained to capture more regional content and temporal dynamics. \citet{wang2022object} proposes an object-aware Transformer to leverage bounding boxes and object tags to guide the training process.

\subsection{Transferring Image-text Models to Video-text Tasks}\label{sec:chp5_it2vt}
The studies of ClipBERT~\citep{lei2021less} and Frozen~\citep{bain2021frozen} 
demonstrate that image-text pre-training is effective in improving downstream video-text performance. 
Recent efforts in image-text modeling have also shown that, when scaled up to hundreds of millions~\citep{radford2021learning} or even billions~\citep{li2021align} of image-text pairs, image-text models can achieve state-of-the-art results on various video-text tasks, including text-to-video retrieval~\citep{luo2021clip4clip,yuan2021florence,yu2022coca}, video question answering~\citep{alayrac2022flamingo}, and video captioning~\citep{tang2021clip4caption,wang2022git}. 

The advantages of transferring image-text models to video-text tasks are twofold. 
First, leveraging image-text pre-training or well-pretrained image-text models can potentially save the computational cost of video-text pre-training. 
Second, compared to video-text data, large-scale image-text data are cleaner (\textit{i.e.}, the text description is usually better-aligned to the image content) and are easier to collect (\textit{e.g.,} there are much more image alt-text pairs than video alt-text pairs existing widely over the internet). 
However, existing approaches~\citep{yu2022coca,wang2022git} 
adapt image-text models to video-text tasks by simply concatenating video frames without explicit temporal modeling. 
Such frame concatenation can only work on short videos with sparsely sampled frames, but is ineffective for long videos~\citep{yu2019activitynet} or more challenging tasks that require temporal reasoning~\citep{lei2020tvr}. It is worth exploring more effective ways to transfer image-text models for more challenging video-text tasks. 

\subsection{Learning from Multi-channel Videos}
Videos are multi-channel in nature, which are composed of visual signals from video frames, language cues from speech-transcribed texts, and audio signals from environmental sound or background musics. 
However, most video-text models~\citep{sun2019videobert,miech2019howto100m,zellers2021merlot} mainly focus on vision-language modeling with video frames only (\textit{i.e.}, single-channel videos) to learn video representations and the joint video-language representations. Although video-text tasks defined on most existing video-text datasets~\citep{xu2016msr,chen-dolan-2011-collecting} can be largely solved with single-channel video inputs, we cannot solely rely on these datasets to test the model's capability of video-text understanding. Recent efforts in learning from multi-channel videos have been made for developing new modeling techniques and benchmark datasets.

\paragraph{Multi-channel Video Encoding.} The challenge in multi-channel video-language modeling is how to encode information from all channels in a video. 
\citet{gabeur2020multi} investigate how multi-channel video inputs can help text-to-video retrieval. The proposed MMT model models multi-channel videos with different expert features (\textit{e.g.}, OCR, Speech, Audio, Face, Scene, Motion and Appearance). 
These expert features require different well-supervised expert models, trained for different purposes (\textit{e.g.}, text recognition, speech transcription, audio recognition, face classification, scene classification, action recognition, image classification). 
Follow-up studies~\citep{alayrac2020self,liu2021hit,chen2021multimodal} remove the hassle of extracting all kinds of vision expert features, with just motion and appearance features to represent the visual frames. Another line of work~\citep{rouditchenko2020avlnet,shvetsova2022everything} focus on a variation of the text-to-video retrieval task, text-to-video-audio retrieval, and strive to learn audio encoders from scratch. These audio encoders are randomly initialized, and learned end-to-end through self-supervised multi-modal pre-training on video-text data. 

The aforementioned methods are evaluated on benchmarks~\citep{xu2016msr,chen-dolan-2011-collecting} designed for single-channel videos. They do not test understanding on the other video channels. For more realistic multi-channel video-text applications (\textit{e.g.}, TVR and TVC~\citep{lei2020tvr}),  HERO~\citep{li2020hero} is an good example prior to the end-to-end era. It uses a hierarchical architecture to learn both local temporal alignments between frames and subtitle sentences, and the global temporal context. 
An interesting finding in HERO is that although pre-trained on multi-channel videos, the model can generalize to single-channel video-text tasks~\citep{xu2016msr,hendricks2017didemo}. In addition, when augmenting these single-channel videos with ASR inputs, the performance can be substantially improved. Recently, 
\citet{seo2022end} propose a generative pre-training framework for multimodal video captioning with both visual frames and subtitles as inputs. The proposed MV-GPT model is based on the Transformer architecture and can be end-to-end trained. MERLOT-Reserve~\citep{zellers2022merlot} similarly adopts the Transformer architecture, but takes all three channel inputs (visual frames, subtitles and audio). 
An important finding of the MERLOT-Reserve study is that  video-text pre-training with audio can help visual commonsense reasoning~\citep{zellers2019recognition}, an audio-less image-text task.
    
\paragraph{Multi-channel Video-text Benchmarks.} TVQA~\citep{lei2018tvqa}, TVQA+~\citep{lei2019tvqa}, TVR and TVC~\citep{lei2020tvr} are attempts to building video-text datasets on multi-channel video inputs in TV show domain. 
Specifically, during data collection, annotators are instructed to write text descriptions/QA pairs given the context provided in visual frames only, or subtitles only, or both visual frames and subtitles. Following the same procedure, How2R and How2QA~\citep{li2020hero} are created to cover additional video domain of instructional videos. 

Another problem in model evaluation is that existing video-text models~\citep{xu2021vlm,yang2021just,tang-etal-2021-decembert} are often evaluated on their own choices of downstream datasets, which makes it hard to compare between models. 
We expect that a general video-language system should do well on diverse tasks/domains/datasets, as we have witnessed in the NLP field that publicly accessible large-scale multi-task benchmarks~\citep{wang2018glue,wang2019superglue} can facilitate advances in modeling. 
With the above motivation, 
VALUE~\citep{li2021value} is a first attempt to build a comprehensive benchmark for video and language understanding evaluation.  
There are four characteristics of VALUE benchmark. ($i$) It features multi-channel videos, with video frames and subtitle as video inputs. ($ii$) The videos in VALUE are collected from diverse video domains, including movie, TV shows, instructional videos and vlogs. ($iii$) VALUE includes 11 datasets over 3 representative tasks: text-to-video retrieval, video question answering and video captioning. ($iv$) VALUE supports a live leaderboard to track the advances in video-and-language research. 

\subsection{VLP for Core Video Tasks}
Contrastive video-text pre-training~\citep{miech19endtoend,yang2021taco,xu2021videoclip} has shown promising results on video action recognition~\citep{Kuehne11,kay2017kinetics}, action localization~\citep{abu2016youtube,zhukov2019cross}, and action segmentation~\citep{tang2019coin}. TAN~\citep{han2022temporal} enhances the dual-encoder architecture trained with VTC by adding a temporal alignment network to tackle long-term video understanding. For procedure planning in instructional videos, \citet{zhao2022p3iv} propose a weakly supervised method of learning models from natural language instructions in HowTo100M~\citep{miech2019howto100m}. Given that contrastive image-text pre-training~\citep{radford2021learning,li2021align} is beneficial to learning image representations, a line of work~\citep{wang2021actionclip,ju2021prompting,li2022bridge} try to prompting CLIP~\citep{radford2021learning} to perform video action recognition.  Leveraging VLP for other core video tasks, such as video object detection~\citep{damen2018scaling} and video object segmentation~\citep{Perazzi2016}, is an interesting direction to explore in future.

\subsection{Analysis on Video-text Benchmarks/Models}
In parallel to improving downstream video-text performance, researchers try to analyze existing video-text benchmarks and models
to building better benchmarks or design better models. 

On the \textbf{benchmark} front, there are several works questioning whether the existing video-language tasks require temporal reasoning.  In Singularity~\citep{lei2022revealing}, the authors show that training a model with a single frame is sufficient in improving state-of-the-art on many existing video-text tasks. Thus, they propose a new dataset built upon Something-Something~\citep{goyal2017something} that requires temporal reasoning. 
\citet{buch2022revisiting} report similar findings, and propose a temporal probing method for identifying temporally challenging data, to disentangle a subset in existing benchmarks. \citet{wray2021semantic} focus on the video retrieval task, and find the assumption that only a single caption is relevant to the query video and vice versa, often does not hold. The instance-based evaluation protocol with Recall@K, can wrongly penalize relevant captions. Hence, they propose an alternative protocol to measure the semantic similarity between the query and retrieved instance to better measure the relevance. 

On the \textbf{modeling} front, 
\citet{park-etal-2022-exposing} identify the weakness in video-language models through some text manipulations, and report that a pre-trained video-language model can be easily fooled, which indicates that the models may rely on some spurious clues in the training data.

\subsection{Unified Modeling for Video-text Understanding} \label{sec:unified_modeling_video}
How to design a unified VL model that can support various downstream VL tasks without introducing task-specific heads is a popular topic in image-text modeling (detailed in Section~\ref{sec:chp3_adv_topics}). 
Similar attempts have been made in video-text understanding along two dimensions.

\begin{itemize}[leftmargin=*]
\item \textbf{One Transformer for all:} Transformer~\citep{vaswani2017attention} and Transformer-based pre-trained models~\citep{devlin2018bert,dosovitskiy2020image} have revolutionized a wide range of research fields, \emph{e.g.}, natural language processing~\citep{devlin2018bert,liu2019roberta}, computer vision~\citep{dosovitskiy2020image, liu2021swin}, and speech processing~\citep{baevski2020wav2vec,chen2022wavlm}. Motivated by this, researchers have tried to further narrow the modeling gaps among different modalities by using a shared Transformer for video-text modeling. 
For example, All-in-one~\citep{wang2022all} encodes raw video and textual signals into joint representations using a unified backbone architecture. VATT~\citep{akbari2021vatt} uses a modality-agnostic Transformer shared across video, text and audio inputs.  Similarly, Uni-Perceiver~\citep{zhu2022uni} shares the backbone weights across various modalities, such as image, video and text.  
OmniVL~\citep{wang2022omnivl} is a universal architecture to support both image-text and video-text tasks. SkillNet~\citep{dai2022one} uses a sparsely activated Transformer (\emph{i.e.,} mixture-of-experts~\citep{shazeer2017outrageously} where different parts of the parameters are specialized to processing different modalities, including text, audio, image, video, and code. 

\item \textbf{Unifying video-text tasks as text generation.} 
Inspired by the unifying efforts in image-text modeling~\citep{cho2021unifying,wang2021simvlm}, LAVENDER~\citep{li2022lavender} focuses on integrating different video-text tasks into a unified format so that a single architecture can be used for all tasks. Specifically, all pre-training and downstream tasks are reformulated as masked language modeling, so that a single task head is used for both pre-training and downstream finetuning, without introducing additional task-specific heads. With the unified architecture, LAVENDER can support all downstream tasks with just a set of shared parameter values when multi-task finetuned, showing strong generalizability on downstream tasks with limited training examples, and enabling zero-shot prediction on video question answering tasks. 
However, LAVENDER has several limitations that suggest two future improvements: ($i$) extensions to fine-grained video-text tasks (\textit{e.g.}, video corpus moment retrieval~\citep{lei2020tvr}), as current LAVENDER only supports video retrieval; and ($ii$) more effective in-context few-shot learning or prompt tuning to better leverage and improve the generalizability of LAVENDER.
\end{itemize}

\subsection{Multi-lingual VLP for Video-text Tasks}
The majority of the literature in video-text understanding focus on English-only video-text tasks, while we live in a multilingual world. Due to the missing of large-scale non-English video-text datasets for both pre-training and downstream evaluation, video-text models in non-English languages are less explored.  As initial attempts, 
~\citet{lei2021understanding} and 
~\citet{zeng2019tencent} have developed large-scale Chinese video-text datasets. 
\citet{huang2021multilingual} crawl multilingual subtitles for each video in HowTo100M~\citep{miech2019howto100m} in 9 languages, covering English, German, French, Russian, Spanish, Swahili, Chinese and Vietnamese, to develop a new multilingual instructional video dataset (MultiHowTo100M).

\chapter{VL Systems in Industry}
\label{chp:industry}


As the technology of vision-language (VL) learning advances rapidly, more and more companies are integrating VL capabilities into their products and services. iPhone can automatically generate image captions which are read by VoiceOver so that vision-impaired users know what is in the image. Chrome OS has the capability to generate image captions in 10 different languages for unlabeled web images. Microsoft offers image captioning as an Azure cloud service, and Microsoft Office applications (\textit{e.g.}, PowerPoint and Word) use this image captioning service to generate image descriptions automatically. Seeing AI, which is a mobile app for the blind and low vision community currently available on iPhone, has a channel to describe a scene using automatically generated image captions. In addition to image captioning, we believe many other VLP-enabled technologies, such as open vocabulary image classification and object detection, will be deployed into products and services in industry.

\section{VL in Commercial Systems}\label{sec:vl_systems}
We review a variety of commercial systems that are empowered by vision-language learning. Due to the proprietary nature of many of these systems, we limit our review to summarizing published material about the systems, including image captioning in Microsoft Office (\emph{e.g.}, Word and PowerPoint) and LinkedIn, Seeing AI, text-image search in Bing, multi-modal modeling via Microsoft Cognitive Services, \emph{etc}. 

\begin{figure*}[t!]
  \centering
   \subfloat[\label{fig:chap6_ppt_a}\centering ]{\includegraphics[width=7cm]{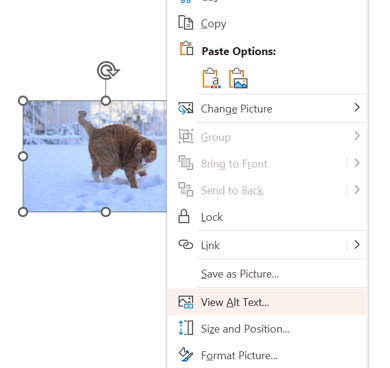}}
  \qquad
  \subfloat[\label{fig:chap6_ppt_b}\centering  ]{\includegraphics[width=5cm]{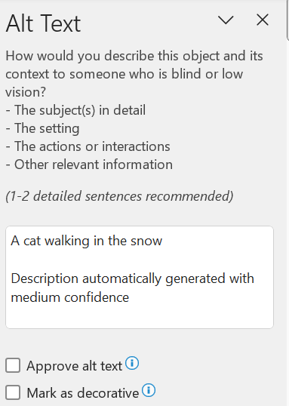}}
  \caption{Microsoft PowerPoint automatically generates image descriptions for user-inserted images. (a) After inserting an image into PowerPoint and right clicking on it, a drop-down menu pops up. Select ``View Alt Text'' to automatically generate image description. (b) The generated image description ``A cat walking in the snow'' is displayed in the text box. A user can also edit the generated image description.}
  \label{fig:chp6_PPT}
\end{figure*}

\begin{figure*}[t!]
  \centering
  \includegraphics[width=7cm]{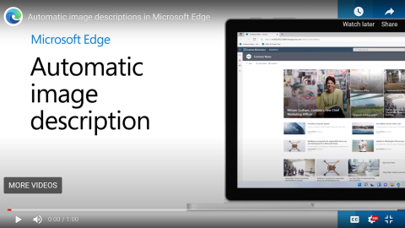}
  \caption{Microsoft Edge browser automatically generates image descriptions (alt text) which are then read out via a text-to-speech engine of a screen reader.}
  \label{fig:chp6_edge}
\end{figure*}

\begin{figure*}[t!]
  \centering
   \subfloat[\label{fig:chp6_seeingai_a}\centering ]{\includegraphics[width=5cm]{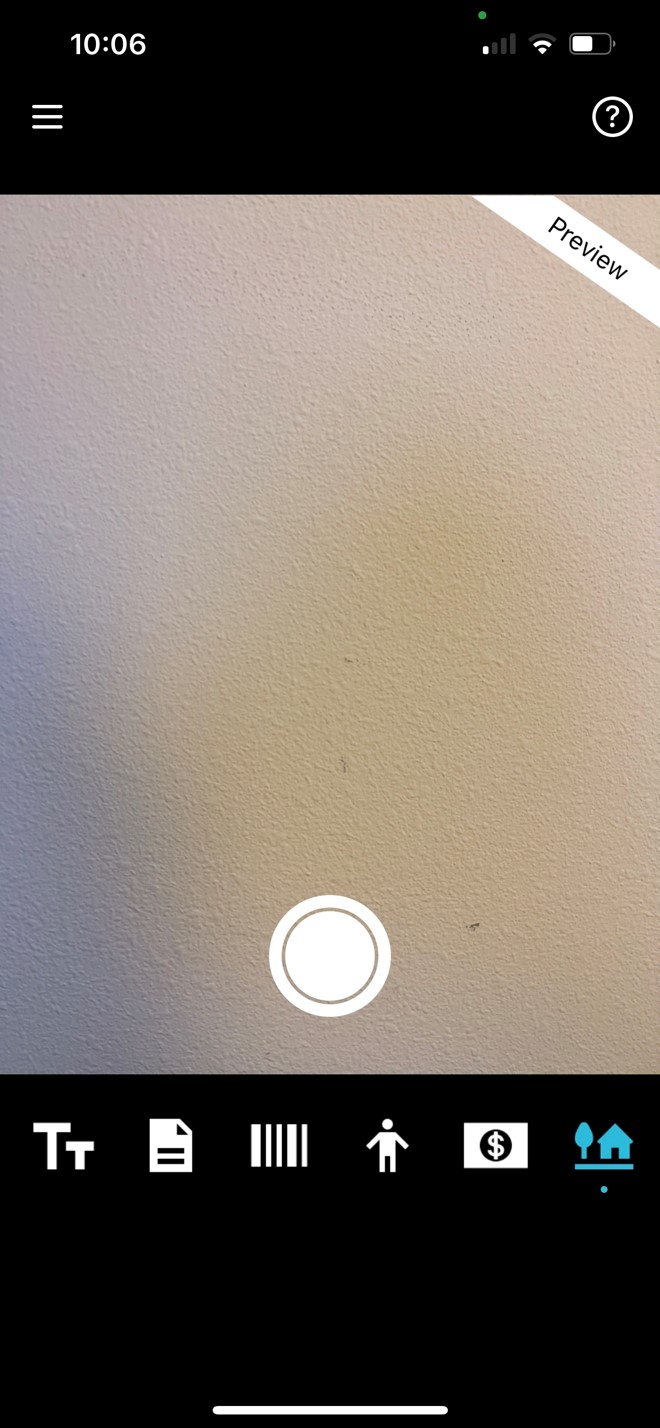}}
  \qquad
  \subfloat[\label{fig:chp6_seeingai_b}\centering  ]{\includegraphics[width=5cm]{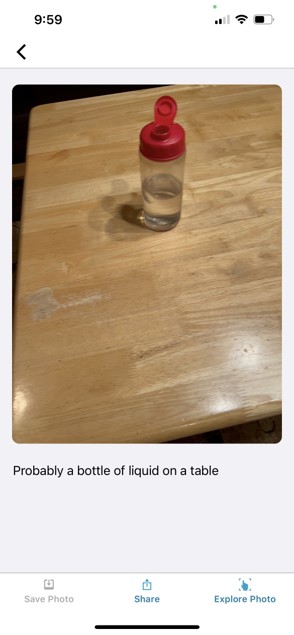}}
  \caption{Seeing AI is designed for those who are blind or have low vision. It is currently available in iPhone App store. (a) A screen shot of Seeing AI where the button at the bottom right invokes Scene Description feature. (b) The generated image caption for the captured image, which will be read out via a text-to-speech engine.}
  \label{fig:chp6_seeingai}
\end{figure*}

\begin{figure*}[t!]
  \centering
   \subfloat[\label{fig:chp6_facebook_a}\centering ]{\includegraphics[width=5cm]{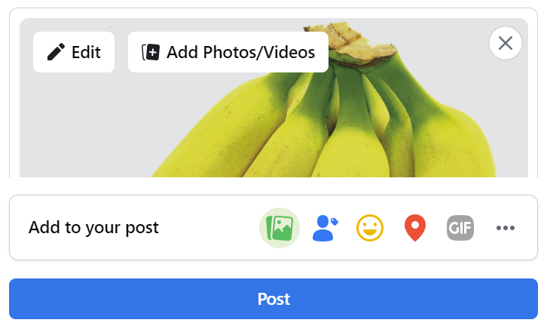}}
  \qquad
  \subfloat[\label{fig:chp6_facebook_b}\centering  ]{\includegraphics[width=5cm]{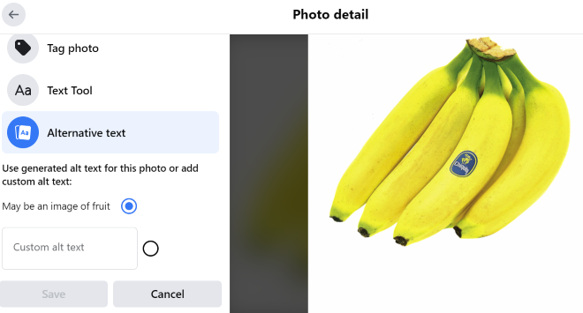}}
  \caption{Facebook automatically generates alt text for images uploaded by users. (a) An image is uploaded to Facebook to create a Post, and an``Edit" button appears. (b) When the ``Edit" button is clicked, the system automatically generates an image description. It also allows the user to manually enter a corrected or preferred description. }
  \label{fig:chp6_facebook}
\end{figure*}

\begin{itemize}[leftmargin=3.0mm]
    \item \textbf{Auto Alt-text in Microsoft PowerPoint/Word/Outlook.} After we insert a picture into Microsoft PowerPoint, we can right click on the image and a dropdown menu will appear (see Figure~\ref{fig:chap6_ppt_a}). With the selection of ``View Alt Text'',  a dialog window will appear as shown in Figure~\ref{fig:chap6_ppt_b}. It will call the image captioning API of Microsoft Cognitive Service, and display the image captioning result (``A cat walking in the snow'' for this example image). 
    
    \item \textbf{Microsoft Edge.} In March 2022, Microsoft announced that its Edge browser will start to provide automatically-generated alt text for images that do not have alt text (see Figure~\ref{fig:chp6_edge}). The image descriptions will be read out via a text-to-speech engine by a screen reader (such as Narrator) when a user is browsing the web. To use this feature, the user needs to change a setting in the Microsoft Edge browser. Detailed instructions can be found from the web page.\footnote{\href{https://blogs.windows.com/msedgedev/2022/03/17/appears-to-say-microsoft-edge-auto-generated-image-labels/}{Appears to say: Microsoft Edge now provides auto-generated image labels}}

    \item \textbf{Google Chrome.} Google Chrome OS provides an accessibility extension that allows users to obtain image descriptions automatically for those images that do not have alt text.  The generated image descriptions will be read out by a screen reader. It supports multiple languages including English, French, German, Italian, Russian, Spanish, \emph{etc}. Detailed instructions can be found from the web page.\footnote{\href{https://support.google.com/chrome/answer/9311597?hl=en&co=GENIE.Platform}{Get image descriptions on Chrome}} 

    \item \textbf{Microsoft Seeing AI.} Seeing AI is a mobile app that is currently available on iPhone app store. The app is designed for those who are blind or have low vision. Figure~\ref{fig:chp6_seeingai_a} shows a screenshot of Seeing AI. The button at the bottom right performs scene description. It calls image captioning API (Microsoft Cognitive Service) to automatically generate alt text which is then read out via a text-to-speech engine. Figure~\ref{fig:chp6_seeingai_b} shows an image captured by the phone, and the generated image caption is shown below.
    
    \item \textbf{Facebook.} Facebook provides the feature to automatically generate alt text for user-uploaded images. When we create a post in Facebook and upload an image as shown in Figure~\ref{fig:chp6_facebook_a}, a button ``Edit'' will pop up. By clicking on the ``Edit'' button, the system will automatically generate an image description as shown in Figure~\ref{fig:chp6_facebook_b} (``May be an image of fruit'' in this example). More detailed instructions can be found from the webpage.\footnote{\href{https://www.facebook.com/help/214124458607871}{How do I edit the alternative text for a photo on Facebook?}}

    \item \textbf{Apple iOS VoiceOver Screen Reader.} VoiceOver is the screen reader built into IOS, the operating system on Apple’s mobile devices. With VoiceOver, users with visual impairment can use a few simple gestures to hear aloud what is displayed on the screen for people who are sighted.
    
\end{itemize}

There are also many VL models hosted in cloud services, detailed below.
\begin{itemize}[leftmargin=3.0mm]
    \item \textbf{Microsoft Azure Computer Vision - Cognitive Services.} Microsoft Azure Computer Vision is an AI service that analyzes and extracts rich information from images and videos. The Image Analysis service extracts many visual features from images, such as objects, faces, adult content, and auto-generated text descriptions. One can follow the Image Analysis quickstart to have a try.
The Spatial Analysis service analyzes the presence and movement of people on a video feed and produces events that other systems can respond to. One can also install the Spatial Analysis container to get started.

    \item \textbf{Google Cloud Vision AI.} Google Cloud Vision AI provides two computer vision products to help you understand images. \textbf{AutoML Vision} automates the training of your own custom machine learning models. Simply upload images and train custom image models with AutoML Vision's easy-to-use graphical interface; optimize your models for accuracy, latency, and size; and export them to your application in the cloud or to an array of devices at the edge. \textbf{Vision API} offers powerful pre-trained machine learning models through REST and RPC APIs. Assign labels to images and quickly classify them into millions of predefined categories. Detect objects and faces, read printed and handwritten text, and build valuable metadata into your image catalog.
    
    \item \textbf{Amazon Rekognition.} Amazon Rekognition offers pre-trained and customizable computer vision capabilities to extract information and insights from your images and videos. The AI service includes key features such as Content moderation, Face compare and search, Face detection and analysis, Labels, Custom labels, Text detection, Celebrity recognition, Video segment detection, and Streaming video events detection.
    
    \item \textbf{Alibaba Cloud Image Search.} Image Search allows users to search by image based on image similarities. Image Search uses deep learning and machine vision to capture characteristics of images and then search for images based on the captured information. \textbf{Search by Product Images} allows customers to use a product image to search for the same product or similar products in your self-managed image library. Then, the system returns information about the product images. \textbf{Search by General-Purpose Image} allows users to use an image to search for images that contain the same elements or objects from a self-managed image library. The system returns the same or similar images based on the captured image information.
    
\end{itemize}

How are the current prevailing big foundation models changing the industry? Besides the fact that big foundation models demonstrate superior performance on a variety of downstream vision and vision-language tasks, more importantly, they are also fundamentally changing the way the industry collects data, develops models, delivers services, and builds their R\&D organizations.



\section{Issues in VL Model Deployment}

On one hand, as discussed in Section~\ref{sec:vl_systems}, it is very encouraging to see that there are already many VL systems deployed in industry. On the other hand, 
there are also many factors that one has to consider when deploying a VL model to real-world applications, including robustness to new domains, inference cost and latency, fairness, and responsible AI issues. Since VL learning is still a relatively new field, research on these practical issues is preliminary. As more and more applications start to deploy VL models, we expect the demand for solutions to these practical issues to become increasingly strong, which will inspire more research in these areas.
In this section, we review solutions to three fundamental issues regarding deploying VL systems for real-world applications, domain adaptation, serving cost, and responsible AI. 

\paragraph{Domain Adaptation.}
The images in the real-world scenarios are usually unpredictable and have large variations, and the model must be robust and generalize well to new domains. Though there has been a lot of work on domain adaptation in other computer vision areas such as image classification and object detection, there has been little research on domain adaptation for VL models. One interesting domain is the non-natural images such as diagrams, tables, and charts. So far, most VL datasets are composed of natural images, how to handle non-natural images remains unaddressed.

\paragraph{Serving Cost.} 

In addition to accuracy, there are also constraints on the inference cost and latency. How to reduce model size without sacrificing accuracy is an important problem for real-world applications, and there has been a lot of research devoted to it. For example,  \cite{wang2020minivlm} developed a small VL model called MiniVLM that reduces model size by 73\% and the inference time cost by 94\% while being able to retrain 94-97\% of the accuracy on multiple VL tasks. \cite{fang2021compressing} developed a knowledge distillation technique to compress a Transformer-based large VL model into a small VL model. Inspired by  the lottery ticket hypothesis~\citep{frankle2019lottery},  \cite{gan2021playing} found that lottery tickets also exist for VLP models such as UNITER, LXMERT, and ViLT. They were able to discover ``relaxed'' winning tickets at 50\%-70\% sparsity that maintain 99\% of the full accuracy.

\paragraph{Fairness and Responsible AI.}

Fairness and responsible AI is another critical issue when deploying a VL model. As VL datasets typically contain biases, the trained models usually have biases as well, which may generate results that appear offensive.  As reported by \cite{zhao2021understanding}, 
the COCO image captioning dataset~\citep{chen2015microsoftcoco} is heavily skewed towards lighter-skinned and male individuals. In addition, there are racial terms including racial slurs in the manually annotated captions.  Given the biases in the training data, it is difficult to avoid biases in the trained models. For example, as reported by \cite{srinivasan2021worst}, 
the VL-BERT model exhibits gender biases where it sometimes may prefer to reinforce a stereotype over faithfully describing the visual scene. 

There has been research work to reduce biases in VL models.
\cite{cadene2019rubi} proposed a learning strategy for VQA to reduce the importance of the most biased examples that can be correctly classified without looking at the image, thus forcing the model to use both input modalities instead of relying on statistical regularities between the question and the answer. \cite{kv2020reducing} proposed a technique to reduce the dependency of a VQA  model on the language prior by using a model-agnostic question encoder that utilizes
both visual and language modalities equally while encoding the question. Despite the progress, how to eliminate biases in general VL models is still an open problem. Since real-world data contains a lot of biases, scalable solutions have yet to  be developed to eliminate data biases during data collection and curation.

Fairness and Responsible AI have also been studied in the context of conversational AI agents (bots). VL systems and conversational AI agents share many problems. Recently, VL models have also been incorporated into AI bots for social chat and task completion \citep[\emph{e.g.},][]{gao2019neural,zhou2020design,gao2022neural}.
The design of AI bots needs to defend against harms and mitigate potential toxicity and bias -- either in training data, introduced through feedback and usage, or simply inappropriate due to local culture or context \citep[\emph{e.g.},][]{breitfeller2019finding,zhang2018conversations}.
The 10 guidelines described in \cite{MicrosoftResponsibleAIBots2018} outlines what need to be considered to develop a responsible AI bot that meets the challenges in real-world situations.


\chapter{Conclusions and Research Trends}
\label{chp:conclusion}

Vision-Language Pre-training (VLP) has attracted rapidly growing attention from both the computer vision and NLP communities, especially due to the emergence of large-scale multimodal foundation models like CLIP~\citep{radford2021learning}, DALL-E~\citep{ramesh2021dalle}, CoCa~\citep{yu2022coca}, Flamingo~\citep{alayrac2022flamingo}, and Florence~\citep{yuan2021florence}. In this chapter, we provide a concise summary of what has been reviewed, and discuss the current research trends. 

\section{Summary and Conclusions}
This paper surveys the most recent advances at the frontier of VLP research, including ($i$) VLP for image-text tasks; ($ii$) VLP for core vision tasks; and ($iii$) VLP for video-text tasks. Specifically, we have discussed the following. 
\begin{itemize}[leftmargin=*]
    \item \textbf{Task-specific models.} To lay a comprehensive foundation for the introduction of VLP models, we have discussed many seminar papers before the era of pre-training. One major theme during this period is the design of various attention mechanisms. 
    We have introduced how the field has been moving from ($i$) \emph{inter-modality} attention design, which aims to capture multimodal alignment and perform multimodal fusion, to ($ii$) \emph{intra-modality} attention design, which aims to capture visual relations among image regions, \textit{e.g.,} via graph attention networks, to ($iii$) the convergence to the Transformer architecture, which models both inter- and intra-modality interactions. Besides the attention design, we have also briefly discussed topics regarding bilinear pooling methods for multimodal fusion, neural module networks for compositional visual reasoning, and so on. 
    
    \item \textbf{VLP for image-text tasks.} We have covered VLP models for image captioning, visual question answering, image-text retrieval, and visual grounding. A general methodology transition is from the development of OD-based VLP models (which requires the extraction of image regional features first from an offline pre-trained object detector) to the prevailing end-to-end VLP models, partially due to the popularity of vision Transformer. Early VLP models are only pre-trained on approximately 4M images (with roughly 10M image-text pairs), while the most recent big VLP models are already pre-trained over 10B image-text pairs. We have also discussed many advanced topics, ranging from unified image-text modeling, few-shot learning, knowledge, robustness, multilingual VLP, to model compression and efficient adaptation.
    
    \item \textbf{VLP for core vision tasks.} We have reviewed VLP models for core vision tasks, including image classification, object detection, and segmentation. These language-augmented visual models demonstrate a strong \emph{zero-shot} transfer capability, since they acquire \emph{open-set} and \emph{open-vocabulary} recognition abilities through problem reformulation, casting image classification as image retrieval or object detection as phrase grounding. Moreover, model generalization is improved as natural language supervision typically contains much richer semantics. We advocate the concept of \emph{computer vision in the wild},\footnote{\href{https:https://github.com/Computer-Vision-in-the-Wild/CVinW_Readings/blob/main/README.md}{Computer-Vision-in-the-Wild Readings.}} and encourage the development and evaluation of future foundation models for this. We have also discussed many advanced topics, ranging from benchmark, knowledge, robustness, efficient adaptation, to open-set video classification.  
    
    \item \textbf{VLP for video-text tasks.} We have discussed VLP models for video-text tasks, including video retrieval, question answering and captioning. We observe the same research trend as that of image-text, a transition from the use of offline extracted video features to end-to-end VLP models via \emph{e.g.,} the use of video Transformers. We have provided an in-depth discussion on the design of model architectures, pre-training tasks, and the widely used pre-training datasets to date. We have also covered diverse advanced topics, such as learning from multi-channel videos, the adaptation of image-text models for video-text tasks, VLP for core video tasks, and unified video-text modeling. 
\end{itemize}

\paragraph{Text-to-Image Generation.}
In the context of VLP, text-to-image generation methods can be classified into two categories: ($i$) VQ-token-based auto-regressive methods, such as DALL-E~\citep{ramesh2021dalle},  Make-A-Scene~\citep{gafni2022make}, NUWA-Infinity~\citep{wu2022nuwa}, and Parti~\citep{yu2022scaling}; and ($ii$) diffusion-based methods, such as  DALL-E 2~\citep{ramesh2022hierarchical},  Imagen~\citep{saharia2022photorealistic}, and Stable Diffusion~\citep{rombach2022high}. We provide a brief discussion on this important topic in Section~\ref{sec:vlp4imggen}. This field is rapidly growing, we leave a detailed survey of this topic to future work.  

\section{Towards Building General-Purpose Multimodal Foundation Models}

In the last section of each chapter, we have discussed many advanced topics. One common theme that stands out is how to build a \emph{general-purpose} multimodal foundation model. That is, we aim to build one foundation model that is \emph{scalable} and \emph{generalizable}, and can be readily adopted to various downstream tasks, ranging from image-level vision tasks (\emph{e.g.}, image classification, retrieval, and captioning), region-level vision tasks (\emph{e.g.}, object detection and phrase grounding), to pixel-level vision tasks (\emph{e.g.}, segmentation and image generation). This also aligns with the grand vision of building a single \emph{Generalist Agent} (\emph{e.g.}, Gato~\citep{reed2022generalist}) that can perform a wide range of tasks with a single set of model weights. In what follows, we highlight a number of research trends towards this goal. 

\paragraph{Unified Modeling.} In order to build a general-purpose foundation model, we need a unified model architecture that can be readily scaled up; and when being pre-trained at scale, it can be readily adopted to various downstream computer vision and vision-language (VL) tasks. From this unification, we envision that new model capabilities will be unlocked.
There are different levels of unification. For example, unification of different VL understanding tasks can be achieved relatively easily (\emph{e.g.}, SimVLM~\citep{wang2021simvlm}, GIT~\citep{wang2022git}, CoCa~\citep{yu2022coca}), while the unification of VL understanding tasks and region-level localization tasks can be much more challenging (\emph{e.g.}, UniTAB~\citep{yang2021crossing}, GLIP~\citep{li2021grounded}, and GLIPv2~\citep{zhang2022glipv2}), not to mention the unification of image generation tasks (\emph{e.g.}, OFA~\citep{wang2022ofa} and Unified-IO~\citep{lu2022unified}). Pix2seqV2~\citep{chen2022unified} and UViM~\citep{kolesnikov2022uvim} also propose unified approaches for computer vision tasks.   
MetaLM~\citep{hao2022language} shows that language models can be a general-purpose interface for many diverse tasks. We envision that more research efforts will be devoted to unified modeling. See Section~\ref{sec: unified_modeling} and \ref{sec:unified_modeling_video} for more detailed discussions.

\paragraph{Computer Vision in the Wild.} How can natural language play a more fundamental role in computer vision tasks? In Chapter~\ref{chp:vlp4vision}, we have shown how visual recognition tasks (\emph{e.g.}, image classification, object detection, and segmentation) can be considered as VL problems, and the unification of computer vision and VL tasks is possible. Distinct from the traditional close-set recognition setting, the use of natural language supervision enables open-set and in-the-wild visual recognition (see Section~\ref{sec:vlp_vision_trend} for a detailed discussion), and we envision more research efforts will be devoted to language-augmented computer vision models, and VLP can have the potential to become a mainstream and impactful direction in computer vision research. This also requires better benchmarks for evaluation of the performance of such computer vision foundation models, ranging from \emph{zero-shot} generalization (\emph{i.e.}, how the model performs ``out of the box''), few-shot evaluation, linear probing, prompting, to model finetuning.

\paragraph{Model Scaling.} In recent years, we have witnessed great successes from scaling up language models, via training large Transformers from massive amounts of text data. Prominent examples include T5~\citep{raffel2020exploring}, GPT-3~\citep{brown2020language}, Megatron-Turing~\citep{shoeybi2019megatron}, Chinchilla~\citep{hoffmann2022training}, OPT~\citep{zhang2022opt}, and PaLM~\citep{chowdhery2022palm}.
A crucial benefit of scaling is the potential of zero-shot and few-shot generalization. 
VLP models have been following a similar trend, with examples including SimVLM~\citep{wang2021simvlm}, Florence~\citep{yuan2021florence}, CoCa~\citep{yu2022coca}, GIT~\citep{wang2022git}, BEiT-3~\citep{wang2022image}, PaLI~\citep{chen2022pali}, and Flamingo~\citep{alayrac2022flamingo}. However, compared with the scale of language models, the scaling for VLP models is still in its infant stage. We envision that bigger VL models, especially open-sourced ones~\citep{ilharco_gabriel_2021_5143773}, will appear in near future. It would also be interesting to investigate the emergent abilities of such big models once they become available. See Section~\ref{sec:big_models} for a more detailed discussion.

\paragraph{In-context Few-shot Learning.} Can we train
a model that can quickly adapt to different downstream tasks with only a few in-context
examples? By inheriting this capability form large \emph{frozen} language models, Flamingo~\citep{alayrac2022flamingo} has shown that this is possible for tasks with text outputs, \textit{e.g.}, question answering, captioning and classification. However, due to the diversity of vision tasks, often we require the model to not just output text sequences. It remains unknown how in-context few-shot learning can be enabled for complex tasks, such as localization, where bounding boxes or even pixel output are needed. See Section~\ref{sec:few_shot} for a more detailed discussion.

\paragraph{Efficient Adaptation.} As the size of VL models has been increasing rapidly, it becomes increasingly important to develop methods to adapt big VLP models efficiently for downstream tasks. By freezing the model weights, different parameter-efficient transfer learning methods have been developed, especially for the few-shot setting (please check Section~\ref{sec:vlp_vision_advanced_topics}). As we have access to more and more large foundation models, this topic becomes timely and urgent. 

\paragraph{Knowledge.} On one hand, big foundation models encapsulate abundant multimodal knowledge about the visual world in their model weights. On the other hand, the knowledge encoded in model weights can be dated soon without timely model update, while various types of knowledge are evolving in real world, for example, factual knowledge in databases.  One solution is to enhance pre-trained models
using external knowledge. Knowledge-enhanced NLP models (also called retrieval-augmented methods) have been widely studied for knowledge-intensive NLP tasks~\citep{guu2020retrieval,lewis2020retrieval}, while the exploration of knowledge-enhanced vision and multimodal models are still in its infant stage. See Section~\ref{sec:knowledge} and \ref{sec:vlp_vision_advanced_topics} for more detailed discussions.

\paragraph{Robustness.} We typically evaluate models on the standard and well established benchmarks, such as ImageNet classification, COCO object detection, and VQAv2. On one hand, these benchmarks have driven tremendous progress in the field, allowing top research teams around the world to advance state of the art on top of each other's work.
On the other hand, we also need to be careful not to \emph{over-claim} the model capabilities, before carefully-designed robustness evaluation is performed. As discussed in Section~\ref{sec:robustness}, better diagnostic tests and more robust methods should be developed.     

\paragraph{Concluding Remarks.}
The aforementioned research directions are deeply connected to achieve the same goal of developing a general-purpose, multi-sense, AI system. For example, the model architectures developed in \textbf{unified modeling} could lead to a better solution to \textbf{computer vision in the wild}, and the techniques we developed for \textbf{model scaling} can be used for scaling the unified model. 
Further, when the model is significantly scaled up, the capability of \textbf{in-context few-shot learning} may emerge naturally. With a few light-weight adapters and a few in-context examples, we envision that the unified multimodal foundation model is capable of \textbf{efficiently adapting} itself to different tasks. In addition, \textbf{external knowledge} is an additional source for further enhancing the performance. Lastly, in order to deploy these state-of-the-art models in real-world applications, we also need to improve \textbf{robustness} and \emph{cost-efficiency}. 

The VLP field is progressing at a rapid speed, with new ideas and methods emerging constantly.
There are many important research topics that are not discussed in this paper, mostly due to an ironic observation that it is impossible for our writing to catch up with the daily-updated research innovation. 
We feel glad and blessed to write this paper, as this is an exciting journey to review the progress that we have made as a community. 
We are optimistic about the future of the VLP field, not only because there are so many new research directions to explore, but also because we are convinced that connecting the two important fields in AI, NLP and computer vision, is going to significantly advance the state of the art of AI in the near future.

\chapter* {Acknowledgments} 
Many people have supported us and provided valuable feedback to the writing of this book. 
This book is largely based on our CVPR 2022 tutorial on vision-language pre-training (VLP). We especially thank Zi-Yi Dou, Jianfeng Wang, Zhengyuan Yang, and Xiaowei Hu  for providing valuable materials on ``VLP for image-text tasks''; Jianwei Yang and Pengchuan Zhang for their inputs to ``VLP for core vision tasks''; Chung-Ching Lin and Kevin Lin for tutorials on ``VLP for video-text tasks''; and Chenfei Wu for contributions to ``VLP for Text-to-Image Synthesis''.
This book is also partially based on our CVPR 2021 and 2020 tutorials, for which we thank Luowei Zhou, Licheng Yu, Yu Cheng, Yen-Chun Chen, Jingjing Liu and Xiaodong He for their contributions.
We are also grateful to the anonymous reviewers for their insightful feedback, and Mark de Jongh for making the publication of this book possible.

\bibliographystyle{apalike}
\bibliography{ref}

\end{document}